\documentclass{article}[a4]
\usepackage{natbib}

\include{header}
\usepackage[utf8]{inputenc} 
\usepackage[T1]{fontenc}    
\usepackage{booktabs}       
\usepackage{amsfonts}       
\usepackage{nicefrac}       
\usepackage{microtype}      
\usepackage{xcolor}         

\usepackage{tabularx}
\usepackage{thmtools}
\usepackage{thm-restate}
\usepackage{fullpage}

\usepackage{comment}
\usepackage{hyperref}
\definecolor{myblue}{rgb}{0,0.2,0.8}
\hypersetup{ %
pdftitle={},
pdfauthor={},
pdfsubject={},
pdfkeywords={},
pdfborder=0 0 0,
pdfpagemode=UseNone,
colorlinks=true,
linkcolor=myblue,
citecolor=myblue,
filecolor=myblue,
urlcolor=myblue,
pdfview=FitH}
\usepackage[utf8]{inputenc} 
\usepackage[T1]{fontenc}    
\usepackage{hyperref}       
\usepackage{url}            
\usepackage{booktabs}       
\usepackage{amsfonts}       
\usepackage{nicefrac}       
\usepackage{microtype}      

\usepackage{microtype}
\usepackage{graphicx}
\usepackage{subfigure}

\usepackage{sidecap}
\sidecaptionvpos{figure}{c}

\usepackage{booktabs}
\usepackage{enumitem}
\usepackage{amsmath, amsfonts, color}

\definecolor{behnam}{rgb}{.6,0,0}
\definecolor{hartmut}{rgb}{0,.6,.6}
\definecolor{robert}{rgb}{.6,0,.6}
\definecolor{todo}{rgb}{0.7,0,0}

\newcommand{\cmt}[1]{}
\newcommand{\authfont}{11}
\usepackage{anyfontsize}
\newcommand{\removed}[1]{}
\usepackage{hyperref}

\title{\bf{Deep Learning Through the Lens of Example Difficulty}}

\author{
\fontsize{\authfont}{20}\selectfont Robert J. N. Baldock\thanks{Work completed as part of the Google AI Residency Program}\\
\fontsize{10.1}{20}\selectfont Google Research, Brain Team\\
\fontsize{10.1}{20}\selectfont \texttt{rjnbaldock@gmail.com}\\
\and \fontsize{\authfont}{20}\selectfont Hartmut Maennel\\
\fontsize{10.1}{20}\selectfont Google Research, Brain Team\\
\fontsize{10.1}{20}\selectfont \texttt{hartmutm@google.com}\\
\and \fontsize{\authfont}{20}\selectfont Behnam Neyshabur\\
\fontsize{10.1}{20}\selectfont Google Research, Blueshift Team\\
\fontsize{10.1}{20}\selectfont \texttt{neyshabur@google.com}\\
}

\date{}
\begin{document}

\setenumerate{itemsep=0pt}
\setitemize{itemsep=1pt}
\setlength{\parskip}{0.1em}

\maketitle
\begin{abstract}
Existing work on understanding deep learning often employs measures that compress all data-dependent information into a few numbers. In this work, we adopt a perspective based on the role of individual examples. We introduce a measure of the computational difficulty of making a prediction for a given input: the \emph{(effective) prediction depth}. Our extensive investigation reveals surprising yet simple relationships between the prediction depth of a given input and the model’s uncertainty, confidence, accuracy and speed of learning for that data point. We further categorize difficult examples into three interpretable groups, demonstrate how these groups are processed differently inside deep models and showcase how this understanding allows us to improve prediction accuracy. Insights from our study lead to a coherent view of a number of separately reported phenomena in the literature: early layers generalize while later layers memorize; early layers converge faster and networks learn easy data and simple functions first.
\end{abstract}
\section{Introduction \label{sec:intro}}
Much of the existing work on understanding deep learning ``integrates out'' the data, viewing the inductive bias of the model, or the properties of the optimizer as central to the success of the approach.
Examples of such work include studies of eigenvalues of the Hessian  and the geometry of the loss landscape~\citep{ghorbani2019investigation, yao2019pyhessian, sagun2016eigenvalues, li2017visualizing, pennington2017geometry, sagun2017empirical}, studies of margin and effective generalization measures~\citep{long2019generalization, unterthiner2020predicting, jiang2019fantastic, jiang2018predicting, kawaguchi2017generalization} and mean-field studies of stochastic optimization~\citep{smith2021origin, stephan2017stochastic, smith2018bayesian}. 
However, in practice, we are rarely concerned with only the average behavior of a model.

One pathway to understanding the principles that govern how deep models  process data is to study the properties of deep models for data points with different ``amounts'' or ``types'' of \emph{example difficulty}.
There are a number of definitions of example difficulty in the literature (E.g. see ~\cite{carlini2019distribution, hooker2019compressed, lalor2018understanding, hooker_exdiff_gradvar}).
Two are particularly relevant to this work.
Firstly, the probability of predicting the ground truth label for an example, when that example is omitted from the training set~\citep{chiyuan_cscores},
which represents a \emph{statistical view} of example difficulty.
Secondly, the difficulty of learning an example, parameterized by the
earliest training iteration after which the model predicts the ground truth class for that example in all subsequent  iterations~\citep{forgetting19}.
This measure represents a~\emph{learning view} of example difficulty~\footnote{We expand on other notions of example difficulty in Appendix~\ref{app:context}.}.

These notions suffer from two fundamental limitations. While early-exit strategies in computer vision~\citep{teerapittayanon2016branchynet, huang2018multi} and NLP~\citep{dehghani2018universal, liu2020fastbert, schwartz2020right, xin2020deebert} suggest predictions for easier examples require less computation, the above example difficulty notions do not encapsulate the processing of data inside a given converged model. Moreover, existing notions of example difficulty (E.g.~\citet{carlini2019distribution}) provide a one-dimensional view of difficulty which can not distinguish between examples that are difficult for different reasons.

In this paper, we take a significant step towards resolving the above shortcomings. To take the processing of the data into account we propose a new measure of example difficulty, the prediction depth, which is determined from the hidden embeddings. To escape the one-dimensional view of difficulty, we introduce three distinct difficulty types by relating the hidden embeddings for an input to high-level concepts about example difficulty: ``Does this example look mislabeled?''; ``Is classifying this example only easy if the label is given?''; ``Is this example ambiguous both with and without its label?''. Furthermore, we show how this enhanced notion of example difficulty can unify our understanding of several seemingly unrelated phenomena in deep learning. We hope that the results presented in this work will aid the
development of models that capture heteroscedastic uncertainty, our understanding of how deep networks respond to distributional shift, and the advancement of curriculum learning approaches and machine learning fairness. These connections are discussed in Section~\ref{sec:discussion}.

\begin{figure}[t]
	\centering
	\includegraphics[trim=0 0 20 0, clip,width=0.62\linewidth]{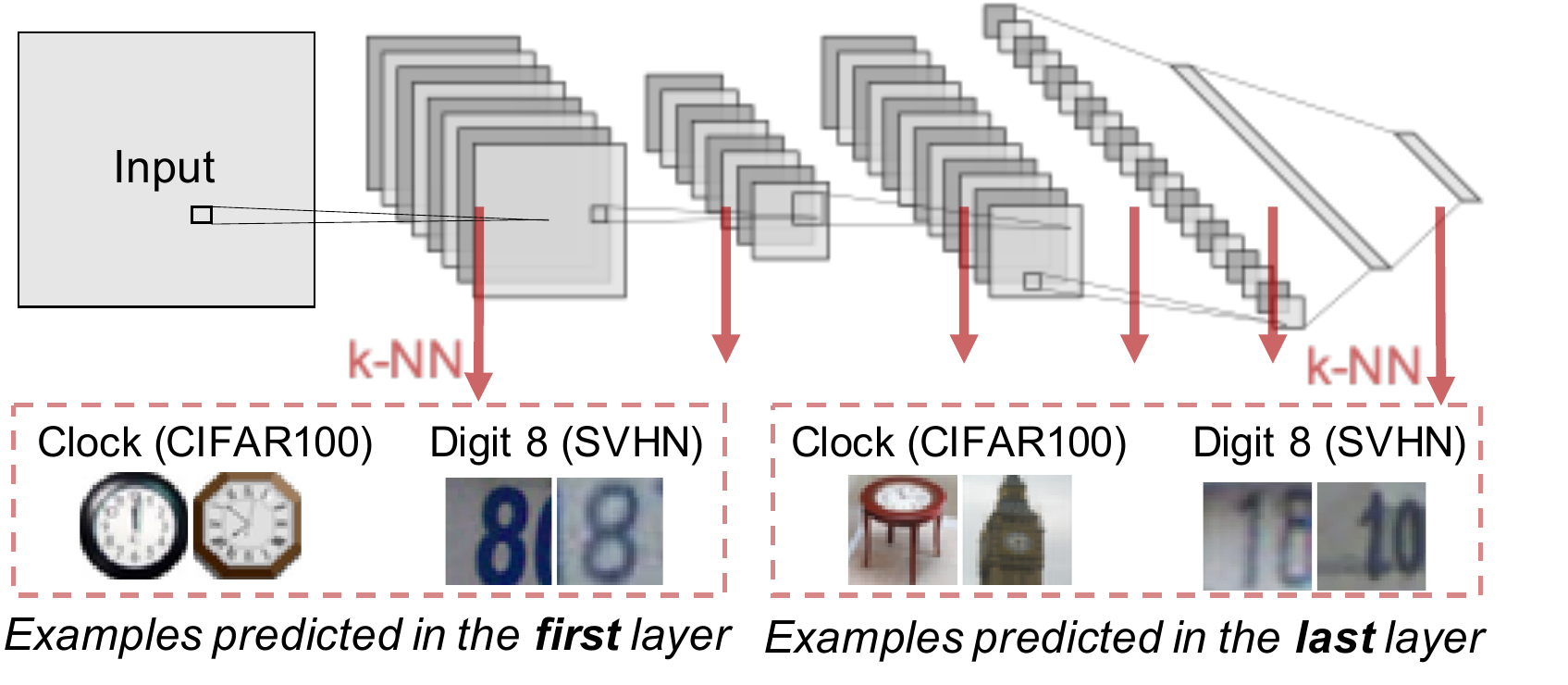}
	\includegraphics[width=0.37\linewidth]{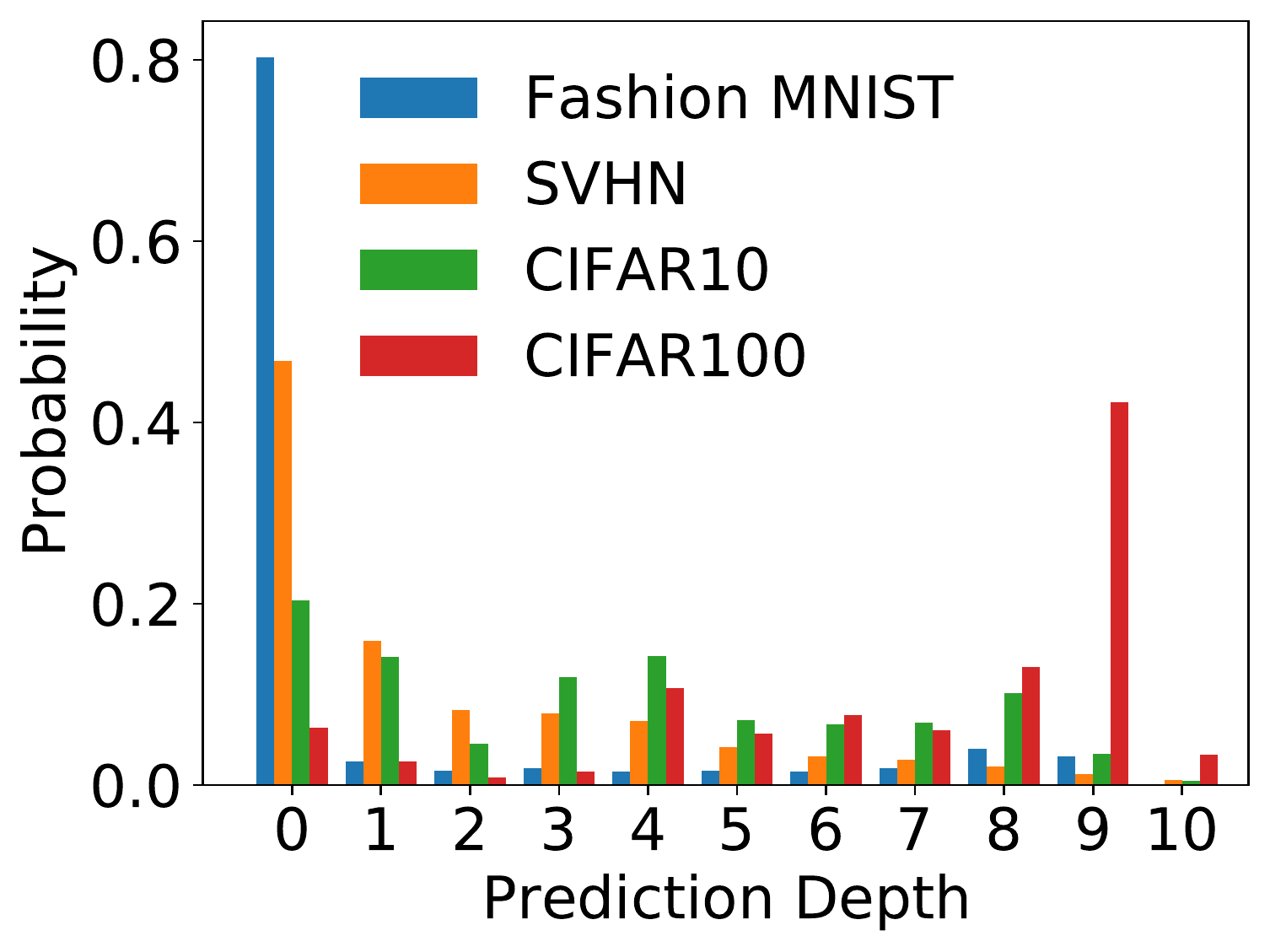}
\caption{\small \emph{ Deep models use fewer layers to (effectively) determine the prediction for easy examples and more layers for hard examples.} 
{\bf Left:} A cartoon illustrating the definition of prediction depth (given in Section~\ref{sec:def_ll}).
Also shown are training examples from CIFAR100 (``Clock'') and SVHN (``Digit 8'').
The examples shown are predicted at the input (first layer) or softmax (last layer) of ResNet18.
The examples predicted in the input are visually typical (``easy''), while those predicted in the softmax are mislabeled and/or visually confusing (``hard'' examples).
To find the prediction depth, we build k-NN classifiers from the embeddings of the training set in different layers of the model. The prediction depth corresponds to the earliest layer at which the predictions of all subsequent k-NN classifiers converge to a fixed label.
{\bf Right:} Probability of prediction depth in ResNet18 models for four datasets (training split).
We see that the four distributions have different characteristic prediction depths.
Ranking the mean prediction depths of these datasets in ascending order, we observe: Fashion MNIST (smallest), SVHN (second), CIFAR10 (third), and CIFAR100 (largest).
This order aligns with how one might intuitively rank the difficulties of these classification tasks.
}
\label{fig:LL_train}
\end{figure}

\paragraph{Contributions} 

Our main contributions are as follows:
\begin{itemize}
    \item We introduce a measure of \emph{computational example difficulty}: the \emph{prediction depth} (PD). The prediction depth, illustrated in Figure~\ref{fig:LL_train}, represents the number of hidden layers after which the network's final prediction is already (effectively) determined (Section~\ref{sec:on_ex_diff}).
    \item We show that the prediction depth is larger for examples that visually appear to be more difficult, and that prediction depth is consistent between architectures and random seeds (Section~\ref{sec:ll_sanity_check}).
    \item Our empirical investigation reveals that prediction depth appears to establish a \emph{linear} lower bound on the consistency of a prediction. We further show that predictions are on average more accurate for validation points with small prediction depths (Section~\ref{sec:ll_decision_consistency}).
    \item We demonstrate that final predictions for data points that converge earlier during training are typically determined in earlier layers which establishes a correspondence between the training history of the network and the processing of data in the hidden layers (Section~\ref{sec:compdiff_vs_learndiff}). 
    \item We show that both the adversarial input margin and the output margin are larger for examples with smaller prediction depths. We further design an intervention to reduce the output margin of a network and show that this leads to predictions being made only in the latest hidden layers (Section~\ref{sec:simplicity_vs_compdiff}).     
    \item We identify three extreme forms of example difficulty by considering the prediction depth in the training and validation splits independently and demonstrate how a simple algorithm that uses the hidden embeddings in one middle layer to make predictions can lead to dramatic improvements in accuracy for inputs that strongly exhibit a specific form of example difficulty (Section~\ref{sec:4_ex_diff}). 
    \item We use our results to present a coherent picture of deep learning that unify four seemingly unrelated deep learning phenomena: early layers generalize while later layers memorize; networks converge from input layer towards output layer; easy examples are learned first and networks present simpler functions earlier in training (Section~\ref{sec:discussion}). 
\end{itemize}

{\bf Experimental Setup:} To ensure that our results are robust to the choice of architectures and datasets, we report empirical findings for ResNet18~\citep{he2016deep}, VGG16~\citep{simonyan2014very} and MLP architectures trained on CIFAR10, CIFAR100 \citep{krizhevsky2009learning},
Fashion MNIST (FMNIST)~\citep{xiao2017fashion} and SVHN \citep{netzer2011reading} datasets.
All models were trained using SGD with momentum. Our MLP comprises 7 hidden layers of width 2048 with ReLU activations. Details of the datasets, architectures, and hyperparameters used can be found in Appendix~\ref{app:experiments_desc}.

{\bf Related Work:} Our work uses hidden layer probes to determine example difficulty. We have discussed how our study relates to prior work on example difficulty. Hidden layer probes have also been used to study deep learning. Deep k-NN methods~\citep{papernot2018deep} determine their predictions and estimate their own uncertainties by comparing the hidden embeddings of an input to those of the training set.  \citet{cohen2018dnn} showed that SVM, k-Nearest Neighbors (k-NN) and logistic regression probes achieve similar accuracies. However, they did not study the processing of individual data points nor did they relate the k-NN accuracy to notions of example difficulty. \citet{alain2016understanding} used linear classifier probes in the hidden layers to interrogate deep models and demonstrated that linear separability of the embeddings increases monotonically with depth. We provide a more detailed discussion of related work in Appendix~\ref{app:context}.

\section{Prediction Depth: a Computational View of Example Difficulty\label{sec:on_ex_diff}}
We discussed the \emph{statistical} and \emph{learning} views of example difficulty  in Section~\ref{sec:intro}. In this section, we introduce a \emph{computational} view of example difficulty parametrized by the prediction depth as defined in Section~\ref{sec:def_ll}. This computational view asserts that, for ``easy'' examples, a deep model's final prediction is effectively made after only a few layers, while more layers are used for ``difficult'' examples.

\begin{figure}[t]
	\centering
	\begin{subfigure}
\centering
   \includegraphics[width=0.48\linewidth]{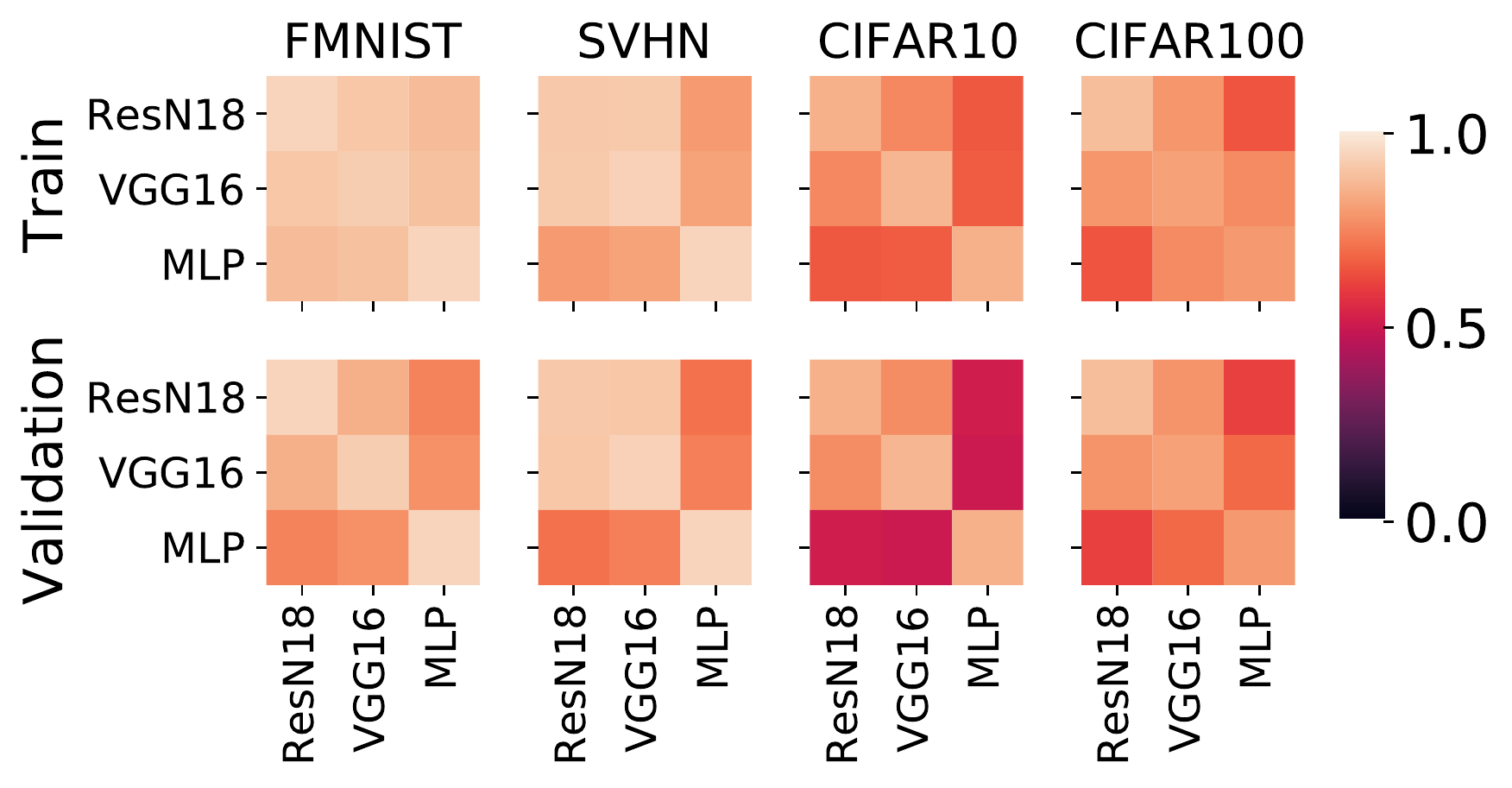}
\end{subfigure}
   \begin{subfigure}
\centering
   \includegraphics[width=0.50\columnwidth]{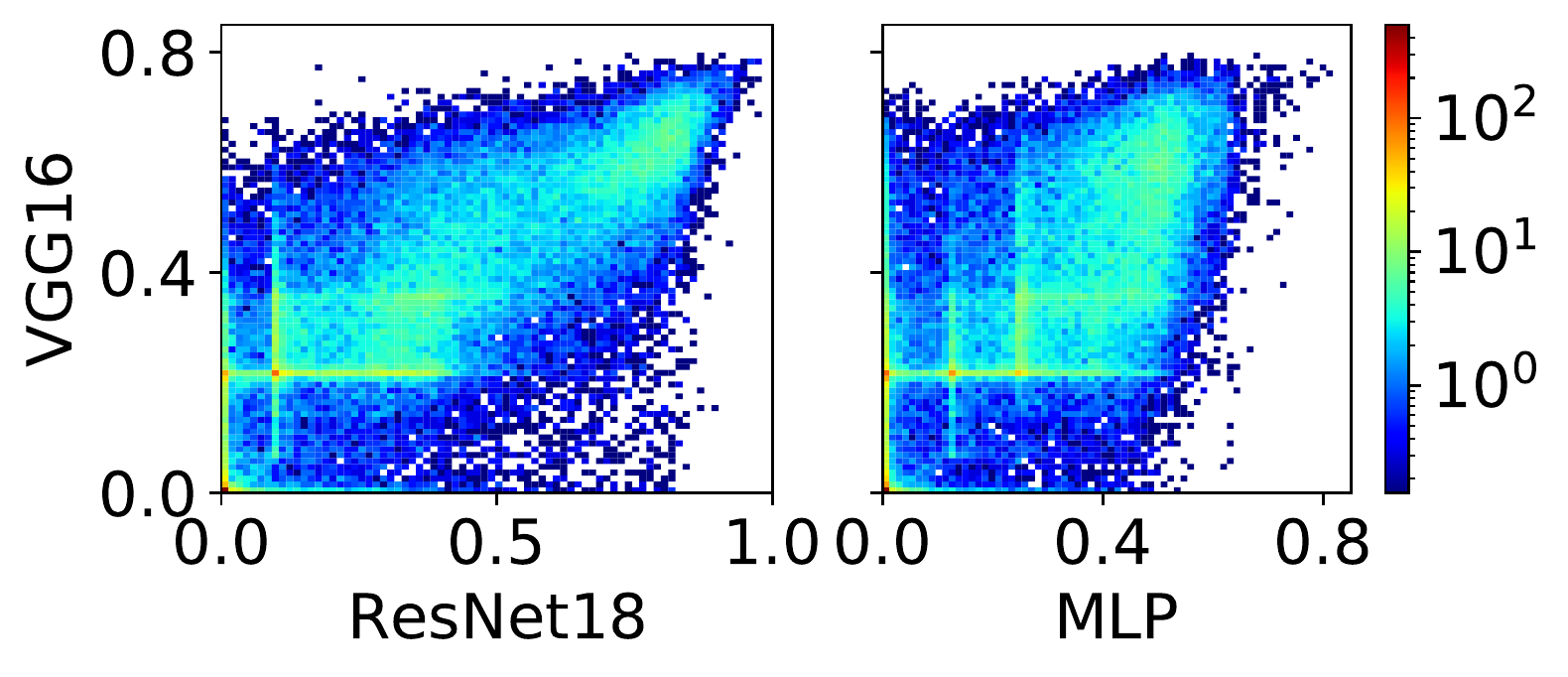}
\end{subfigure}
\caption{%
        \small \emph{ Consistency of prediction depth between architectures and random seeds.} {\bf Left:}  The panel shows the correlation coefficient between prediction depths in different architectures, for both train and validation splits in four datasets. Diagonal comparisons between an architecture and itself show the correlation for the same architecture trained with different random seeds. {\bf Right:} Histograms comparing the mean value of prediction depth obtained for each data point in the training set of CIFAR10 from an ensemble of 250 trained models. In this plot, for visual simplicity, we rescale prediction depth to the interval $[0,1]$ for each network. Similar results for all other datasets are presented in Appendix~\ref{app:consistency_ll_archs}.
\label{fig:ll_between_archs_svhn_main}
}
\end{figure}

\subsection{Definition \label{sec:def_ll}}

Asserting that the final prediction is effectively determined in earlier layers of a model, before the output, we estimate the depth at which a prediction is made for a given input as follows~\footnote{In the process of arriving at this definition of the prediction depth we considered several alternatives, including using the ground truth class in place of the predicted class and using logistic regression probes in place of k-NN probes. See Appendix~\ref{app:ll_alt_defs} for a discussion on the choices we made in our definition.}: 
\begin{enumerate}
    \item We construct k-NN classifier probes from the embeddings of the training set after particular layers of the network, including the input and the final softmax. 
    The placement of k-NN probes is described in Appendix~\ref{app:knn_probe_placement}.    
    We use $k=30$ in the k-NN probes.
    Appendix~\ref{app:knn_conv} establishes that  the k-NN accuracies we report are insensitive to $k$ over a wide range. 
    
    \item A prediction is defined to be made at a depth $L = l$ if the k-NN classification after layer $L= l-1$ is different from the network's final classification, but the classifications of k-NN probes after every layer $L \ge l$ are all equal to the final classification of the network. Data points consistently
    classified by all k-NN probes are determined to be (effectively) predicted in layer $0$ (the input)~\footnote{Implementation details can be found in Appendix~\ref{app:knn_matches_model}.}.
\end{enumerate}

It is worth noting that the prediction depth can be calculated for all data points: both in the training and validation splits.
This leads to two notions of computational difficulty:
\begin{itemize}
    \item The difficulty of predicting the (given) class for an input (in the training split)
    \item The difficulty of making a prediction for an input, unseen in advance (from the validation split)
\end{itemize}
We examine both notions of computational difficulty in this paper and use the distinction between them to describe different forms of example difficulty in Section~\ref{sec:4_ex_diff}.

\subsection{Prediction depth is a meaningful and robust notion of example difficulty\label{sec:ll_sanity_check}}

In this section we show that prediction depth agrees with intuitive notions of example difficulty and that it is consistent between different training runs and similar architectures.

\paragraph{Prediction depth is higher for examples and datasets that seem more difficult}If prediction depth is a sensible measure of example difficulty then we would expect the following sanity checks to be observed:
\begin{enumerate}
    \item Individual data points that are visually confusing or mislabeled should have larger prediction depths as compared to images that are clear examples of their class.
    \item Data points from tasks that are intuitively simpler should have lower prediction depths on average.
\end{enumerate}
Figure~\ref{fig:LL_train} shows that the prediction depth passes both of these sanity checks.

\paragraph{Prediction depth is consistent across random seeds and similar architectures} Figure~\ref{fig:ll_between_archs_svhn_main} shows that the prediction depth is highly consistent between different architectures and random seeds for all datasets. Perfect agreement is not expected as different deep learning algorithms have different inductive biases which affects the perceived difficulty of examples. 
We observe stronger correlation between prediction depth for ResNet18 and VGG16, than between VGG16 and MLP.
This may be explained by the fact that ResNet18 and VGG16 are both convolutional networks and we expect their inductive biases to be more similar to one another than to MLP.

\begin{figure}[t]
\begin{center}
     \begin{subfigure}
\centering
   \includegraphics[width=0.324\columnwidth]{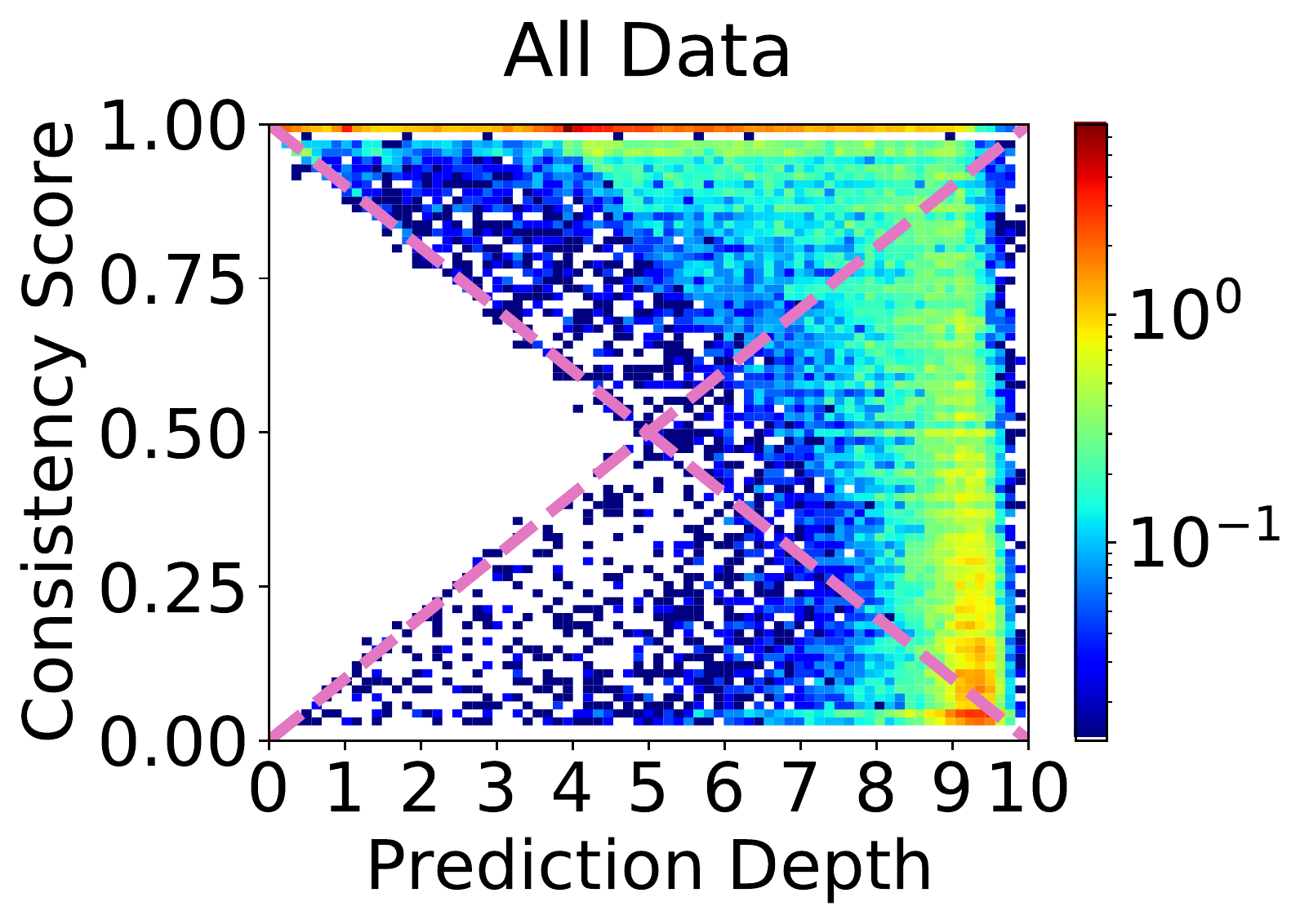}
\end{subfigure}
   \begin{subfigure}
\centering
   \includegraphics[width=0.324\columnwidth]{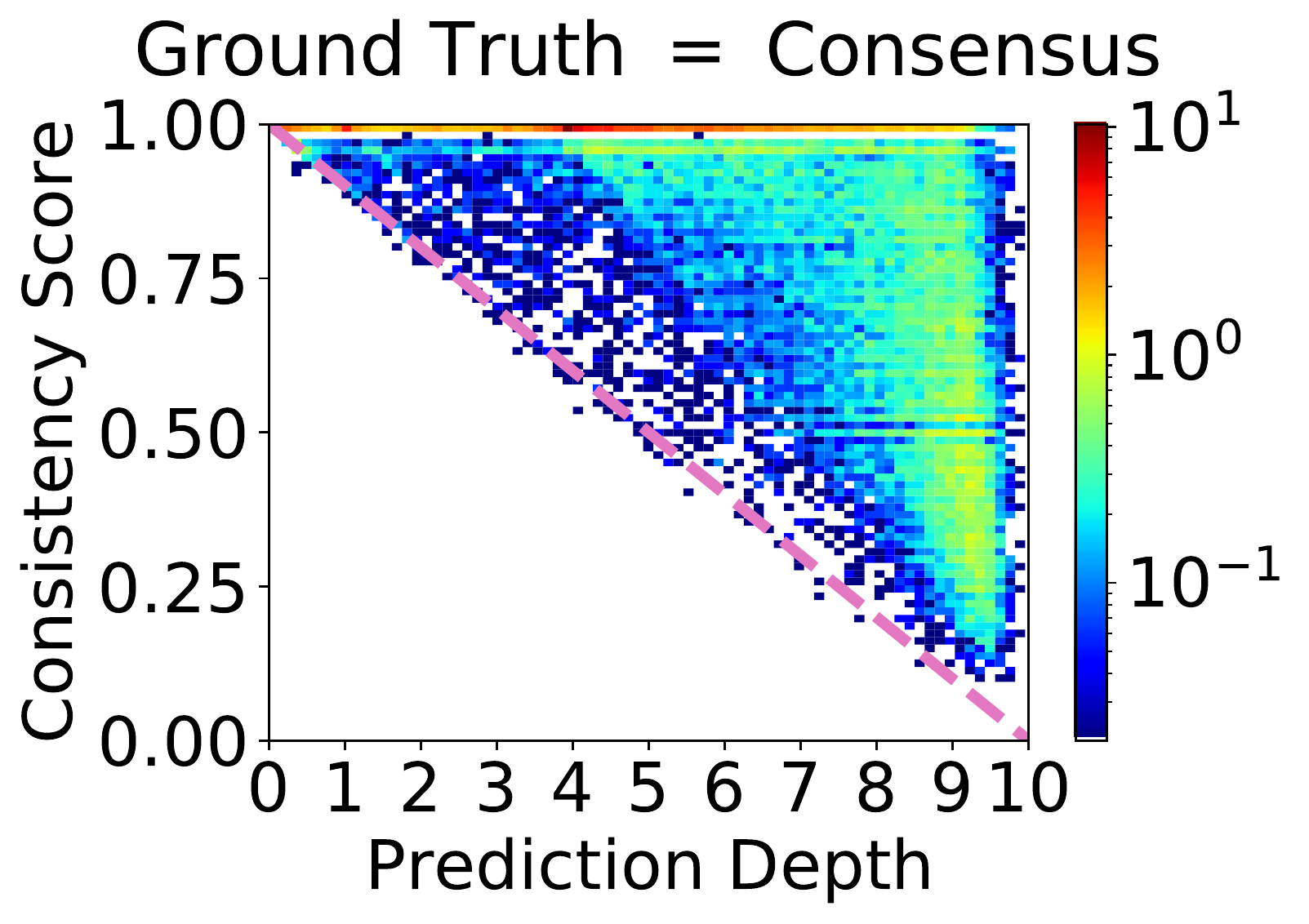}
\end{subfigure}
   \begin{subfigure}
\centering
   \includegraphics[width=0.324\columnwidth]{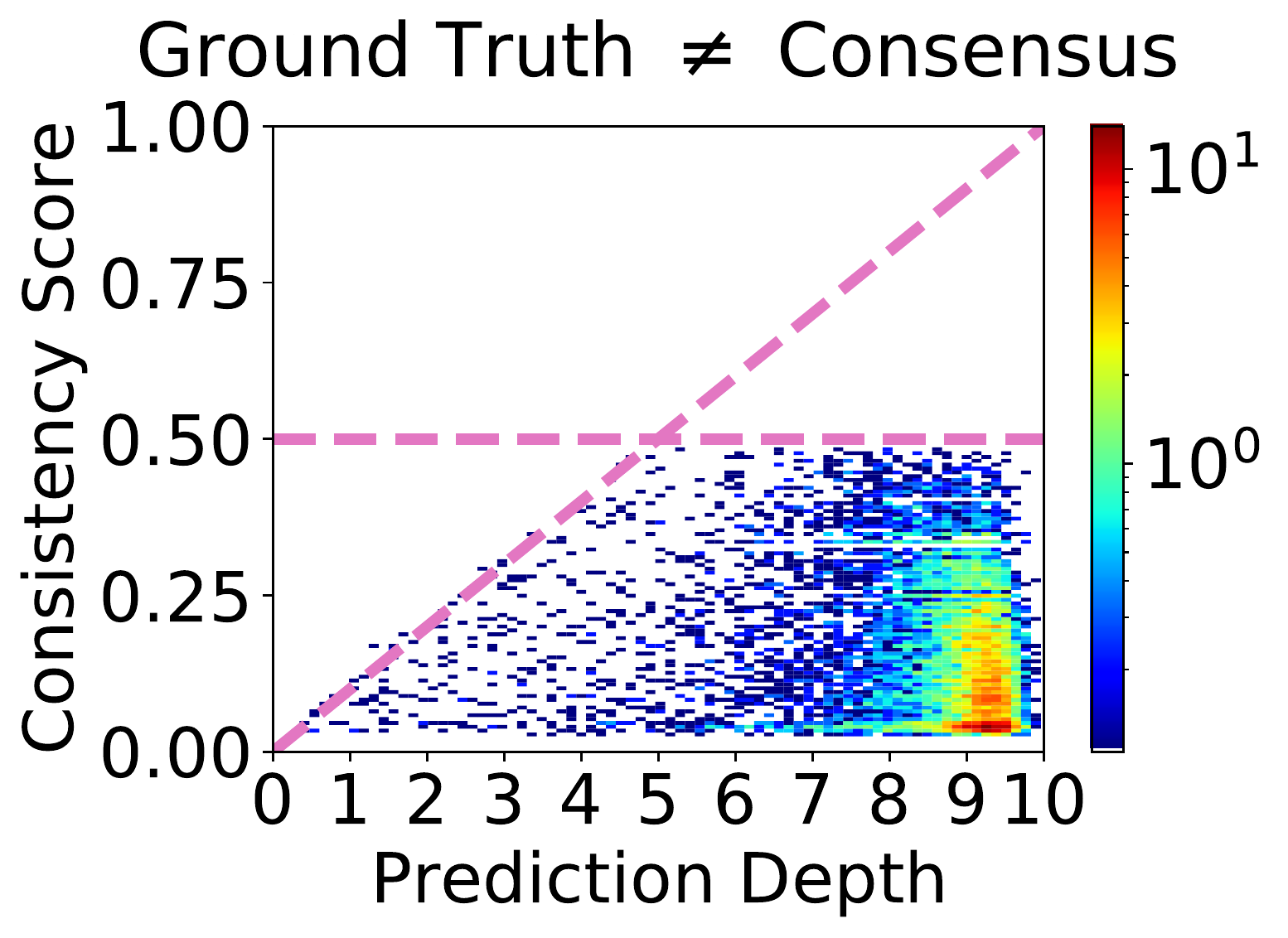}
\end{subfigure}
     \end{center}
\caption{\small \emph{ Consistency score vs. prediction depth in the validation split (left) can be understood as the superposition of two simple functions (middle and right).}
We trained 250 ResNet18 models on CIFAR10, with 90:10\% random train:validation splits as described in Appendix~\ref{app:experiments_desc}.
These histograms compare the frequency of correct predictions to the average prediction depth for a data point when it occurs in the validation split.
The average prediction depth forms two, surprisingly simple, linear bounds on the consistency score (see Section~\ref{sec:ll_decision_consistency} for a full description.)
This Figure is reproduced for all datasets and architectures in Appendix~\ref{app:ll_pcpm}, illustrating the consistency of this result.
\label{fig:generalization_vs_depth}
}
\end{figure}

\section{Deep Learning Phenomena Through the Lens of Prediction Depth\label{sec:dl_phenomena}}
\begin{figure}[t]
\begin{center}
         \begin{subfigure}
         \centering
         \includegraphics[width=0.35\columnwidth]{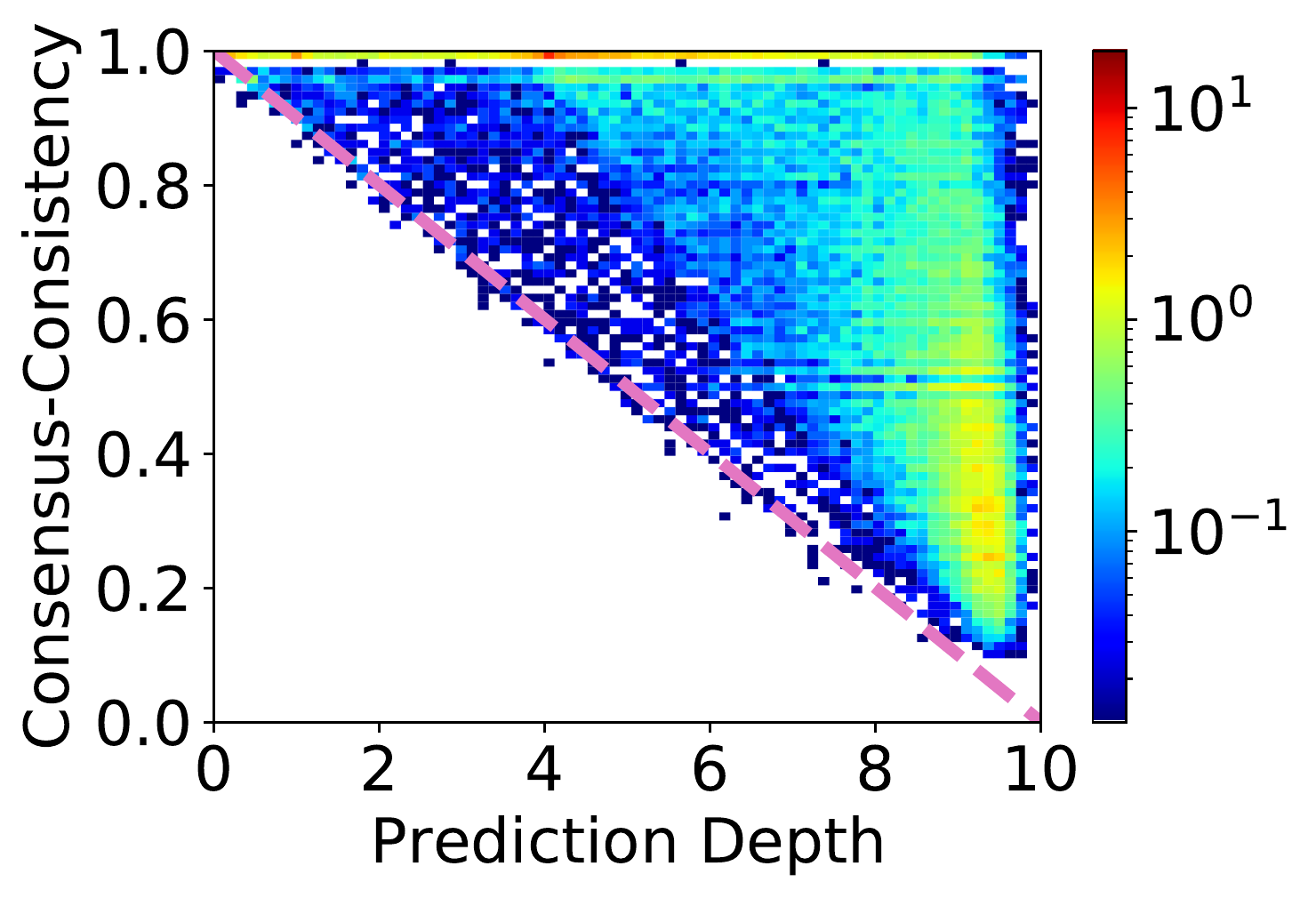}
\end{subfigure}
\begin{subfigure}
         \centering
         \includegraphics[width=0.31\columnwidth]{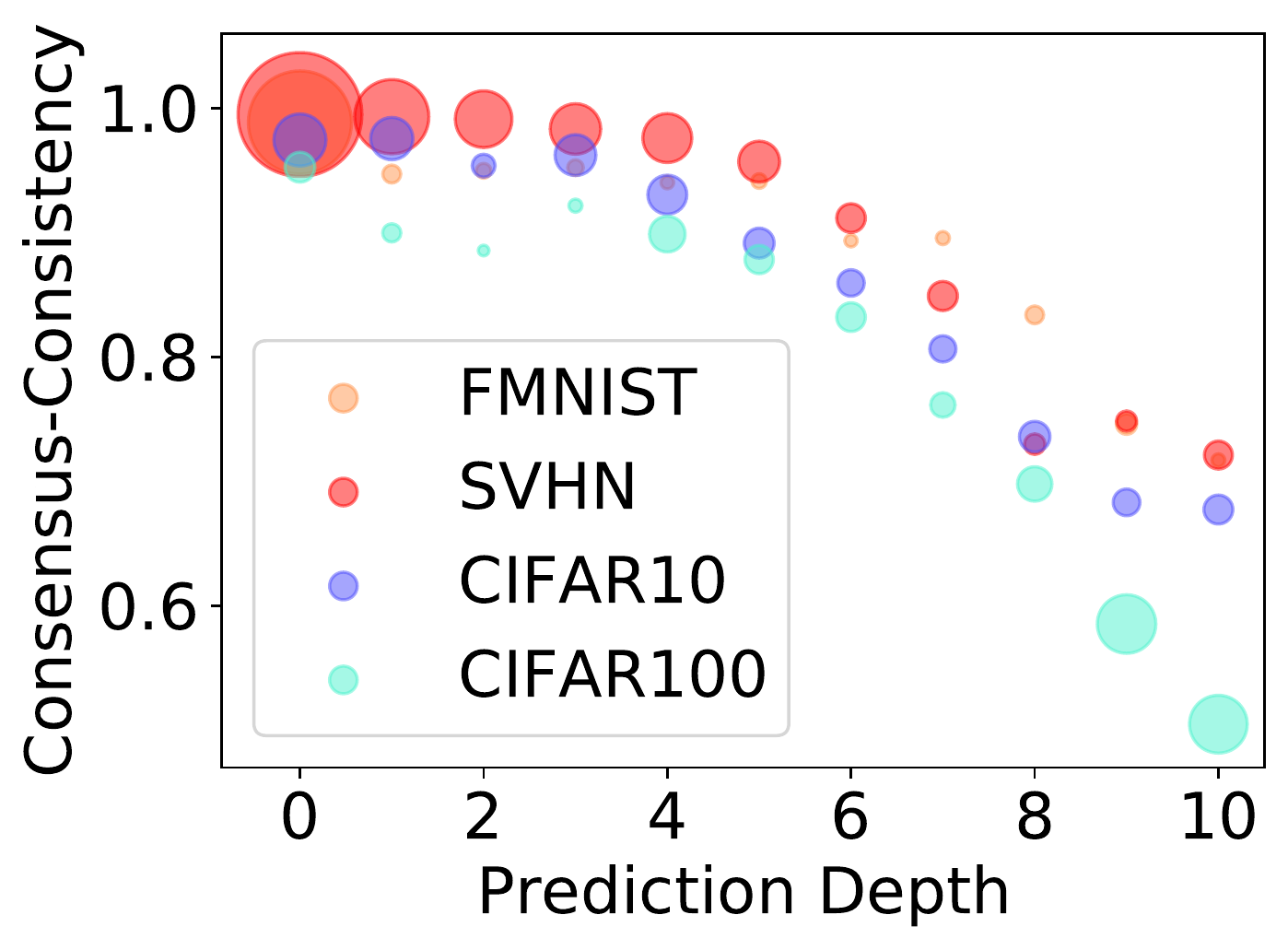}
\end{subfigure}
\begin{subfigure}
         \centering
         \includegraphics[width=0.31\columnwidth]{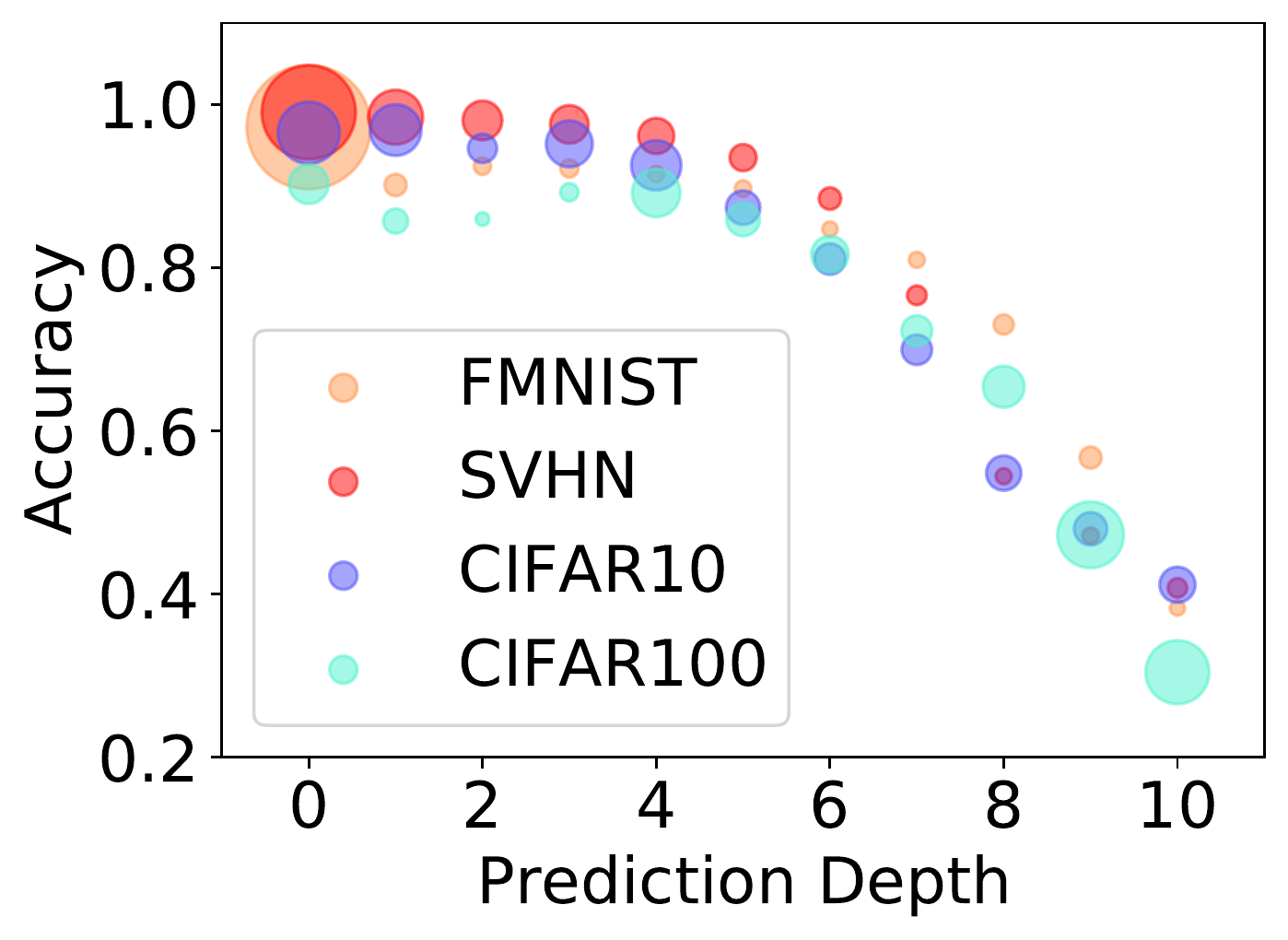}
\end{subfigure}
     \end{center}
\caption{\small  {\bf Left:} \emph{Prediction depth provides us with a linear lower bound on consensus-consistency.} 
Results for CIFAR100 with ResNet18.
We train 250 models (90:10\% random train:validation splits) and compare the average prediction depth when a point occurs in the validation set, to the consensus-consistency of the corresponding predictions.
Predictions made for points with low mean prediction depths are highly consistent.
Conversely, predictions for points with high mean prediction depths are typically more sensitive to the particular training split and random seed used during training.
This left plot shows the result for CIFAR100 with ResNet18.
{\bf Middle:} \emph{Prediction depth in one model predicts the consensus-consistency of an ensemble that does not include that model.}
For each dataset we train 25 ResNet18 models with the full training set (see Appendix~\ref{app:experiments_desc}).
The consensus-consistency of each test point is obtained from 24 of the models, while the prediction depth is obtained from the remaining 1 model.
We see that prediction depth in one model predicts the consensus-consistency of a separate ensemble: a measure of the uncertainty of the prediction.
The size of each marker in the middle and right plots shows the fraction of the dataset with each prediction depth.
{\bf Right:} \emph{Prediction depth predicts accuracy.} For each dataset we train 250 ResNet18 models (90:10\% random train:validation splits). Each time a point appears in the validation split we record the prediction depth and whether the prediction was correct. Predictions made in earlier layers are more likely to be correct.
Consistency of these plots is demonstrated for all datasets and architectures in Appendix~\ref{app:ll_pcpm} where we also describe the relationship between the prediction depth and the \emph{entropy of the predictions} for an ensemble.}
\label{fig:ent_vs_depth}
\end{figure}

In this section, we explore how the prediction depth can be used to better understand three important aspects of Deep Learning: accuracy and consistency of a prediction; the order in which data is learned and the simplicity of the learned function (as measured by the margin) in the vicinity of a data point.

\subsection{Depth of a prediction gives a linear lower bound on its consistency\label{sec:ll_decision_consistency}}

Adopting a \emph{statistical view} of example difficulty, \citet{chiyuan_cscores} identified example difficulty with the expected accuracy of the learning algorithm for a given input, averaged over models trained on different random subsets of the training set with different random seeds.
In this section, we clarify the relationship between the prediction depth and the expected accuracy by disentangling the accuracy from the sensitivity of predictions to the particular training split and random seed.
Following \citet{chiyuan_cscores}, we measure the expected accuracy using the consistency score.

\begin{description}
\item[Consistency score $\hat{C}$:]
Consistency score is the frequency of classifying an example correctly when it is omitted from the training set.
An empirical estimator of the consistency score for a validation point $(x,y)$ is given by~\citep{chiyuan_cscores}:
\begin{equation}
    \hat{C}_{A,\mathcal{S}}(x,y) = \hat{\mathbb{E}}^r_{\tilde{\mathcal{S}}\stackrel{n}{\sim} \mathcal{S}\backslash\{(x,y)\}}     \left[\delta_{y_{A},y}\right] 
    \label{eq:empirical_cscore}
\end{equation}
where $A$ is a deep learning algorithm (architecture, loss and optimizer), $y$ is the ground truth class for $x$, $\tilde{\mathcal{S}}$ is a random subset of $n$ points sampled from a training dataset $\mathcal{S}$ excluding $(x,y)$, $y_{A}$ is the predicted class of $x$ for $A$ trained with data $\tilde{\mathcal{S}}$, $\delta$ is the Kronecker delta and $\hat{\mathbb{E}}^r$ denotes empirical averaging with $r$ i.i.d. samples of such subsets $\tilde{\mathcal{S}}$.
\end{description}
Figure~\ref{fig:generalization_vs_depth} (left panel) shows the relationship between consistency score and prediction depth. This plot indicates a surprising piecewise linear boundary which is symmetric around consistency score $\frac{1}{2}$. This suggests the existence of a missing concept that could simplify the picture. We next show that the missing concept is the notion of a consensus class which is defined below.

\begin{description}
\item[Consensus class $\hat{y}_A$:]The \emph{consensus class} of $x$ is defined as the predicted class for input $x$ by a majority voting ensemble of $r$ models each of which is trained on a randomly chosen subset $\tilde{\mathcal{S}}\stackrel{n}{\sim} \mathcal{S}\backslash\{(x,y)\}$~\footnote{Implementation details can be found in Appendix~\ref{app:mode_note}}.
\end{description}

Figure~\ref{fig:generalization_vs_depth} (middle and right) shows how conditioning on whether consensus class matches the ground truth can change the relationship between consistency score and the prediction depth.
For points where the consensus class matches the ground truth (middle) we see that the prediction depth forms a, surprisingly simple, linear lower bound on the consistency score.
For points where the  consensus class differs from the ground truth (right) at low prediction depth the consistency score is bounded from above by a line that reflects the bound from the middle plot in $\hat{C}=\frac{1}{2}$, suggesting that such points are repeatedly mislabeled with a wrong class label.
At high prediction depth, the consistency score is low, which suggests highly inconsistent predictions and low accuracy.
This result suggests a simple hypothesis:
that predictions with low prediction depth are consistent with the \emph{consensus class}, whether that matches the ground truth class or not, while predictions made in later layers depend strongly on the specific training split and random seed used for training and initialization.
We measure consistency with the consensus class using the consensus-consistency score.

\begin{description}
\item[Consensus-consistency score $C^*$:]
The fraction of models in an ensemble that predict the ensemble's consensus class $\hat{y}_A\left(x\right)$ for an unseen input $x$.
\begin{equation}\label{eq:empirical_modal_cscore}
     C^*_{A,\mathcal{S}}(x) =  \hat{\mathbb{E}}^r_{\tilde{\mathcal{S}}\stackrel{n}{\sim} \mathcal{S}\backslash\{(x,y)\}} \left[\delta_{y_A,\hat{y}_A\left(x\right)}\right]
\end{equation}
where the notation is the same as in~\eqref{eq:empirical_cscore}~\footnote{Consensus-consistency score is a measure of uncertainty and can be used for calibration~\citep{lakshminarayanan2017simple,wenzel2020hyperparameter, wen2019batchensemble}. See Appendix~\ref{app:est_CC} for details of our implementation.}.
\end{description}

Figure~\ref{fig:ent_vs_depth} (left) establishes that our simple hypothesis is indeed correct: the prediction depth forms a linear lower bound on the consensus-consistency score for all data points, irrespective of whether the consensus class matches or differs from the ground truth. 
Interestingly, Figure~\ref{fig:ent_vs_depth} (middle and right) shows how the prediction depth in a single model, can be used to estimate both of these quantities. That is, predictions of data points with lower prediction depth are both more likely to be consistent and more likely to be correct.

\begin{figure}[t]
\begin{center}
         \begin{subfigure}
         \centering
         \includegraphics[width=0.26\columnwidth]{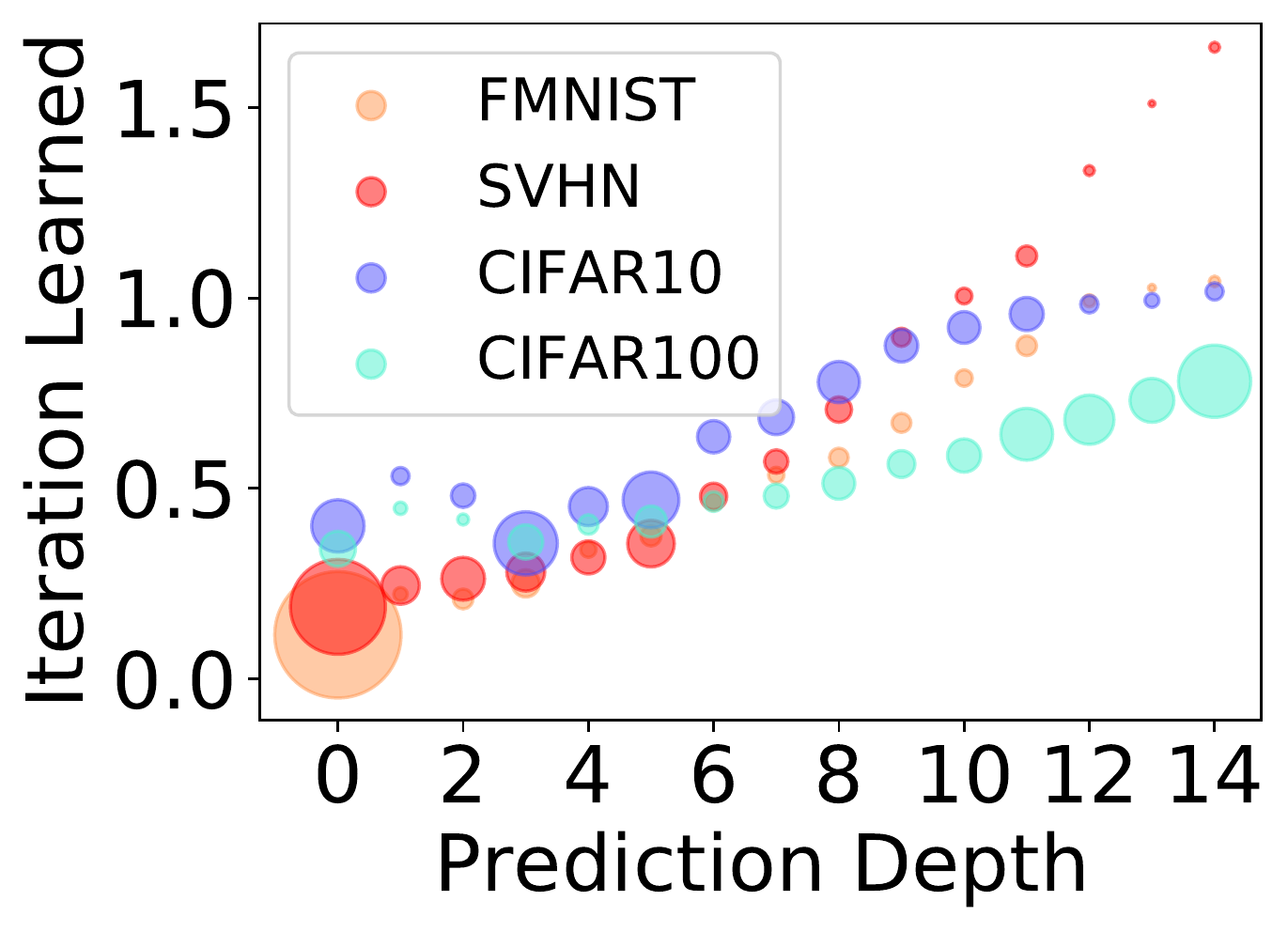}
\end{subfigure}
\begin{subfigure}
         \centering
         \includegraphics[width=0.26\columnwidth]{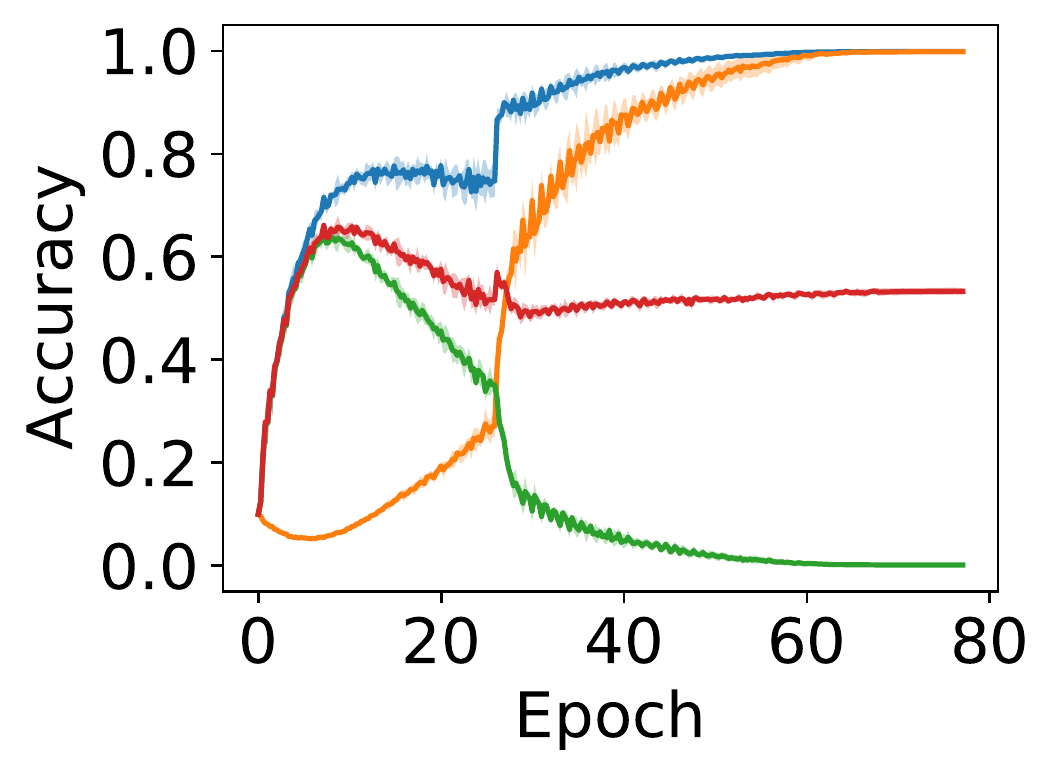}
\end{subfigure}
\begin{subfigure}
         \centering
         \includegraphics[width=0.43\columnwidth]{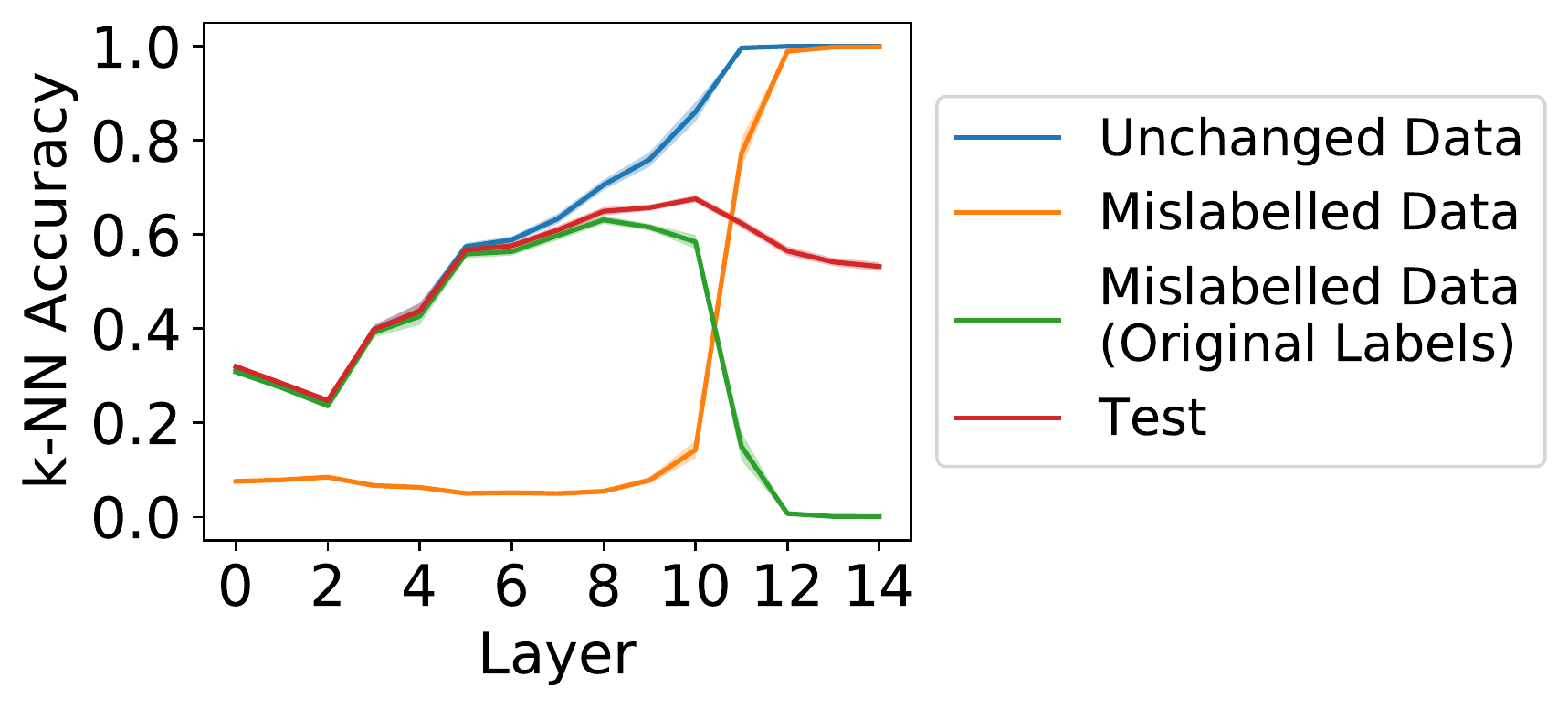}
\end{subfigure}
     \end{center}
\caption{ \small 
{\bf Left:} \emph{Data points with small prediction depths are on average learned before data points with higher prediction depths.} We train 250 VGG16 models for each dataset, using a 90:10\% random train:validation split as described in Appendix~\ref{app:experiments_desc}.
Each time an input appears in the validation split we record the prediction depth and the iteration learned in that model.
This plot shows the average iteration learned for data points at each prediction depth. Marker size shows the fraction of the dataset with each prediction depth.
{\bf Middle and right:} \emph{The training learning curve (middle) shares several important features with the inference learning curve (right).}
Blue, yellow and green curves represent different components of the CIFAR10 training split, in which we have randomized (and fixed) 40\% of the labels, and red curves show the test split. The middle and right plots show results from 5 random seeds.
The inference learning curve (right) is the sequence of k-NN probe accuracy values for each split. All three plots show results for VGG16. The hyperparameters used are given in Appendix~\ref{app:experiments_desc}. 
\label{fig:visual_correspondance}
}
\end{figure}

\subsection{The prediction depth of an input is correlated with its learning difficulty\label{sec:compdiff_vs_learndiff}}
In Section~\ref{sec:ll_decision_consistency}, we describe the relationship between the prediction depth, which represents a \emph{computational view} of example difficulty and the consistency and consensus-consistency scores, which represent a \emph{statistical view}. In this section we compare prediction depth to a \emph{learning view} of example difficulty. We measure the difficulty of learning an example by the speed at which the model's prediction converges for that input during training. The following definition is adapted from~\citet{forgetting19}: 
\begin{description}
\item[Iteration learned] A data point is said to be learned by a classifier at training iteration $ t = \tau $ if the predicted class at iteration $t = \tau - 1$ is different from the final prediction of the converged network and the predictions at all iterations $ t \ge \tau$ are equal to the final prediction of the converged network.
Data points consistently classified after all training steps and at the moment of initialization, are said to be learned in step $t=0$~\footnote{Note that this definition can be applied to points in both training and validation splits. In order to compare different models and datasets we rescale the iteration learned in each model so that the 95th percentile occurs at 1.0 and network initialization at 0.}.
\end{description}

Figure~\ref{fig:visual_correspondance} (left plot) shows the positive correlation between the prediction depth and the iteration learned, for all four datasets in VGG16. Consistent results are presented for all architectures and datasets, in both the validation and training splits in Appendix~\ref{app:ll_itl}.
As a result of the reported correlation, we anticipate that many of the data points correctly classified by the k-NN probe in a particular layer should also be correctly classified by the network at a corresponding interval of training steps.
If this is correct then we would expect there to be a visual correspondence between the \emph{training learning curve} (which shows how the accuracy of the network changes during training) and the accuracy of the k-NN probes as data passes from input, through the network, towards the output layer.
We call the series of k-NN probe accuracies the \emph{inference learning curve}.

To test this hypothesis we train a model on a training split where a subset of labels are corrupted and compare the training and inference learning curves on four splits of the data: unchanged training data; mislabeled training data; the original labels of the mislabeled training data and the test split.
In Figure~\ref{fig:visual_correspondance} (middle and right plots) we see that many of the important features of the training learning curve are indeed present in the inference learning curve.
During training (middle), mislabeled data are initially processed as though they are a member of their original class (before they were mislabeled)~\citep{liu2020early}.
After an initial period of learning, the network begins to learn the new (random) labels that have been assigned to those data points, so the orange curve moves upwards, and the green curve downwards. At this point, a maximum is observed in the training accuracy~\citep{arpit2017closer}. In the right plot we see that these same phenomena occur in the inference learning curve.

\begin{figure}[t]
\begin{center}
\includegraphics[width=\linewidth]{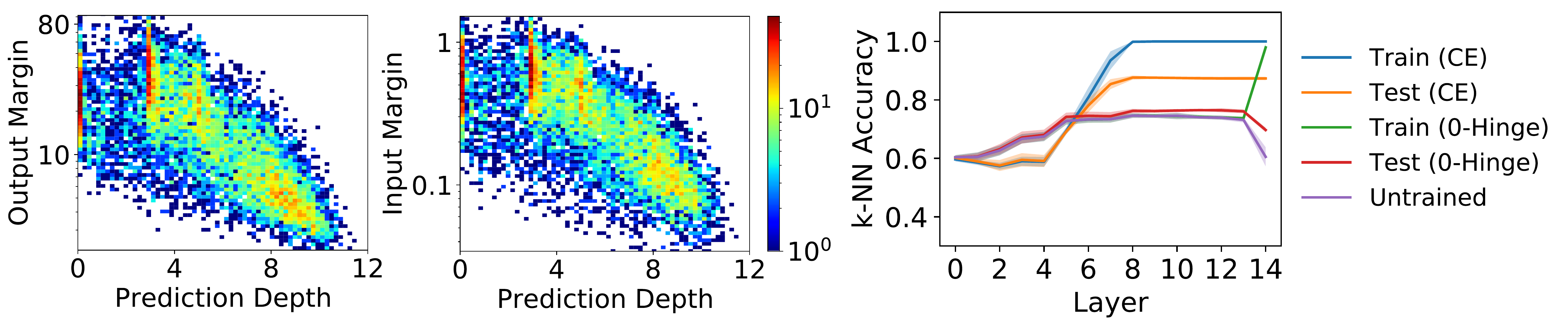}
\end{center}
\caption{\small {\bf Left and Middle:} \emph{Test examples with smaller prediction depths, on average, have larger output and input margins}. We train 25 VGG16 models with different random seeds on CIFAR10 (see Appendix~\ref{app:experiments_desc} for details) and compare the mean prediction depth of each test point in these 25 runs to its mean output and input margins (log scales).
Correlation coefficients are $-0.70$ (output margin) and $-0.69$ (input margin). Although the prediction depth could be at most 14, no data point has an average prediction depth greater than 12. 
{\bf Right:} \emph{An intervention that does not encourage large output margin (``0-Hinge'') results, as predicted, in models where the predictions are effectively determined in higher layers in the network compared to the standard training (``CE'')}.
\label{fig:margins_vs_depth}
} 
\end{figure}

\subsection{Deep models exhibit larger margins for inputs with lower prediction depth\label{sec:simplicity_vs_compdiff}}

It is reported in the literature that deep networks learn 
functions of increasing complexity during training~\citep{hu2020surprising, nakkiran2019sgd}.
We frame this observation differently: the learned function is ``locally simpler'' in the vicinity of data points with smaller prediction depths, and these points are typically learned earlier in training (Section~\ref{sec:compdiff_vs_learndiff}).

Two known measures of the simplicity of a learned function are the output margin (the difference between the largest and second-largest logits) and the adversarial input margin (the smallest norm required for an adversarial perturbation in the input to change the model's class prediction).
We estimate the adversarial input margin, $\gamma$, with a linear approximation~\citep{jiang2018predicting}:
for an input $x$ with predicted class $i$, $\gamma \simeq \min_{j \neq i}{\frac{|z_i - z_j|}{|\nabla_x\left( z_i - z_j \right)|}}$ where $z_j$ is the logit returned by the network for class $j$.
Figure~\ref{fig:margins_vs_depth} (left and middle plots) show that data points with smaller prediction depths have both larger input and output margins on average and that variances of the input and output margins decrease as the prediction depth increases.

To illustrate the strength of the relationship between the prediction depth and output margin, we demonstrate that
reducing the output margin of the learned function results in a model that clusters the data only in the latest layers:
such a solution has a very high average prediction depth.
We do not minimize the output margin directly but rather use a loss and an optimizer that do not encourage high output margin.
Naturally there are many unknowns that may contribute to this effect. We simply report the intervention and the outcome.

The intervention is performed as follows: we construct a loss function that does not promote confidence: a zero-margin hinge loss (``0-Hinge''), 
 and optimize the network using \emph{full-batch} gradient descent with momentum and \emph{very small learning rate}.
For an input $x$ with label $i$ the 0-Hinge loss is given by $l(x) = \sum_{j \neq i}\max(0, z_i - z_j)$
where $z_j$ represents the logit for class $j$.
The form of this intervention is justified in Appendix~\ref{app:fsgd_sgd}.
As a control, we additionally train a model in the standard fashion using the cross-entropy loss and SGD with momentum and large initial learning rate.
Since full-batch gradients are computationally expensive, we train on a subset of CIFAR10 (see Appendix~\ref{app:fsgd_sgd}, where we also give the hyperparameters and learning curves.).
The output margin
obtained with the intervention is 5 orders of magnitude smaller than in the control experiment: $2.0\times 10^{-4} \pm 2.0\times 10^{-4}$ for the 0-Hinge loss and $1.6\times 10^{1} \pm 0.50\times 10^{1}$ for cross-entropy loss. 
Figure~\ref{fig:margins_vs_depth} (right) compares the accuracies of the k-NN probes resulting from these training approaches.
The 0-Hinge loss training achieves only a marginal improvement in accuracy (red) over an untrained network (purple), and the training split is accurately clustered only in the latest layers. This confirms the predicted behavior: the intervention leads to a model that exhibits both very small average output margins and very late clustering of the data. Very late clustering of the data implies high prediction depths since the k-NN probe classifications change in the latest layers for many data points.

\section{Beyond a One-Dimensional Picture of Example Difficulty\label{sec:4_ex_diff}}

\begin{figure*}[t]
\centering
\includegraphics[width=\linewidth]{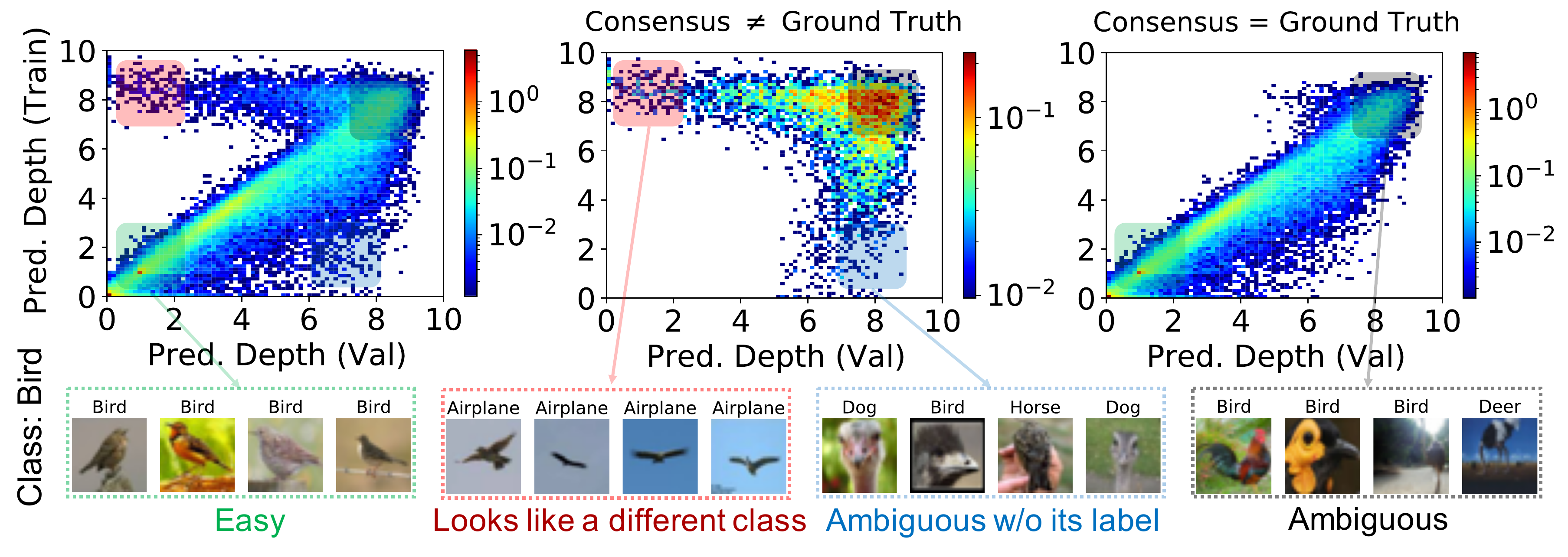}
\caption{\small \emph{ The prediction depth can be the same, or very different for the same input when it occurs in the train and validation splits. Corners of this plot correspond to different forms of example difficulty.} (See Section~\ref{sec:4_ex_diff} for discussion.) 
We train 250 ResNet18 models on CIFAR10 with random 90:10\% train:validation splits as described in Appendix~\ref{app:experiments_desc}.
These histograms compare average prediction depth for each data point when it occurs in the validation split vs the training split. This behavior is consistently reproduced for all datasets and architectures in Appendix~\ref{app:ll_tvt}. Below we show extreme (not hand-chosen) images of ``Birds'' that appear closest to the corners of this plot. The consensus class is given above each image (tiebreaks favor the class ``Bird''.).
}
\label{fig:ll_test_v_train}
\end{figure*}
In this section we transcend the one-dimensional picture of example difficulty by identifying different underlying reasons behind the difficulty of an example, in a way that is general to different architectures and datasets.

Figure~\ref{fig:ll_test_v_train} shows that the prediction depth can be different when an input occurs in the training split vs. the validation split.
Thus, there are two axes of example difficulty:
\begin{enumerate}
    \item Difficulty of making a prediction when an input is in the validation set
    \item Difficulty of finding commonalities during training with other examples of the same ground truth class 
\end{enumerate}
{\bf Both axes have a range from ``clear'' to ``ambiguous''.}
In Section~\ref{sec:ll_decision_consistency} we show that predictions made for validation points with later prediction depths are often inconsistent, with low consensus-consistency.
Conversely, a low prediction depth typically indicates an input with high consensus-consistency.
For Axis 1 we will identify validation points with low prediction depths as ``clear'' and those with high prediction depths as ``ambiguous''.
We will additionally identify a low or high prediction depth in the training split with examples that are respectively ``clear'' and ``ambiguous'' on Axis 2.
By making combinations of low/high values of $(\mathrm{PD}_\mathrm{Val.}, \mathrm{PD}_\mathrm{Train})$ we obtain four extremes of example difficulty:
\begin{description}
\item[Easy examples:] (Low $\mathrm{PD}_\mathrm{Val.}$, Low $\mathrm{PD}_\mathrm{Train}$). 
Such examples are often visually typical members of their class and the predicted label nearly always matches the ground truth.
\item[Looks like a different class:]
(Low $\mathrm{PD}_\mathrm{Val.}$, High $\mathrm{PD}_\mathrm{Train}$). In the validation set, there is a clear (and nearly always incorrect) classification for such an input, but it is difficult to connect such inputs to other examples of their ground truth class during training. Mislabeled examples are of this kind, as are visually confusing images which at first appear to show something else.
\item[Ambiguous unless the label is given:] 
(High $\mathrm{PD}_\mathrm{Val.}$, Low $\mathrm{PD}_\mathrm{Train}$). These examples are difficult to connect to their predicted class in the validation split but easy to connect to their ground truth class during training.
These points may, for example, visually resemble both their own class and another class. They are likely to be misclassified.
\item[Ambiguous:] 
(High $\mathrm{PD}_\mathrm{Val.}$, High $\mathrm{PD}_\mathrm{Train}$). These examples may be corrupted or show an example of a rare sub-class. Predictions for these inputs can depend strongly on the random seed used for training and initialization.
\end{description}

In Figure~\ref{fig:ll_test_v_train} we visualize CIFAR10 ``Bird'' images with the extreme forms of example difficulty for ResNet18, as identified using the prediction depth in the training and validation splits. In the full dataset (left panel) we see that the prediction depth can be very different in the training and validation splits: the two prediction depths are typically similar for points where the consensus class is equal to the ground truth (right panel), but can be very different when the consensus class is different from the ground truth (middle panel). This behavior is consistently reproduced for all datasets and architectures in Appendix~\ref{app:ll_tvt}. 

Looking at these examples of the class ``Bird'' with different difficulty types, we observe that ResNet18 finds small garden birds easiest, while birds in flight against a blue background ``look like airplanes'', ostriches are ``ambiguous without their label'' and the ``ambiguous'' examples are either unclear photographs or examples of rare sub-groups that don't appear frequently in the data. We found the consensus-consistency of inputs that are ``Ambiguous'' or ``Ambiguous without its label'' to be significantly lower than those of examples that are ``Easy'' or ``Look like a different class''.

\begin{figure}[t]
          \resizebox{1.\textwidth}{!}{%
\includegraphics[trim=0 0 0 0, clip,height=5cm]{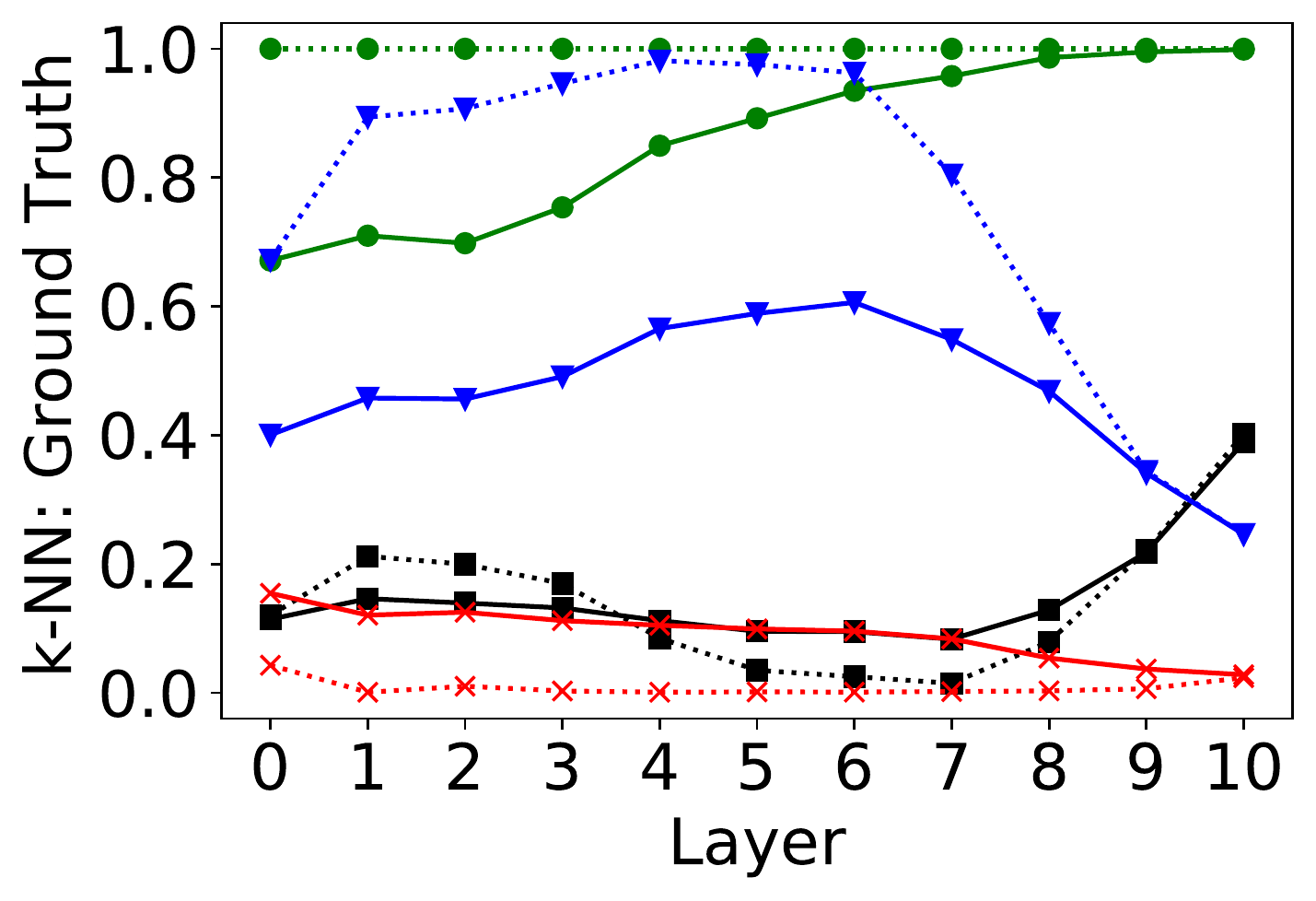}%
\quad
\includegraphics[trim=0 0 0 0, clip,height=5cm]{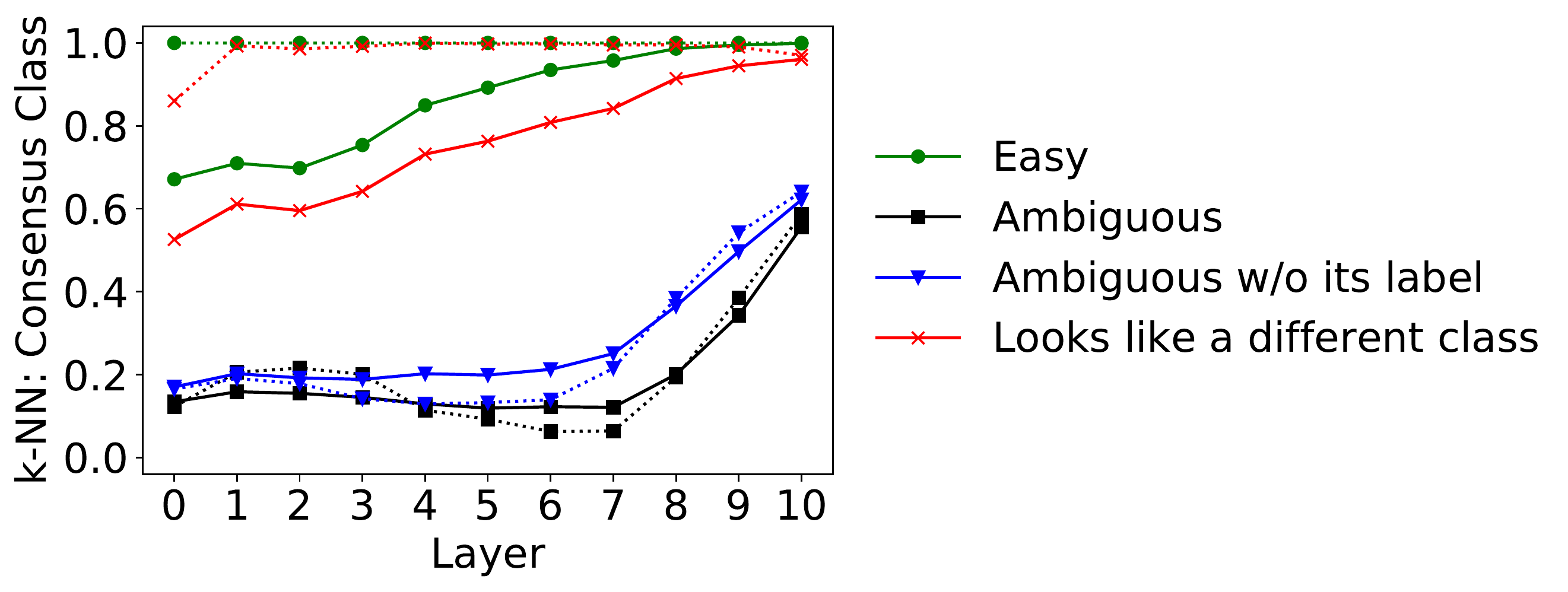}%
}
\caption{\small \emph{ Average k-NN probe confidence (solid lines) and accuracy (dotted lines) for the ground truth class (left) and consensus class (right), in the validation split for examples exhibiting extreme forms of difficulty.}
Mean values for 100 examples with each form of difficulty, identified as the 100 examples closest to the corners in Figure~\ref{fig:ll_test_v_train} (left). This result is for CIFAR10 with ResNet18: similar plots for all datasets and architectures are shown in Appendix~\ref{app:cluster_4forms}. See Section~\ref{sec:4_ex_diff} for the discussion of the result and how it can be used to improve prediction accuracy.\label{fig:knn_4forms}}
\end{figure}

In order to better understand how networks process examples with different, extreme forms of example difficulty,
Fig.~\ref{fig:knn_4forms} examines how the k-NN confidence (fraction of votes) and accuracy of the ground truth class and of the consensus class progress, as validation points pass through the network. ``Easy'' examples are classified as their consensus class (which is equal to their ground truth class) in all k-NN probes and the confidence in the consensus class steadily increases as data points proceed through the hidden layers. Examples that ``look like a different class'' are also processed as members of their consensus class, similarly to ``easy'' examples. However, unlike ``easy'' examples, their consensus classes do not match their ground truth classes.
Examples that are ``ambiguous without their labels'' are initially processed as members of their ground truth classes with intermediate confidence, but in later layers become mistaken for their consensus class. ``Ambiguous'' examples are processed with low confidence and accuracy in the early layers, for both ground truth and consensus classes.
In later layers ``ambiguous'' examples are recognized, with intermediate confidence and accuracy, as members of the consensus class, which matches the ground truth class for a sizeable fraction of ``ambiguous'' examples.

\paragraph{Improving the prediction accuracy} Can the prediction accuracy be improved using our understanding of how each class of difficult examples are processed by deep models? Figure~\ref{fig:knn_4forms} suggest that k-NN probes in intermediate layers may be more accurate than the full deep model for examples that are ``ambiguous without their label'' (data points closest to the lower right corner of Figure~\ref{fig:ll_test_v_train}). In order to test this hypothesis, we compare the accuracy of the k-NN probe in layer 4 to the full model's prediction for the 100 examples closest to the lower right corner of Figure~\ref{fig:knn_4forms}. We obtain a striking improvement in accuracy from 25\% to 98\% for these examples. This showcases how insights from this study can be directly used to improve prediction accuracy.

\section{Discussion \label{sec:discussion}}
\paragraph{Summary} We have introduced a notion of example difficulty called the prediction depth, which uses the processing of data inside the network to score the difficulty of an example.
We have shown how the prediction depth is related to the accuracy and uncertainty of a prediction, the adversarial input margin and the output margin of the learned solution, and that data points that are easier according to the prediction depth are also typically learned earlier in training.  We have also shown that the difficulty of an example can be both similar, or very different depending on whether an input appears in the validation split or the training split, and described four extremes of example difficulty. For data points that are ``ambiguous without their label'', we have demonstrated how returning the k-NN prediction in a middle layer can lead to impressive increases in model accuracy: for CIFAR10 in ResNet18 we obtained an increase in accuracy from 25\% to 98\% for the inputs that are most ``ambiguous without their label''.

\paragraph{Connecting known phenomena \label{sec:connect_phenomena}}
In the literature, the following phenomena are separately reported from different experimental paradigms:
\begin{enumerate}
    \item Early layers generalize while later layers memorize~\citep{stephenson2021on}.
    \item Model layers converge from input layer towards output layer~\citep{raghu2017svcca, morcos2018insights}.
    \item Deep models learn easy data~\citep{chiyuan_cscores, forgetting19} and simple functions first~\citep{hu2020surprising, nakkiran2019sgd}.
\end{enumerate}

Following this paper, a coherent and closely related picture emerges: 
\begin{enumerate}
\item Predictions made in early layers are more likely to be consistent than those made in later layers. Consistent predictions are likely to be correct and the expected accuracy of inconsistent predictions is naturally low (Section~\ref{sec:ll_decision_consistency}).
\item Data points learned early in training typically have smaller prediction depths than those learned later during training (Section~\ref{sec:compdiff_vs_learndiff}).
\item On average, deep neural networks exhibit wider input and output margins (common measures of “local simplicity”) in the vicinity of data with smaller prediction depths (Section~\ref{sec:simplicity_vs_compdiff}).
\end{enumerate}

\paragraph{Pertinence of example difficulty to topics in machine learning}
Curriculum Learning attempts to treat hard examples differently from easy examples during training. Robustness to distribution shifts that change the relative frequencies of common and rare subgroups in the test set (which we have shown can have different forms of example difficulty) is important for ML Fairness.
Methods developed to address heteroscedastic uncertainty typically address example difficulty as a one-dimensional quantity.
We expand upon the relevance of our work to these three topics in Appendix~\ref{app:exdiff_pertinence}.

\paragraph{Limitations} We believe that the results we report stem from a deep model's representation, which is hierarchical by construction. We expect that the same results will therefore apply in larger models, larger datasets, and tasks other than image classification, but testing this remains as further work.
Although we demonstrate that returning the results of a hidden k-NN can yield dramatic increases in accuracy for examples that are ``ambiguous without their label'', we otherwise do not explore ways to practically apply the insights we present.
In particular, we expressly do not claim that all that is required for good accuracy is to reduce the prediction depth:
freezing later layers of the network
would not be expected to result in good generalization.

\section*{Acknowledgment}
We would like to thank Hanie Sedghi, Ilya Tolstikhin, Ibrahim Alabdulmohsin, Daniel Keysers and Julian Eisenschlos for valuable discussions on the topic and Arthur Baldock for proofreading the manuscript.

\bibliographystyle{apalike}
\bibliography{arxiv}

\clearpage

\appendix
\renewcommand\thefigure{\thesection.\arabic{figure}}

\section{Detailed Description of Experiments, Architectures and Hyperparameter Optimization\label{app:experiments_desc}}

For each combination of dataset (CIFAR10, CIFAR100, Fashion MNIST, SVHN) and architecture (ResNet18, VGG16, MLP) we train 250 models with a 10\% validation split selected at random each time, and an additional 25 models on the full training set.

\subsection{Datasets}
\label{app:data_sets}
CIFAR10 / CIFAR100:\\ 
Reference: \citep{krizhevsky2009learning}. License: MIT.\\
URL: \url{https://www.cs.toronto.edu/~kriz/cifar.html}\\
\\
Fashion MNIST:\\
Reference: \citep{xiao2017fashion}. License: MIT.\\
URL: \url{https://github.com/zalandoresearch/fashion-mnist}\\
\\
Street View House Numbers:\\
Reference: \citep{netzer2011reading}. License: CC0.\\
URLs: \url{http://ufldl.stanford.edu/housenumbers/} \\
\url{https://www.kaggle.com/stanfordu/street-view-house-numbers}

\subsection{Architectures}
\subsubsection{ResNet18}
We implemented the standard ResNet18 architecture for CIFAR10~\citep{he2016deep}, except that we replaced Batch Norm with Group Norm and applied Weight Standardization, following recent state of the art~\citep{bit_gzh}.

\subsubsection{VGG16}

We used VGG16~\citep{simonyan2014very}, except that we removed the final three dense layers: a standard modification for datasets smaller than ImageNet.
We also did not use batch-norm or dropout: our focus is on understanding trends in example difficulty and we do not expect the results to be dependent on these devices.

\subsubsection{MLP \label{app:mlp_arch}}

\begin{figure}[ht!]
\begin{center}
\begin{subfigure}
         \centering
         \includegraphics[width=0.49\columnwidth]{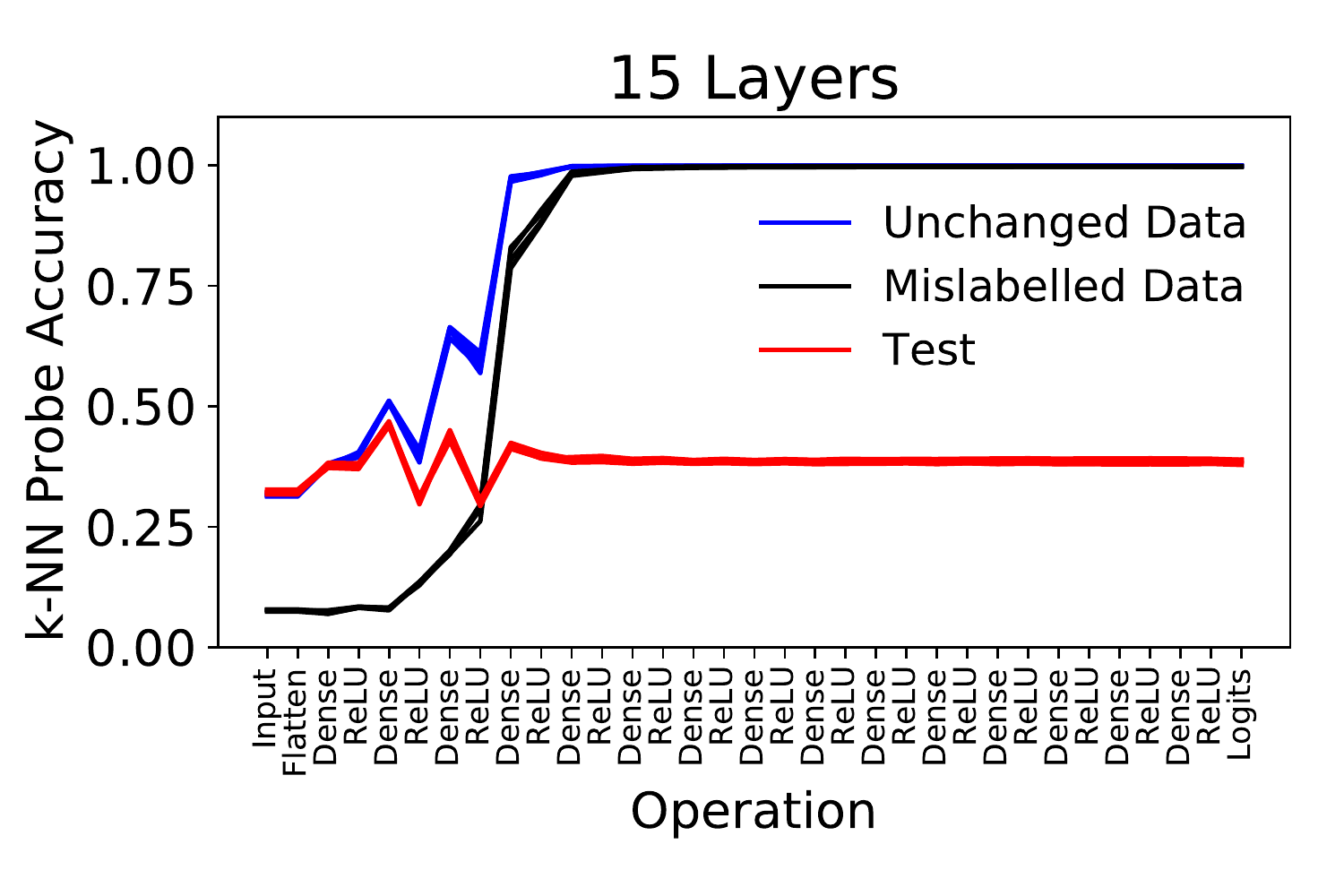}
\end{subfigure}
\begin{subfigure}
         \centering
         \includegraphics[width=0.49\columnwidth]{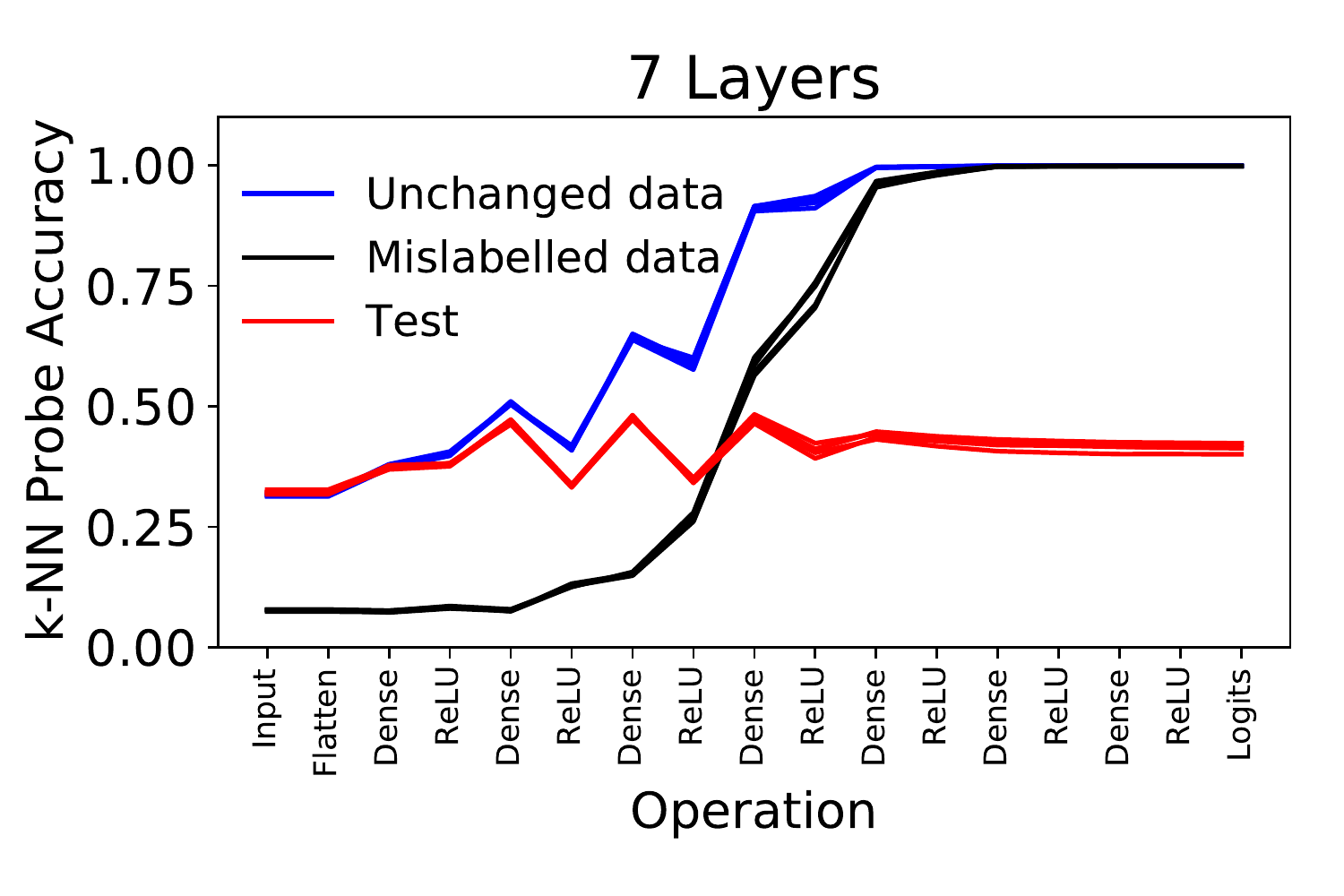}
\end{subfigure}
         \caption{\emph{Seven layers are sufficient in MLP for CIFAR10 with 40\% random label noise.}
         CIFAR10 with 40\% random label noise. For this plot, k-NN probes were placed after every operation in two MLP architectures of the same width (2048) but different depths. {\bf Left:} 15 dense layers; {\bf Right:} 7 dense layers.  Separate accuracies are reported for the test split, those data points in the training split with unchanged labels and the randomly mislabeled data in the training split.
\label{fig:mlp_arch_choice}}
\end{center}
\end{figure}

Our MLP architecture comprises seven hidden layers with ReLU activations.
We chose seven layers after performing the experiments shown in Figure~\ref{fig:mlp_arch_choice}. 
There we show the accuracies of k-NN probes placed after each operation of two MLP architectures, depths 15 layers and 7 layers, both of width 2048.
We used CIFAR10 with 40\% fixed random label noise as a reasonably difficult model classification task, to choose the depth.

\subsubsection{Data augmentation}

We did not apply data augmentation: different data augmentation schemes could be expected to have disparate effects on different examples, but we do not expect them to change the overall phenomena that we report here.
We leave the use of data augmentation to subsequent studies.

\subsection{Hyperparameter optimization \label{app:hyperparameters}}

For each architecture and dataset we initially performed $10^4$ steps of SGD with momentum, using all combinations of the following hyperparameters: learning rate $\in [4\times10^1, 1\times10^{-1}, 4\times10^{-2}, 1\times10^{-2}, 4\times10^{-3}, 1\times10^{-3}, 4\times10^{-4}, 1\times10^{-4}]$ ; momentum $\in [0.0, 0.5, 0.9, 0.95]$; weight decay $\in [0, 5\times10^{-4}]$.
In CIFAR10, we additionally considered a learning rate of $2\times10^{-2}$.
For each dataset and architecture we selected the 7 most accurate and stable training curves, extended the number of training steps and added a learning rate schedule, reducing the learning rate in steps of $\frac{1}{5}$. At least two rounds of optimization were performed to adapt the learning rate schedule for each combination of architecture and dataset.
In each case a mini-batch size of 256 was used.
The final parameters obtained are shown in Table~\ref{tab:hyperparams}, which also gives the hyperparameters used in Sec.~\ref{sec:simplicity_vs_compdiff} and Appendix~\ref{app:mlp_arch} for CIFAR10 with 40\% label noise.
Final accuracies of the trained models are given in Table~\ref{tab:model_accs}.

\begin{table*}[ht]
\begin{center}
    \begin{tabular}{l||c|c|c|c}
              & Learning Rate & Momentum & Weight Decay & Schedule / steps \\
\hline\hline
 & \multicolumn{4}{c}{SVHN} \\ \hline
ResNet18 & $4\times10^{-2}$ & 0.95 & 0.0 & [7000]    \\
VGG16       & $4\times10^{-2}$ & 0.9  & 0.0 & [3000, 6000, 1000]  \\
MLP         & $4\times10^{-2}$ & 0.9  & 0.0 & [2500, 5500, 2000] \\ \hline
 & \multicolumn{4}{c}{Fashion MNIST} \\ \hline
ResNet18 & $1\times10^{-2}$ & 0.95 & 0.0 & [4000, 3000]    \\
VGG16       & $1\times10^{-2}$ & 0.95  & 0.0 & [3000, 6000, 1000]  \\
MLP         & $4\times10^{-2}$ & 0.5  & 0.0 & [10000, 2500] \\ \hline
 & \multicolumn{4}{c}{CIFAR10} \\ \hline
ResNet18 & $4\times10^{-2}$ & 0.95 & 0.0 & [7000]    \\
VGG16       & $4\times10^{-2}$ & 0.9  & 0.0 & [5000, 1000]  \\
MLP         & $2\times10^{-2}$ & 0.9  & 0.0 & [5000, 1250, 1000] \\ \hline
 & \multicolumn{4}{c}{CIFAR10 w/ 40\% (Fixed) Randomized Labels} \\ \hline
VGG16       & $4\times10^{-2}$ & 0.9  & 0.0 & [5000, 10000]  \\ 
MLP       & $2\times10^{-2}$ & 0.9  & 0.0 & [12000, 1250, 4000  \\ \hline
 & \multicolumn{4}{c}{CIFAR100} \\ \hline
ResNet18 & $1\times10^{-1}$ & 0.95 & 0.0 & [6000]    \\
VGG16       & $4\times10^{-2}$ & 0.9  & 0.0 & [2500, 7500]  \\
MLP         & $1\times10^{-1}$ & 0.95  & 0.0 & [2500, 6000, 1500] \\ \hline
\end{tabular}
\caption{Training parameters for each model and dataset.}
\label{tab:hyperparams}
\end{center}
\end{table*}

\begin{table*}[ht]
\begin{center}
    \begin{tabular}{l||c|c|c|c}
              &SVHN & Fashion MNIST & CIFAR10 & CIFAR100   \\
\hline\hline
ResNet18 & 95\% & 93\% & 83\% & 56\%    \\
VGG16       & 95\% & 93\%  & 83\% & 45\%  \\
MLP         & 85\% & 90\%  & 59\% & 29\%  
\end{tabular}
\caption{Final accuracies of the trained models.}
\label{tab:model_accs}
\end{center}
\end{table*}

\subsection{Convergence and consistency of k-NN probe accuracies \label{app:knn_conv}}

\begin{figure}[ht]
\begin{center}
         \includegraphics[width=0.55
         \columnwidth]{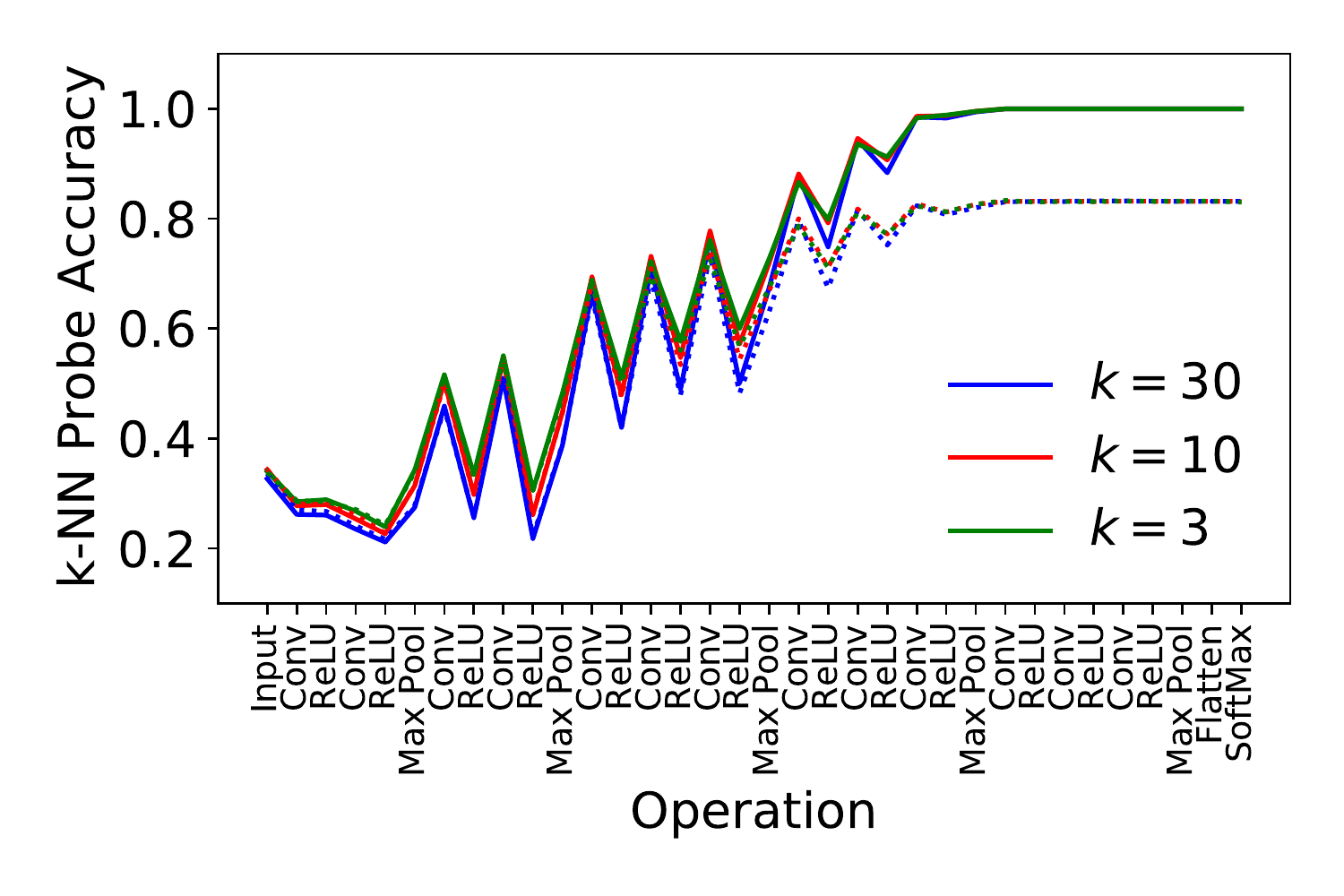}
         \caption{
         CIFAR10, VGG16. k-NN probe accuracies after each operation for $k \in [3, 10, 30]$. Solid lines: training set. Dotted lines: test set. Differences in these results are comparable to the scatter observed for networks trained with  different random seeds at $k=30$.
\label{fig:k_convergence}}
\end{center}
\end{figure}

We tested the convergence of k in k-NN for VGG16 on CIFAR10.
Figure~\ref{fig:k_convergence} shows the accuracies of k-NN probes after every operation of the network for $k \in [3, 10, 30]$. We see that these k-NN probe accuracies are insensitive to $k$ for $k=30$. 

Figure~\ref{fig:mlp_arch_choice} shows separate results for five independent training runs.
Similarly, Figure~\ref{fig:visual_correspondance} (right) and Figure~\ref{fig:margins_vs_depth} (right) each show the mean and uncertainty on the k-NN probe accuracies from 5 independent runs.
The spread of results in these figures is tight, demonstrating consistency of the results.

\subsection{Placement of k-NN probes \label{app:knn_probe_placement}}

For prediction depth, in MLP we constructed k-NN probes after the dense operations and the softmax, in VGG16 after the convolutions and softmax, and in ResNet18 we constructed the probes after the initial Group Norm operation, the sum operations at the end of each block and after the softmax operation.

From figures~\ref{fig:mlp_arch_choice} and~\ref{fig:k_convergence} it is clear that there are upper and lower envelopes that bound the k-NN probe accuracies: the lower envelope corresponds to the ReLU activations and the upper envelope to the operations immediately preceding them.
We chose the preceding operations which, in effect, conceptually shifts the ReLU activations to the ``start'' of a layer rather than the ``end'' of the preceding layer.

\subsection{Notes on definitions\label{app:defs_notes}}

\subsubsection{Consistency of the model's prediction with the k-NN probe after the softmax layer\label{app:knn_matches_model}}

Deep classifier models are trained to create linear separation of the classes in the softmax layer.
There is nearly perfect agreement between the k-NN probe after the softmax layer and predictions of the full model.
In the rare case that the k-NN probe after the softmax predicts a different class from the full network we do not assign a prediction depth. Such data points are extremely rare: we found zero such data points in the large majority of models and always fewer than 1 in $10^4$.

\subsubsection{Tiebreaks in the consensus class\label{app:mode_note}}

When obtaining the consensus class, if predictions are tied between more than one class and the ground truth is in the tiebreak, then we break the tie in favor of the ground truth class. If the ground truth is not in the tiebreak then we report the tied class with the lowest integer index.
This choice was motivated by ease of implementation. 
We are confident that the overall results we report are unaffected by this choice.

\subsubsection{Estimating the consensus-consistency\label{app:est_CC}}

We used the same ensemble to obtain both the consensus class $\hat{y}_A\left(x\right)$ and the consensus-consistency score. Thus we are reporting relationships between observables for a given ensemble.
This is a biased estimator of~\eqref{eq:empirical_modal_cscore}: an unbiased estimator could have been constructed by training an additional set of models to obtain the consensus class, but at greater cost. We are confident that this does not affect the conclusions of this study.

\subsection{Justification and hyperparameters for the output margin intervention \label{app:fsgd_sgd}}

A number of published works informed the design of our intervention. Firstly,~\citet{soudry2018implicit} demonstrate that the cross-entropy (CE) loss leads to large margins. In contrast to the cross-entropy, the 0-Hinge loss has zero gradient if the prediction is correct, so it does not push the model to become arbitrarily confident.
Secondly,~\citet{keskar2017large} show that smaller batch sizes lead to the discovery of flatter minima, which also corresponds to a wider margin~\citep{neyshabur2017exploring}.
Thirdly,~\citet{keskar2017large},~\citet{smith2018bayesian} and~\citet{smith2018don} show that the gradient noise level in stochastic gradient descent is proportional to $\frac{\mathrm{Learning\: Rate}}{\mathrm{Batch\:Size}}$.
Having an appreciable noise level early in training plays an important role in finding the flatter minima with larger output margins reported in~\citet{keskar2017large}.
Our intervention to minimize the margin therefore combines both of the following changes:
\begin{enumerate}
    \item Changing the loss from cross-entropy to the 0-Hinge loss
    \item Minimizing the learning rate and making the batch size as large as possible 
\end{enumerate}
To test whether both or only one of these changes is required to obtain small output margins, we performed separate runs, without any intervention, applying the changes individually and applying them together.
The starting point (the control) is training with cross-entropy loss and SGD with momentum and large initial learning rate.

\begin{figure}[t]
\begin{center}
         \begin{subfigure}
         \centering
         \includegraphics[width=0.25\columnwidth]{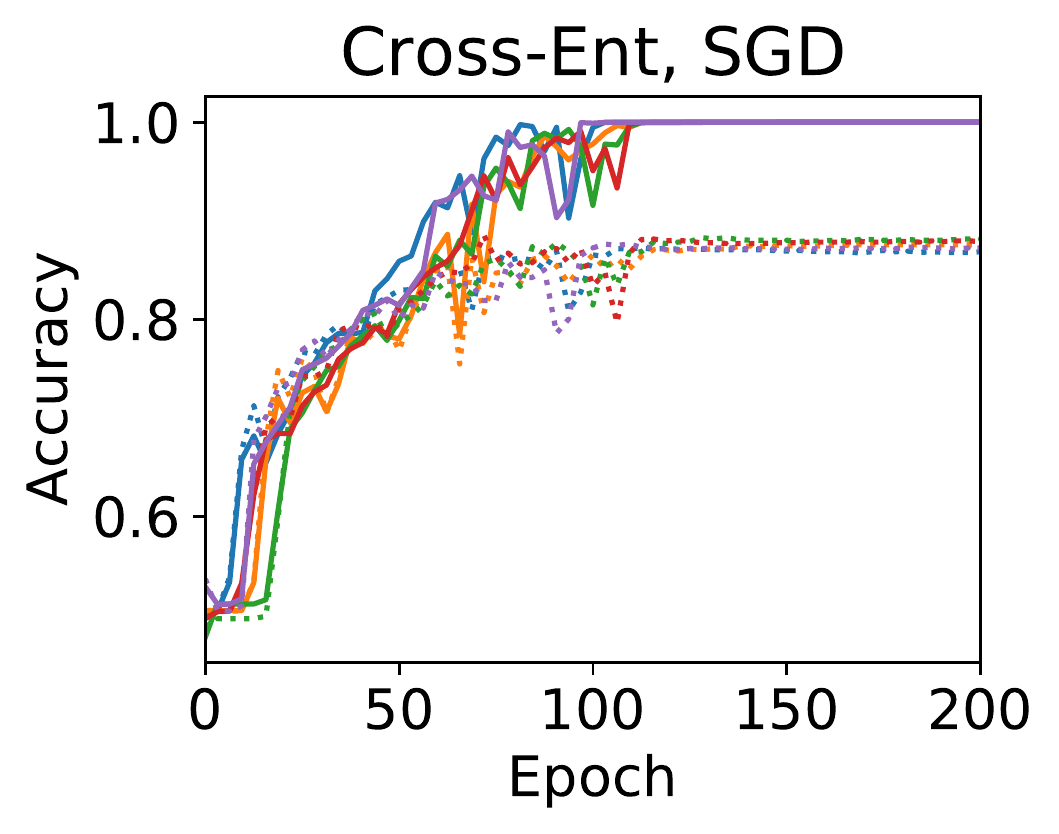}
\end{subfigure}
\hspace{-0.5cm}
\begin{subfigure}
         \centering
         \includegraphics[width=0.25\columnwidth]{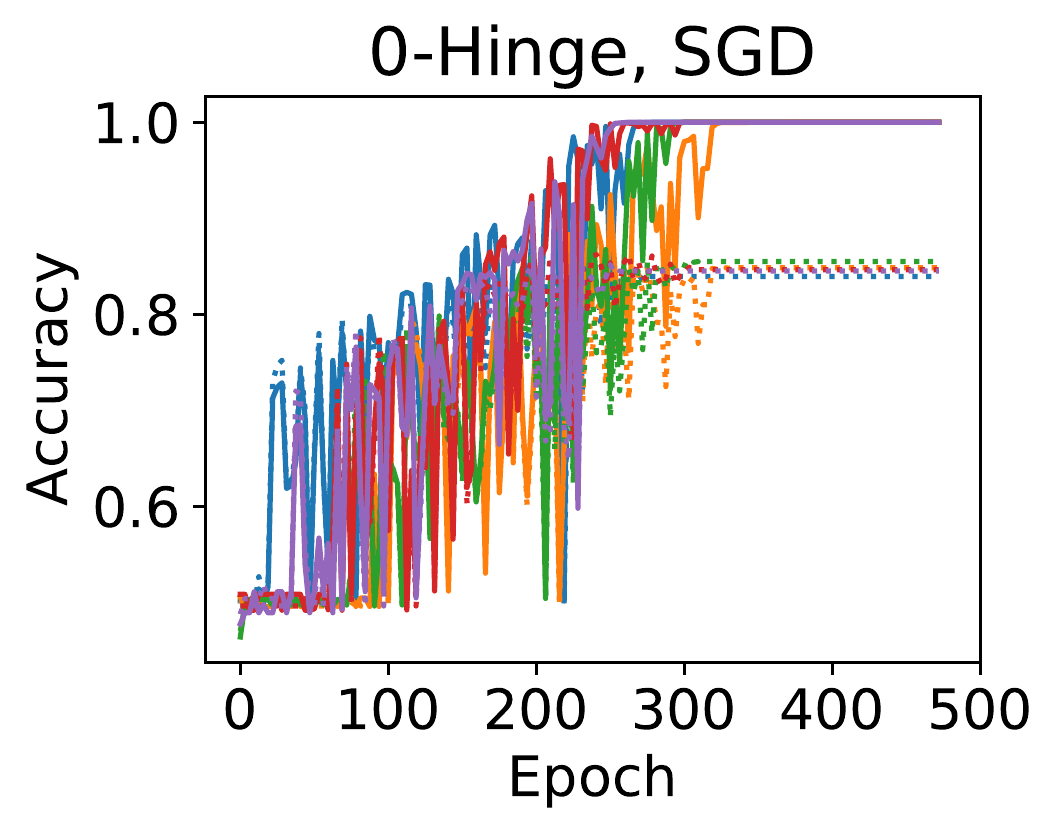}
\end{subfigure}
\hspace{-0.5cm}
\begin{subfigure}
         \centering
         \includegraphics[width=0.25\columnwidth]{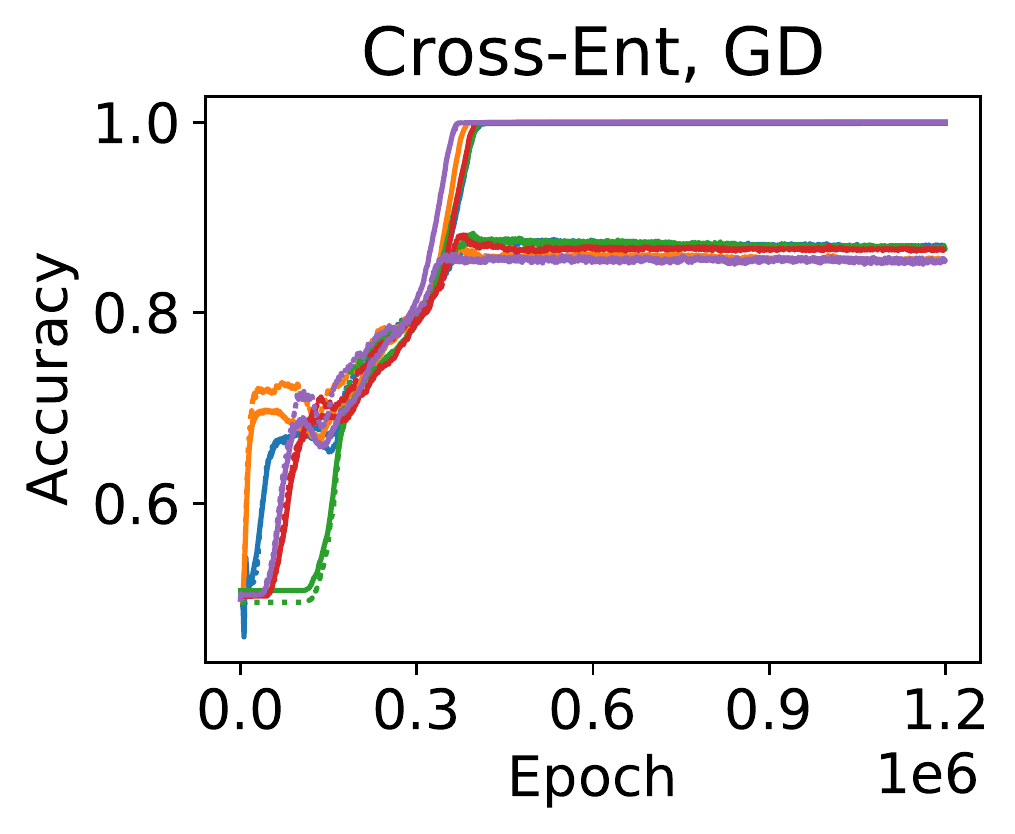}
\end{subfigure}
\hspace{-0.3cm}
\begin{subfigure}
         \centering
         \includegraphics[width=0.25\columnwidth]{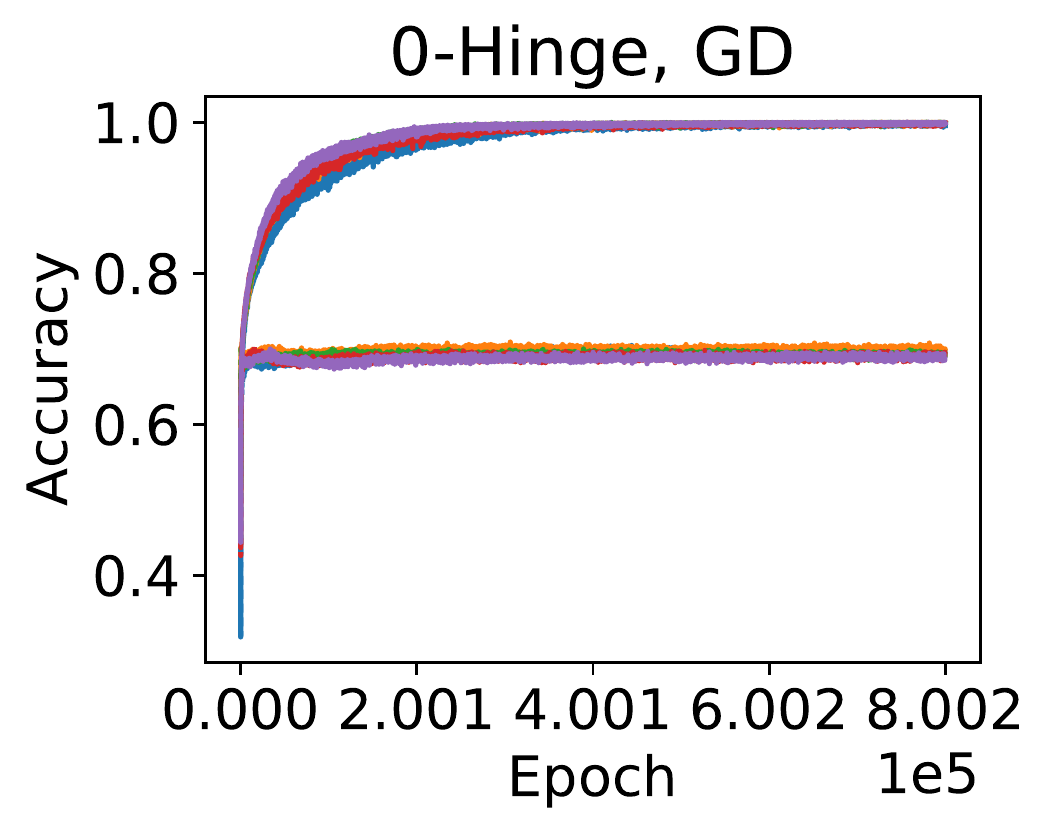}
\end{subfigure}
     \end{center}
\vspace{-.5cm}
\caption{\small \emph{Training curves for Cross-Entropy and 0-Hinge Losses, with either SGD with momentum and large initial learning rate, or GD with momentum and a small learning rate.} The initial learning rates and schedules are set to obtain nearly smooth learning curves for GD and noisy learning curves for SGD. Each plot shows five separate learning curves. Solid lines show training accuracies and dotted lines show test accuracies. 
}
\label{fig:learning_curves_expC}
\end{figure}

We trained VGG16 on CIFAR10.
The hyperparameters, presented in Table~\ref{tab:expC_hp}, were set for each loss, to obtain nearly smooth learning curves for full-batch gradient descent and very noisy learning curves for SGD.
In Figure~\ref{fig:learning_curves_expC} we show the learning curves for these models.
Since full-batch gradients are expensive to compute we restricted the experiments to separating two classes (``Horse'' and ``Deer'') with 4096 training images in total (evenly split).

\begin{table*}[ht]
\begin{center}
    \begin{tabular}{l||c|c|c|c}
              Name  & Batch Size & Initial Learning Rate & Schedule / Steps & Momentum     \\
\hline\hline
CE, SGD & 256 & $4\times 10^{-3}$ & $[3200]$ & 0.9    \\
CE, GD & 4096 & $ 6.4\times 10^{-6}$ & $[1.2\times 10^6]$ & 0.9    \\
0-Hinge, SGD & 256 & $4\times 10^{-2}$ & $[5000,2500]$ & 0.9    \\
0-Hinge, GD & 4096 & $ 6.4\times 10^{-5}$ & $[8\times 10^5]$ & 0.95 
\end{tabular}
\caption{Hyperparameters for all combinations of CE vs. 0-Hinge loss and SGD with momentum and large initial learning rate vs. GD with momentum and small learning rate. In the learning rate schedules we reduced the learning rate by a factor of $\frac{1}{5}$ for each new set of training steps. Weight decay was not employed in these calculations since we do not expect typical, modest amounts of weight decay to qualitatively affect the results.}
\label{tab:expC_hp}
\end{center}
\end{table*}

\begin{figure}[t]
\begin{center}
         \begin{subfigure}
         \centering
         \includegraphics[width=0.49\columnwidth]{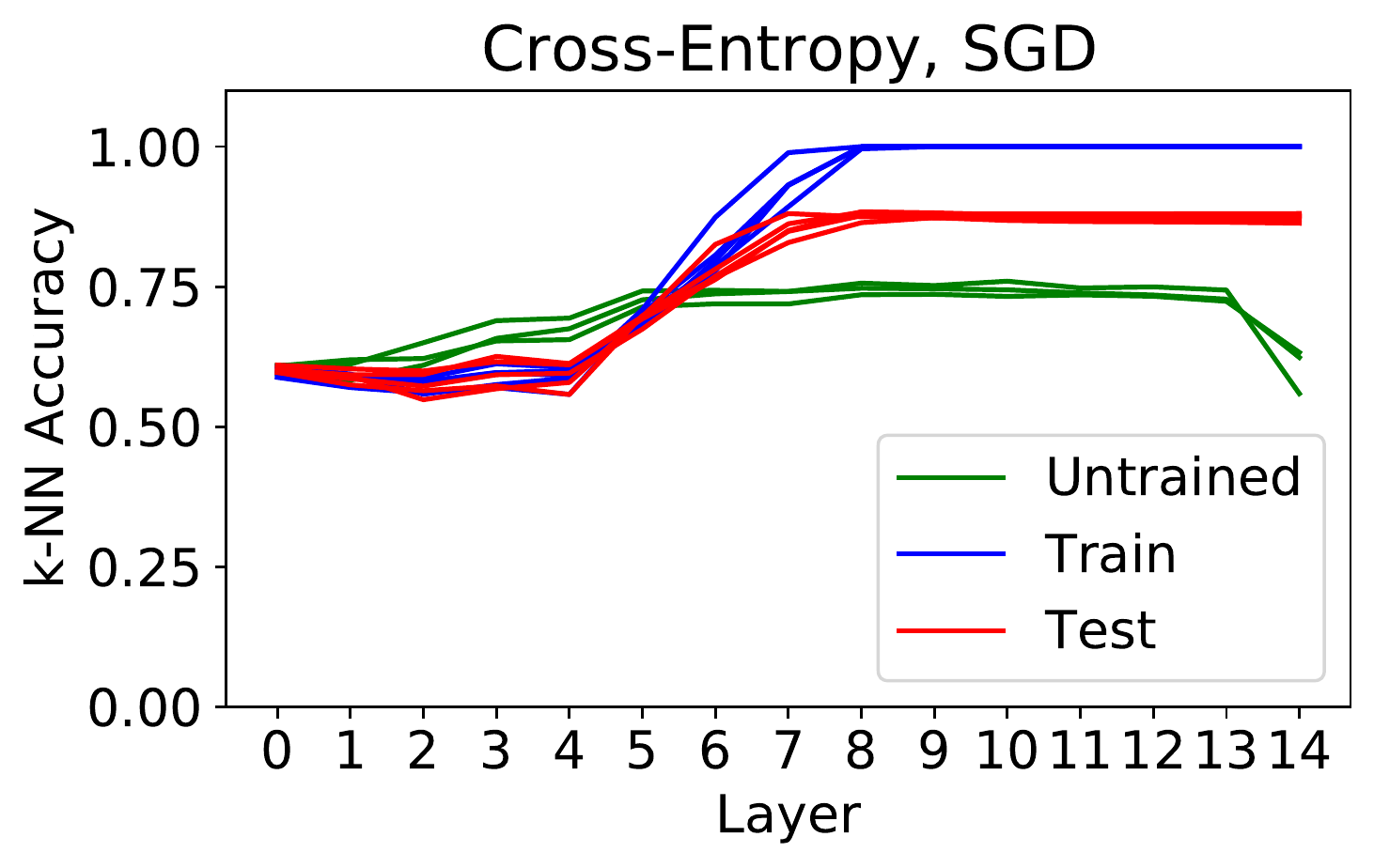}
\end{subfigure}
\begin{subfigure}
         \centering
         \includegraphics[width=0.49\columnwidth]{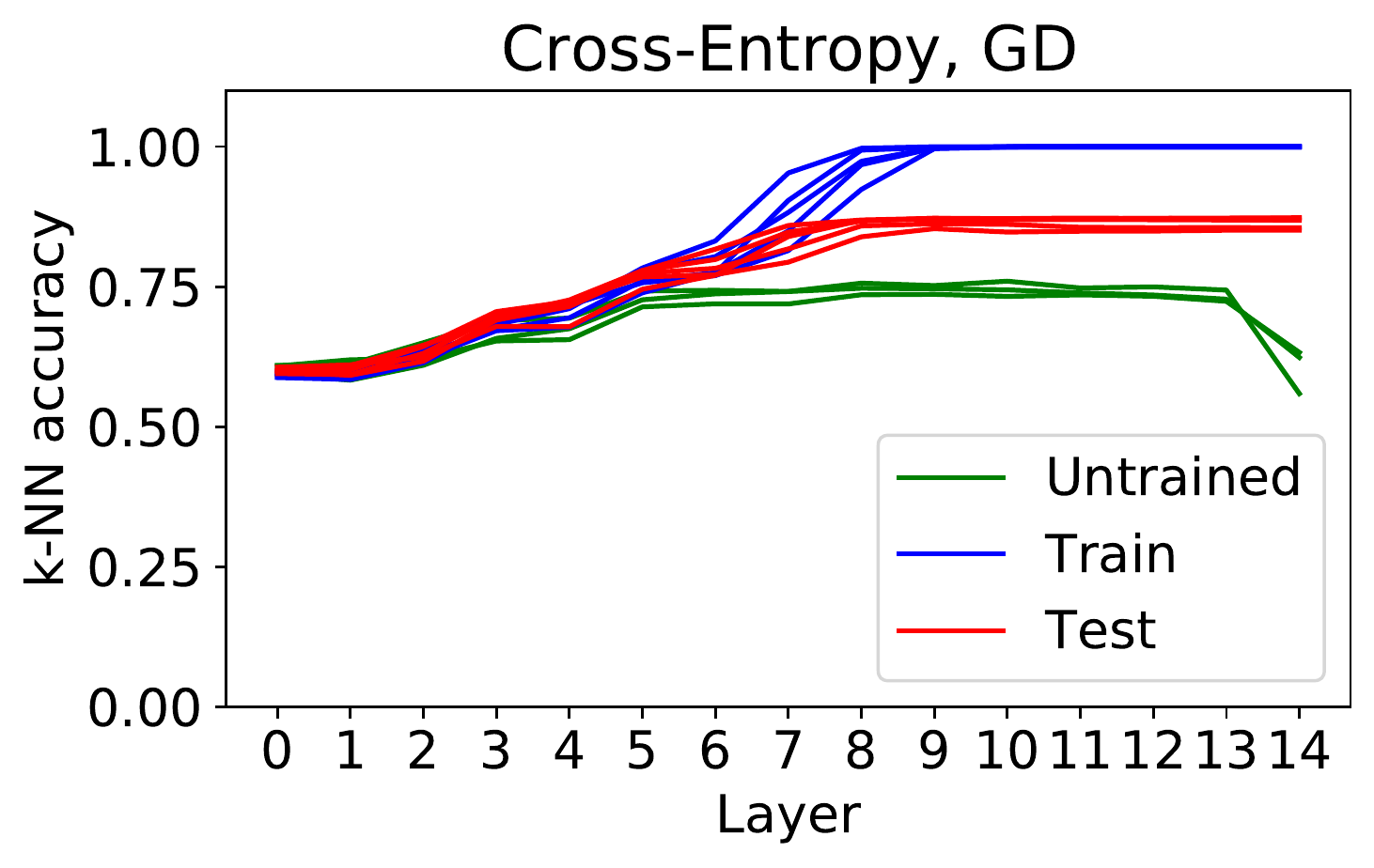}
\end{subfigure}
\begin{subfigure}
         \centering
         \includegraphics[width=0.49\columnwidth]{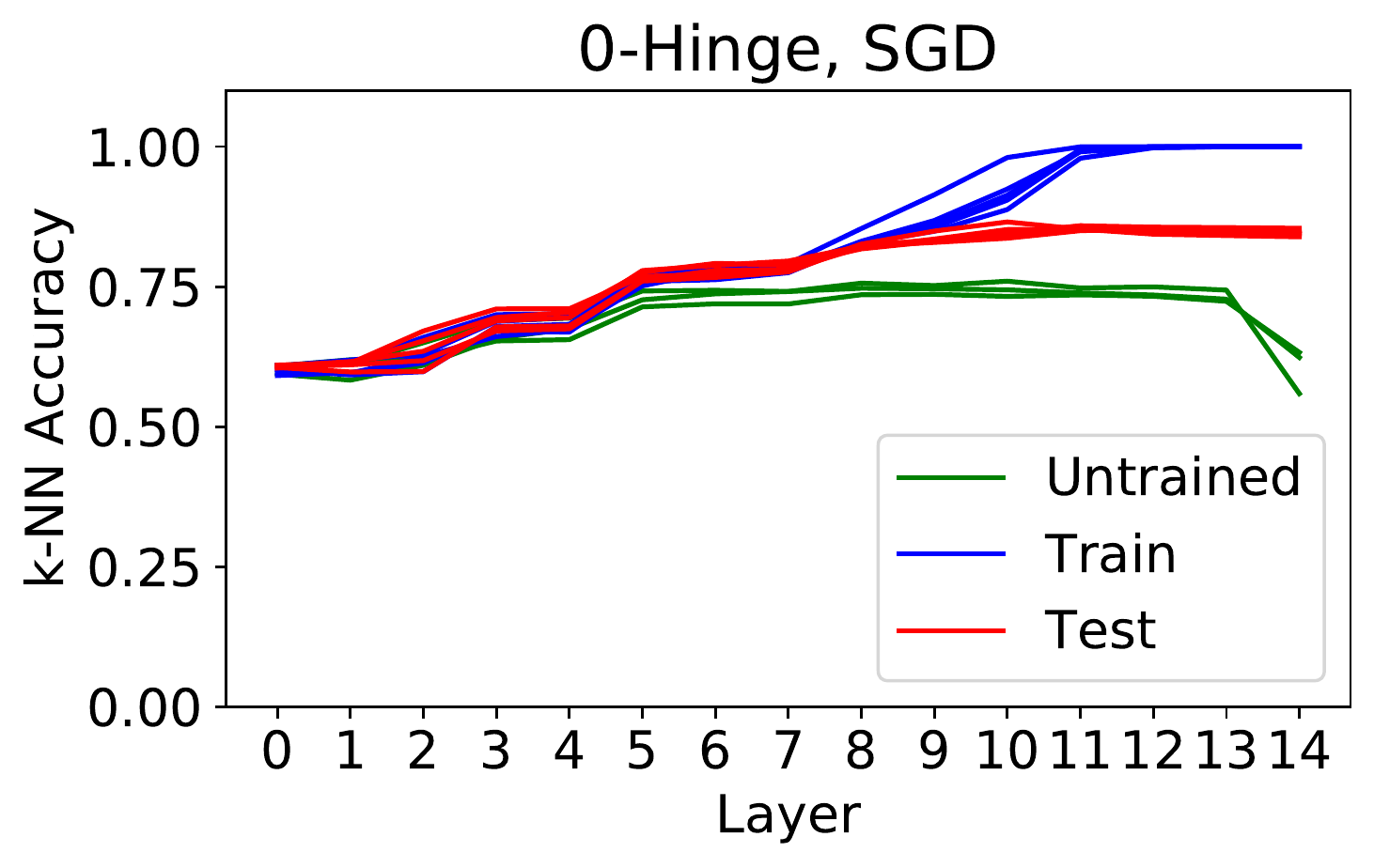}
\end{subfigure}
\begin{subfigure}
         \centering
         \includegraphics[width=0.49\columnwidth]{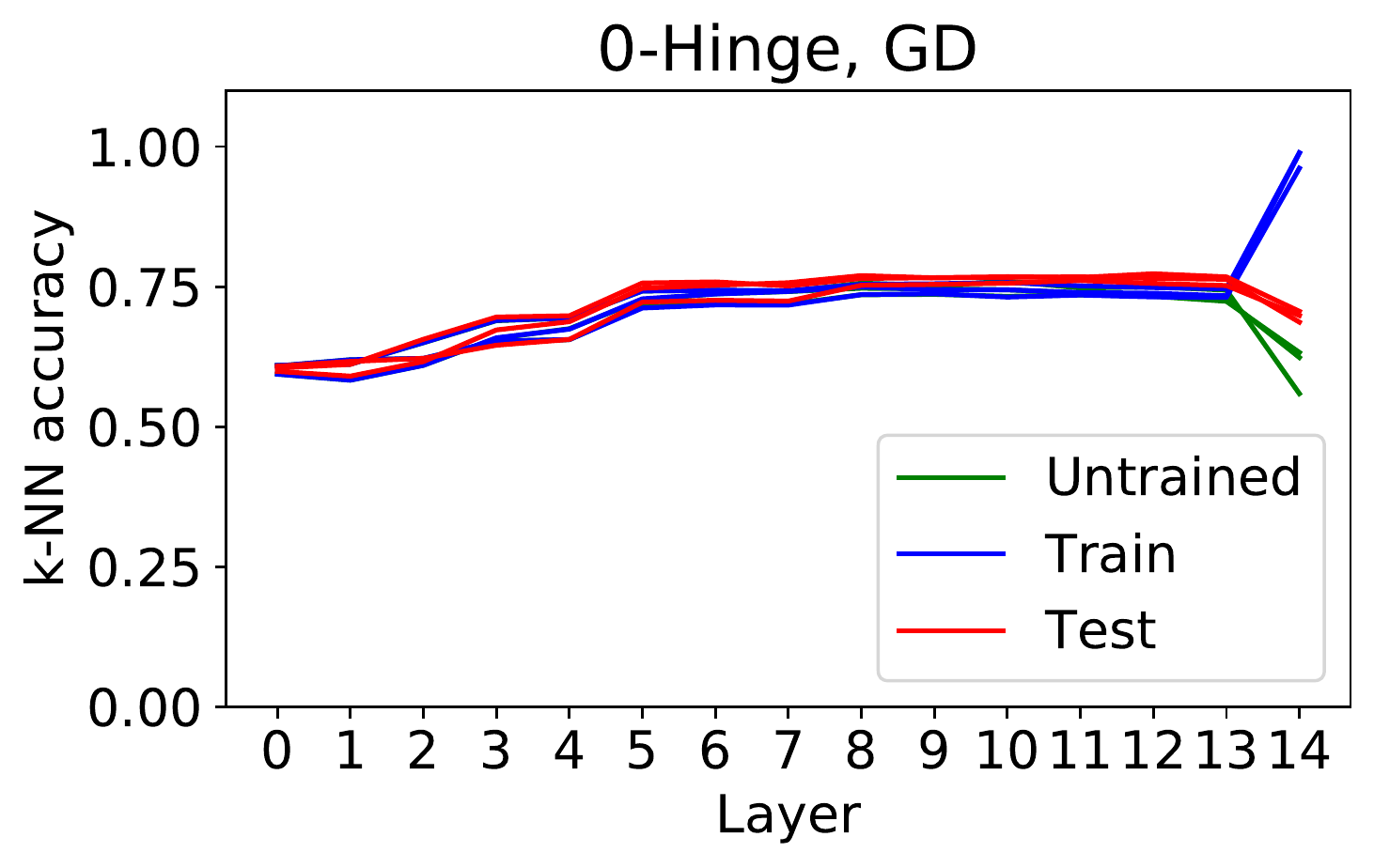}
\end{subfigure}
     \end{center}
\caption{\small \emph{ Accuracies of k-NN probes in the hidden layers of VGG16, resulting from each combination of Cross-Entropy vs. 0-Hinge loss and SGD with momentum and large initial learning rate vs. GD with momentum and small learning rate.}
In each case we compare to the probes for untrained (freshly initialized) networks.
Only the 0-Hinge with gradient descent using momentum and small learning rate (``0-Hinge, GD'') leads to clustering in the latest layers.
}
\label{fig:knn_expC}
\end{figure}

Table~\ref{tab:expC_margs} lists the mean accuracy and output margin for all four combinations of loss function and optimizer. 
We can see that the combination of both changes yields the smallest mean output margin, ~$10^2$ times smaller than the next smallest margin. 
Figure~\ref{fig:knn_expC} presents the k-NN probe accuracies in the hidden layers for all four combinations of loss and optimizer.
The combined intervention, which has the smallest margin, leads to the data being accurately clustered in the very latest layers. 

\begin{table*}[ht]
\begin{center}
    \begin{tabular}{l||c|c}
              Name  & Mean Accuracy & Mean Output Margin    \\
\hline\hline
CE, SGD & 87.6\%
 & $1.6\times 10^{1}$   \\
CE, GD & 86.7\% & $1.1\times 10^{1}$   \\
0-Hinge, SGD & 83.9\% & $6\times 10^{-2}$ \\
0-Hinge, GD & 69.5\% & $2.0\times 10^{-4}$ 
\end{tabular}
\caption{Mean accuracy and output margin for CE vs. 0-Hinge losses and SGD with momentum and large initial learning rate vs. GD with momentum and small learning rate.}
\label{tab:expC_margs}
\end{center}
\end{table*}

\section{Further Related Work\label{app:context}}

Previous studies of deep learning on the level of individual data points have: 
sought to explain its accuracy 
by focusing on the interference of per-example gradients during training~\citep{chatterjee2019coherent, zielinski2020weak}; 
improved our understanding of deep learning by studying its performance on datasets with partially randomized labels, which corresponds to a specific binary partitioning of example difficulty~\cite{arpit2017closer};
quantified example difficulty using 5 different observables: 1) the change in a network's output for elements of the training set after subsequent fine-tuning on a disjoint dataset, 2) the adversarial input margin of an example, 3) the agreement of models in an ensemble, 4) the average confidence of models in an ensemble, and 5) the disparate impact of differential privacy~\cite{carlini2019distribution}; identified difficult examples with those disproportionately impacted by pruning and compression~\cite{hooker2019compressed}, with those whose classifications are more often forgotten during training~\cite{forgetting19}, and with those that are least likely to be correctly classified in the validation set~\cite{chiyuan_cscores}; 
demonstrated a correspondence between those examples that a human finds difficult and examples a machine finds difficult~\cite{lalor2018understanding}. 
In contrast to these works, we study the computational difficulty of inferring the class of an input: the amount of computation used to connect that input with its class label inside the network.
Our definition of example difficulty is precisely described in Section~\ref{sec:on_ex_diff}.

In~\cite{hacohen2020let} the authors report that the order during training in which data points are learned is common between different architectures and random seeds in deep learning. 
In light of the correlation between prediction depth and the order of learning data points (as reported in Section~\ref{sec:compdiff_vs_learndiff}), their result reflects the sanity checks performed in Section~\ref{sec:ll_sanity_check}: that prediction depth is consistent between architectures and random seeds.

Distinct from the forms of example difficulty we describe in Section~\ref{sec:4_ex_diff},~\citet{hooker2019compressed} propose four different forms of example difficulty: ``ground truth label incorrect or inadequate'', ``multiple-object image'', ``corrupted image'', ``fine-grained classification''.
The forms of difficulty we describe in this paper follow directly from the computational difficulty of the examples, derived from the model's behavior. In contrast,~\citet{hooker2019compressed} employ intuitive notions of difficulty to define their four forms and ask humans to assign difficult examples to these categories. 

The Deep k-Nearest Neighbors method~\cite{papernot2018deep} builds a series of k-NN probes in the hidden spaces of the network.
When a test example is processed by the network, Deep k-NN identifies the nearest neighbors of the example in every layer, and then classifies the example according to the class labels of the aggregated nearest neighbors.
By comparing the number of neighbors the example has of the predicted class to the number of similarly labeled nearest neighbors that were recorded (across all layers) for examples in a hold-out test set, Deep k-NN is able to quantify the probability that the prediction is correct and to identify OOD examples.
However, the authors do not report the phenomena reported here.
Our results may yet enable the development of new Deep k-NN methods.
Another algorithm~\cite{bahri2020deep} constructs a k-NN probe in the logit space of a network, and demonstrates that this enables improved detection of mislabeled data.

\section{Consistency of the Main Results Reported in the Paper}

\subsection{Consistency of prediction depth between architectures\label{app:consistency_ll_archs}}

\begin{figure}[ht]
\begin{center}
\begin{subfigure}
         \centering
         \includegraphics[width=0.49\columnwidth]{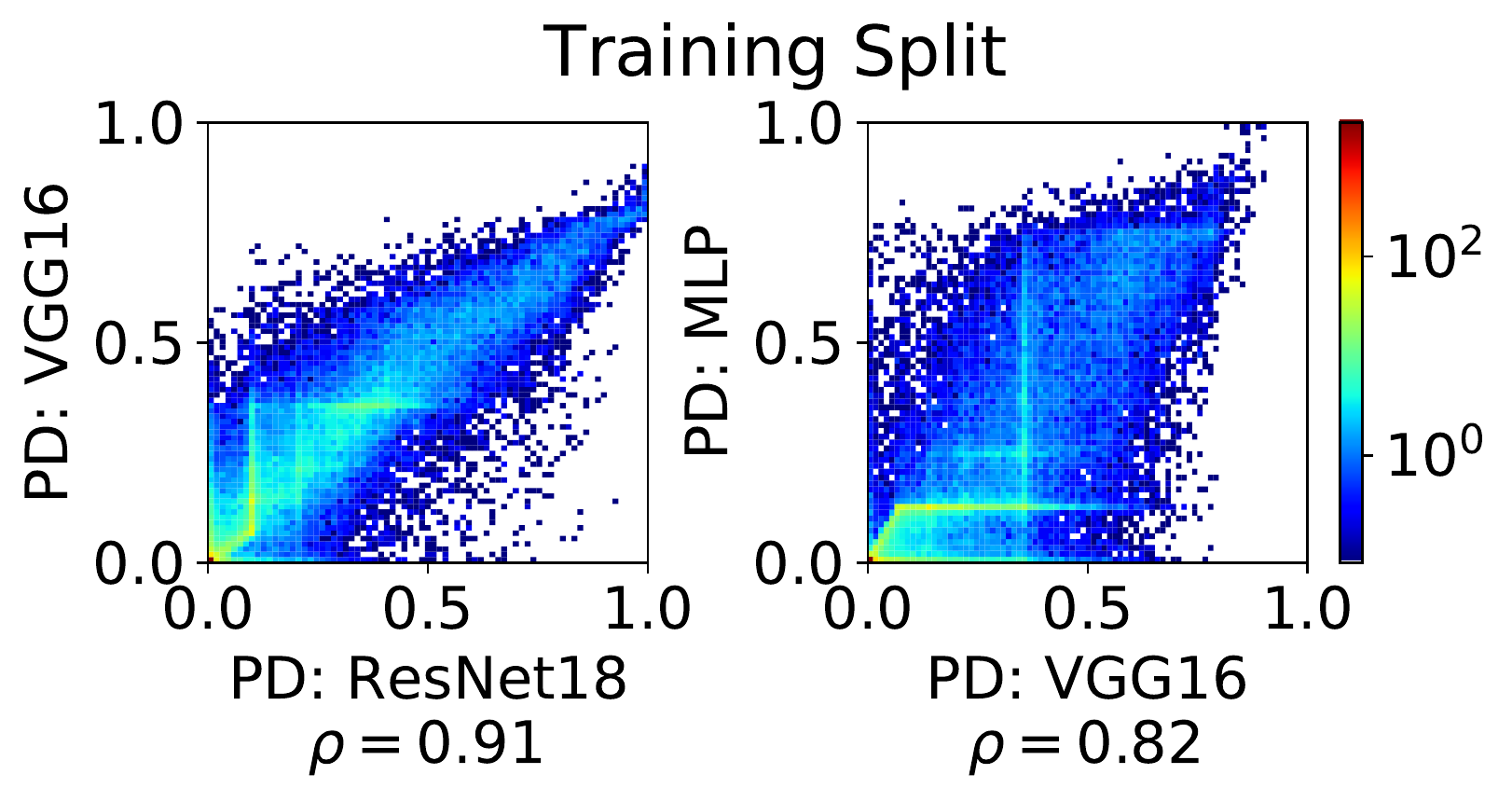}
\end{subfigure}
\begin{subfigure}
         \centering
         \includegraphics[width=0.49\columnwidth]{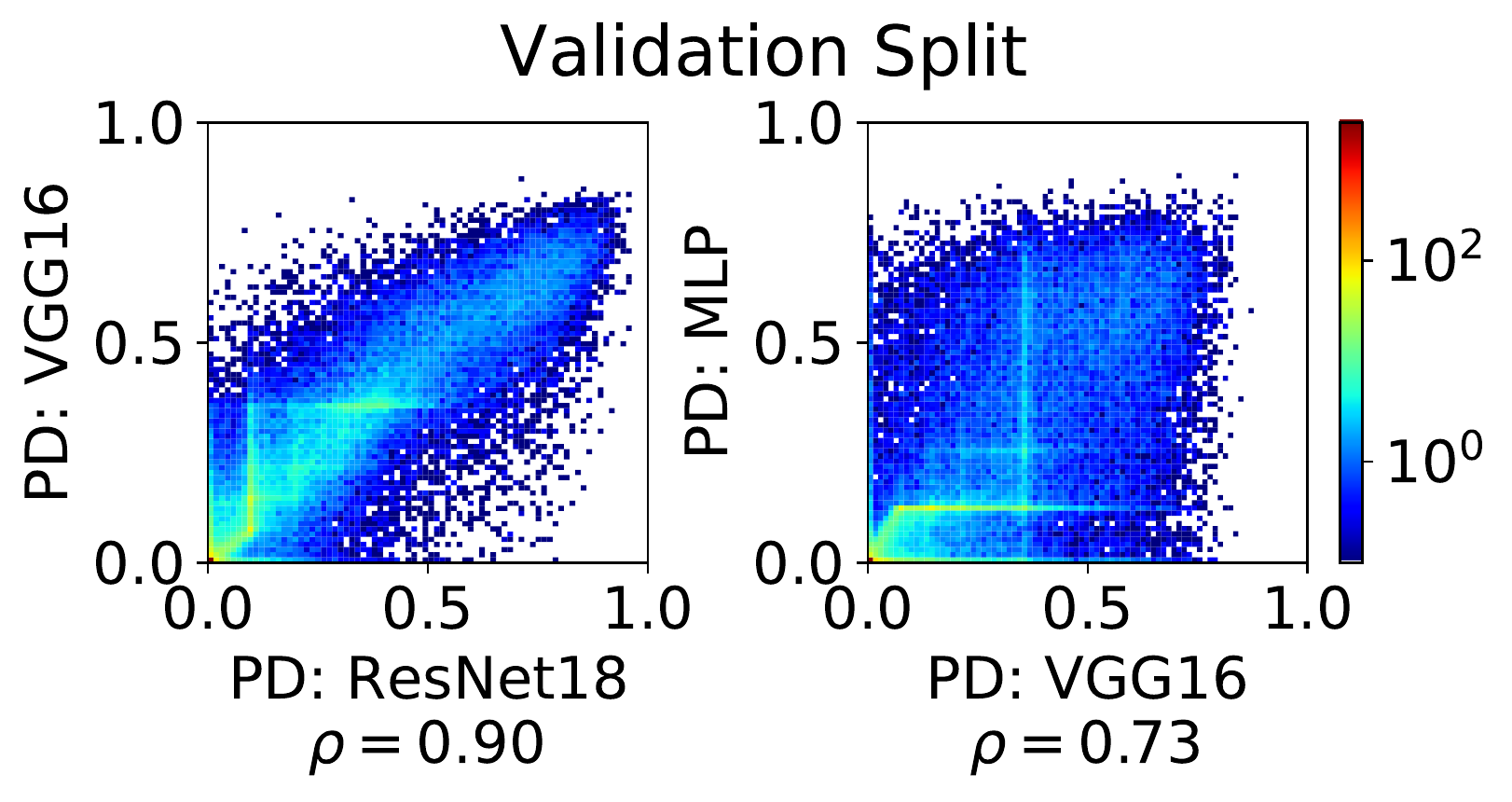}
\end{subfigure}
\end{center}
\caption{\emph{ Consistency of prediction depth between architectures for SVHN.} Histograms comparing the mean value of prediction depth obtained for each data point, across the ensemble of trained models. Left pair: training split. Right pair: validation split. Spearman's Correlation Coefficient is given beneath each plot. See Appendix~\ref{app:consistency_ll_archs} for details.}
\label{fig:ll_between_archs_svhn}
\end{figure}

\begin{figure}[ht]
\begin{center}
\begin{subfigure}
         \centering
         \includegraphics[width=0.49\columnwidth]{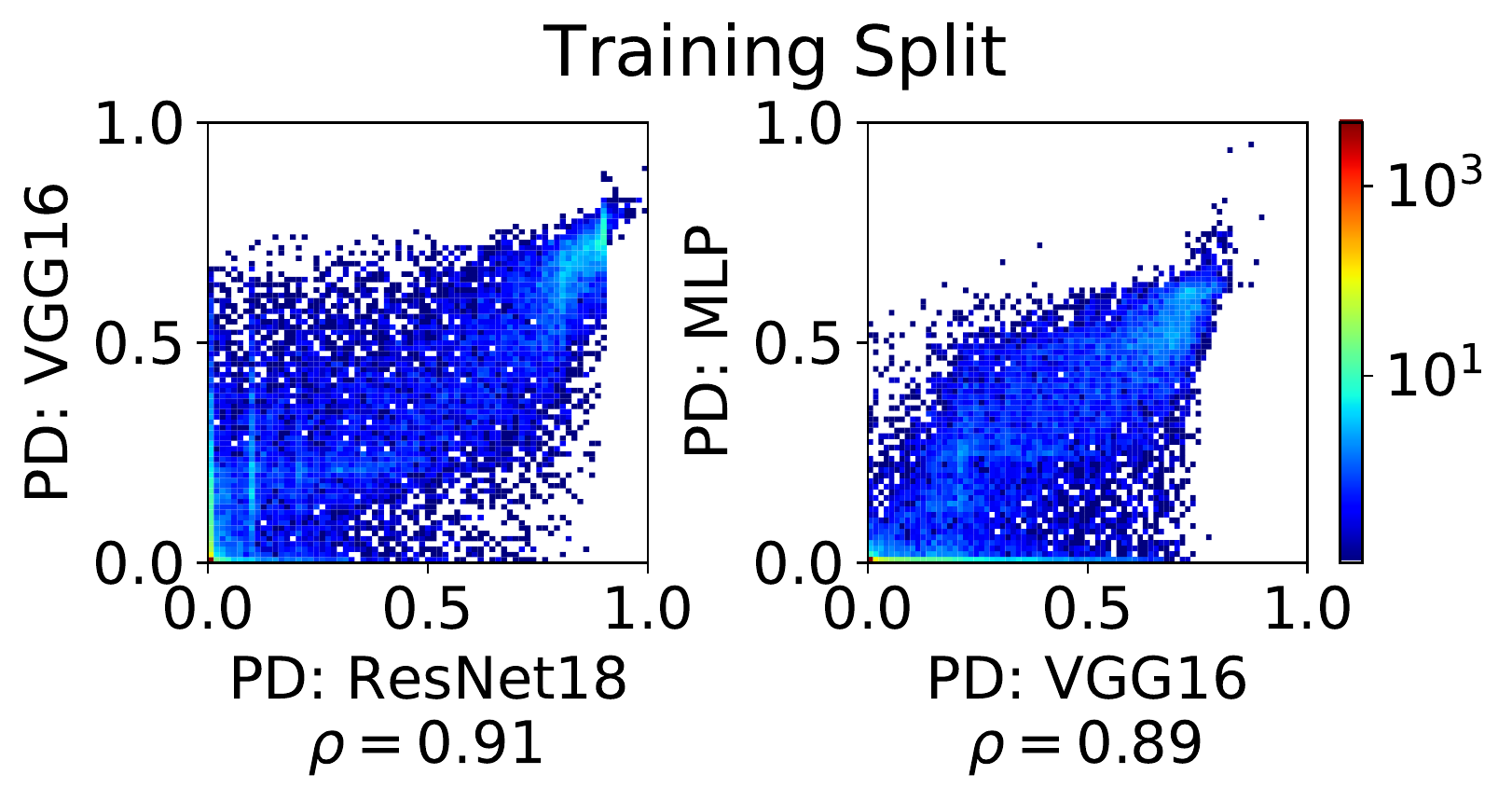}
\end{subfigure}
\begin{subfigure}
         \centering
         \includegraphics[width=0.49\columnwidth]{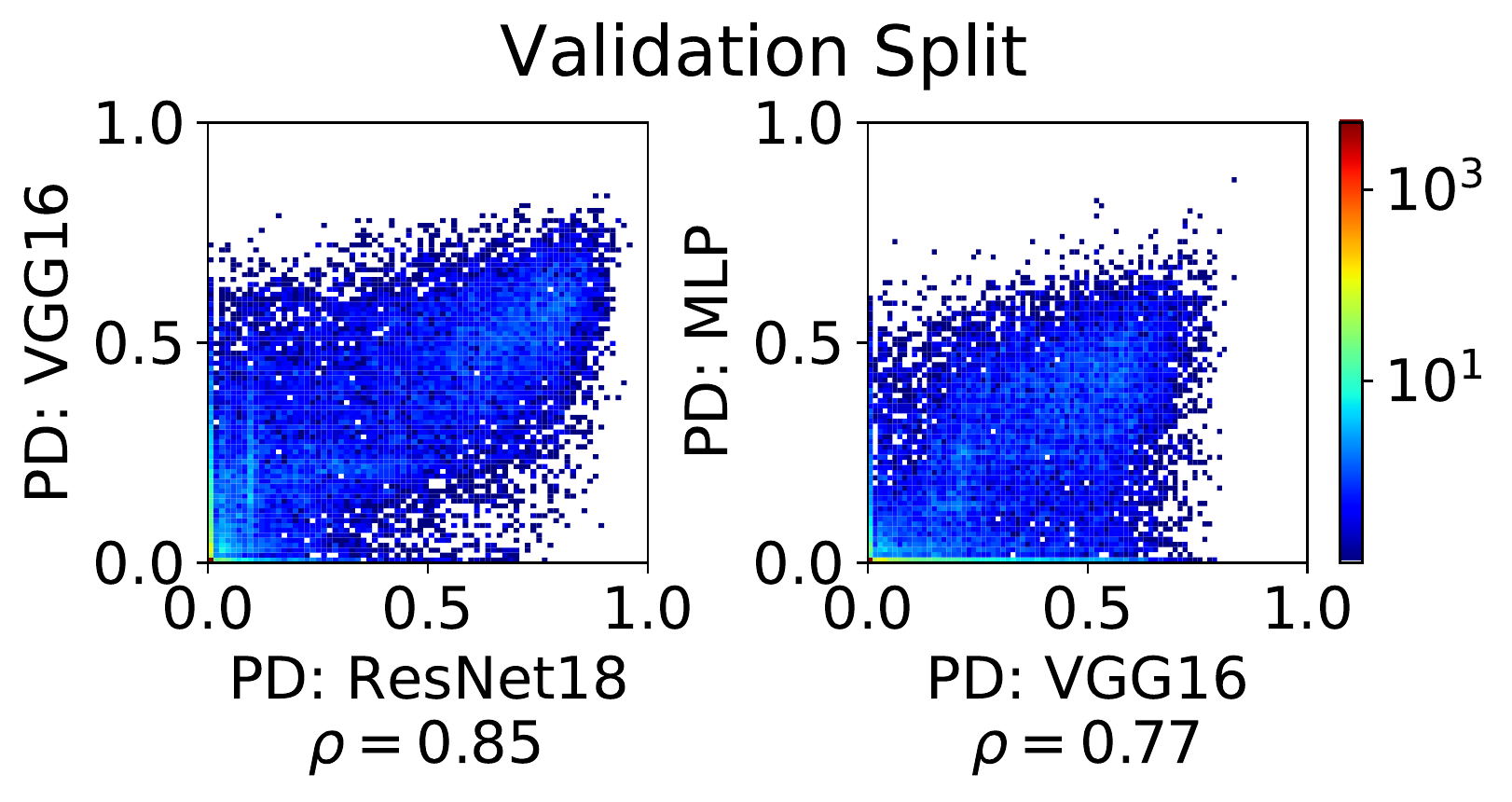}
\end{subfigure}
\end{center}
\caption{\emph{Consistency of prediction depth between architectures for Fashion MNIST.} Histograms comparing the mean value of prediction depth obtained for each data point, across the ensemble of trained models. Left pair: training split. Right pair: validation split. Spearman's Correlation Coefficient is given beneath each plot. See Appendix~\ref{app:consistency_ll_archs} for details.}
\label{fig:ll_between_archs_fmnist}
\end{figure}

\begin{figure}[ht]
\begin{center}
\begin{subfigure}
         \centering
         \includegraphics[width=0.49\columnwidth]{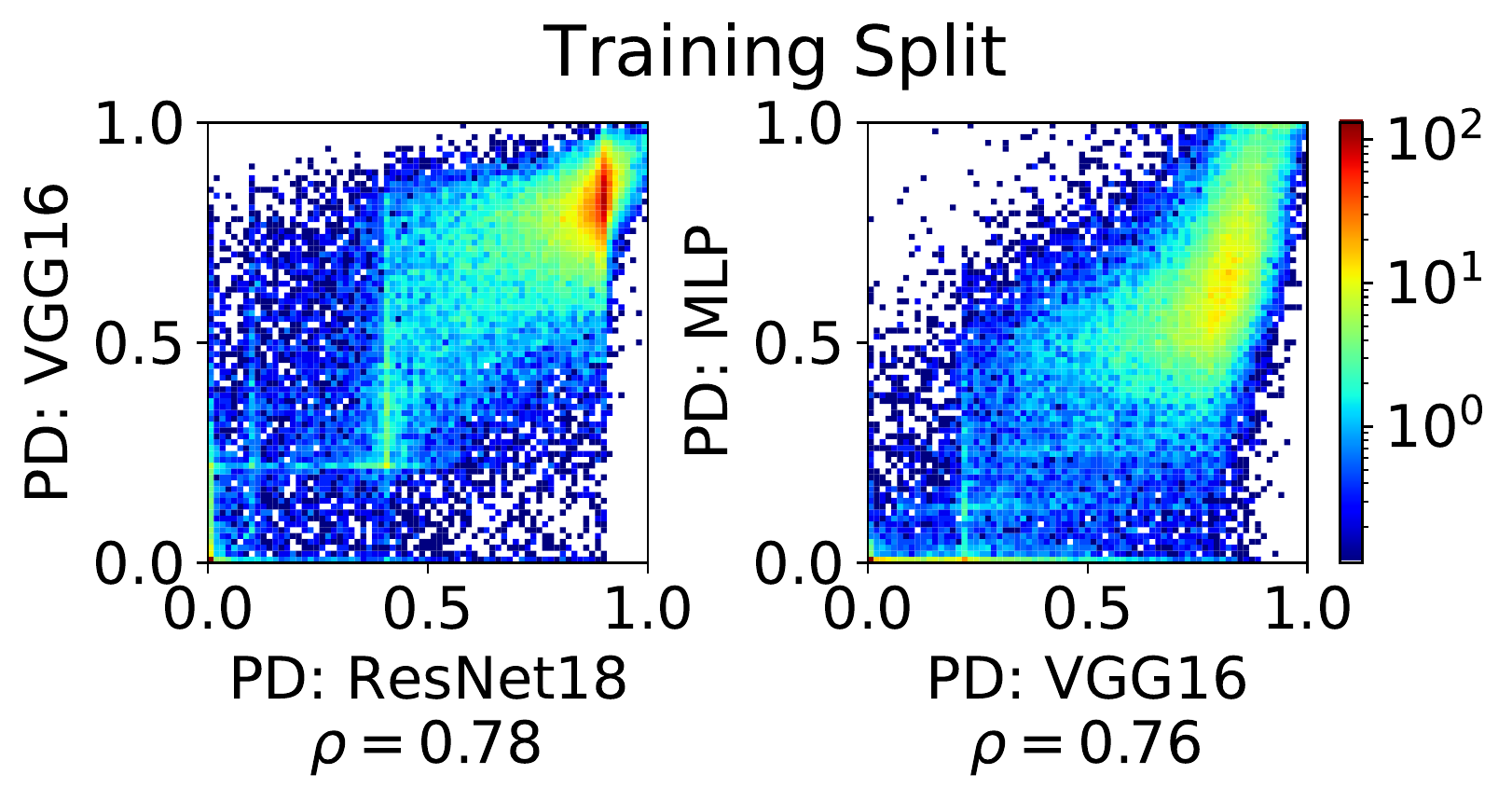}
\end{subfigure}
\begin{subfigure}
         \centering
         \includegraphics[width=0.49\columnwidth]{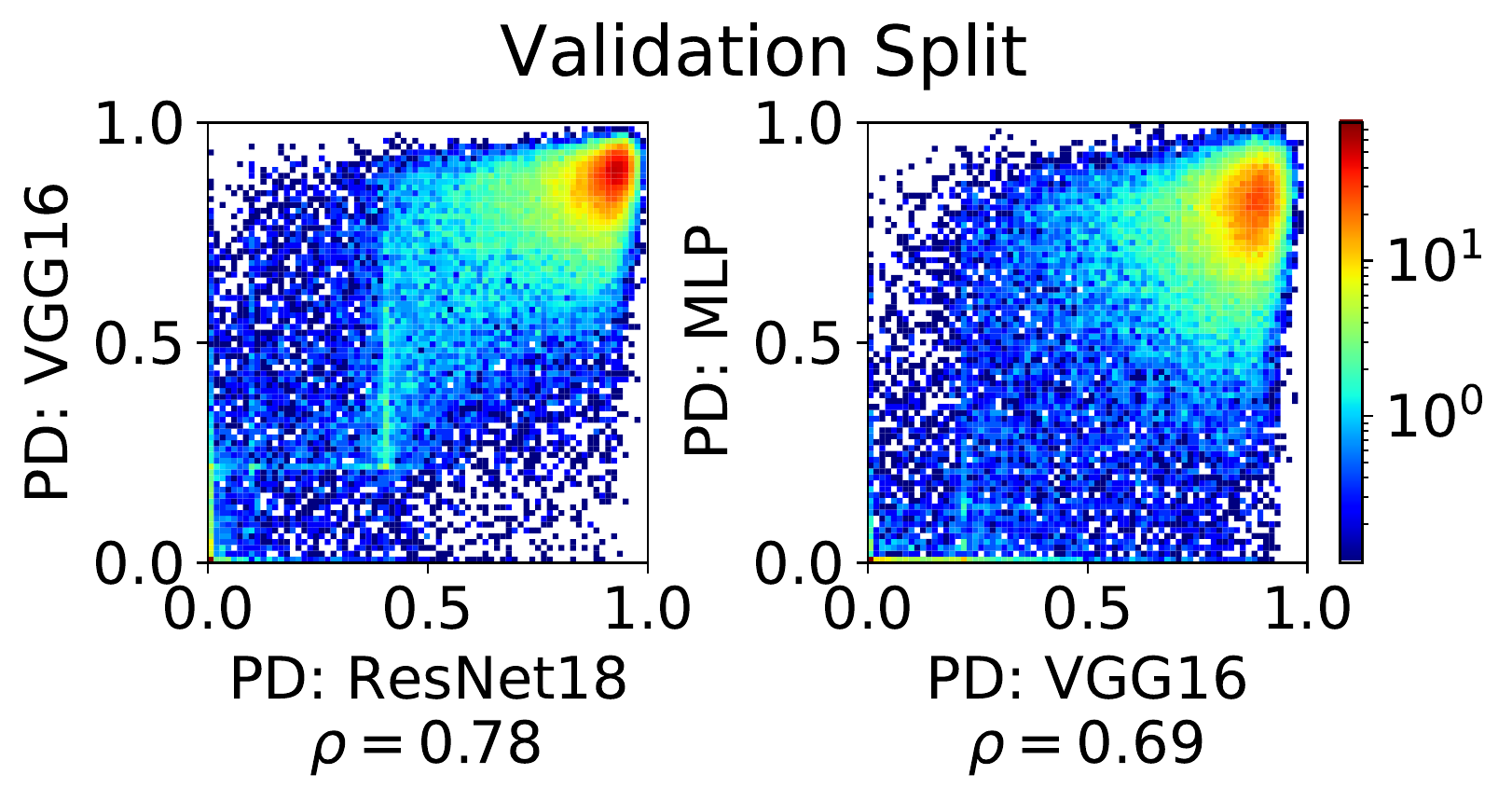}
\end{subfigure}
\end{center}
\caption{\emph{ Consistency of prediction depth between architectures for CIFAR100.} Histograms comparing the mean value of prediction depth obtained for each data point, across the ensemble of trained models. Left pair: training split. Right pair: validation split. Spearman's Correlation Coefficient is given beneath each plot. See Appendix~\ref{app:consistency_ll_archs} for details.}
\label{fig:ll_between_archs_cifar100}
\end{figure}

\begin{figure}[ht]
\begin{center}
\begin{subfigure}
         \centering
         \includegraphics[width=0.49\columnwidth]{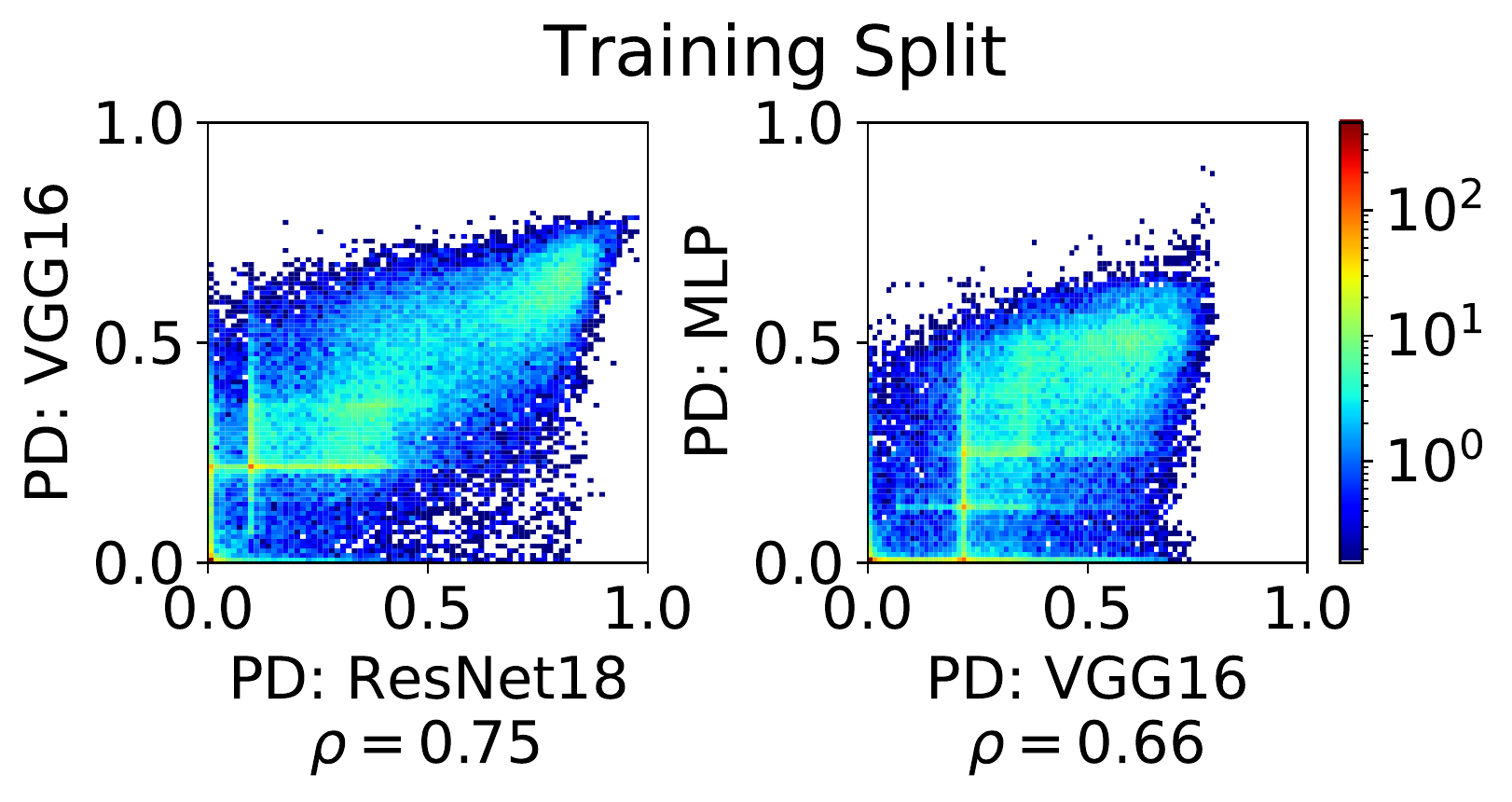}
\end{subfigure}
\begin{subfigure}
         \centering
         \includegraphics[width=0.49\columnwidth]{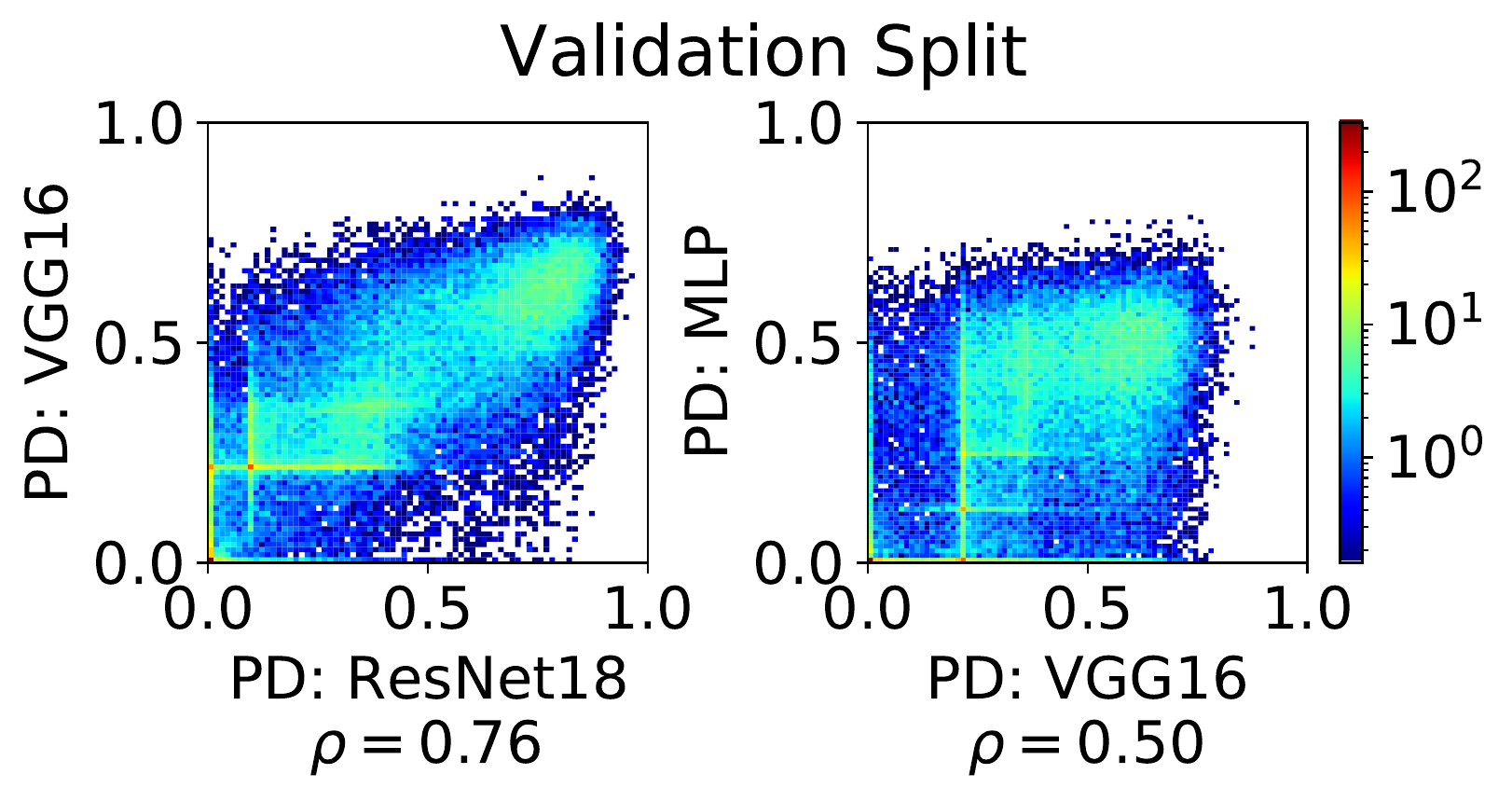}
\end{subfigure}
\end{center}
\caption{\emph{Consistency of prediction depth between architectures for CIFAR10.} Histograms comparing the mean value of prediction depth obtained for each data point, across the ensemble of trained models. Left pair: training split. Right pair: validation split. Spearman's Correlation Coefficient is given beneath each plot. See Appendix~\ref{app:consistency_ll_archs} for details.}
\label{fig:ll_between_archs_cifar10}
\end{figure}

To visually reinforce the correlations reported in Figure~\ref{fig:ll_between_archs_svhn_main} (right), Figures~\ref{fig:ll_between_archs_svhn} to~\ref{fig:ll_between_archs_cifar10} reproduce the result from Figure~\ref{fig:ll_between_archs_svhn_main} (right) for all datasets in both the training and validation splits.
For each combination of dataset and architecture we trained 250 models with random 90:10\% training:validation splits as described in Appendix~\ref{app:experiments_desc}.
These histograms compare the mean prediction depths of the data points between different architectures.
Separate plots are shown for the training and validation splits.
In each case we've rescaled prediction depth to the interval $[0, 1]$ for visual ease of comparison between datasets.
Each histogram is accompanied by the corresponding Spearman's Correlation Coefficient.

\subsection{Relationship between prediction depth and prediction consistency \label{app:ll_pcpm}}

\begin{figure*}[ht!]
\begin{center}
\begin{subfigure}
         \centering
         \includegraphics[width=0.49\columnwidth]{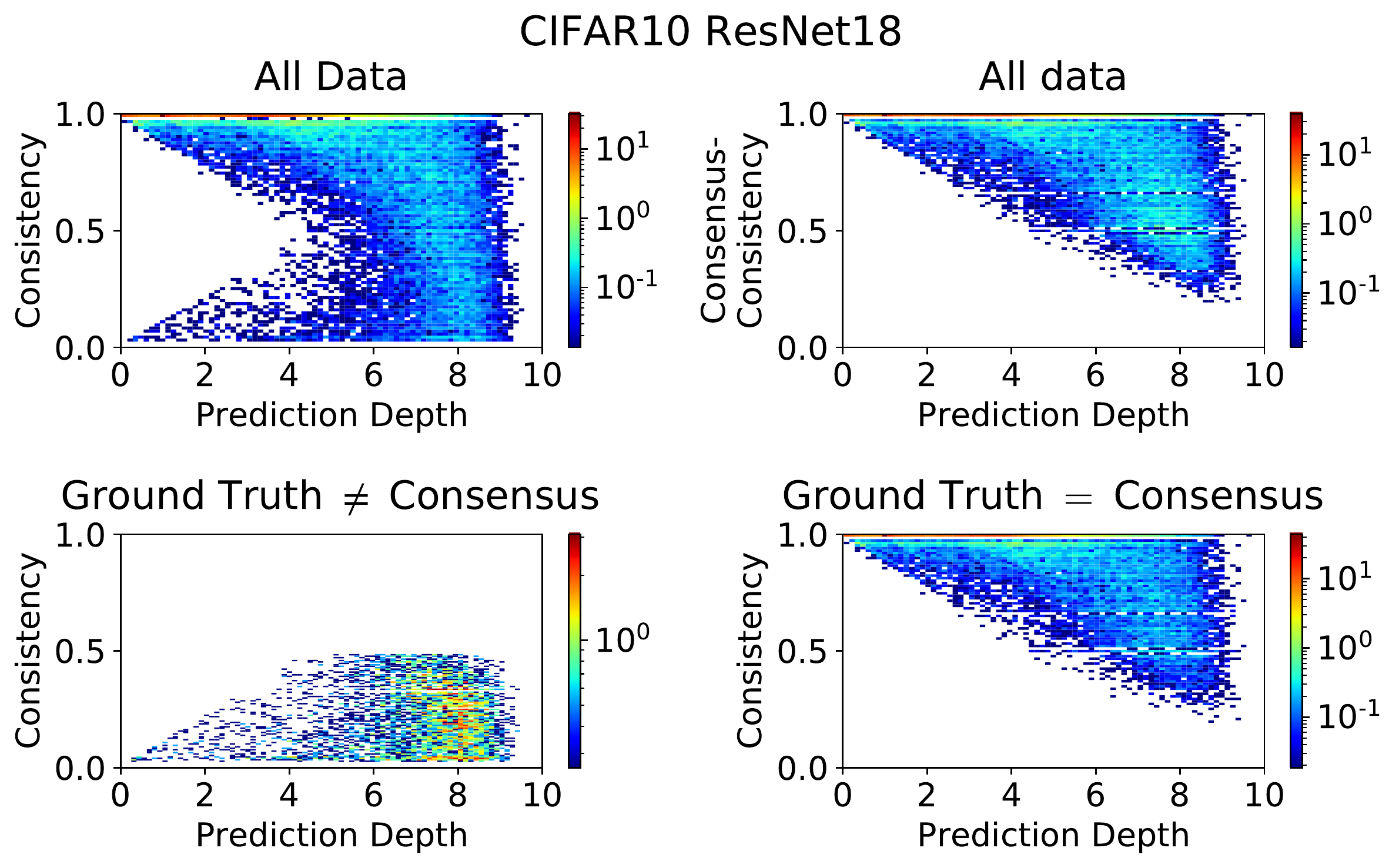}
\end{subfigure}
\hfill
\begin{subfigure}
         \centering
         \includegraphics[width=0.49\columnwidth]{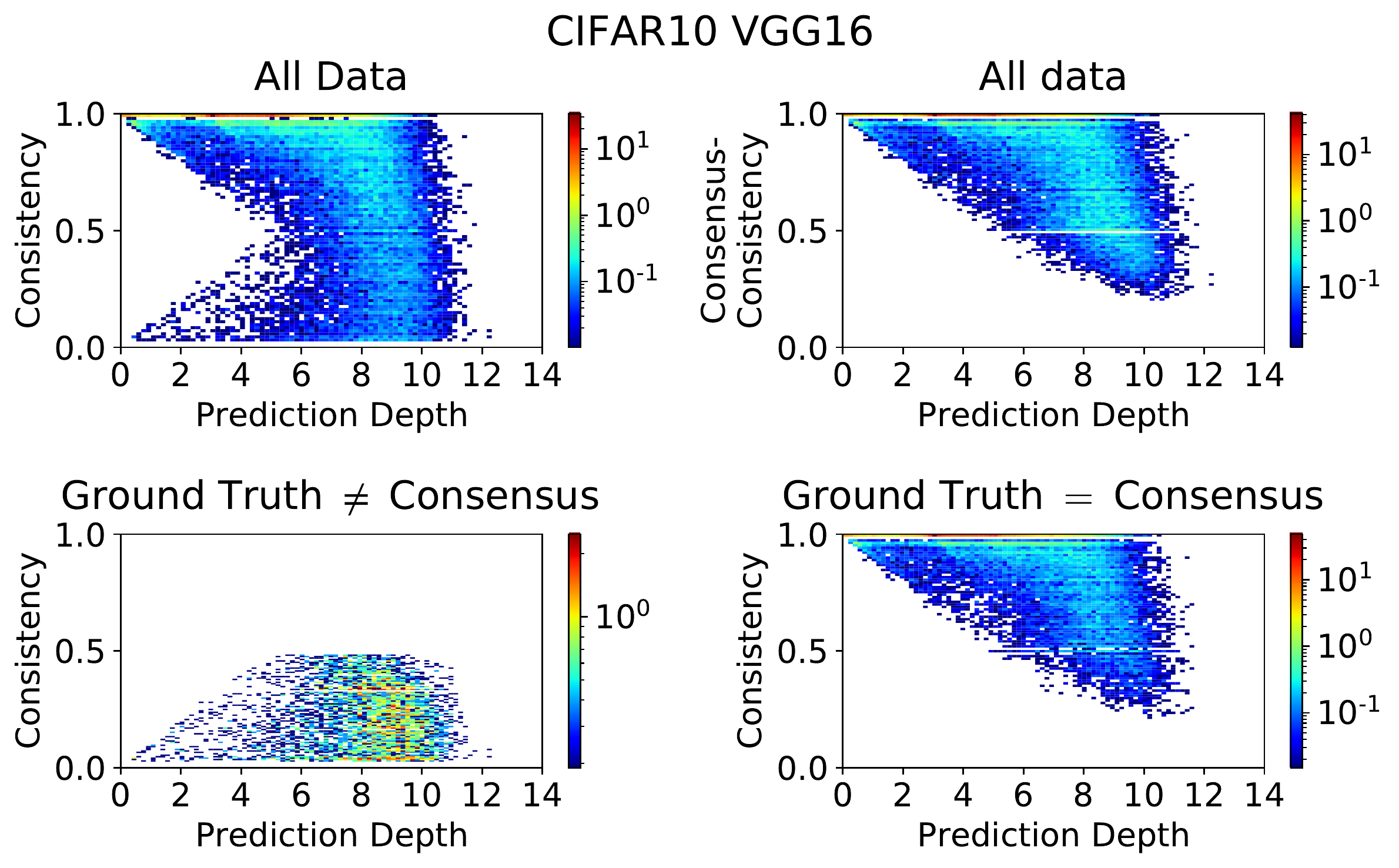}
\end{subfigure}
\hfill
\begin{subfigure}
         \centering
         \includegraphics[width=0.49\columnwidth]{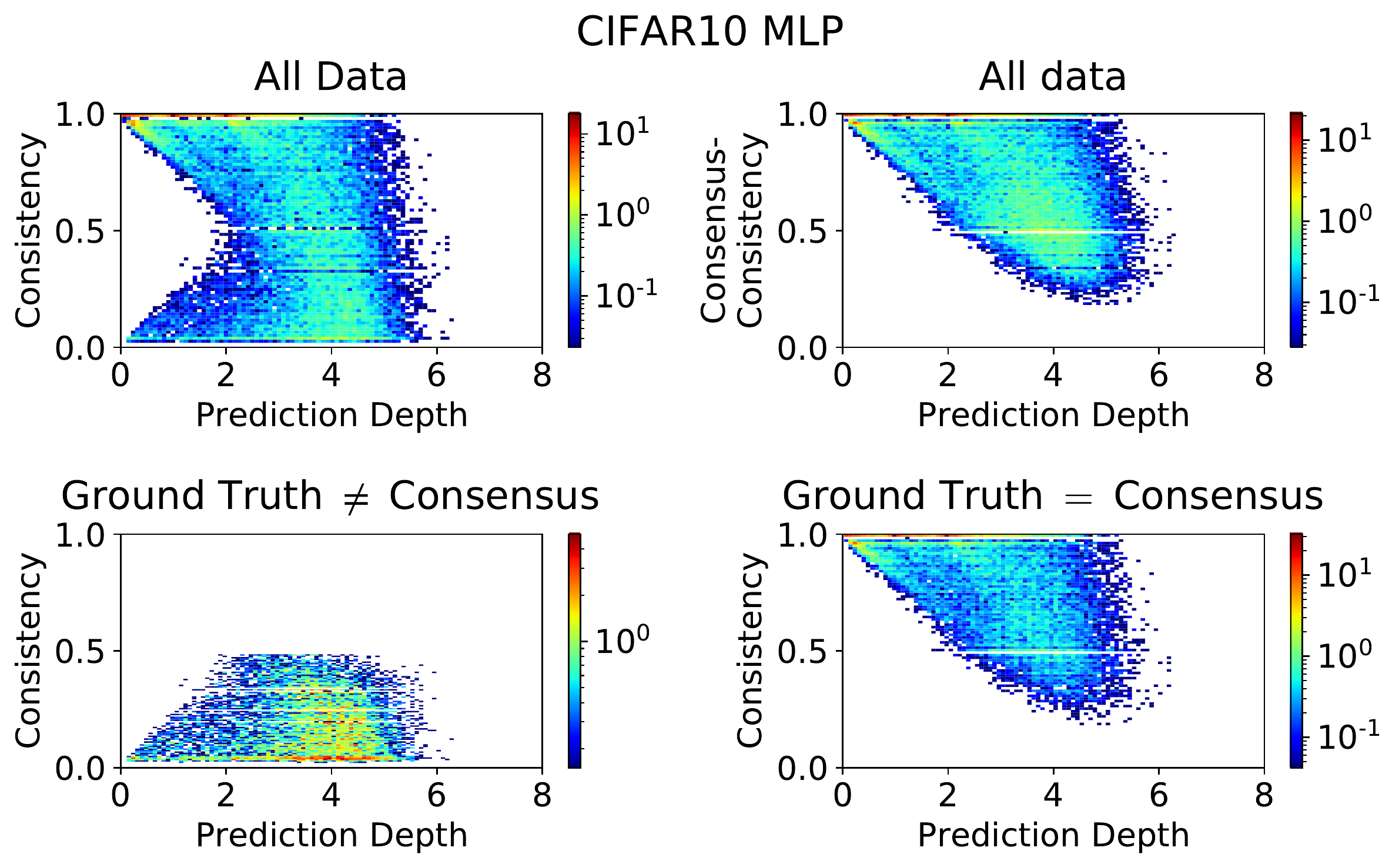}
\end{subfigure}
\hfill
\begin{subfigure}
         \centering
         \includegraphics[width=0.49\columnwidth]{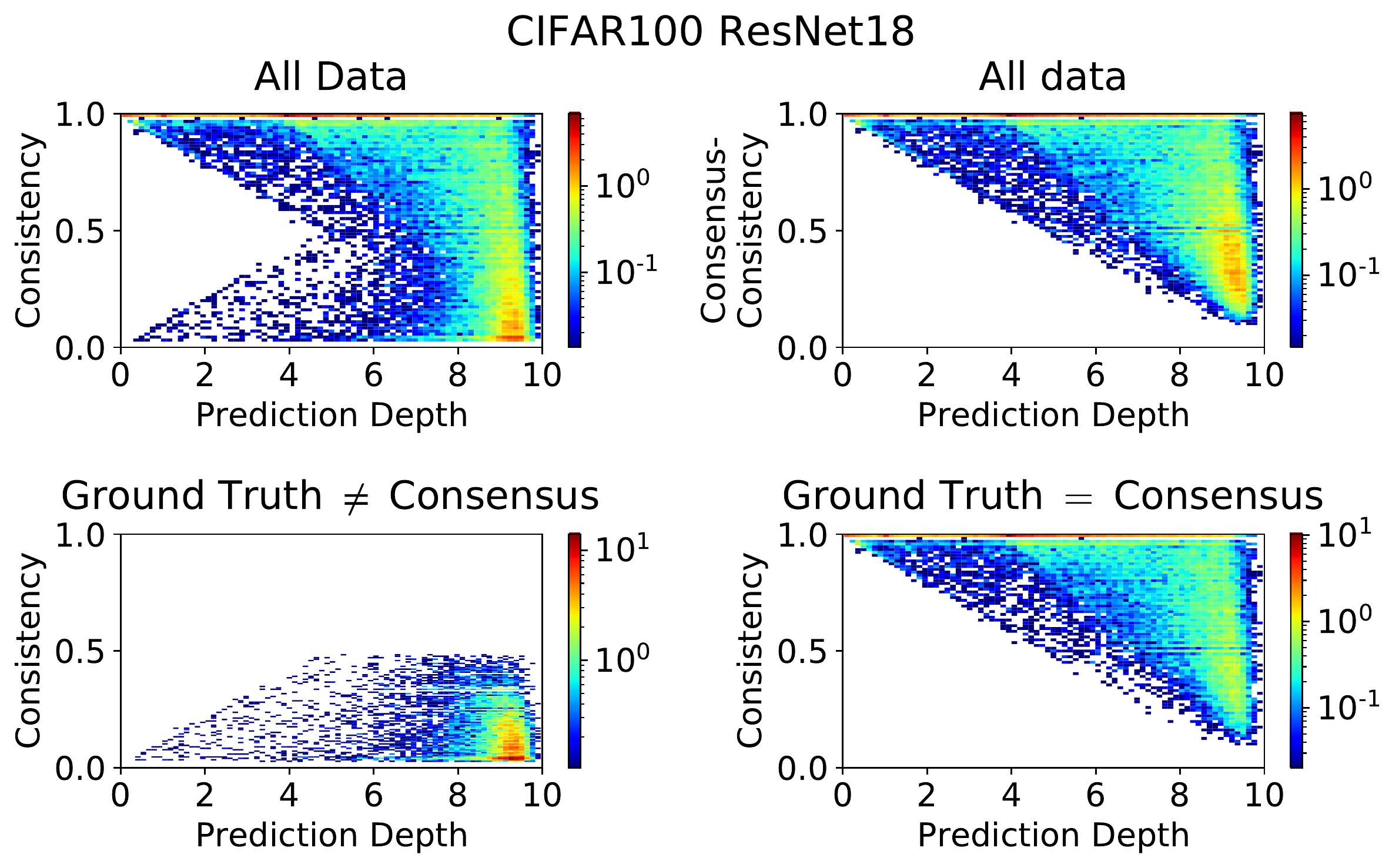}
\end{subfigure}
\begin{subfigure}
         \centering
         \includegraphics[width=0.49\columnwidth]{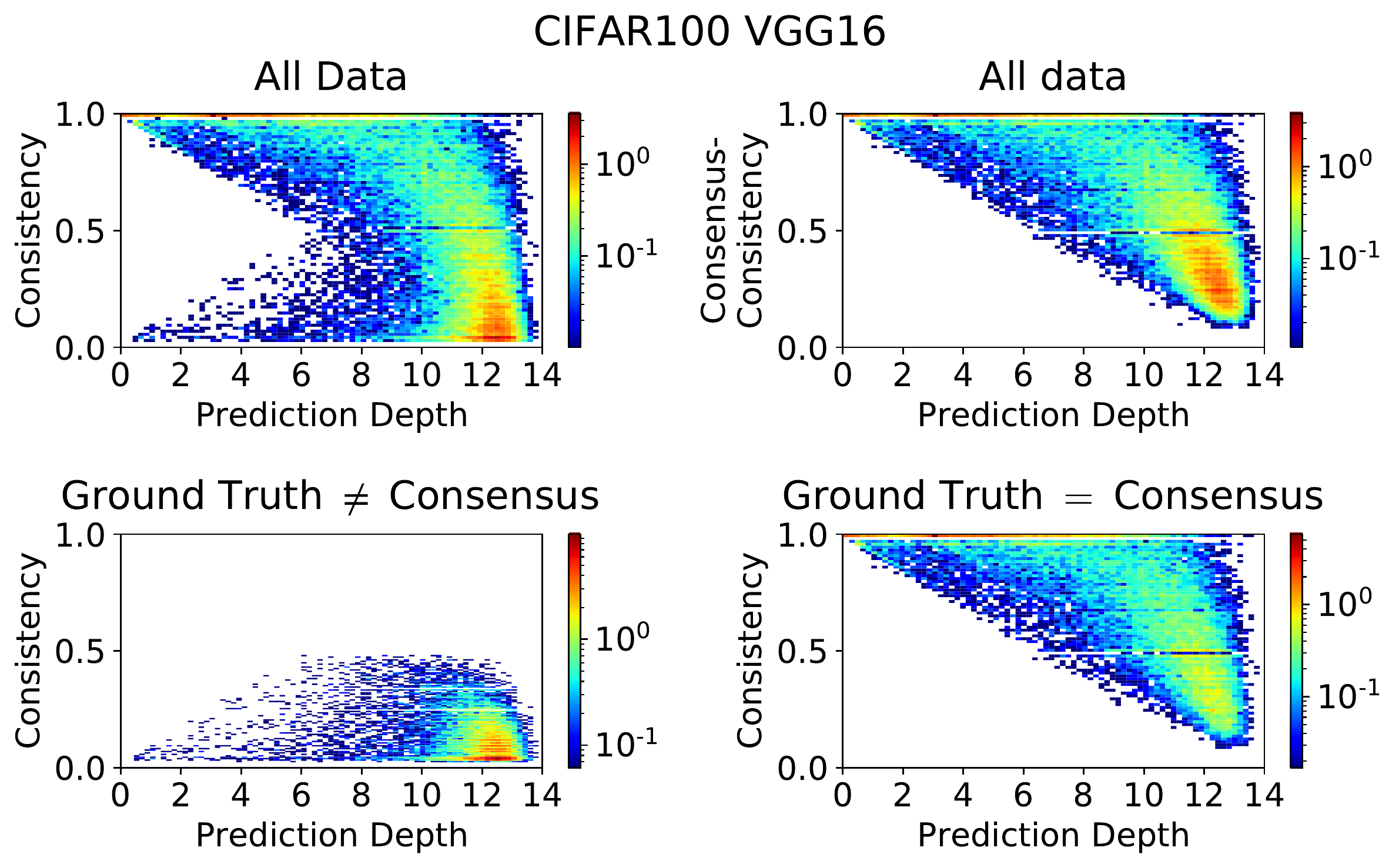}
\end{subfigure}
\begin{subfigure}
         \centering
         \includegraphics[width=0.49\columnwidth]{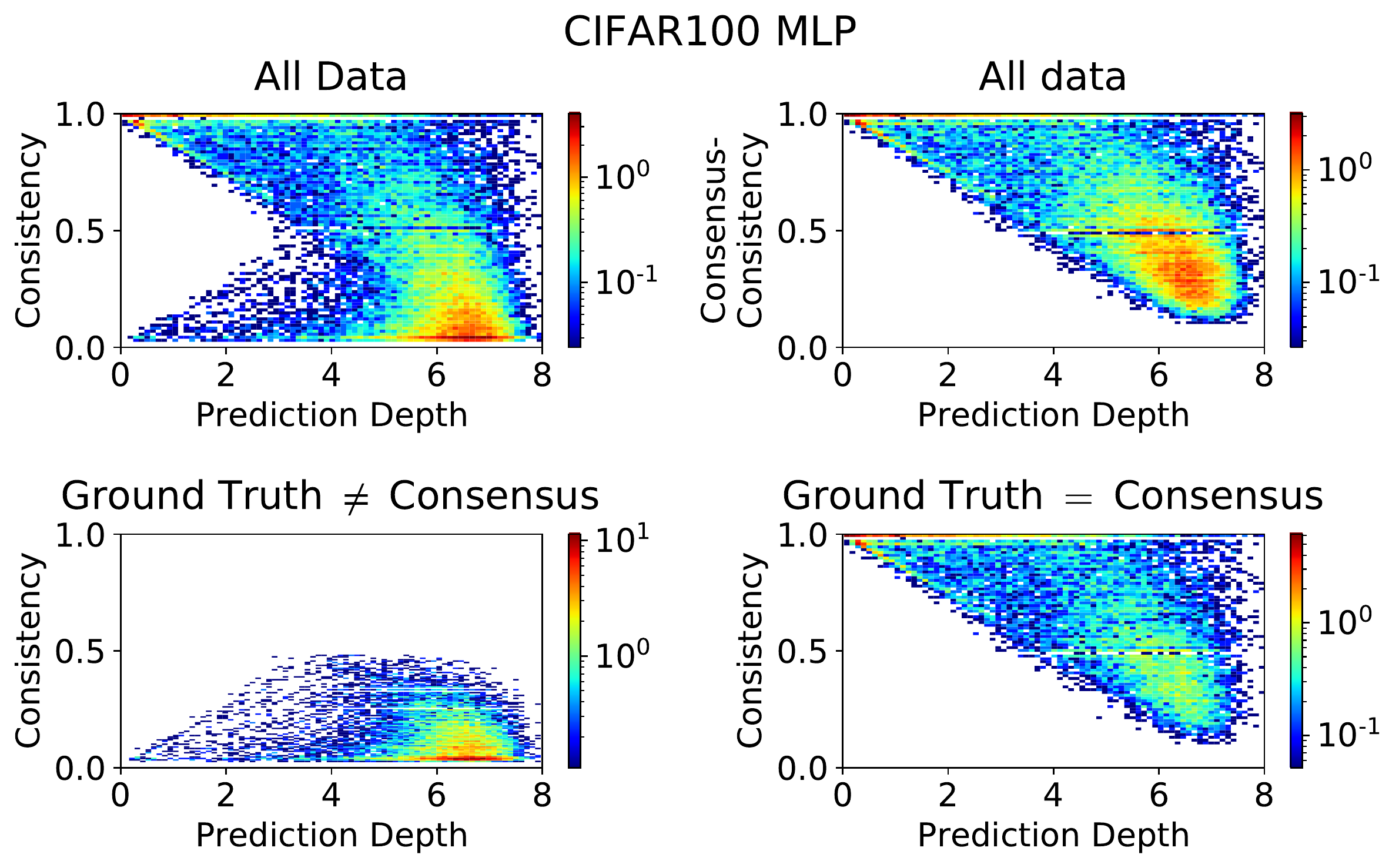}
\end{subfigure}
\end{center}
\caption{This figure demonstrates the consistency of the behavior shown in Figure~\ref{fig:generalization_vs_depth} and Figure~\ref{fig:ent_vs_depth} (left) for all architectures with CIFAR10 and CIFAR100.
\label{fig:ll_vs_pcpm1}
}
\end{figure*}

\begin{figure*}[ht!]
\begin{center}
\begin{subfigure}
         \centering
         \includegraphics[width=0.49\columnwidth]{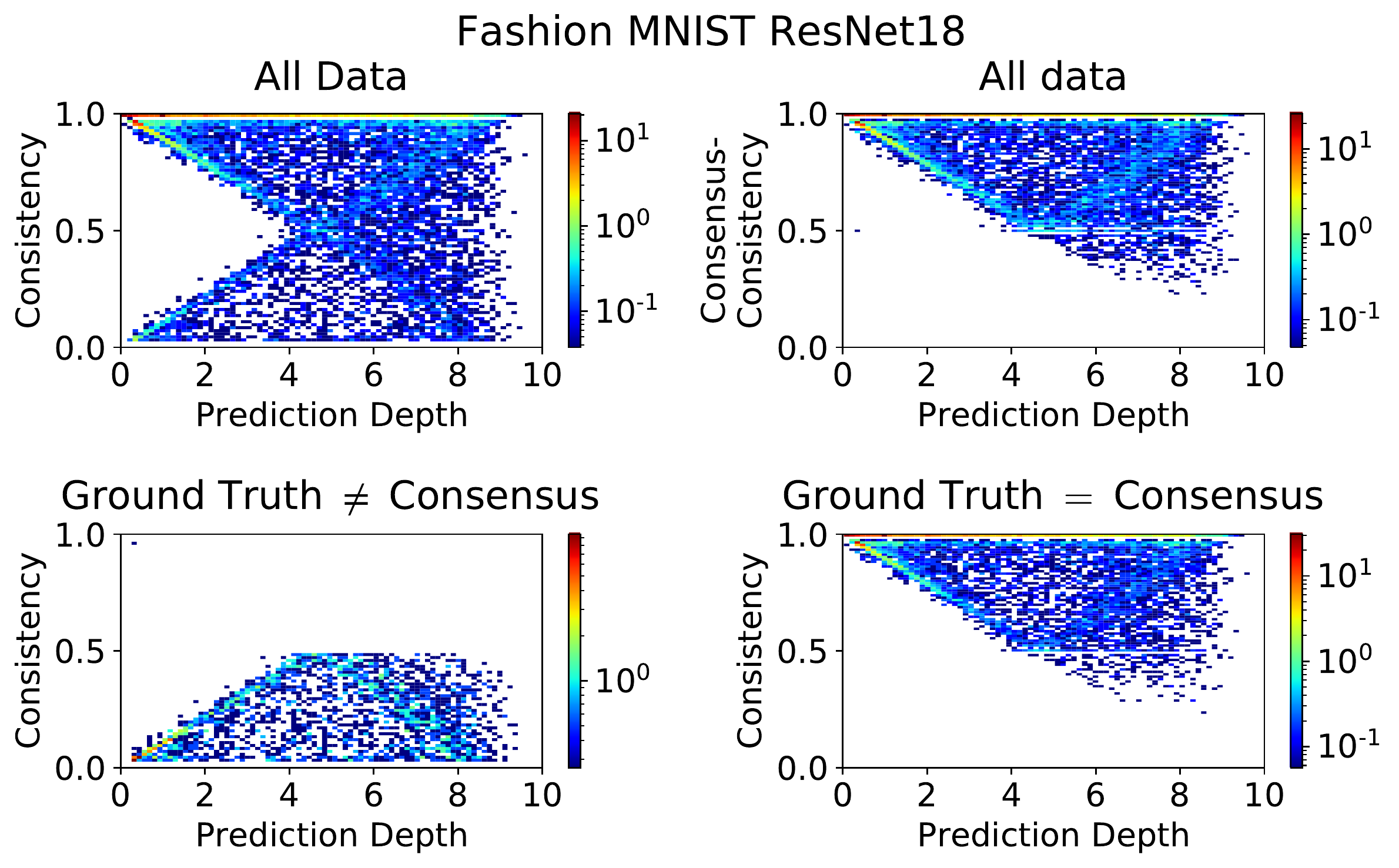}
\end{subfigure}
\begin{subfigure}
         \centering
         \includegraphics[width=0.49\columnwidth]{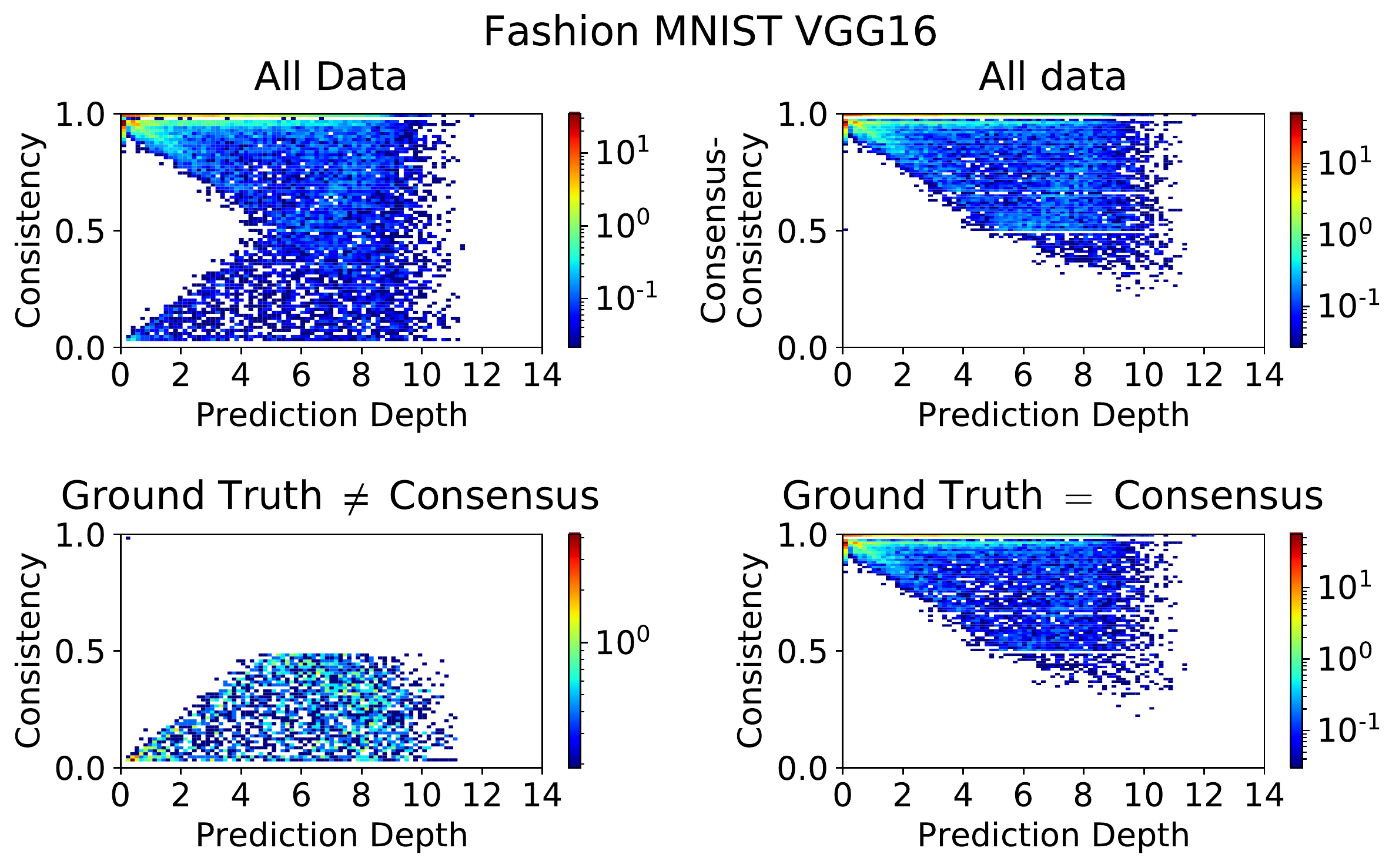}
\end{subfigure}
\begin{subfigure}
         \centering
         \includegraphics[width=0.49\columnwidth]{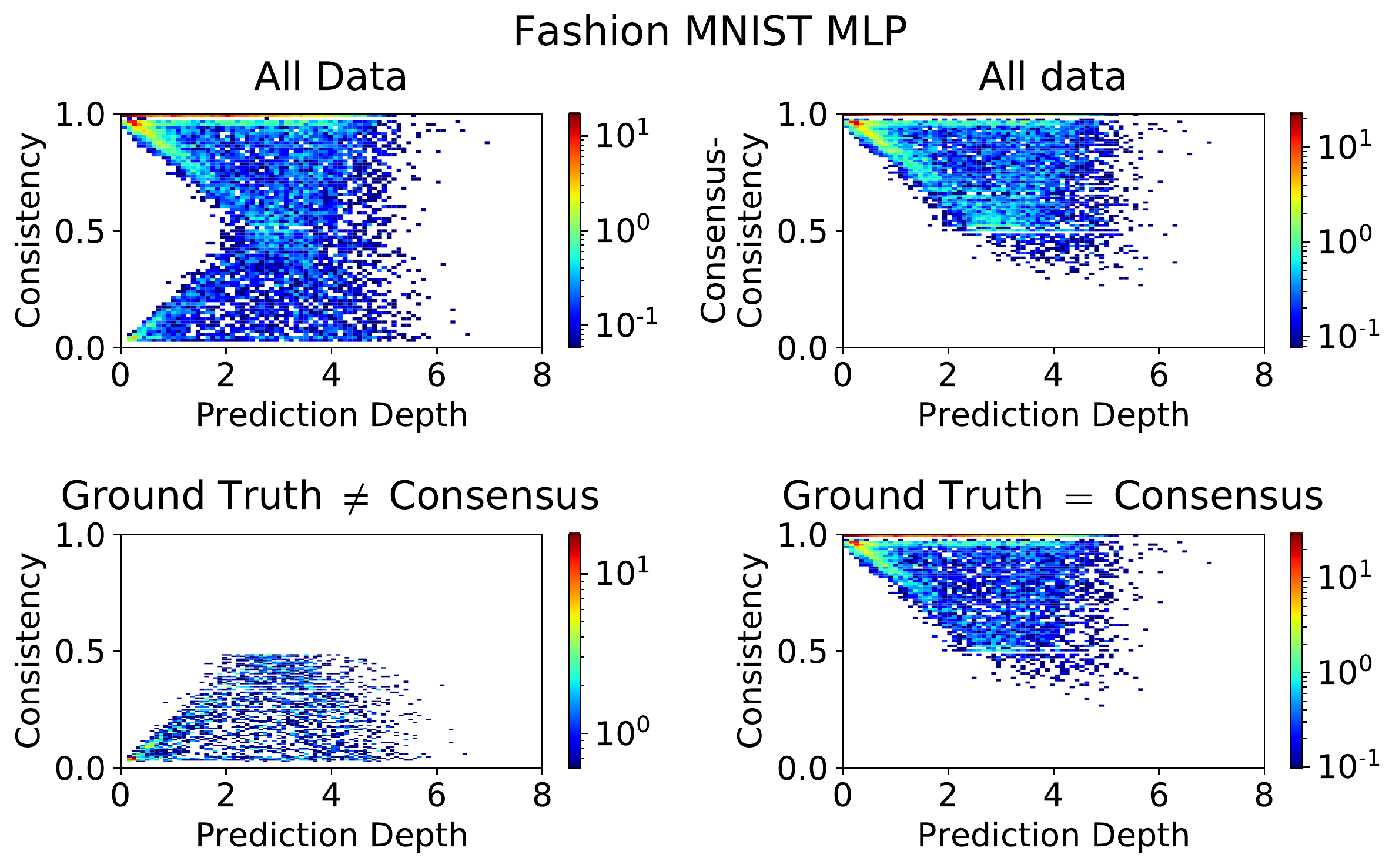}
\end{subfigure}
\begin{subfigure}
         \centering
         \includegraphics[width=0.49\columnwidth]{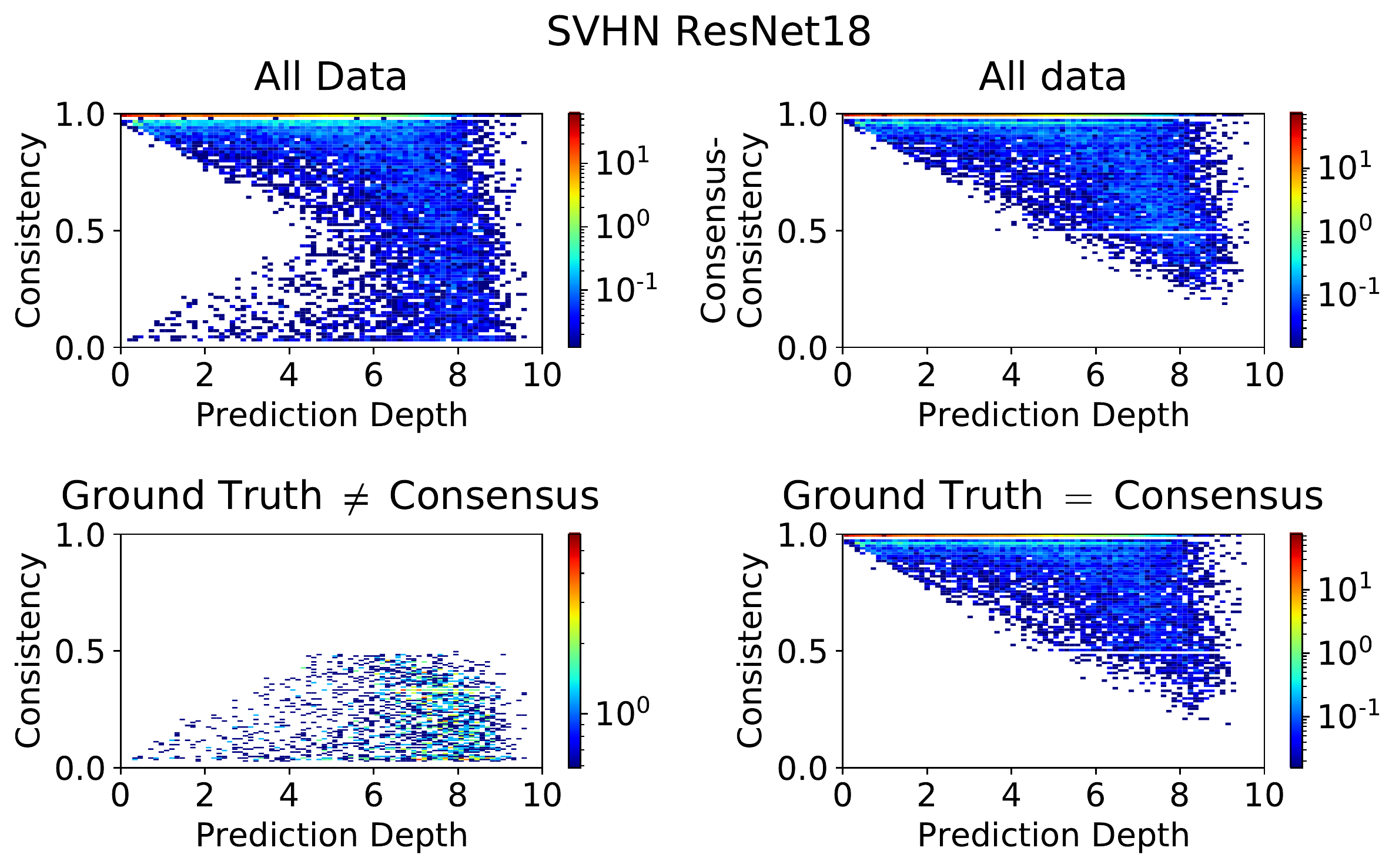}
\end{subfigure}
\begin{subfigure}
         \centering
         \includegraphics[width=0.49\columnwidth]{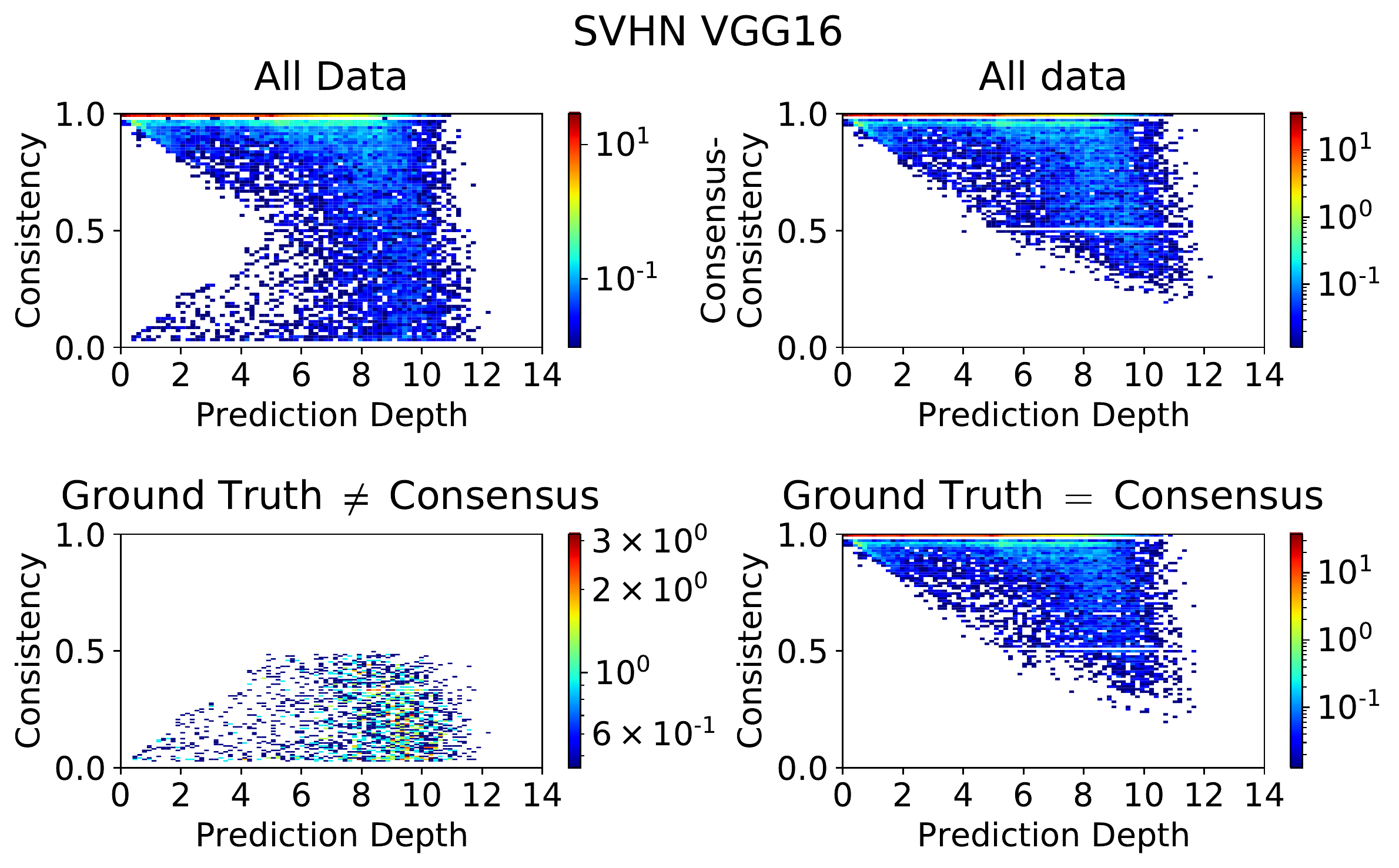}
\end{subfigure}
\begin{subfigure}
         \centering
         \includegraphics[width=0.49\columnwidth]{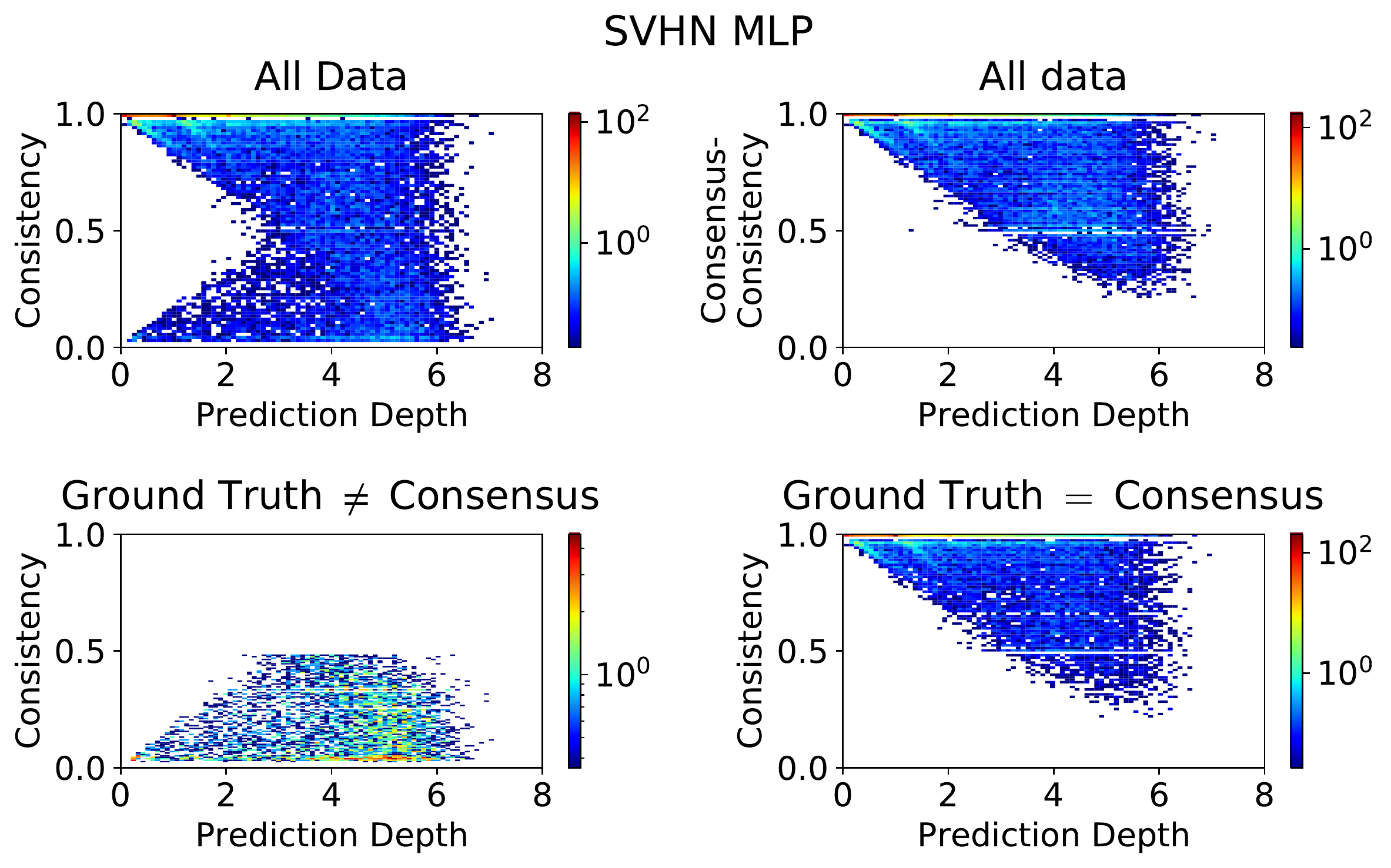}
\end{subfigure}
\end{center}
\caption{This figure demonstrates the consistency of the behavior shown in Figure~\ref{fig:generalization_vs_depth} and Figure~\ref{fig:ent_vs_depth} (left) for all architectures with Fashion MNIST and SVHN.}
\label{fig:ll_vs_pcpm2}
\end{figure*}

Figures~\ref{fig:ll_vs_pcpm1} and~\ref{fig:ll_vs_pcpm2} reproduce the results of Figure~\ref{fig:generalization_vs_depth} and Figure~\ref{fig:ent_vs_depth} (left) for every dataset and architecture.
The gradients of the linear bounds reported in the paper depend on the difficulty of the classification task: easier tasks are solved after fewer layers.

\begin{figure*}[ht!]
\begin{center}
\begin{subfigure}
         \centering
         \includegraphics[width=0.32\columnwidth]{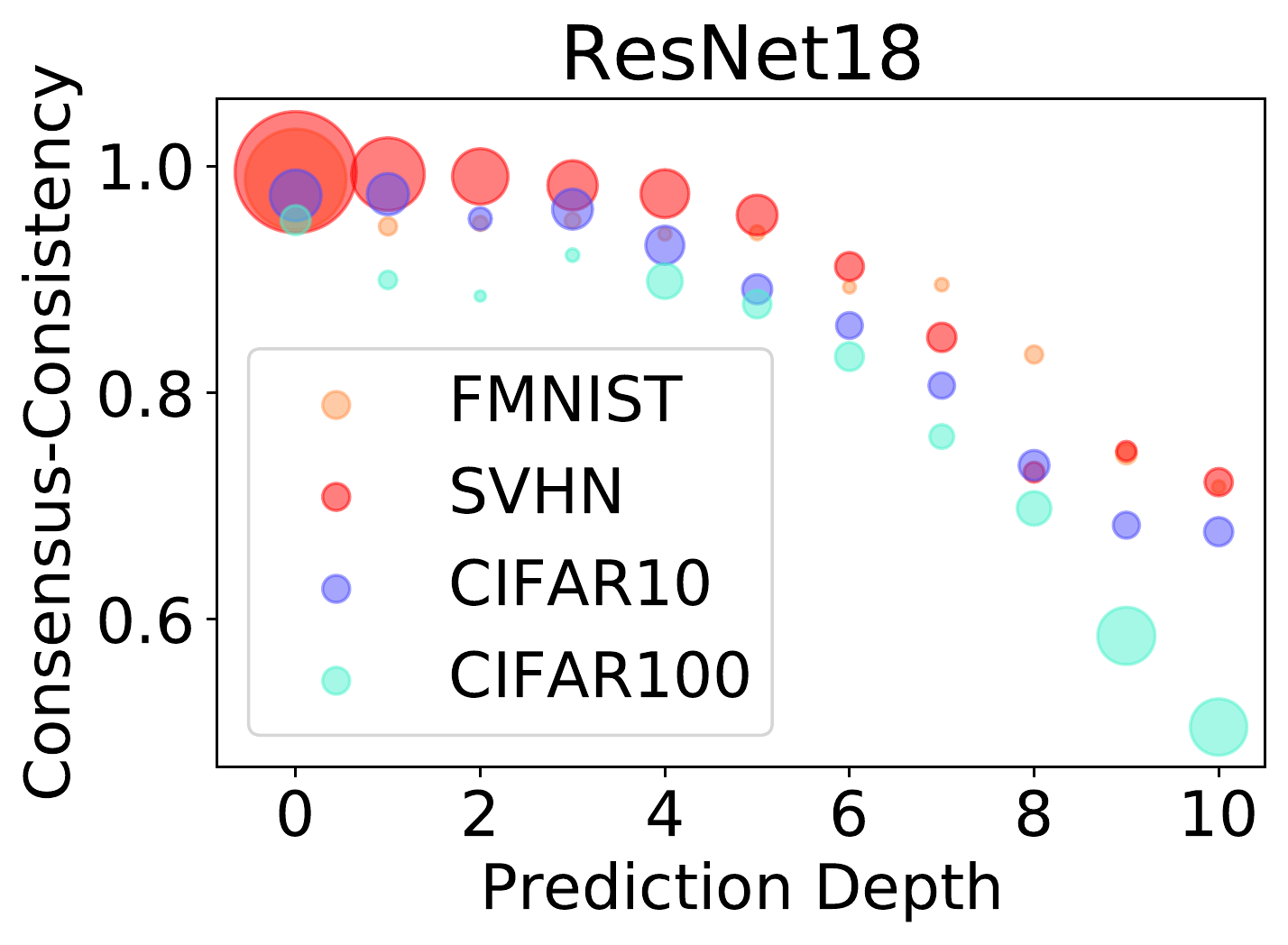}
\end{subfigure}
\begin{subfigure}
         \centering
         \includegraphics[width=0.32\columnwidth]{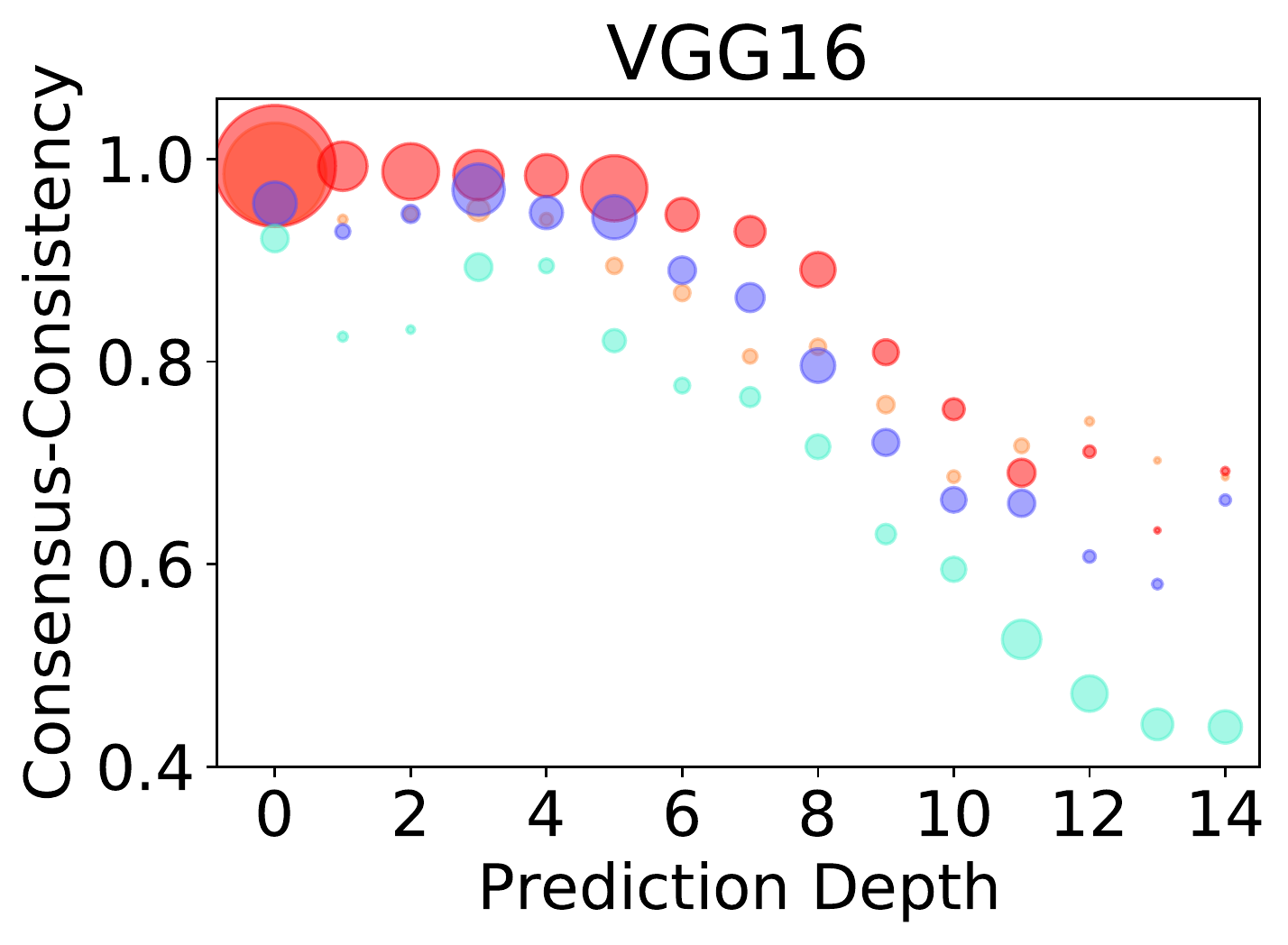}
\end{subfigure}
\begin{subfigure}
         \centering
         \includegraphics[width=0.32\columnwidth]{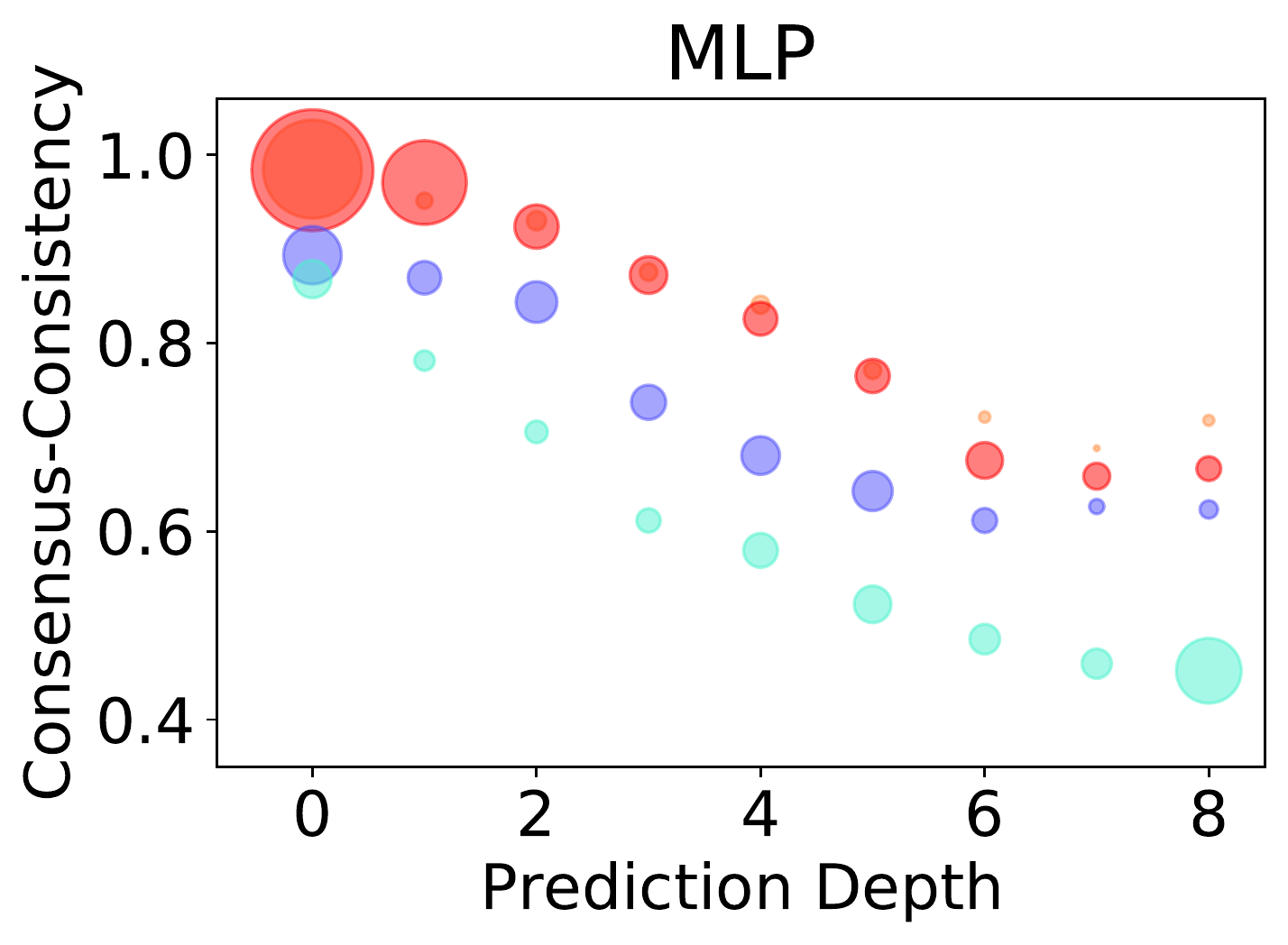}
\end{subfigure}
\end{center}
\caption{This figure demonstrates the consistency of the result shown in Figure~\ref{fig:ent_vs_depth} (middle) for all datasets and architectures.}
\label{fig:ll_vs_mean_cc_all}
\end{figure*}

\begin{figure}[ht]
\begin{center}
\begin{subfigure}
\centering
   \includegraphics[width=0.32\columnwidth]{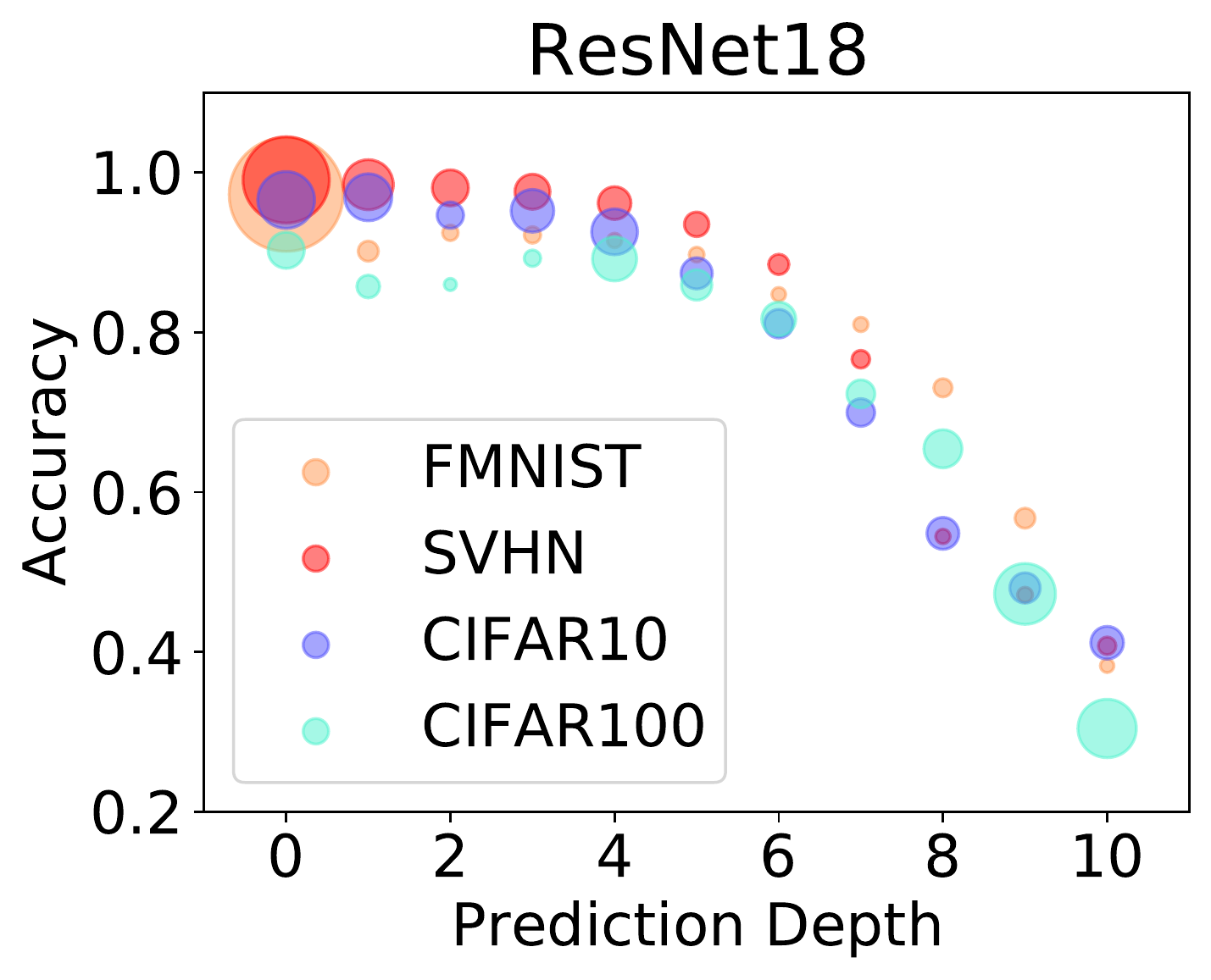}
\end{subfigure}
\begin{subfigure}
\centering
   \includegraphics[width=0.32\columnwidth]{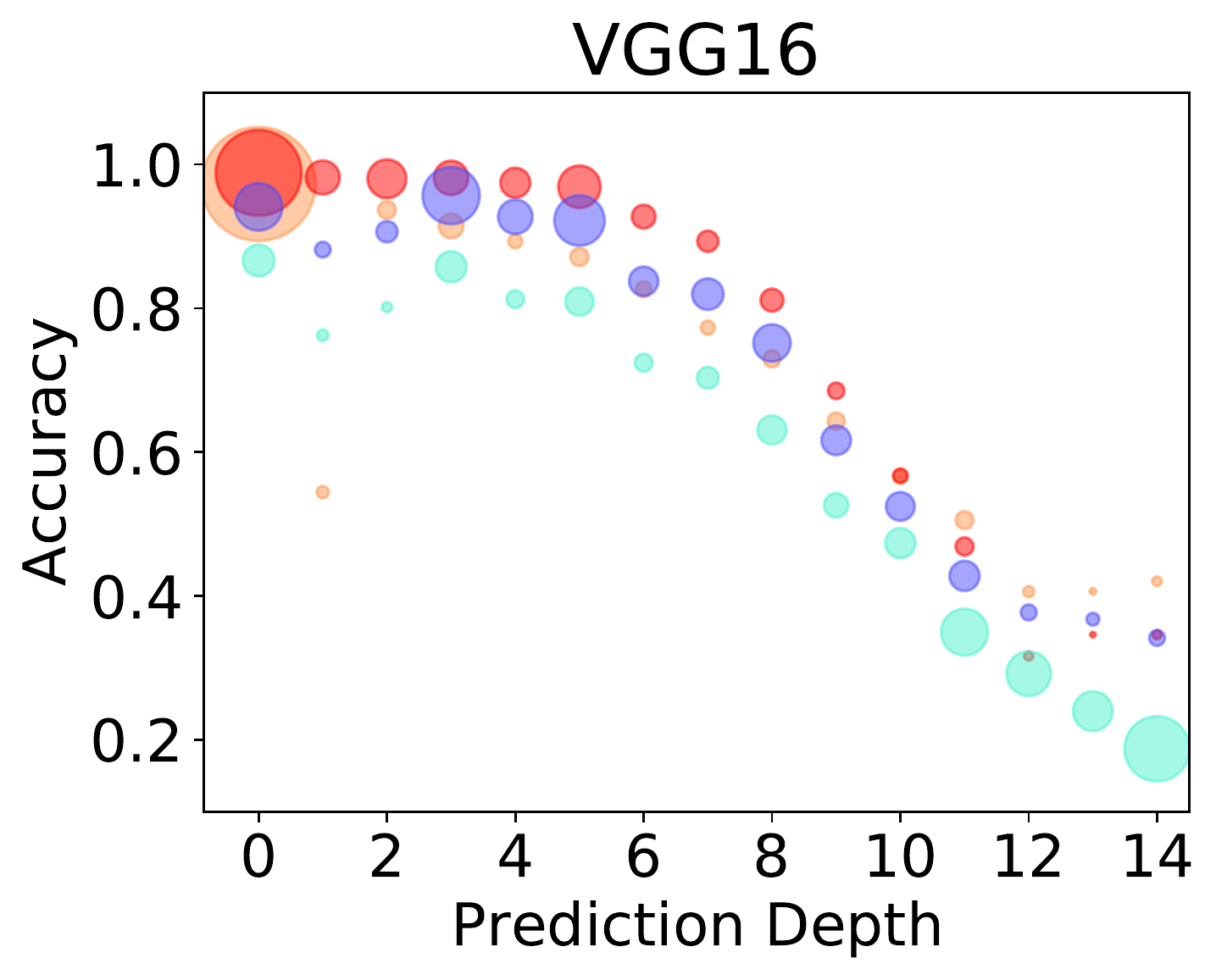}
\end{subfigure}
\begin{subfigure}
\centering
   \includegraphics[width=0.32\columnwidth]{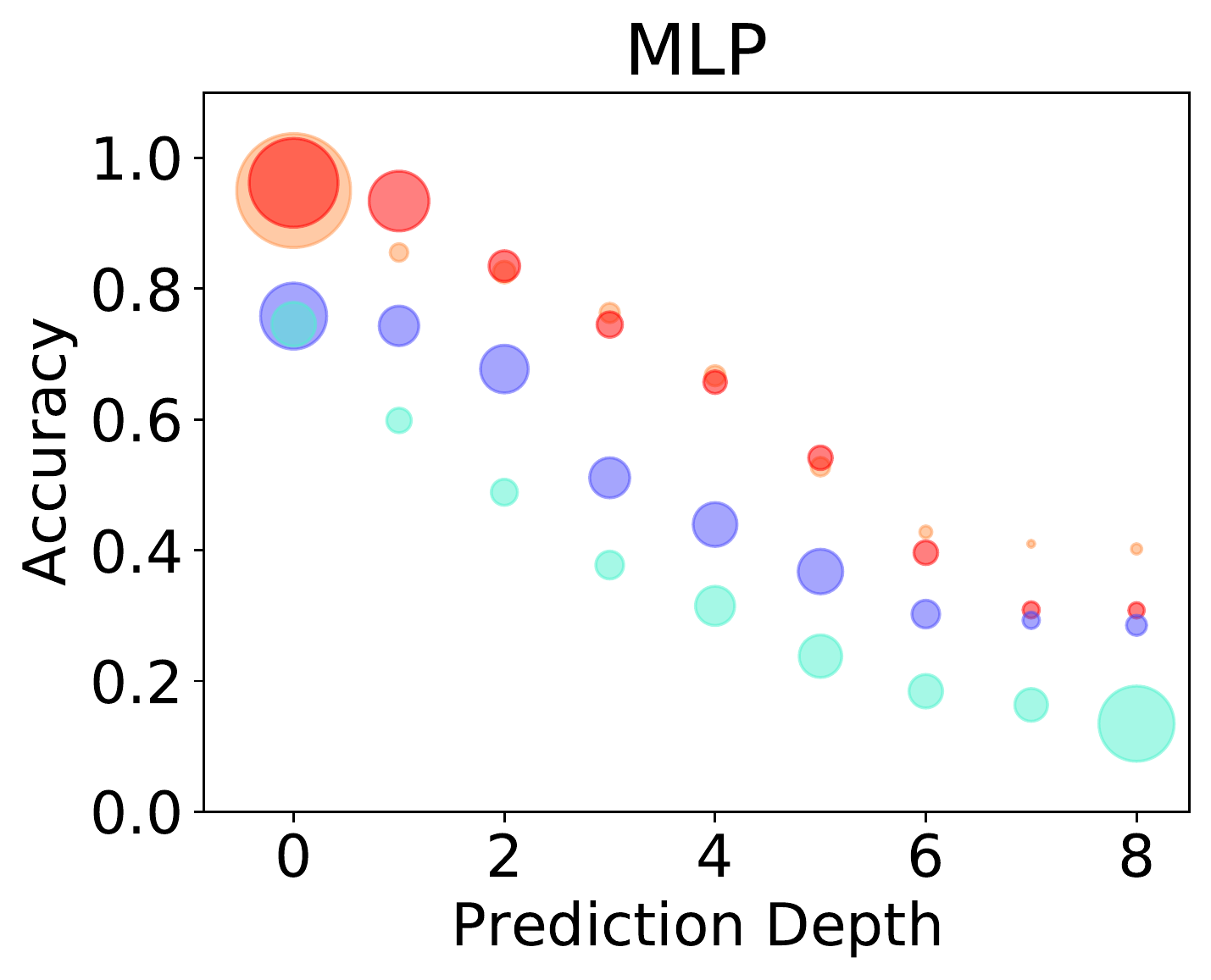}
\end{subfigure}
\end{center}
\caption{This figure demonstrates the consistency of the result shown in Figure~\ref{fig:ent_vs_depth} (right) for all datasets and architectures.
}
\label{fig:acc_vs_depth}
\end{figure}

\begin{figure*}[ht!]
\begin{center}
\begin{subfigure}
         \centering
         \includegraphics[width=0.32\columnwidth]{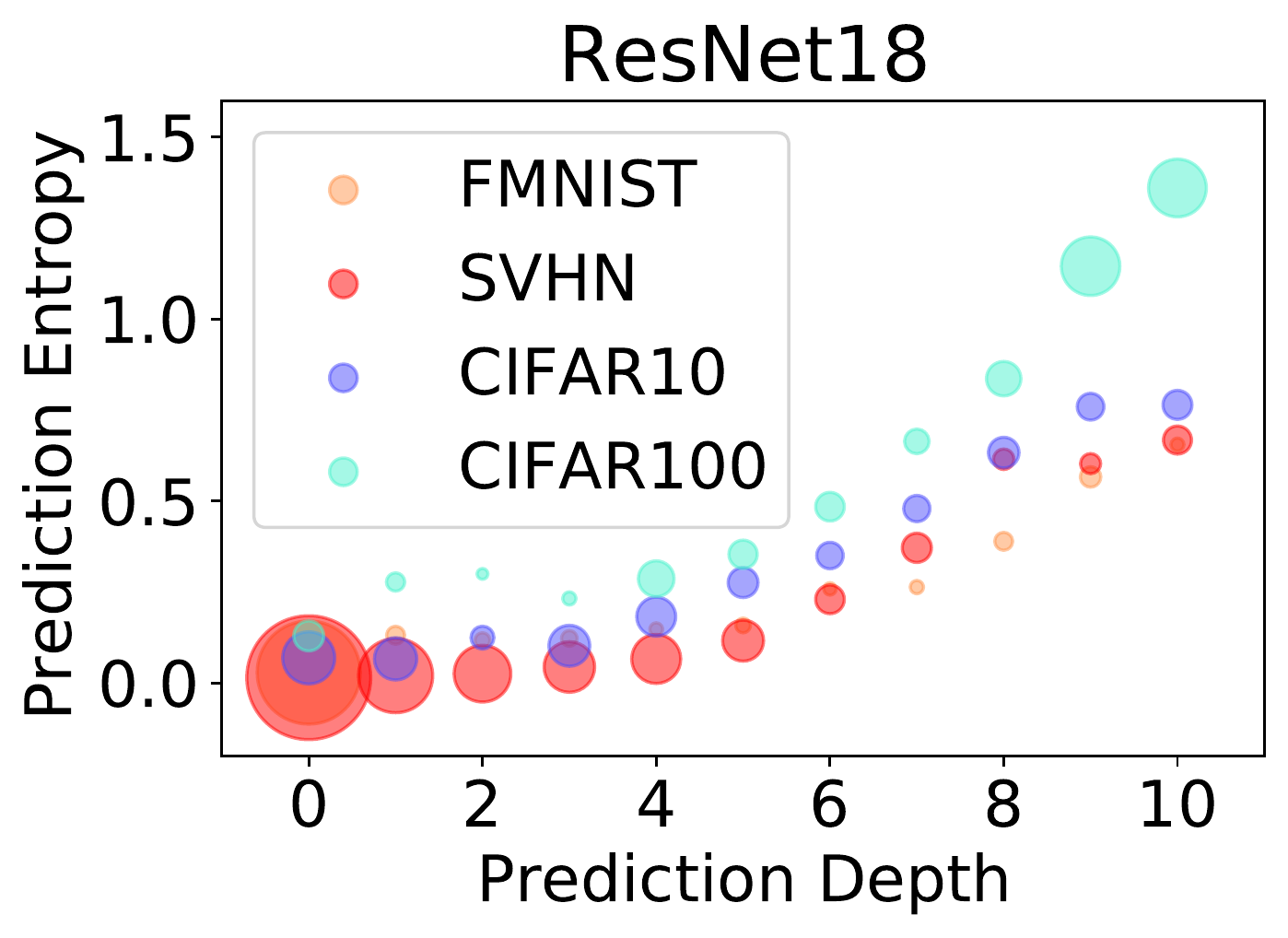}
\end{subfigure}
\begin{subfigure}
         \centering
         \includegraphics[width=0.32\columnwidth]{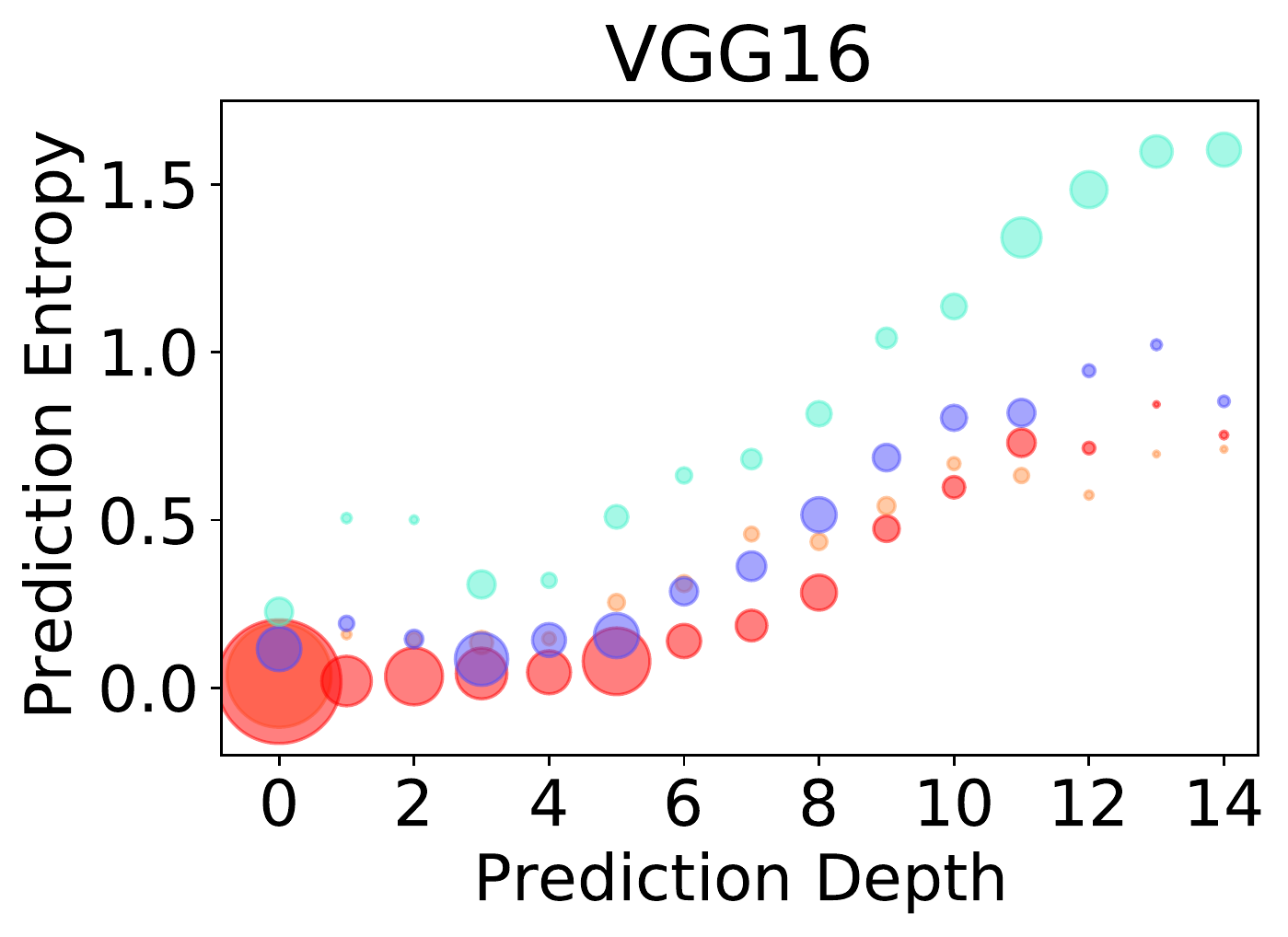}
\end{subfigure}
\begin{subfigure}
         \centering
         \includegraphics[width=0.32\columnwidth]{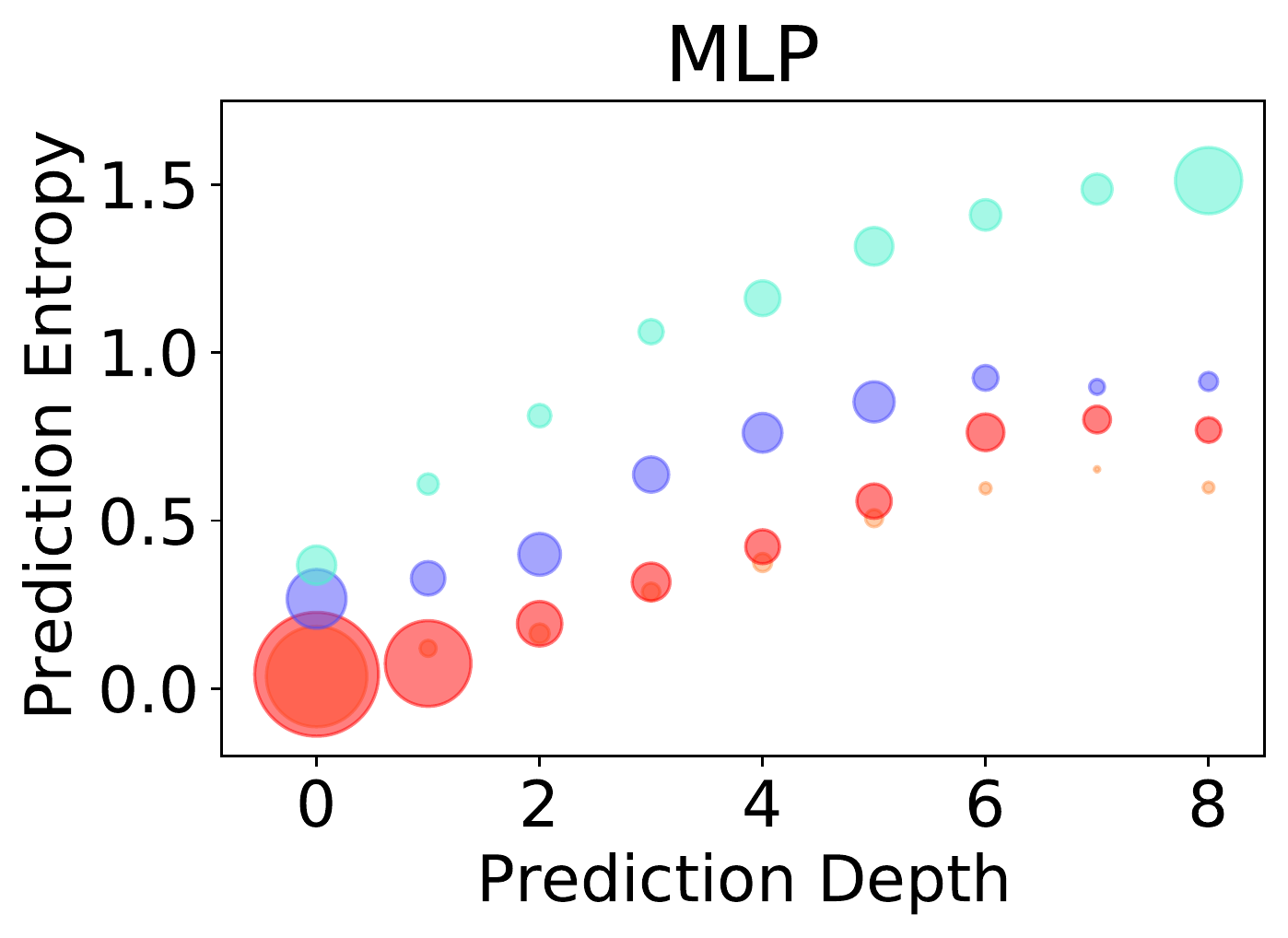}
\end{subfigure}
\end{center}
\caption{
\emph{ The prediction depth in one model can be used to estimate the prediction entropy of an ensemble.}
The size of the marker indicates the fraction of data points with each prediction depth.
We trained 25 models on each dataset and architecture with different random seeds. 
We take the prediction depth from one trained model and report the average prediction entropy of the corresponding data points, where the prediction entropy is determined from the remaining 24 models.
As in Figure~\ref{fig:ll_vs_mean_cc_all}, predictions for data points with smaller prediction depths have lower mean entropy (are more consistent) than those of data points with larger prediction depths.}
\label{fig:ll_vs_mean_ent_all}
\end{figure*}

Figure~\ref{fig:ll_vs_mean_cc_all} reproduces Figure~\ref{fig:ent_vs_depth} (middle) for every dataset and architecture. 
Similarly,
Figure~\ref{fig:acc_vs_depth} reproduces Figure~\ref{fig:ent_vs_depth} (right) for all datasets and architectures. 
Related to Figure~\ref{fig:ll_vs_mean_cc_all}, in Figure~\ref{fig:ll_vs_mean_ent_all} we show that the prediction depth in one model can be used to estimate the prediction entropy of an ensemble of models, where members of the ensemble have the same architecture and are trained using the same hyperparameters but with different random seeds.
\begin{description}
\item[Prediction entropy:] The entropy of predictions in an ensemble for an unseen input $x$.
Consider an ensemble of models trained on $r$ random subsets of the complete dataset $ \tilde{\mathcal{S}} {\sim} \mathcal{S}\backslash\{(x,y)\}$ (which explicitly do not include $(x, y)$).
We obtain the normalized histogram of the one-hot predictions of this ensemble for the input $x$.
The prediction entropy is the entropy of that histogram.
For $N$ classes the entropy of the prediction histogram is given by
\begin{equation}
S(x) = -\sum_{i=1}^N{p_i(x) \log{p_i(x)}} 
\label{eq:pred_entropy}
\end{equation}
where $p_i(x)$ represents the fraction of models that predicted the class $i$ for input $x$.
\end{description}

\begin{figure*}[ht!]
\begin{center}
\begin{subfigure}
         \centering
         \includegraphics[width=0.32\columnwidth]{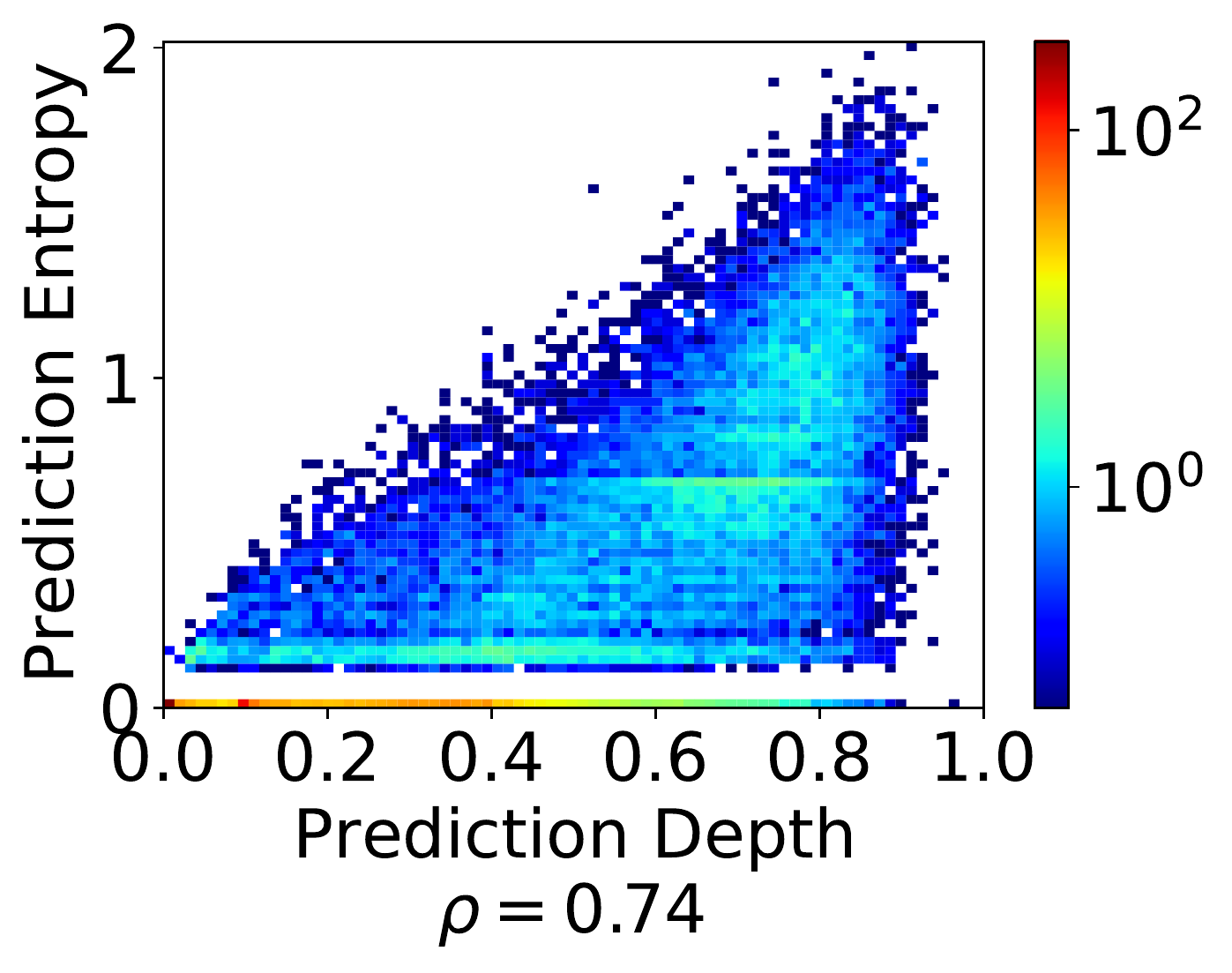}
\end{subfigure}
\hfill
\begin{subfigure}
         \centering
         \includegraphics[width=0.32\columnwidth]{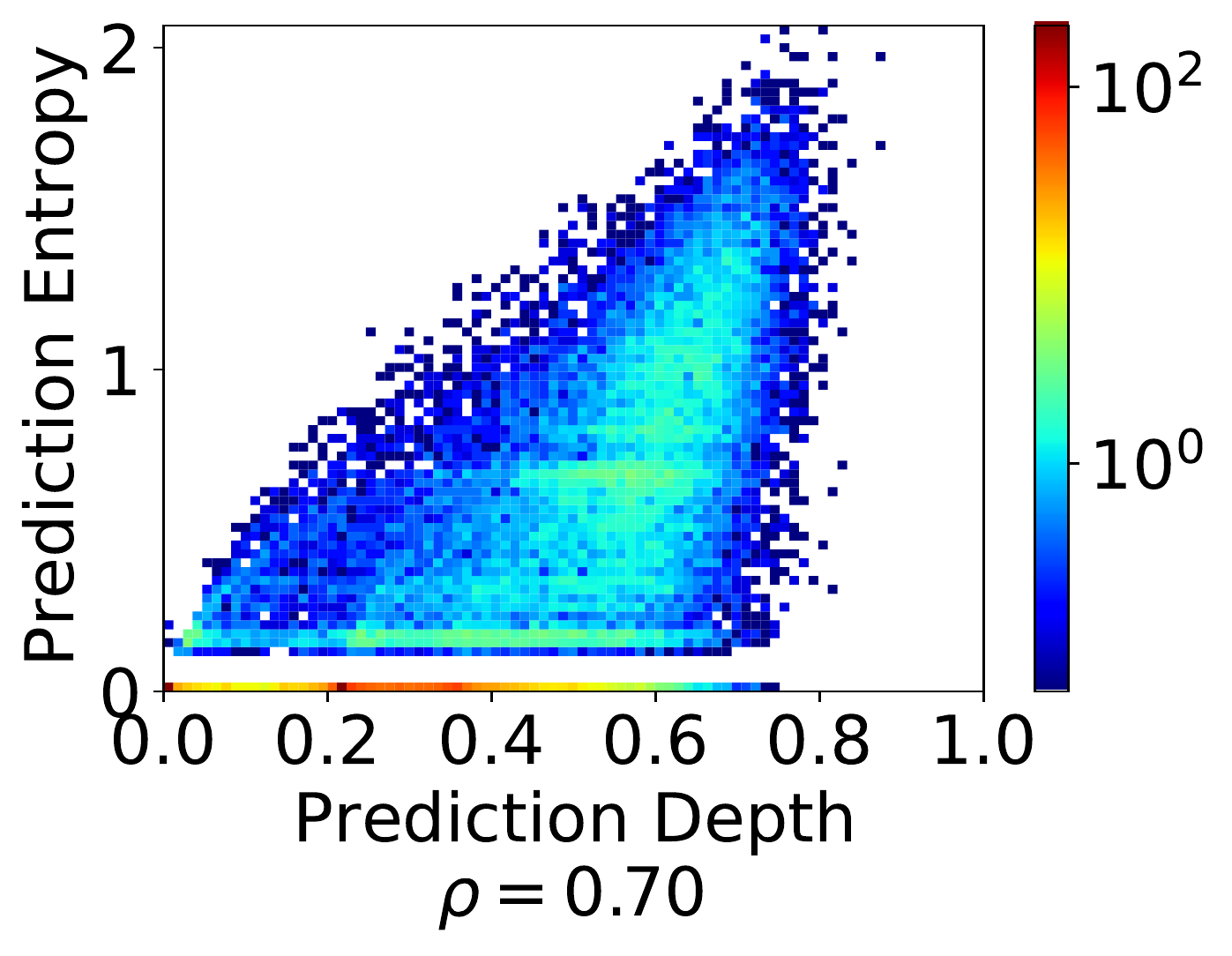}
\end{subfigure}
\hfill
\begin{subfigure}
         \centering
         \includegraphics[width=0.32\columnwidth]{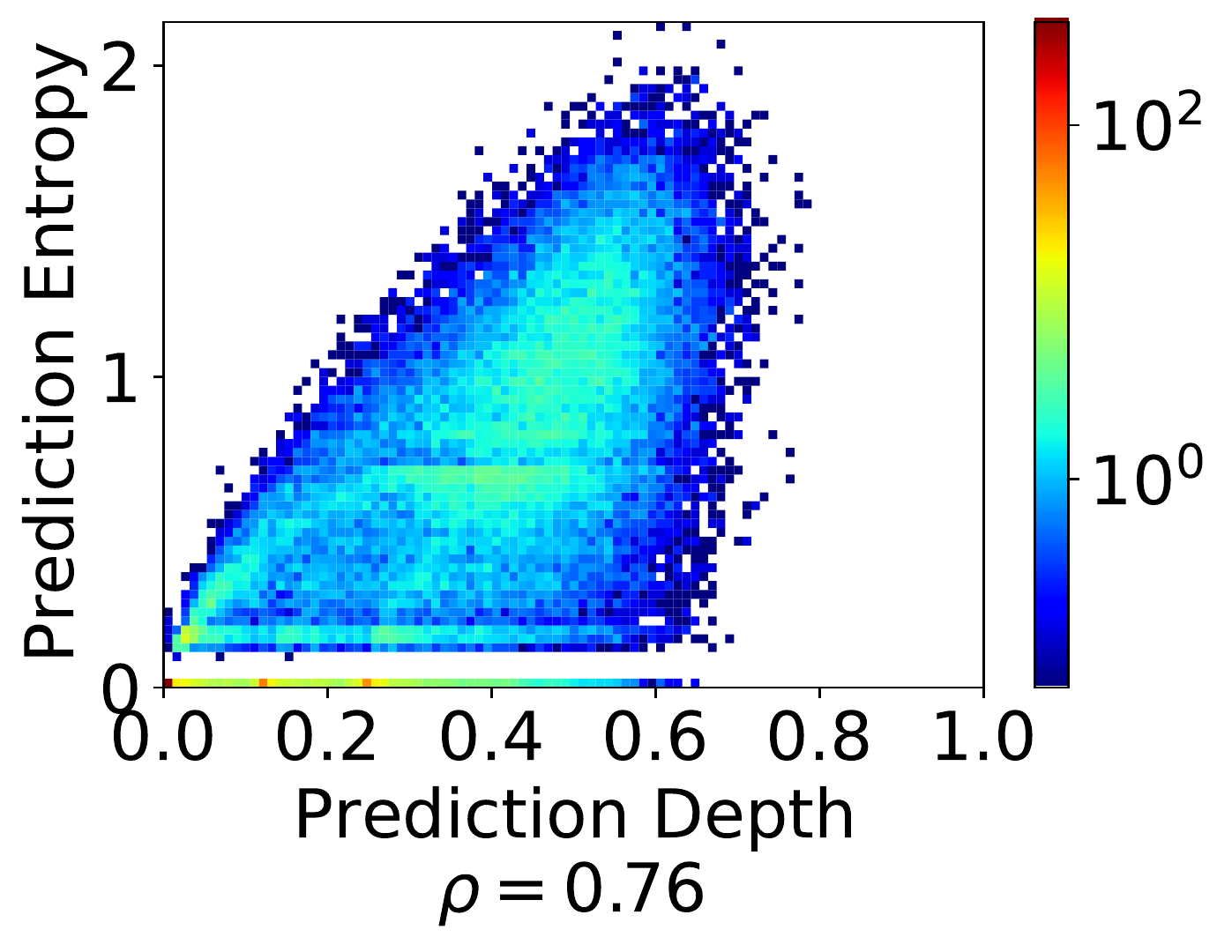}
\end{subfigure}
\hfill
\begin{subfigure}
         \centering
         \includegraphics[width=0.32\columnwidth]{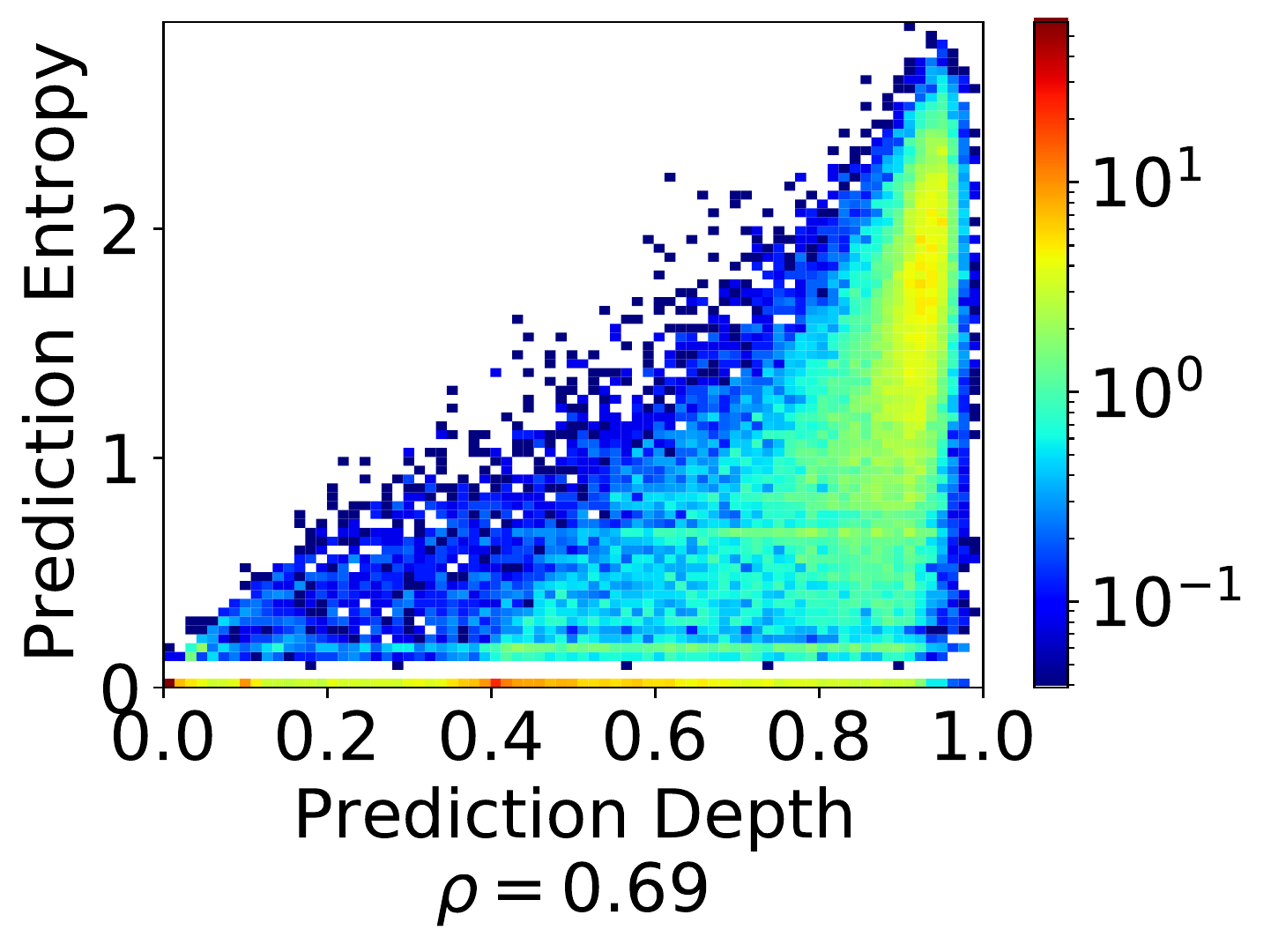}
\end{subfigure}
\begin{subfigure}
         \centering
         \includegraphics[width=0.32\columnwidth]{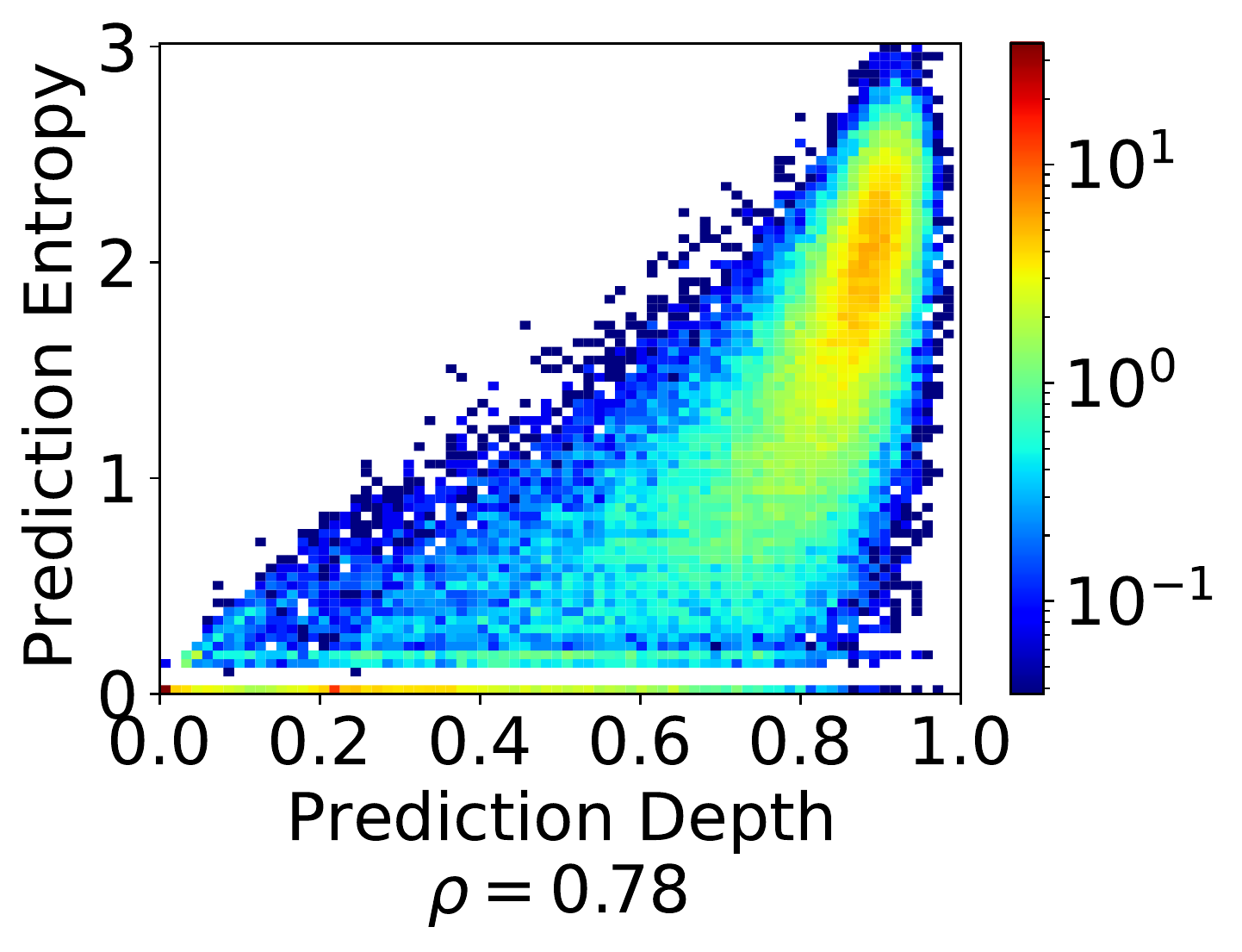}
\end{subfigure}
\begin{subfigure}
         \centering
         \includegraphics[width=0.32\columnwidth]{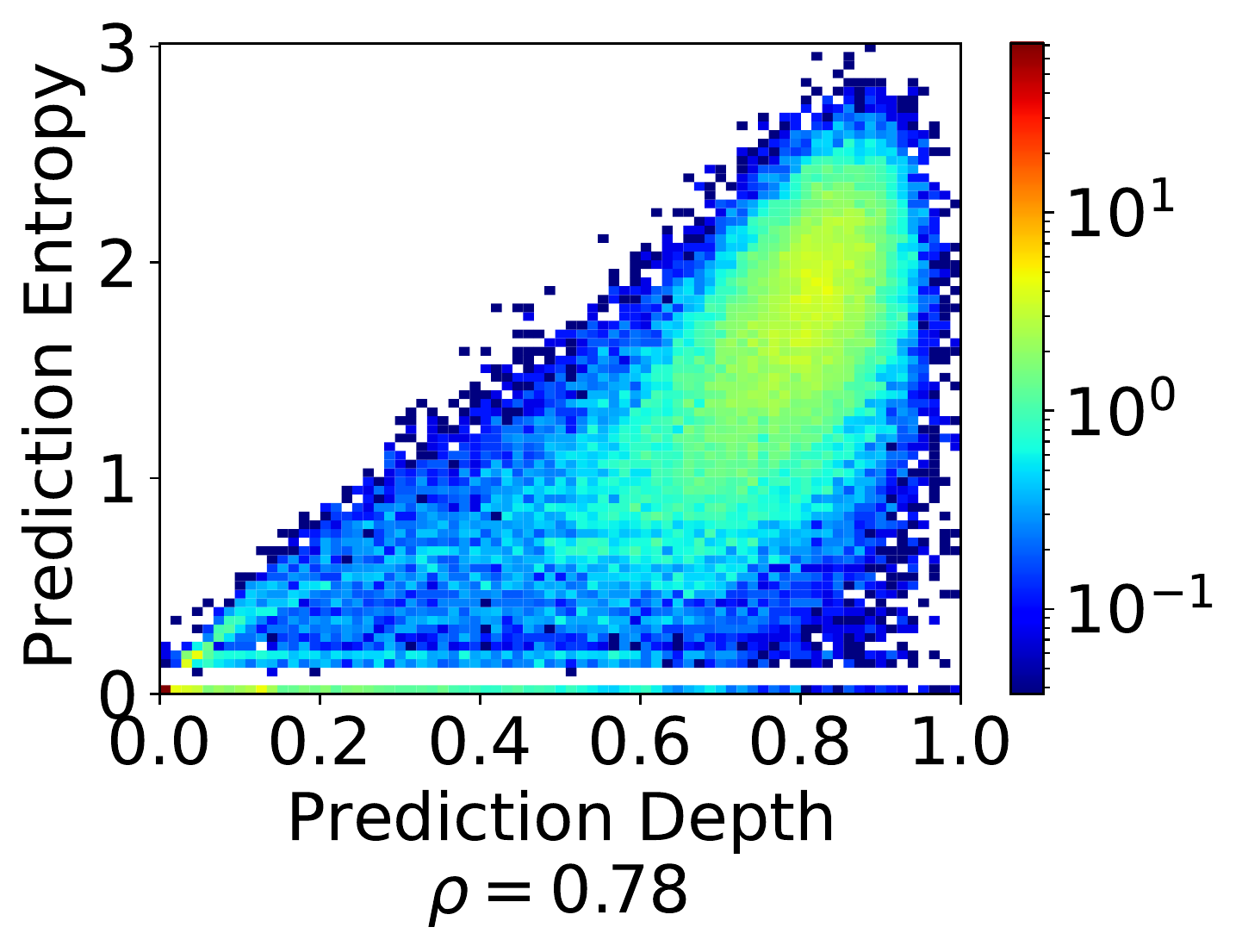}
\end{subfigure}
\begin{subfigure}
         \centering
         \includegraphics[width=0.32\columnwidth]{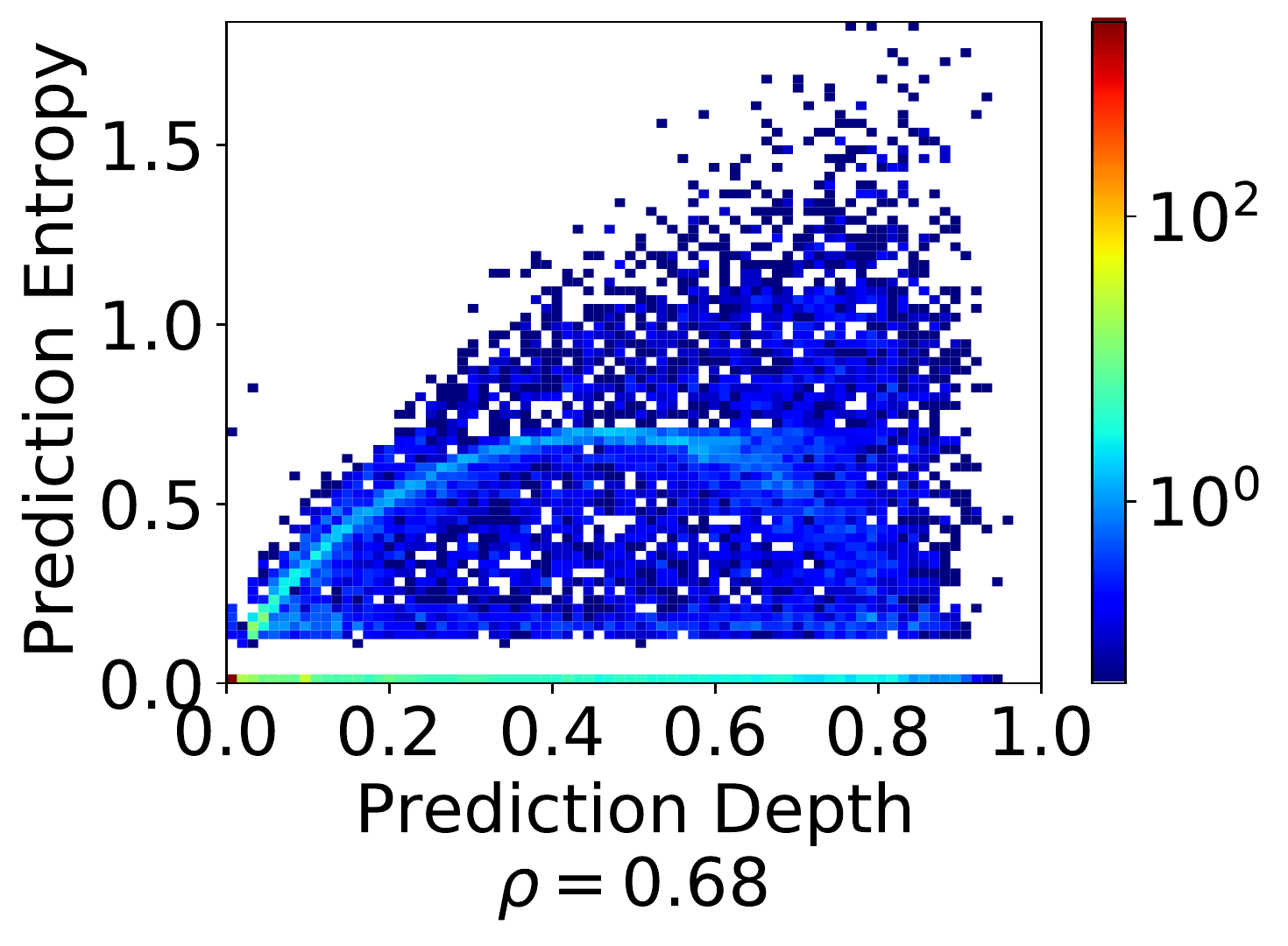}
\end{subfigure}
\begin{subfigure}
         \centering
         \includegraphics[width=0.32\columnwidth]{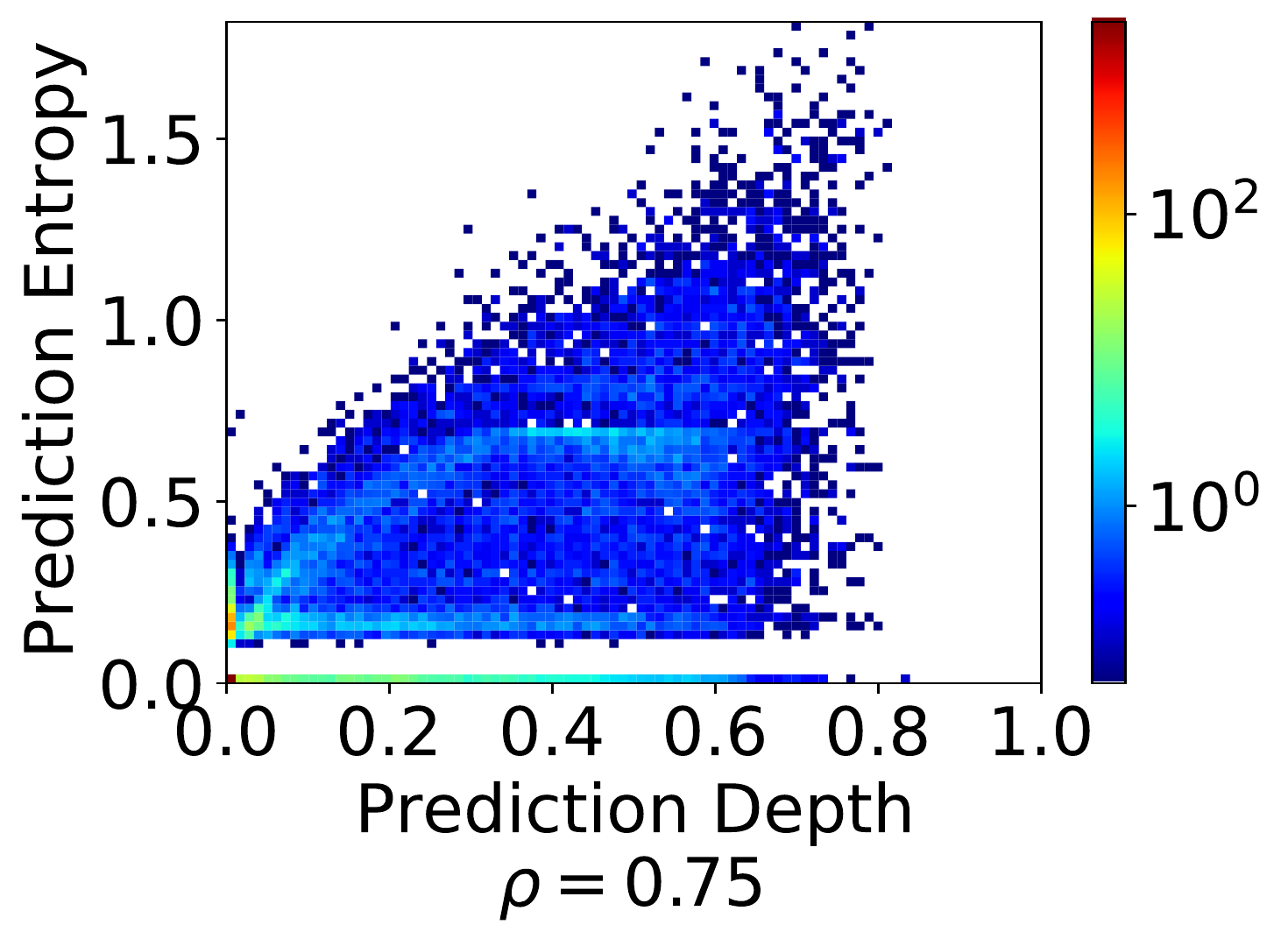}
\end{subfigure}
\begin{subfigure}
         \centering
         \includegraphics[width=0.32\columnwidth]{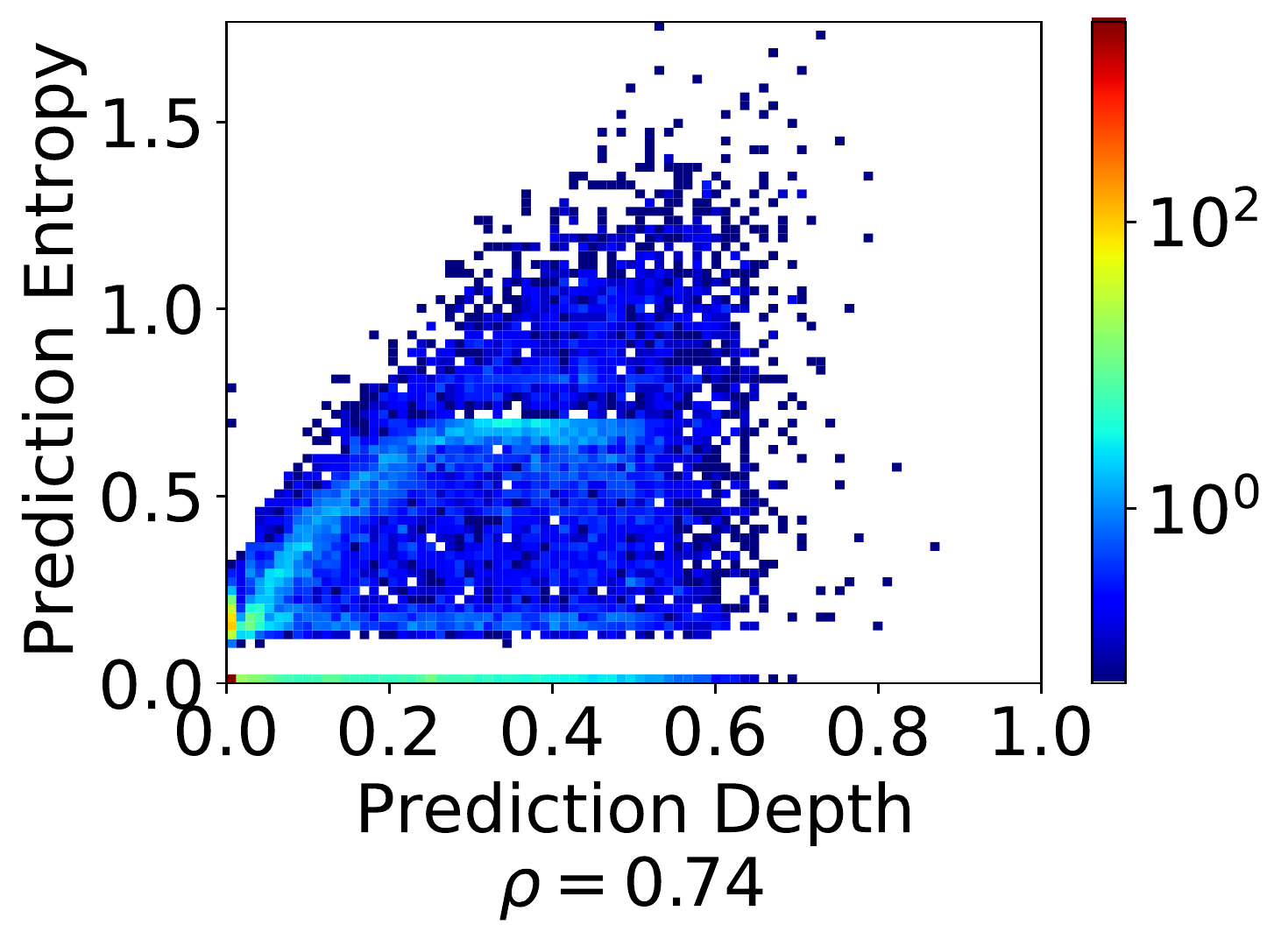}
\end{subfigure}
\begin{subfigure}
         \centering
         \includegraphics[width=0.32\columnwidth]{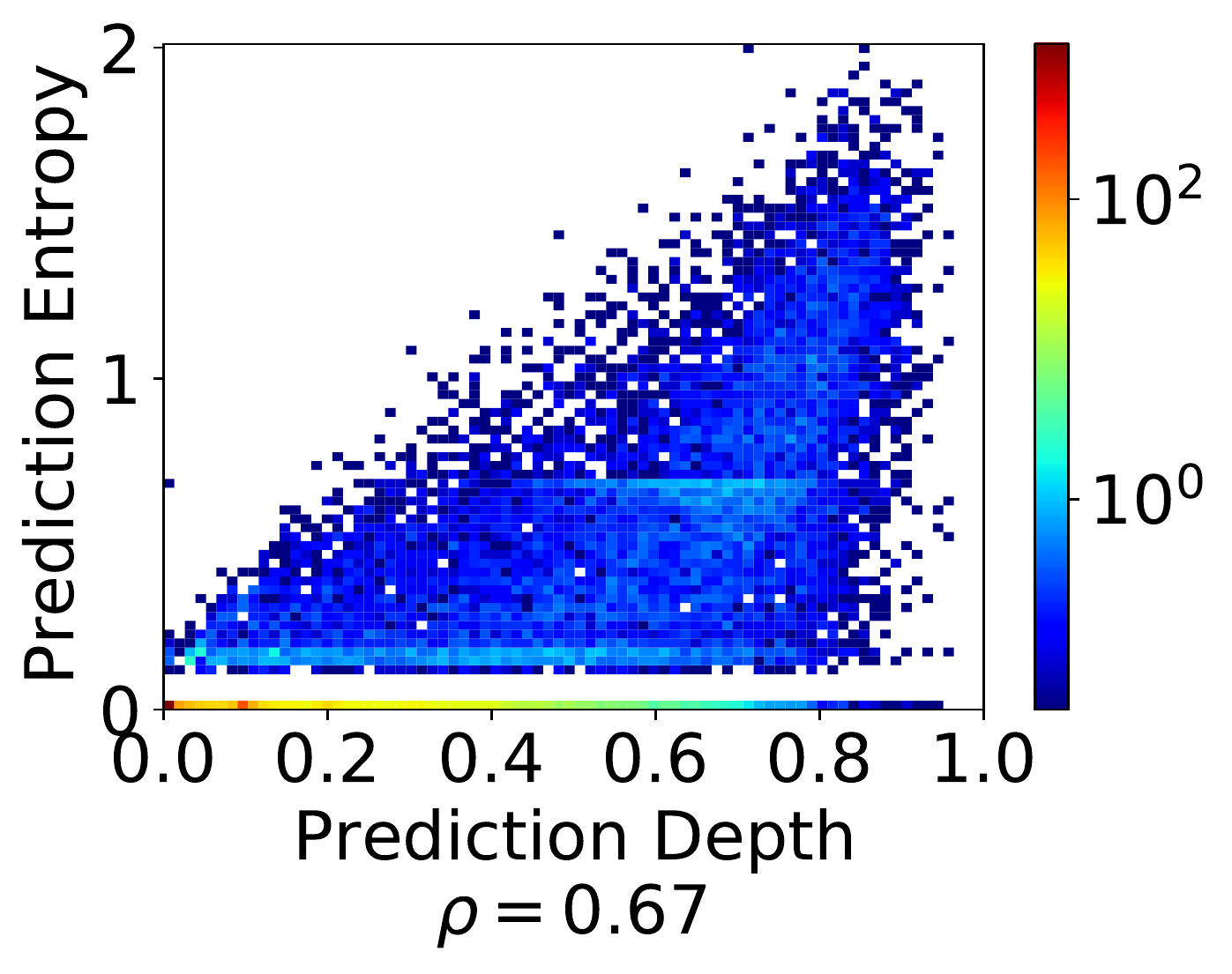}
\end{subfigure}
\begin{subfigure}
         \centering
         \includegraphics[width=0.32\columnwidth]{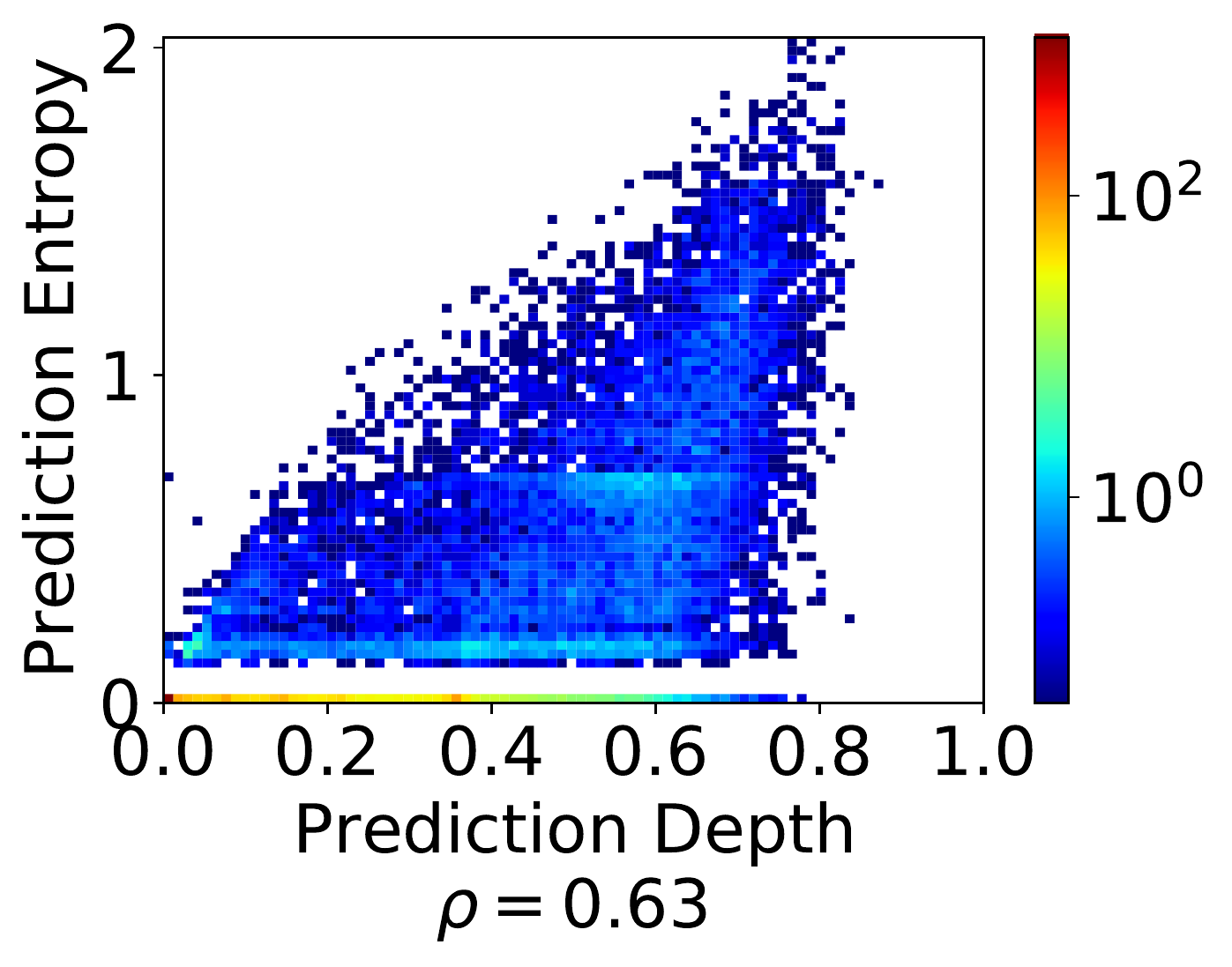}
\end{subfigure}
\begin{subfigure}
         \centering
         \includegraphics[width=0.32\columnwidth]{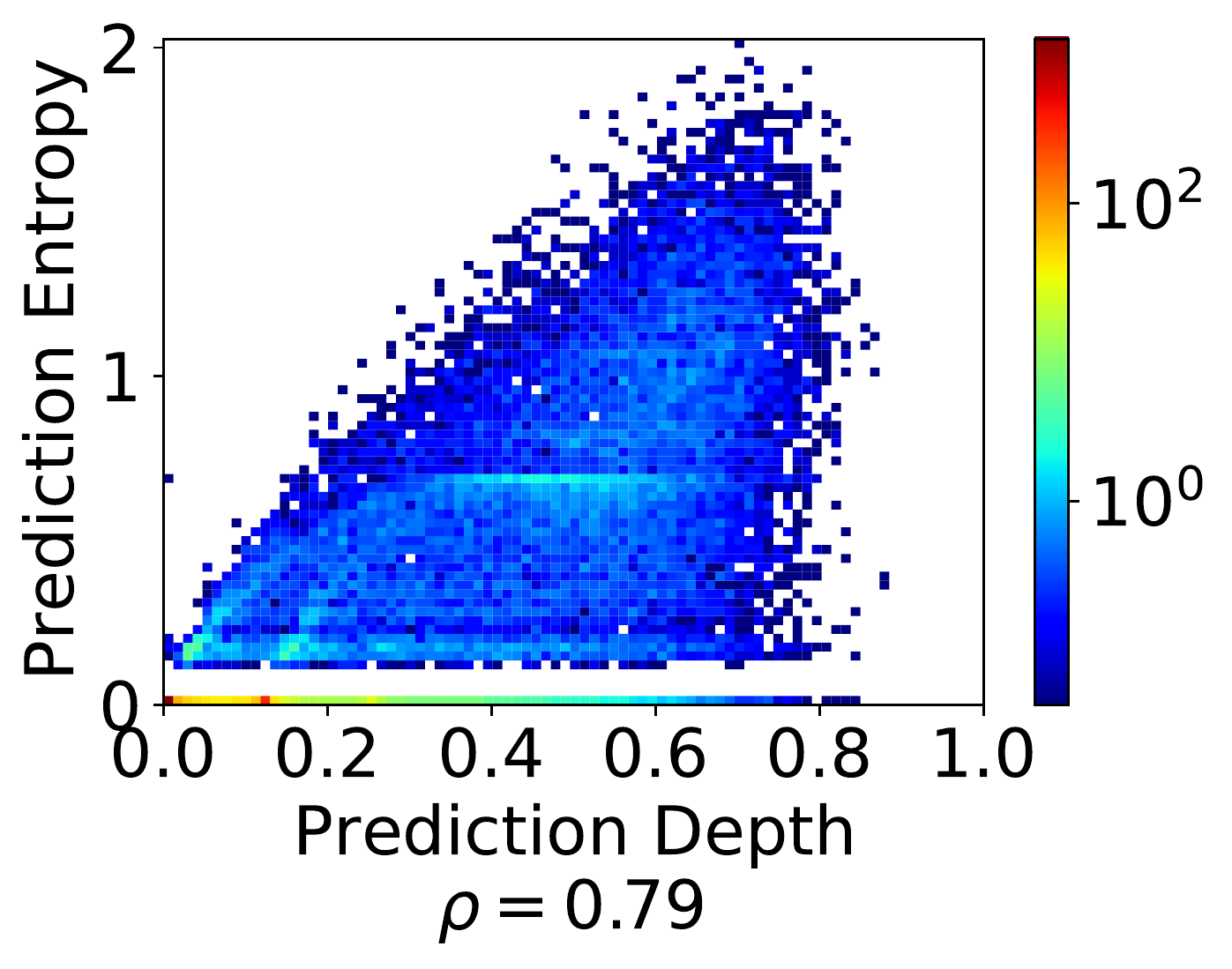}
\end{subfigure}
\end{center}
\caption{First (Top) Row: CIFAR10. Second Row: CIFAR100. Third Row: Fashion MNIST. Fourth (Bottom) Row: SVHN. Left Column: ResNet18. Middle Column: VGG16. Right Column: MLP.
Histograms showing consistency of the relationship between prediction depth in the validation set and prediction entropy of an ensemble. 
As described in Appendix~\ref{app:experiments_desc}, for each dataset and architecture we trained 250 models with random 90:10\% validation:train splits. Each time a data point appears in the validation split we record the prediction depth and the prediction.
These histograms compare the average prediction depth for each data point to its prediction entropy.
We observe that the prediction depth gives linear upper bounds for the prediction entropy as it does linear lower bounds for the consensus-consistency (Figures~\ref{fig:ll_vs_pcpm1} and~\ref{fig:ll_vs_pcpm2}).
\label{fig:ll_vs_ent_all}
}
\end{figure*}

Figure~\ref{fig:ll_vs_ent_all} shows the histogram of average prediction depth (validation set) vs. prediction entropy for each dataset and architecture.
We remark that the mean prediction depth defines a linear upper bound on the prediction entropy similar to the corresponding linear lower bound on the consensus-consistency score (Figures~\ref{fig:ll_vs_pcpm1} and~\ref{fig:ll_vs_pcpm2}).

\subsection{Comparison of prediction depth and iteration learned \label{app:ll_itl}}

\begin{figure*}[t]
\begin{center}
\begin{subfigure}
         \centering
         \includegraphics[width=0.32\columnwidth]{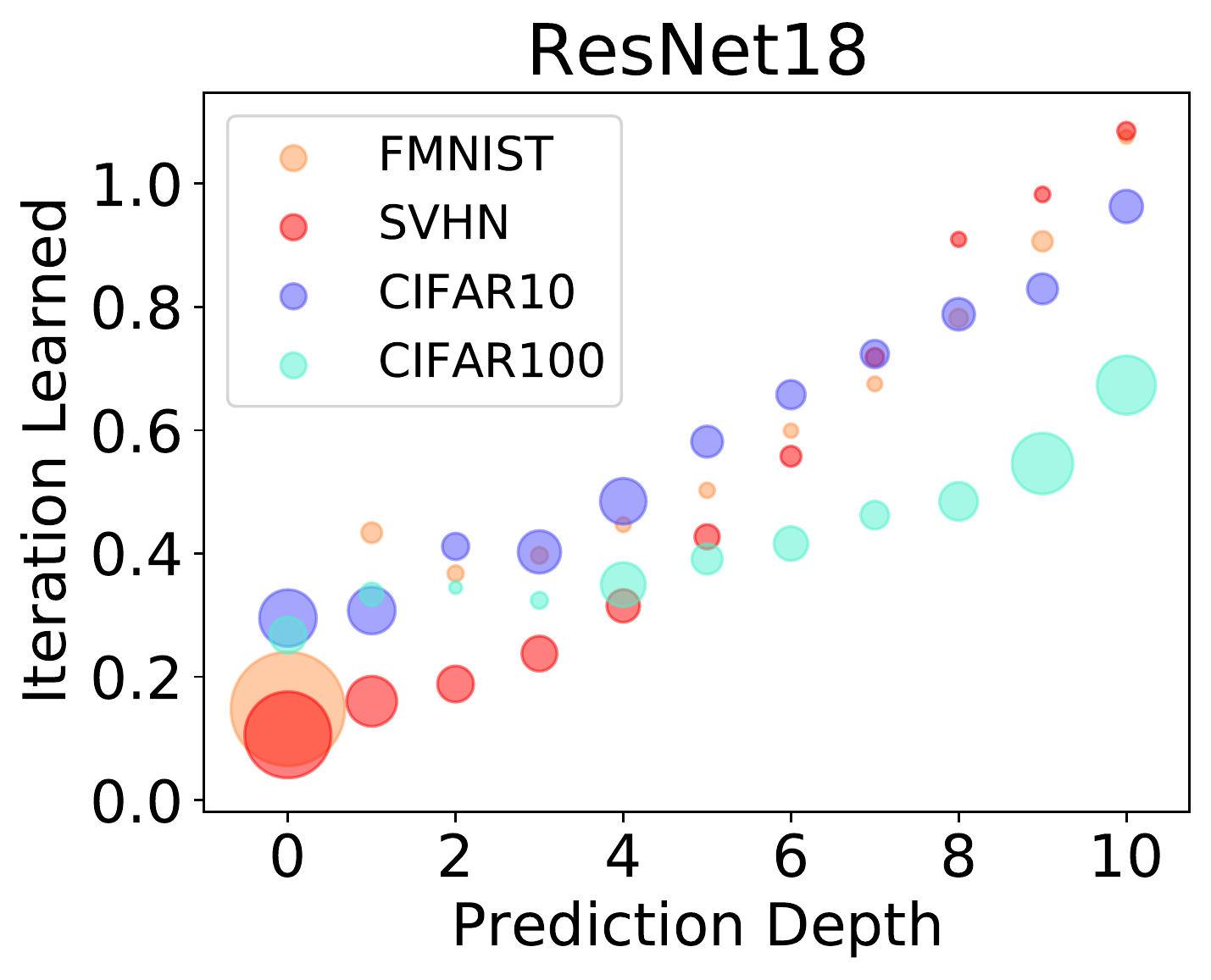}
\end{subfigure}
\begin{subfigure}
         \centering
         \includegraphics[width=0.32\columnwidth]{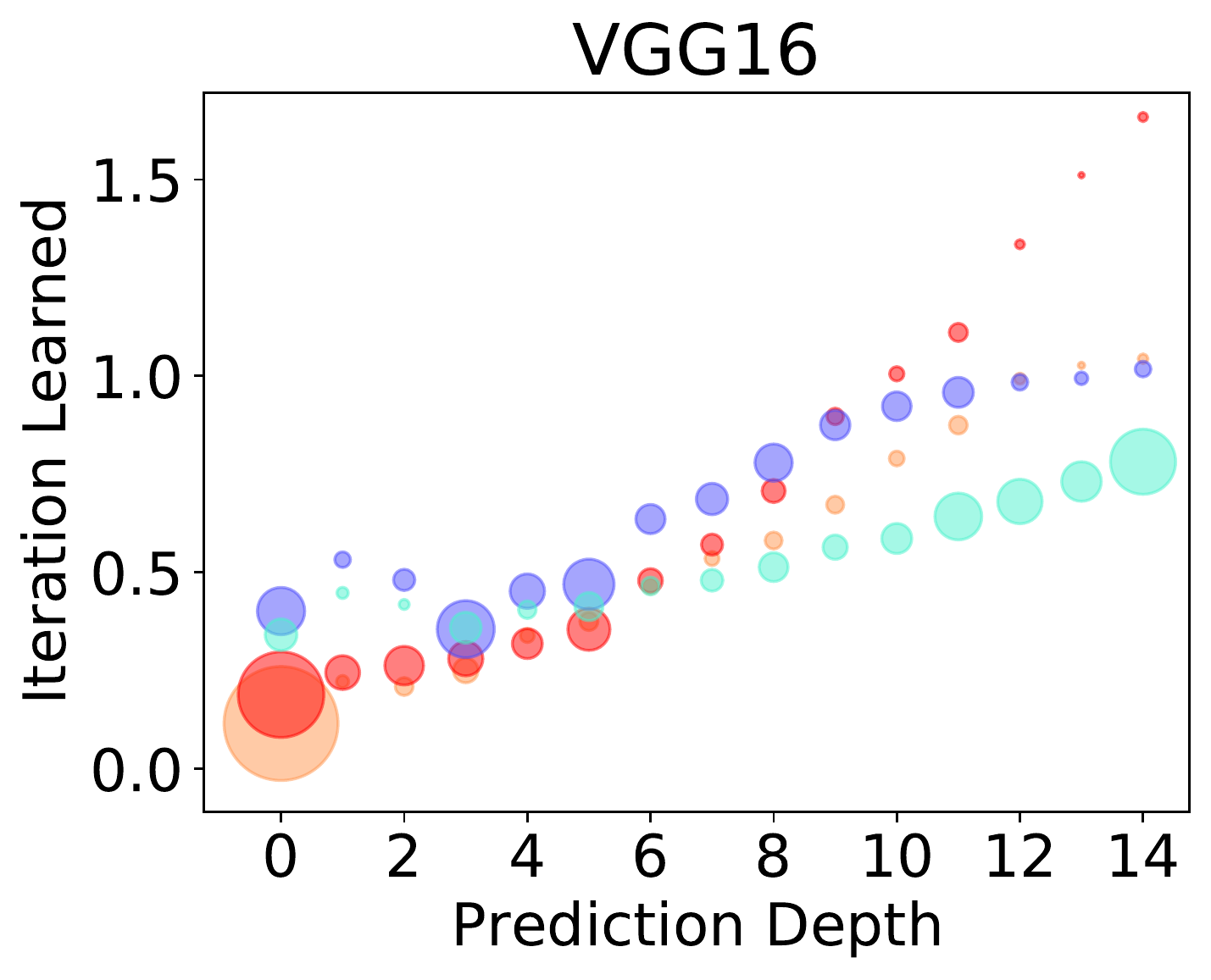}
\end{subfigure}
\begin{subfigure}
         \centering
         \includegraphics[width=0.32\columnwidth]{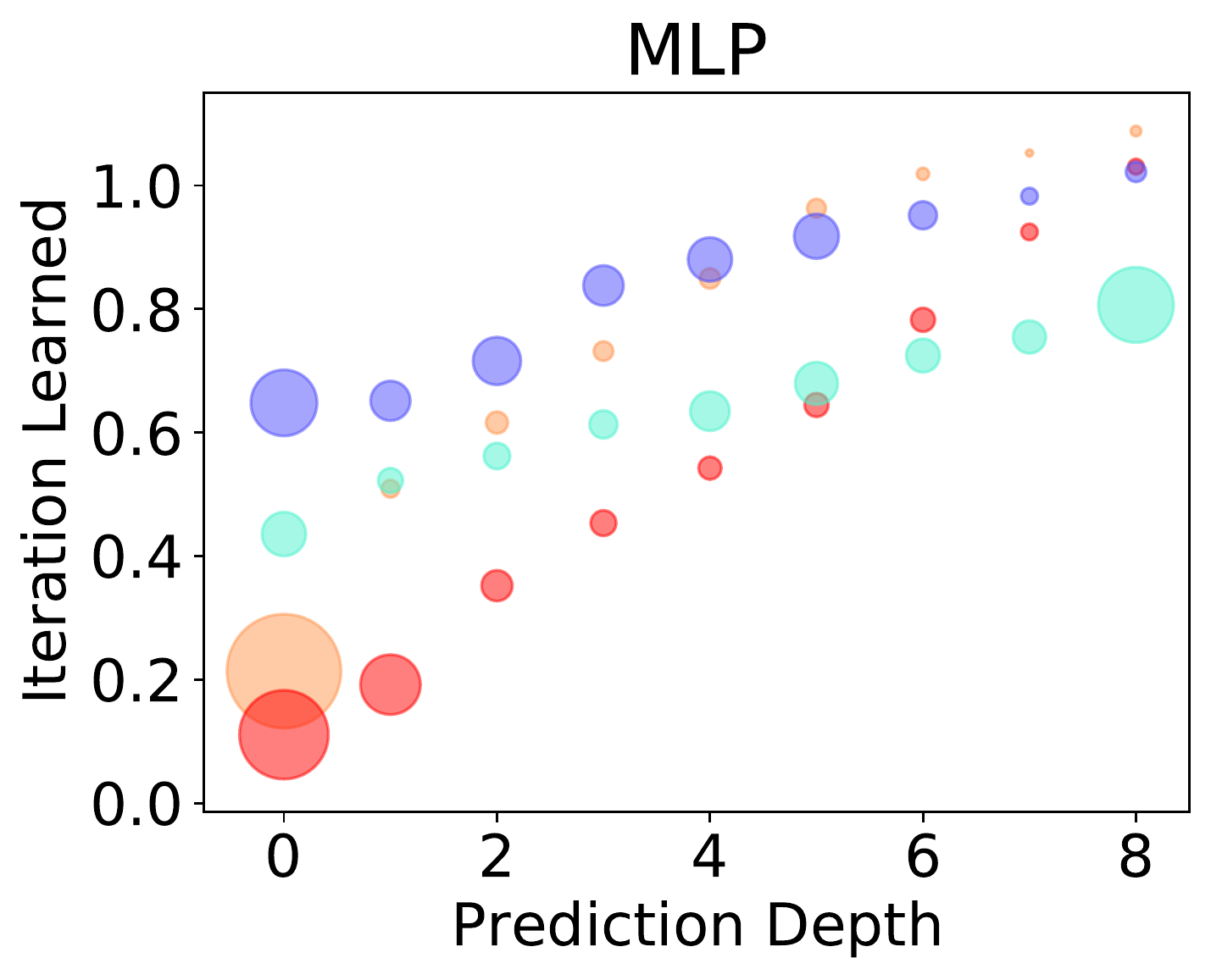}
\end{subfigure}
\end{center}
\caption{This figure demonstrates the consistency of the result shown in Figure~\ref{fig:visual_correspondance} (left) for all datasets and architectures.}
\label{fig:ll_vs_iter_all}
\end{figure*}

\begin{figure}[ht!]
\begin{center}
\begin{subfigure}
         \centering
         \includegraphics[width=0.49\columnwidth]{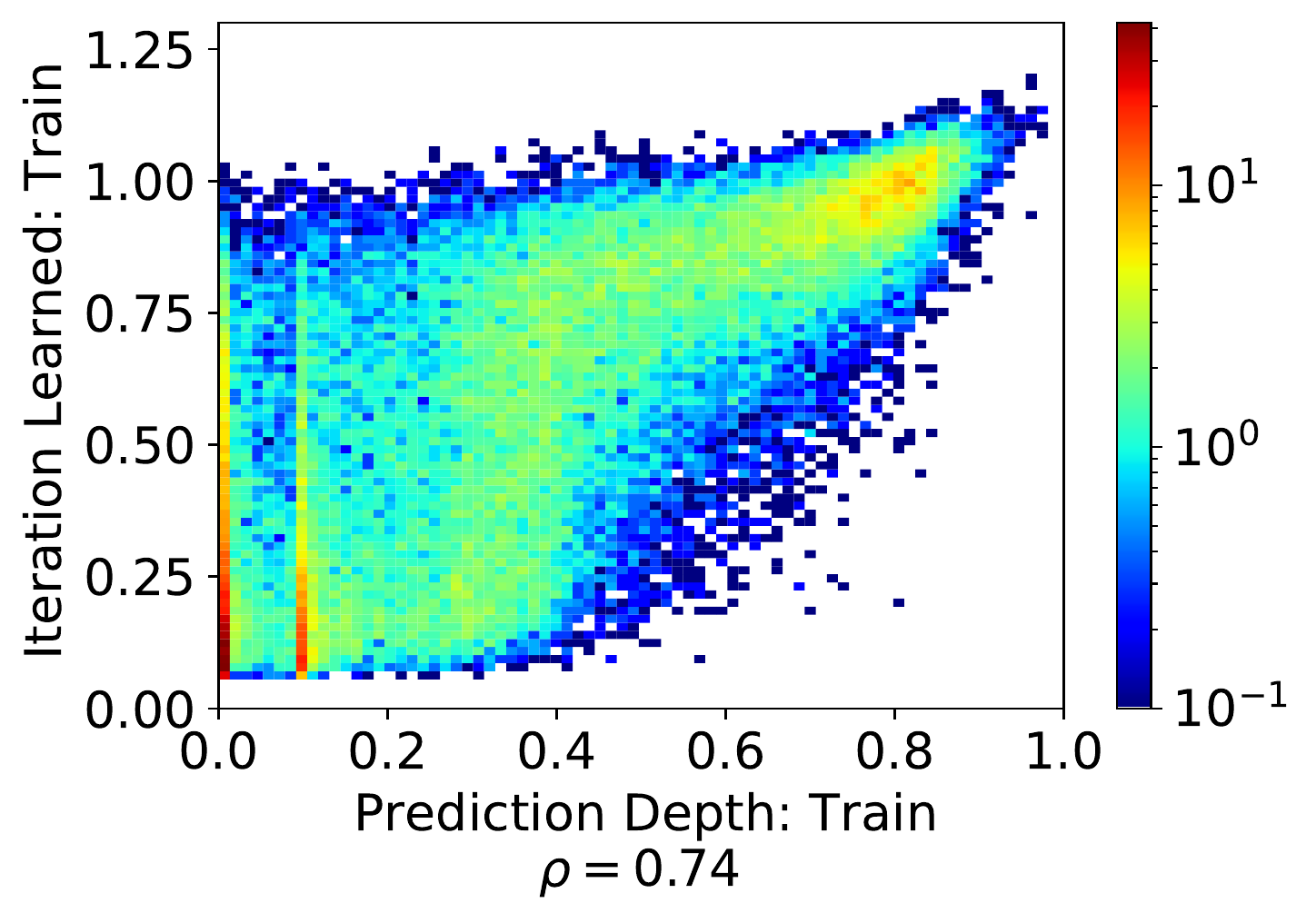}
\end{subfigure}
\begin{subfigure}
         \centering
         \includegraphics[width=0.49\columnwidth]{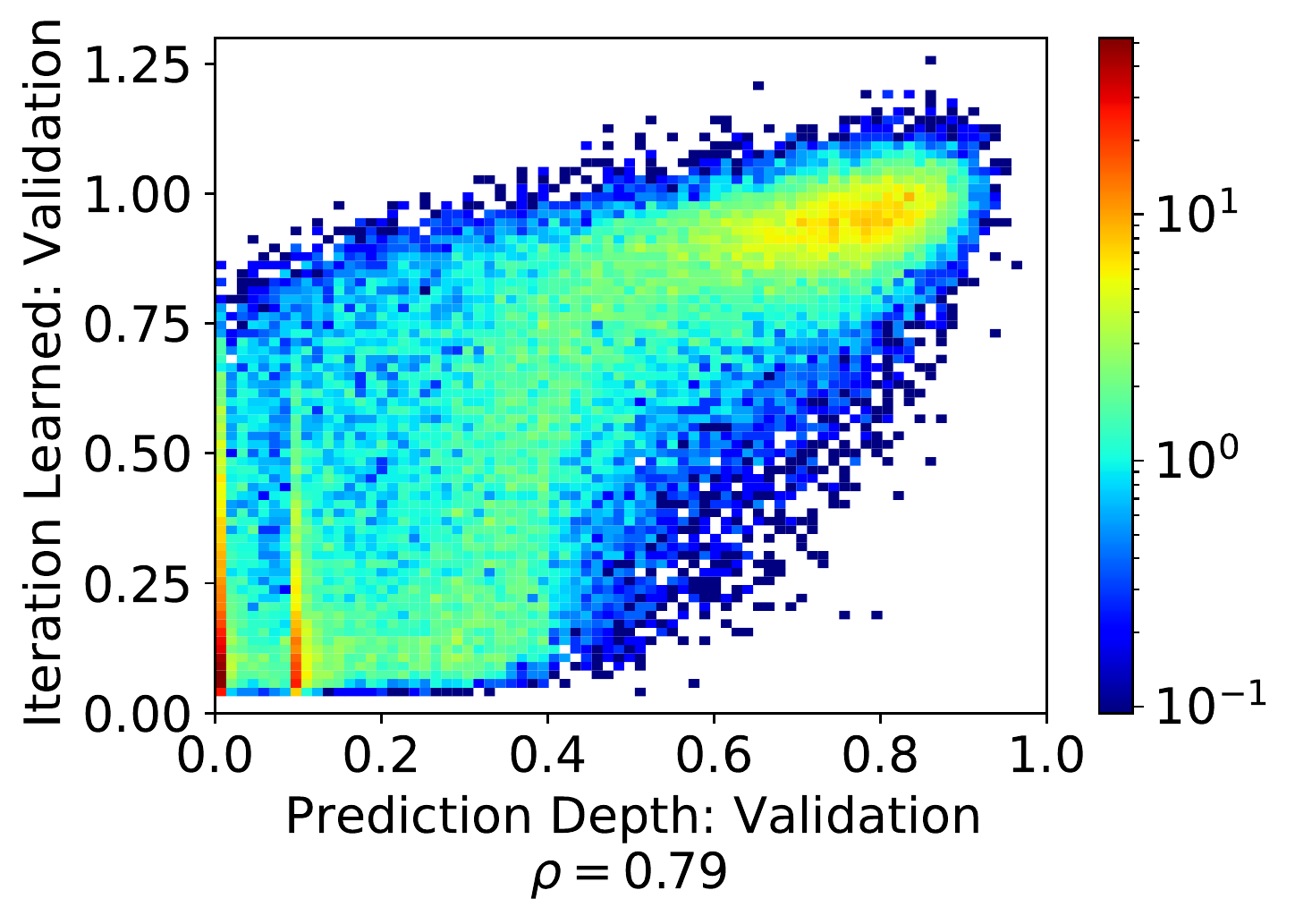}
\end{subfigure}
\begin{subfigure}
         \centering
         \includegraphics[width=0.49\columnwidth]{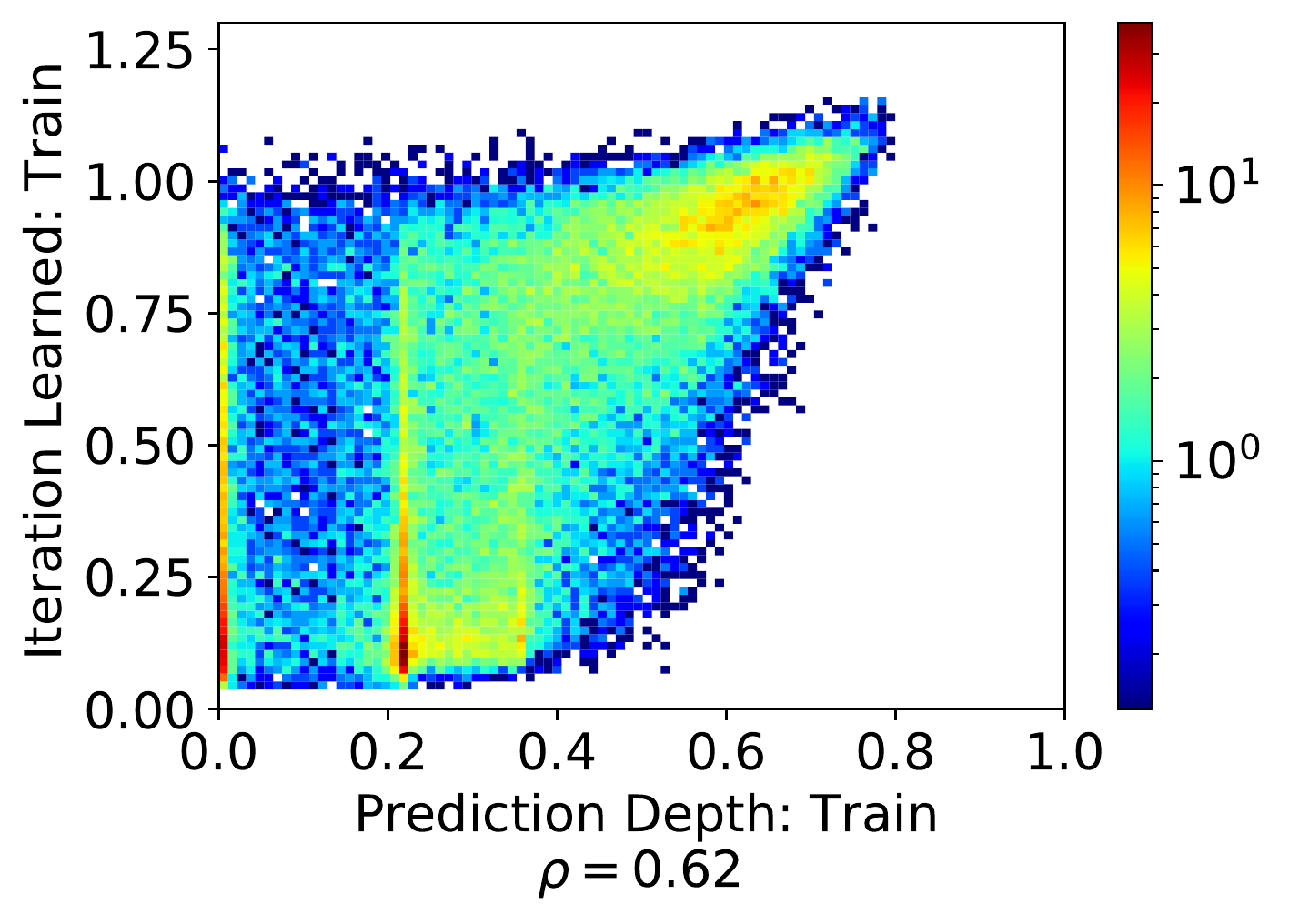}
\end{subfigure}
\begin{subfigure}
         \centering
         \includegraphics[width=0.49\columnwidth]{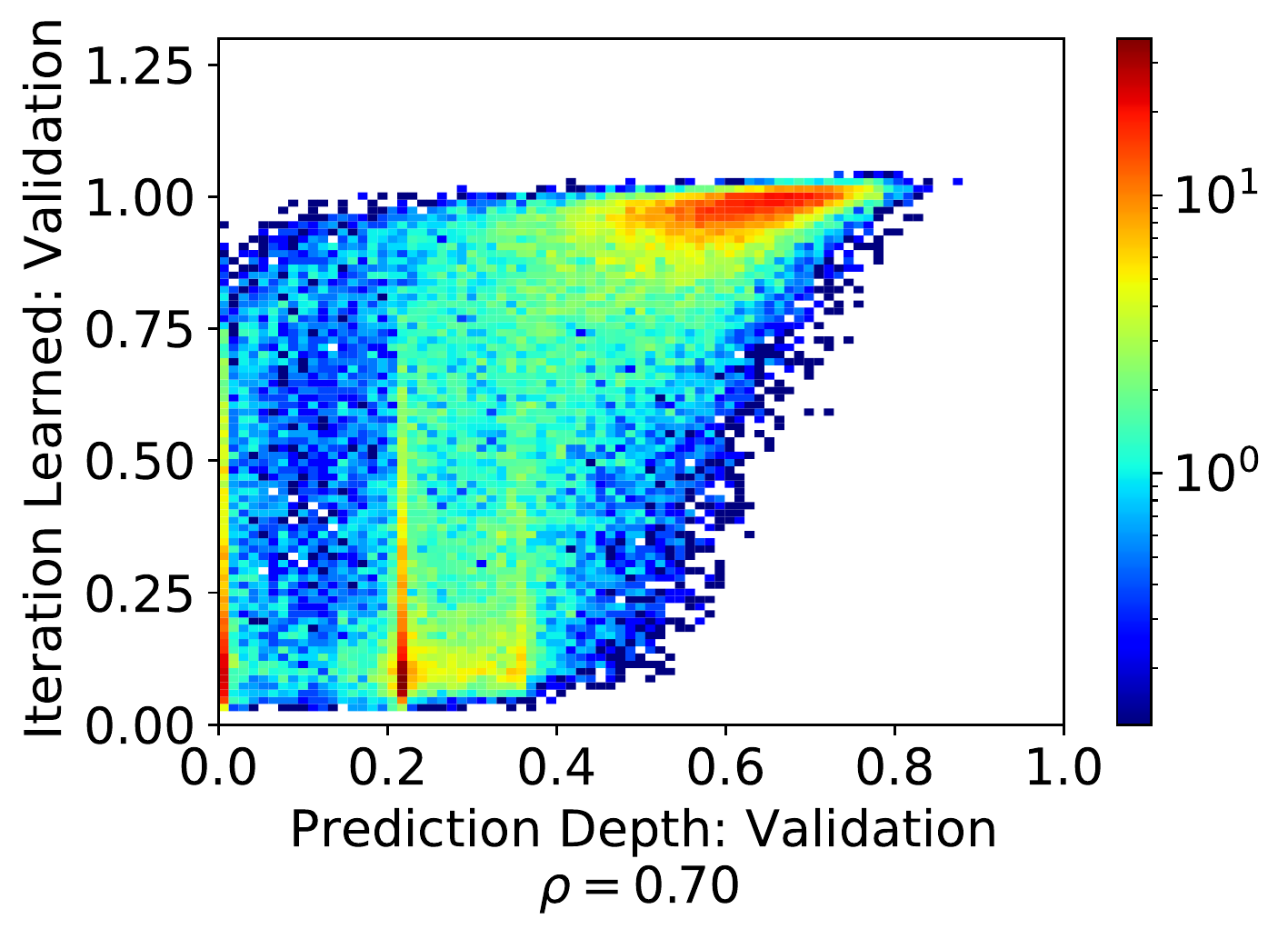}
\end{subfigure}
\begin{subfigure}
         \centering
         \includegraphics[width=0.49\columnwidth]{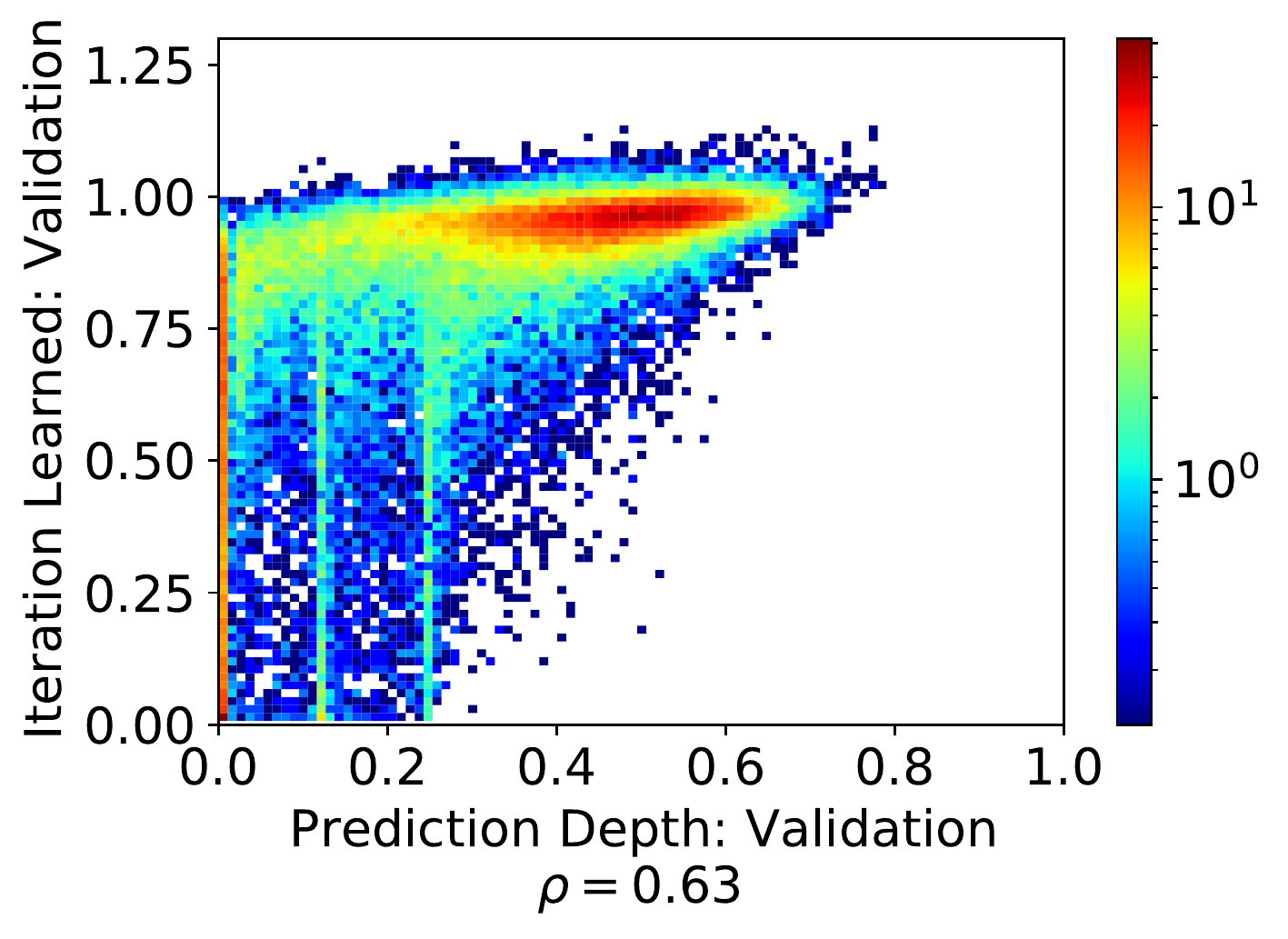}
\end{subfigure}
\begin{subfigure}
         \centering
         \includegraphics[width=0.49\columnwidth]{figures/del_mlp_cifar10_ll_vs_itl.pdf}
         \caption{CIFAR10. Top row: ResNet18. Middle row: VGG16. Bottom row: MLP. Histogram comparing the mean prediction depth to the mean iteration learned when each data point occurs in either the training split (left column) or the validation split (right column). See Appendix~\ref{app:ll_itl} for a description of the experiments performed.
\label{fig:ll_v_itl_1}}
\end{subfigure}
\end{center}
\end{figure}

\begin{figure}[ht!]
\begin{center}
\begin{subfigure}
         \centering
         \includegraphics[width=0.49\columnwidth]{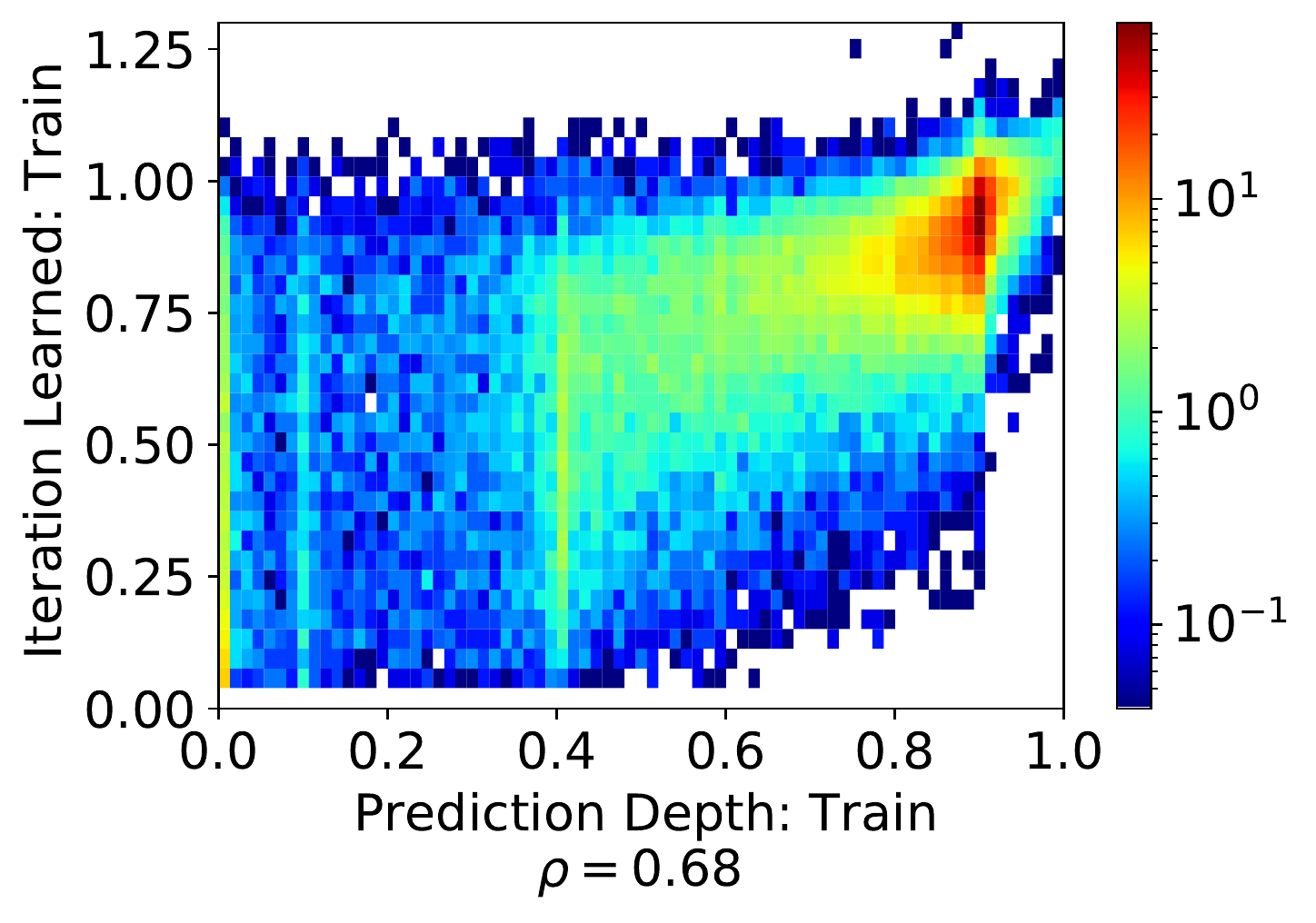}
\end{subfigure}
\begin{subfigure}
         \centering
         \includegraphics[width=0.49\columnwidth]{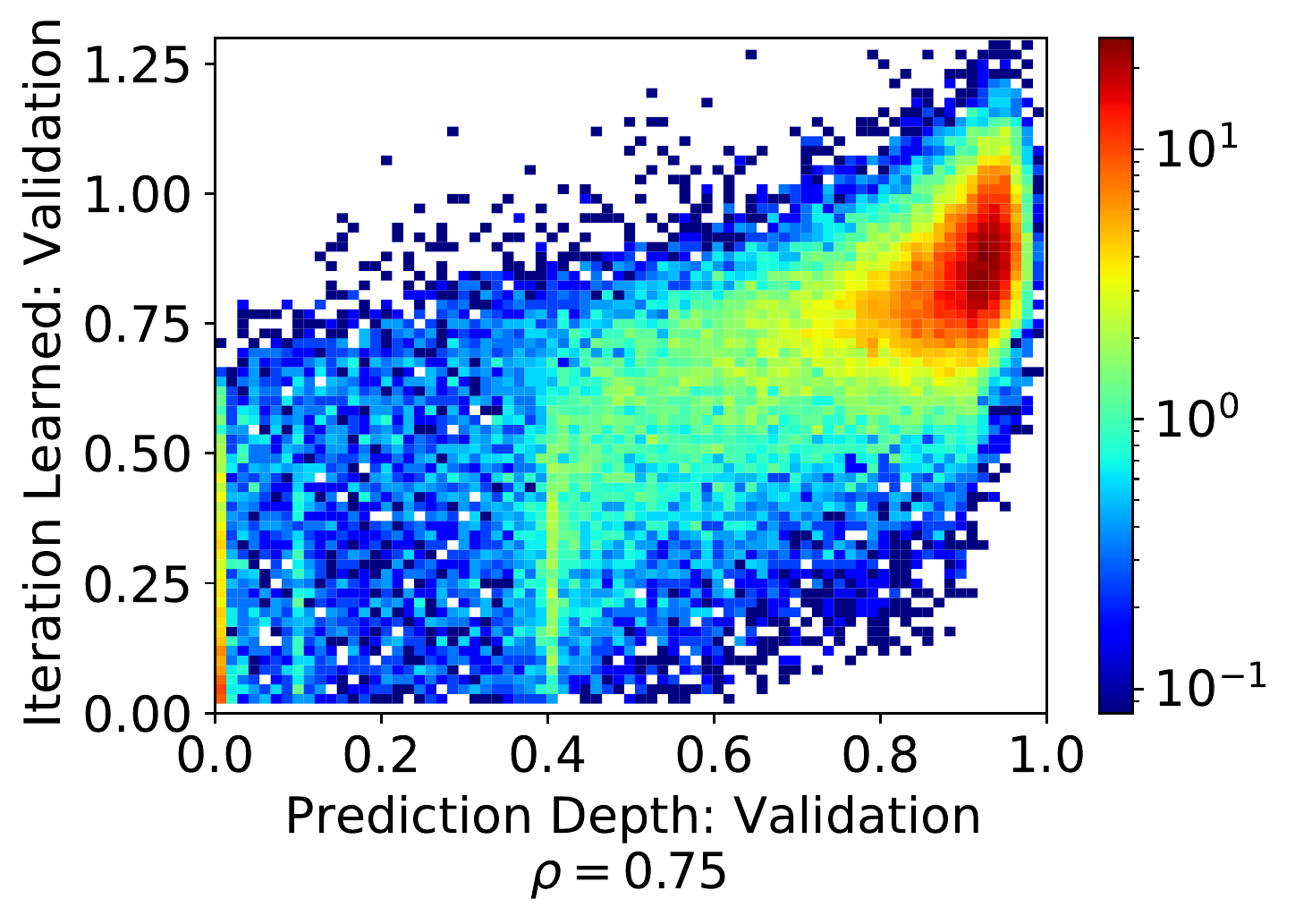}
\end{subfigure}
\begin{subfigure}
         \centering
         \includegraphics[width=0.49\columnwidth]{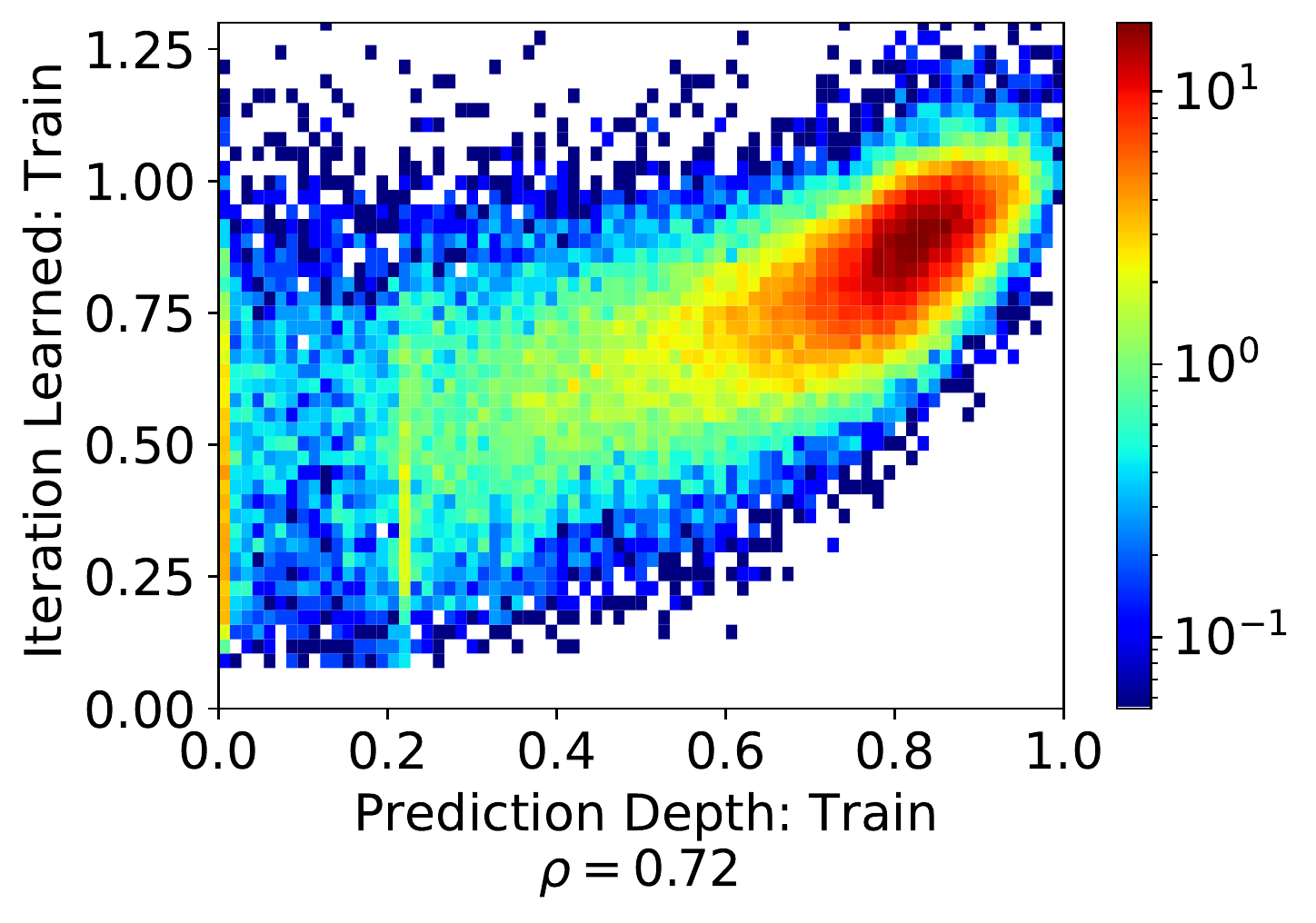}
\end{subfigure}
\begin{subfigure}
         \centering
         \includegraphics[width=0.49\columnwidth]{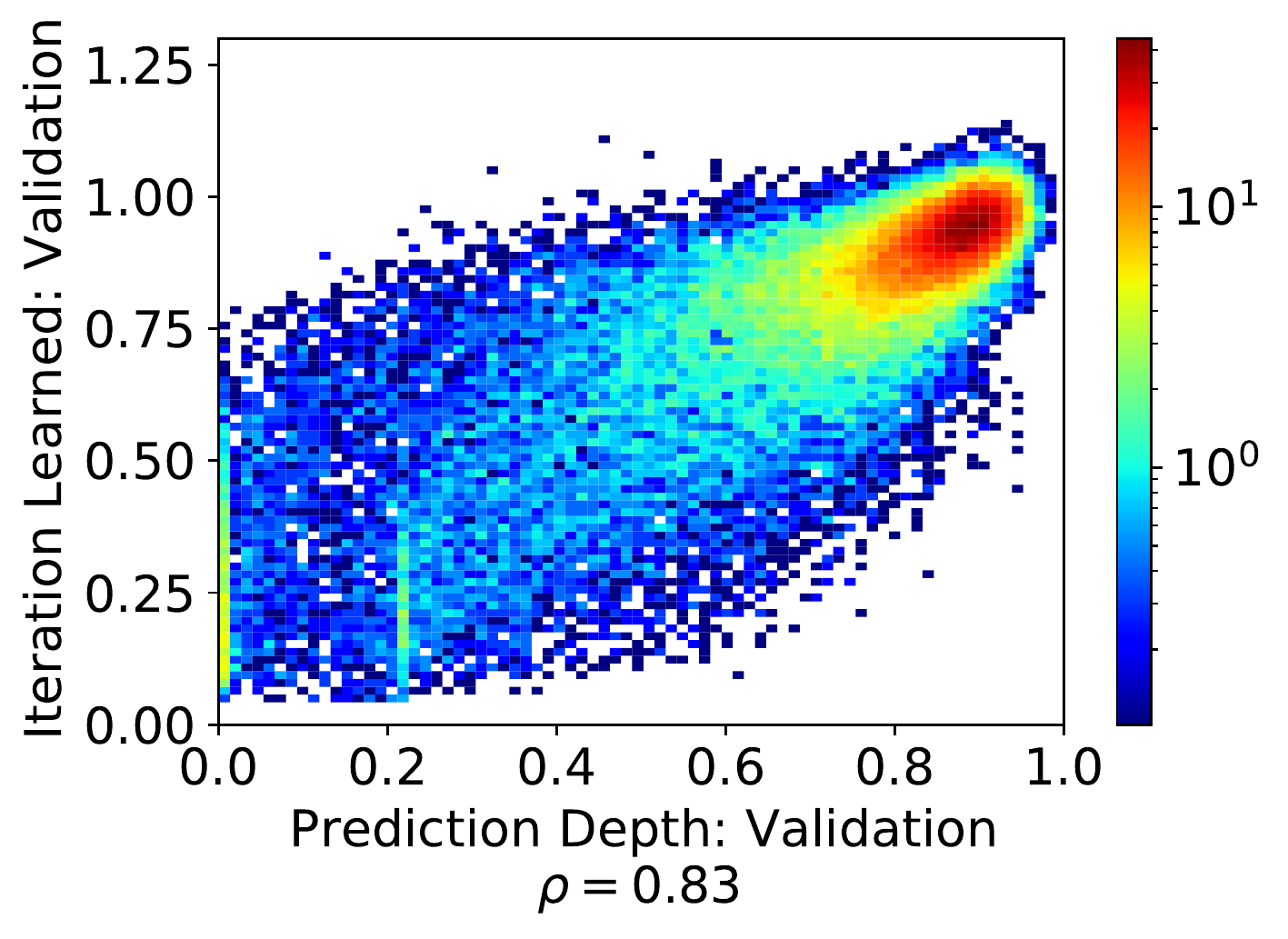}
\end{subfigure}
\begin{subfigure}
         \centering
         \includegraphics[width=0.49\columnwidth]{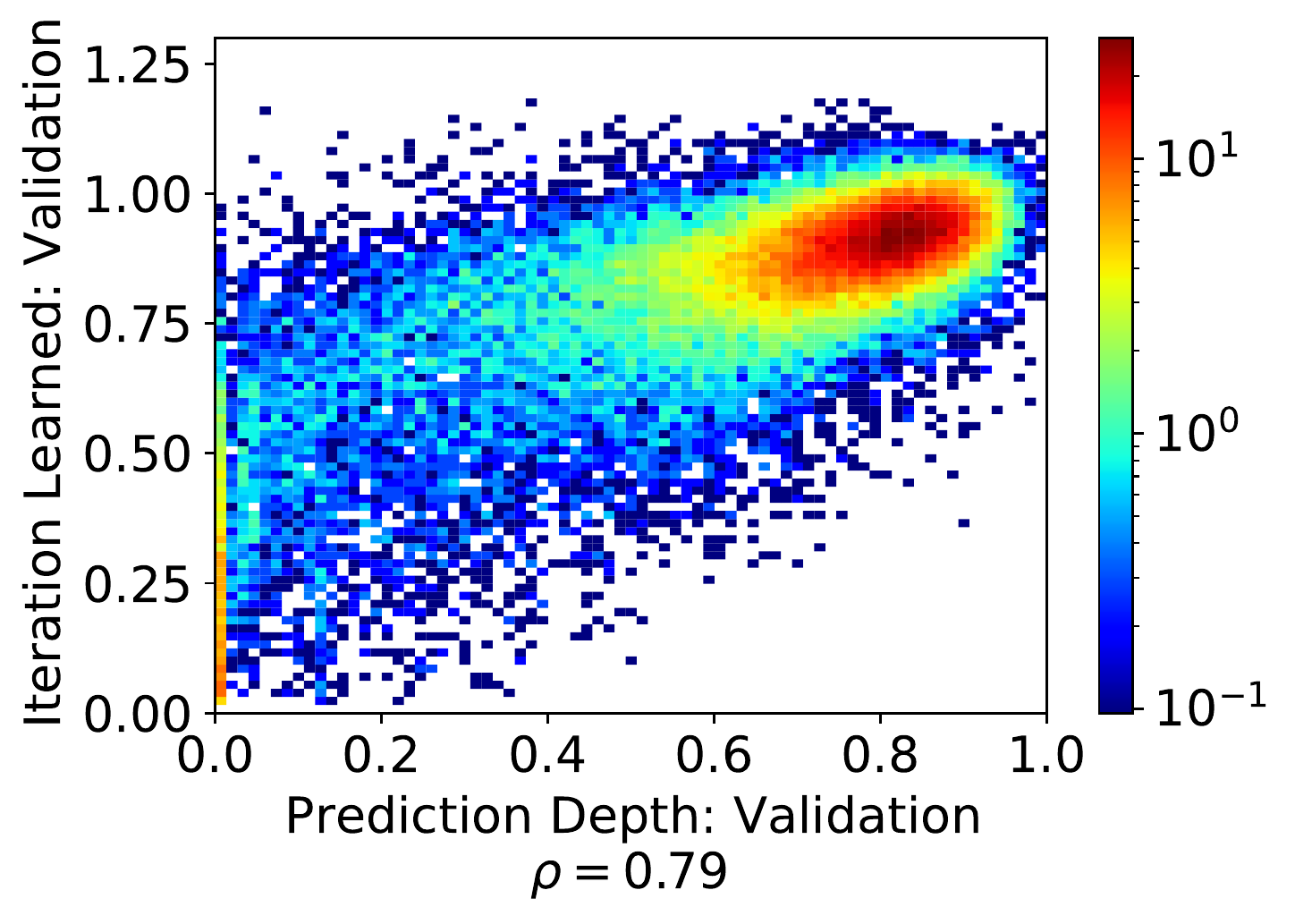}
\end{subfigure}
\begin{subfigure}
         \centering
         \includegraphics[width=0.49\columnwidth]{figures/del_mlp_cifar100_ll_vs_itl.pdf}
         \caption{CIFAR100. Top row: ResNet18. Middle row: VGG16. Bottom row: MLP. Histogram comparing the mean prediction depth to the mean iteration learned when each data point occurs in either the training split (left column) or the validation split (right column). See Appendix~\ref{app:ll_itl} for a description of the experiments performed.
\label{fig:ll_v_itl_2}}
\end{subfigure}
\end{center}
\end{figure}

\begin{figure}[ht!]
\begin{center}
\begin{subfigure}
         \centering
         \includegraphics[width=0.49\columnwidth]{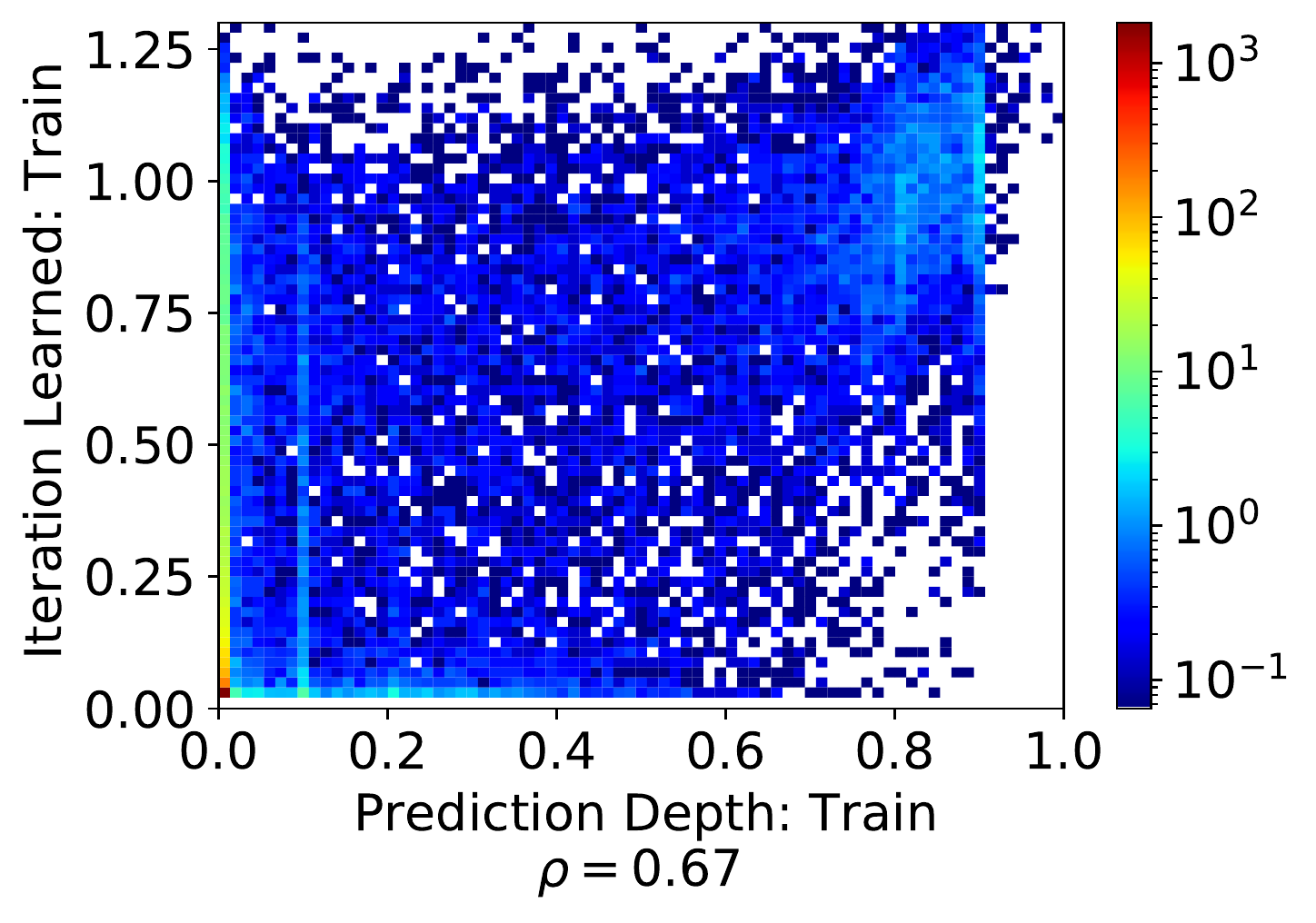}
\end{subfigure}
\begin{subfigure}
         \centering
         \includegraphics[width=0.49\columnwidth]{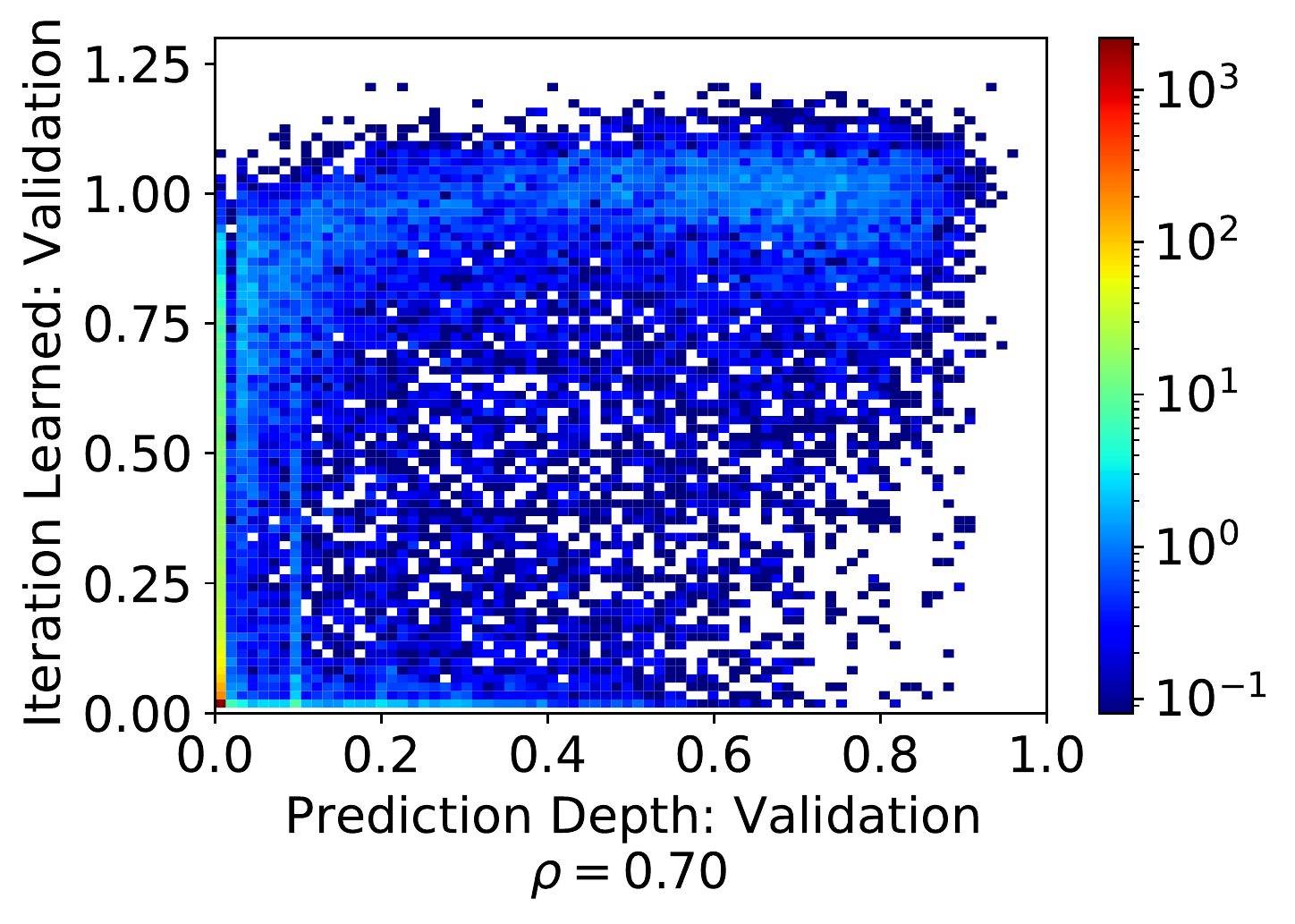}
\end{subfigure}
\begin{subfigure}
         \centering
         \includegraphics[width=0.49\columnwidth]{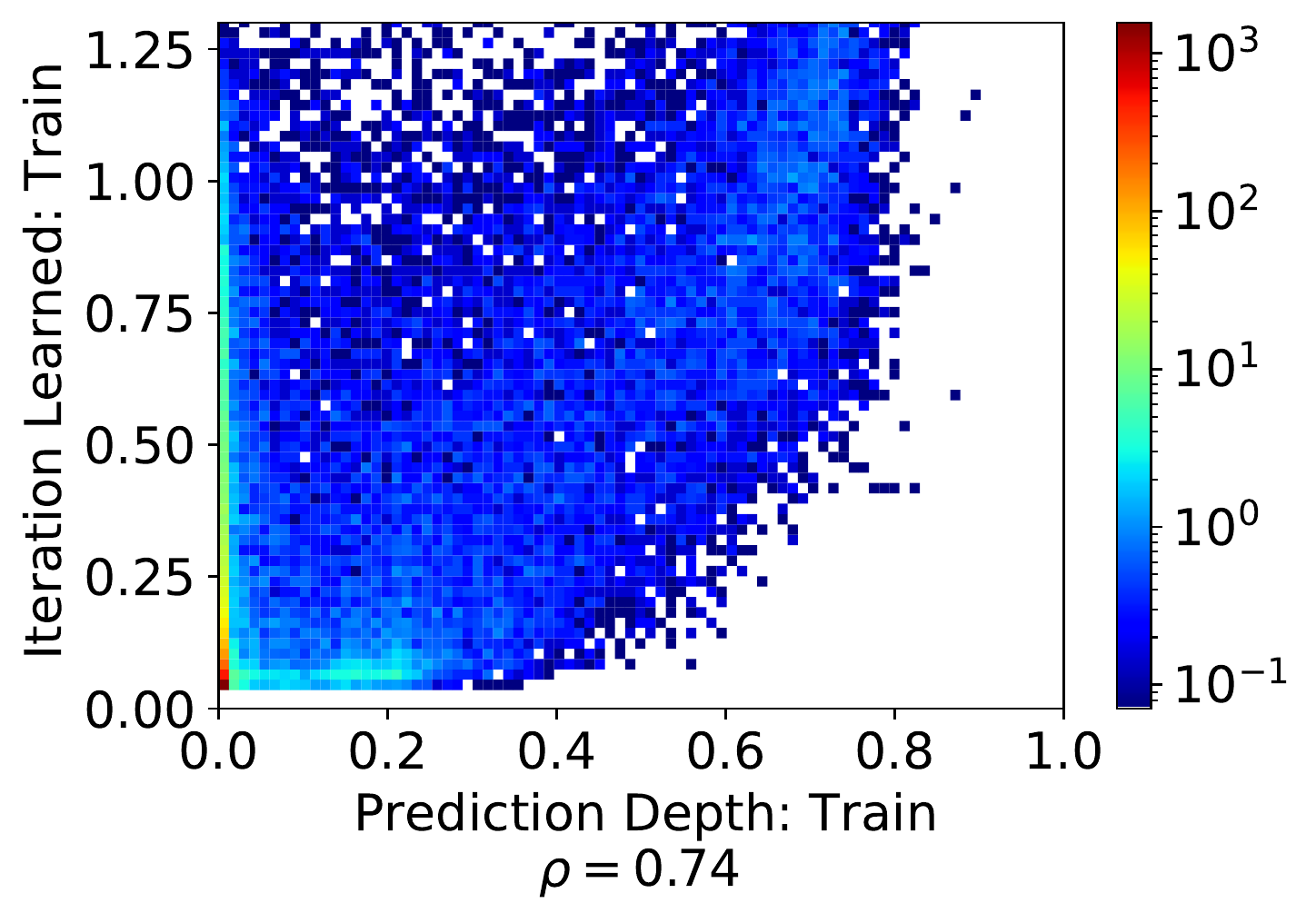}
\end{subfigure}
\begin{subfigure}
         \centering
         \includegraphics[width=0.49\columnwidth]{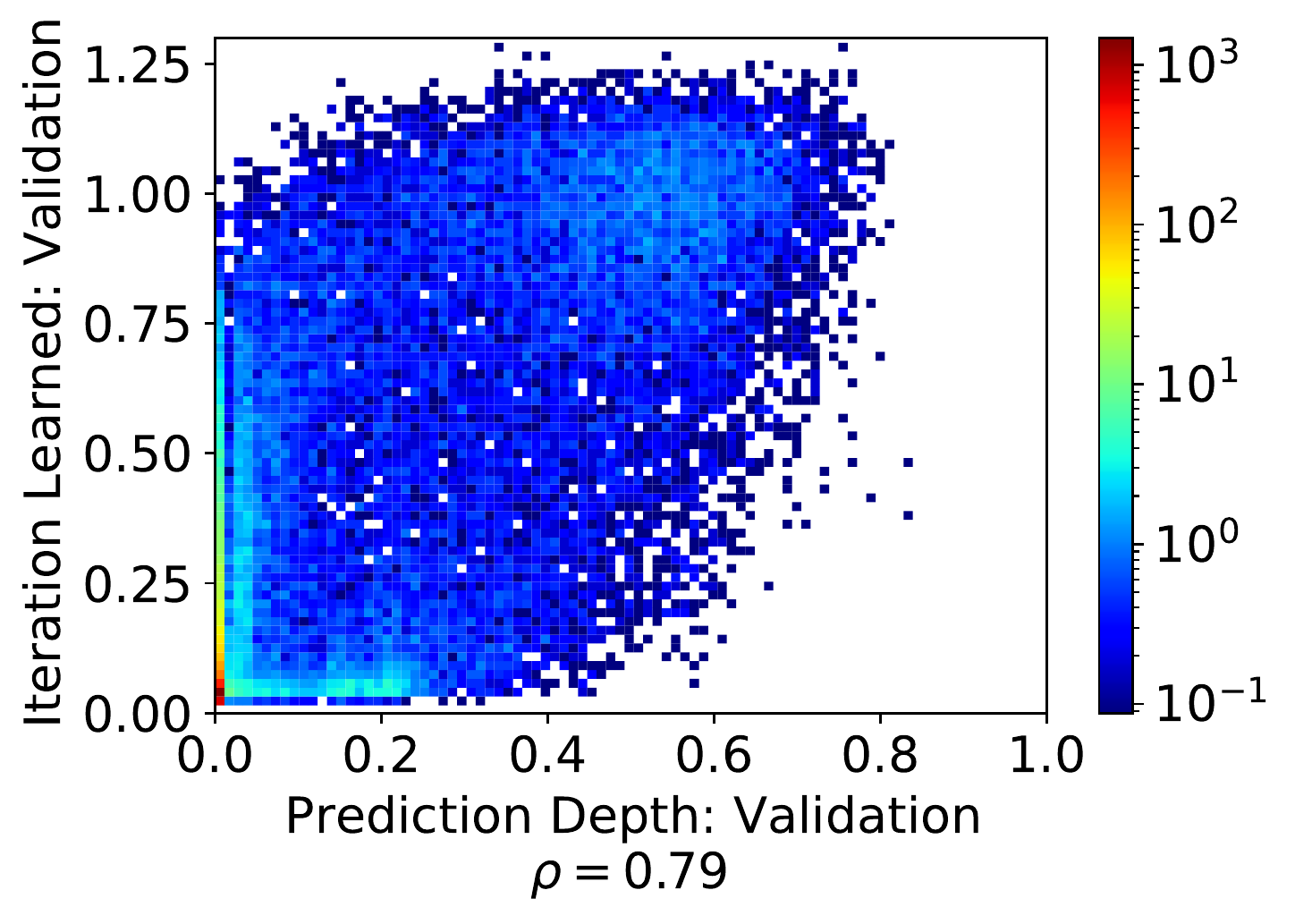}
\end{subfigure}
\begin{subfigure}
         \centering
         \includegraphics[width=0.49\columnwidth]{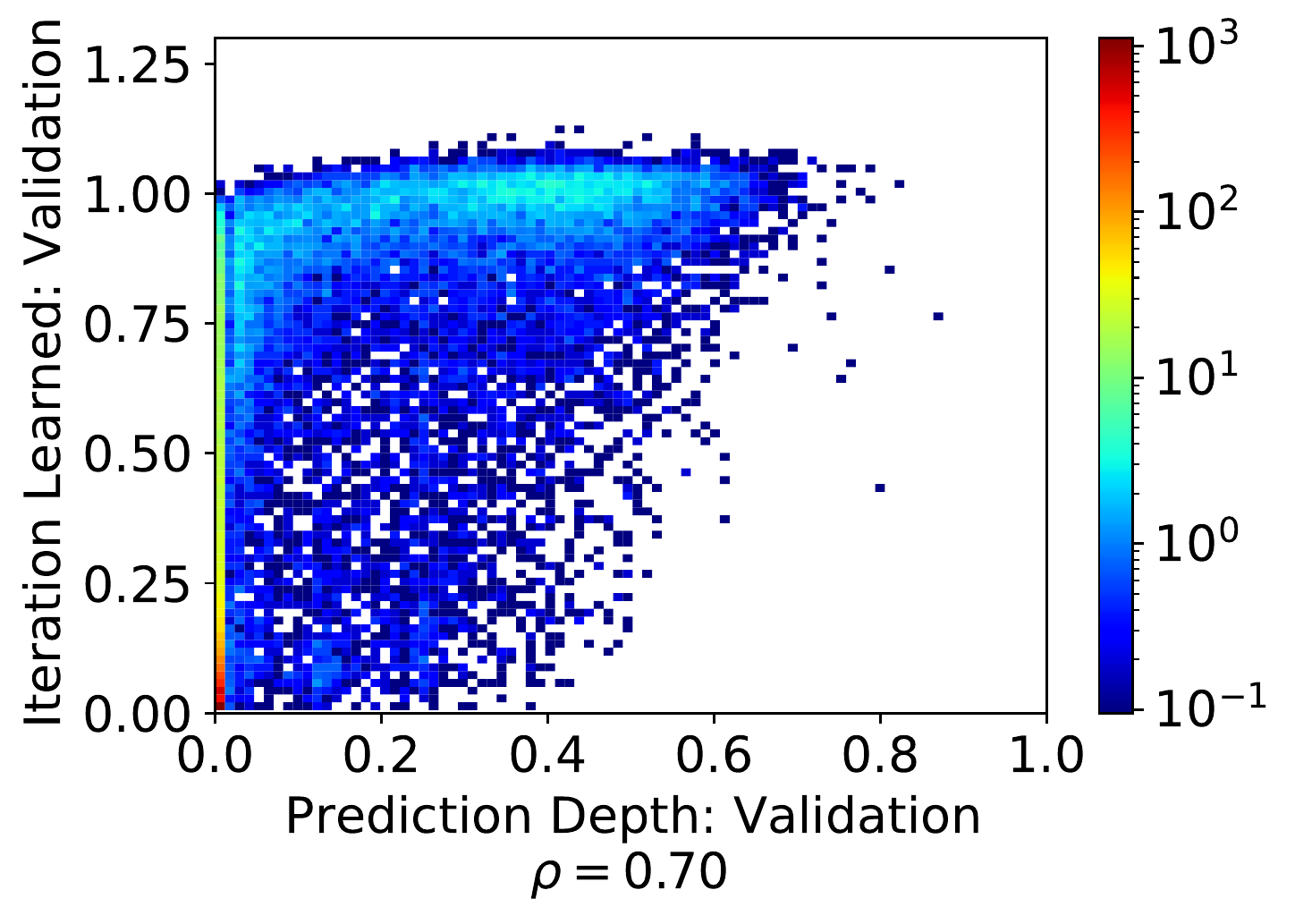}
\end{subfigure}
\begin{subfigure}
         \centering
         \includegraphics[width=0.49\columnwidth]{figures/del_mlp_fmnist_ll_vs_itl.pdf}
         \caption{Fashion MNIST. Top row: ResNet18. Middle row: VGG16. Bottom row: MLP. Histogram comparing the mean prediction depth to the mean iteration learned when each data point occurs in either the training split (left column) or the validation split (right column). In this case, the large majority of the data is already learned in the input layer. See Appendix~\ref{app:ll_itl} for a description of the experiments performed.
\label{fig:ll_v_itl_3}}
\end{subfigure}
\end{center}
\end{figure}

\begin{figure}[ht!]
\begin{center}
\begin{subfigure}
         \centering
         \includegraphics[width=0.49\columnwidth]{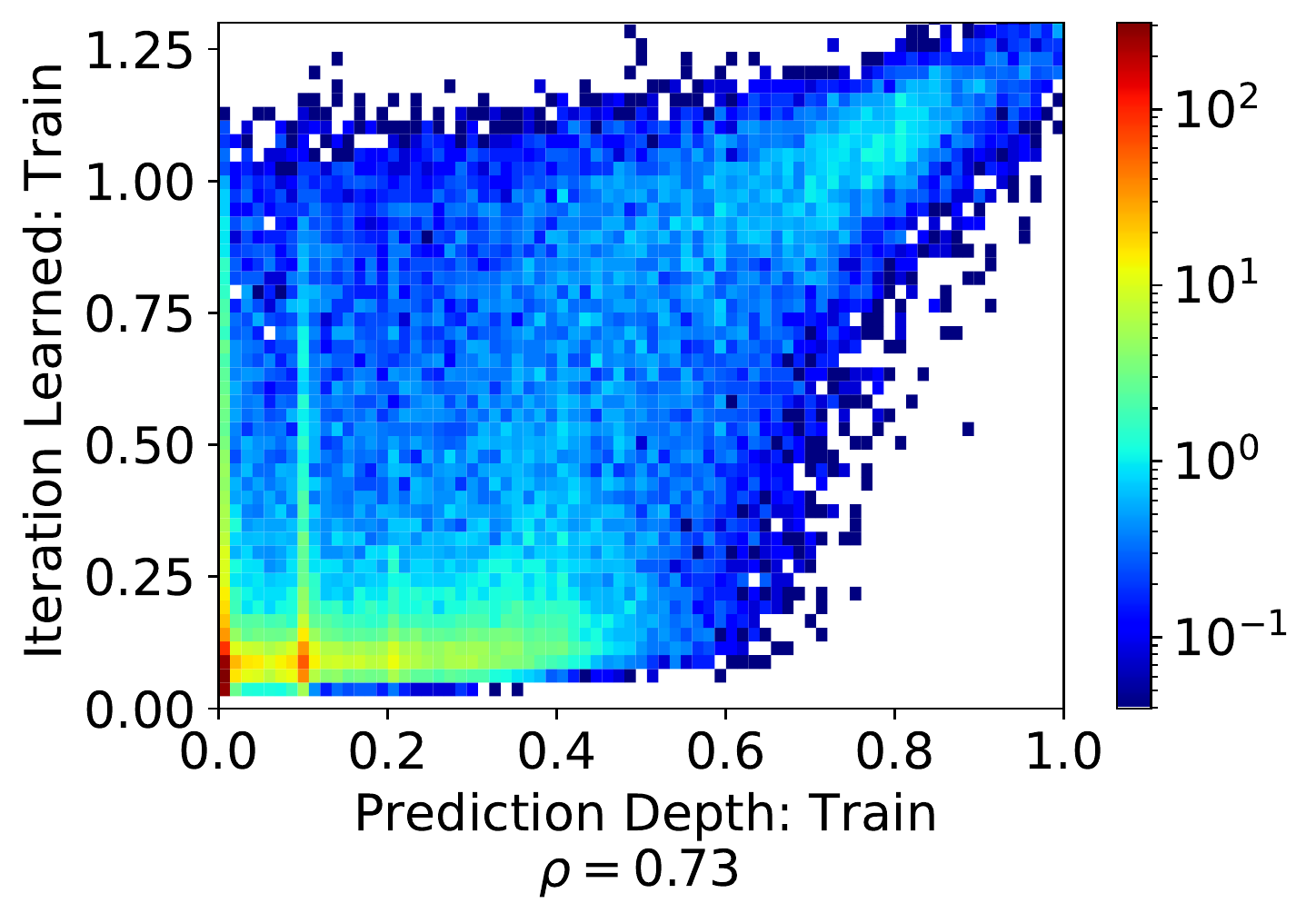}
\end{subfigure}
\begin{subfigure}
         \centering
         \includegraphics[width=0.49\columnwidth]{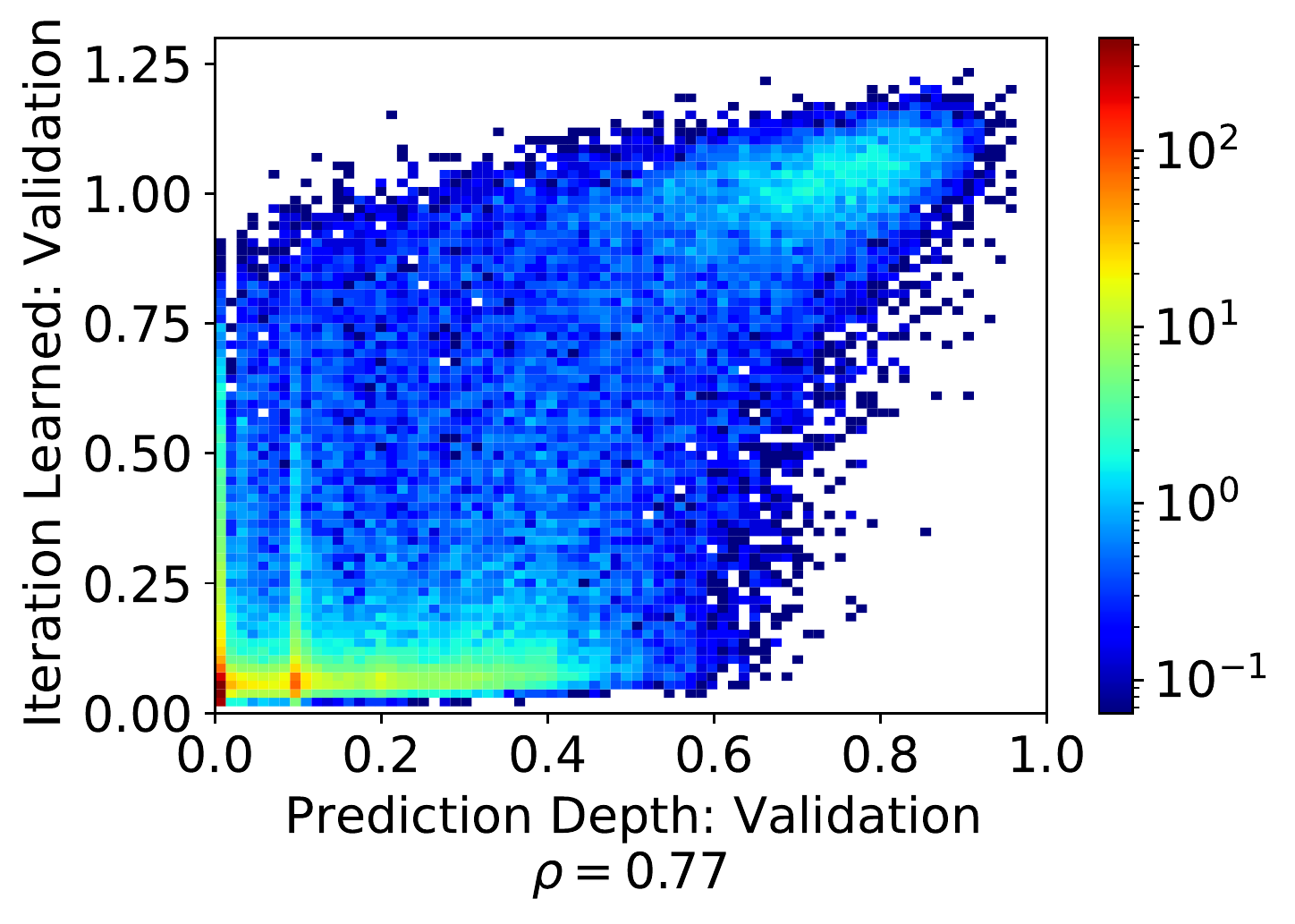}
\end{subfigure}
\begin{subfigure}
         \centering
         \includegraphics[width=0.49\columnwidth]{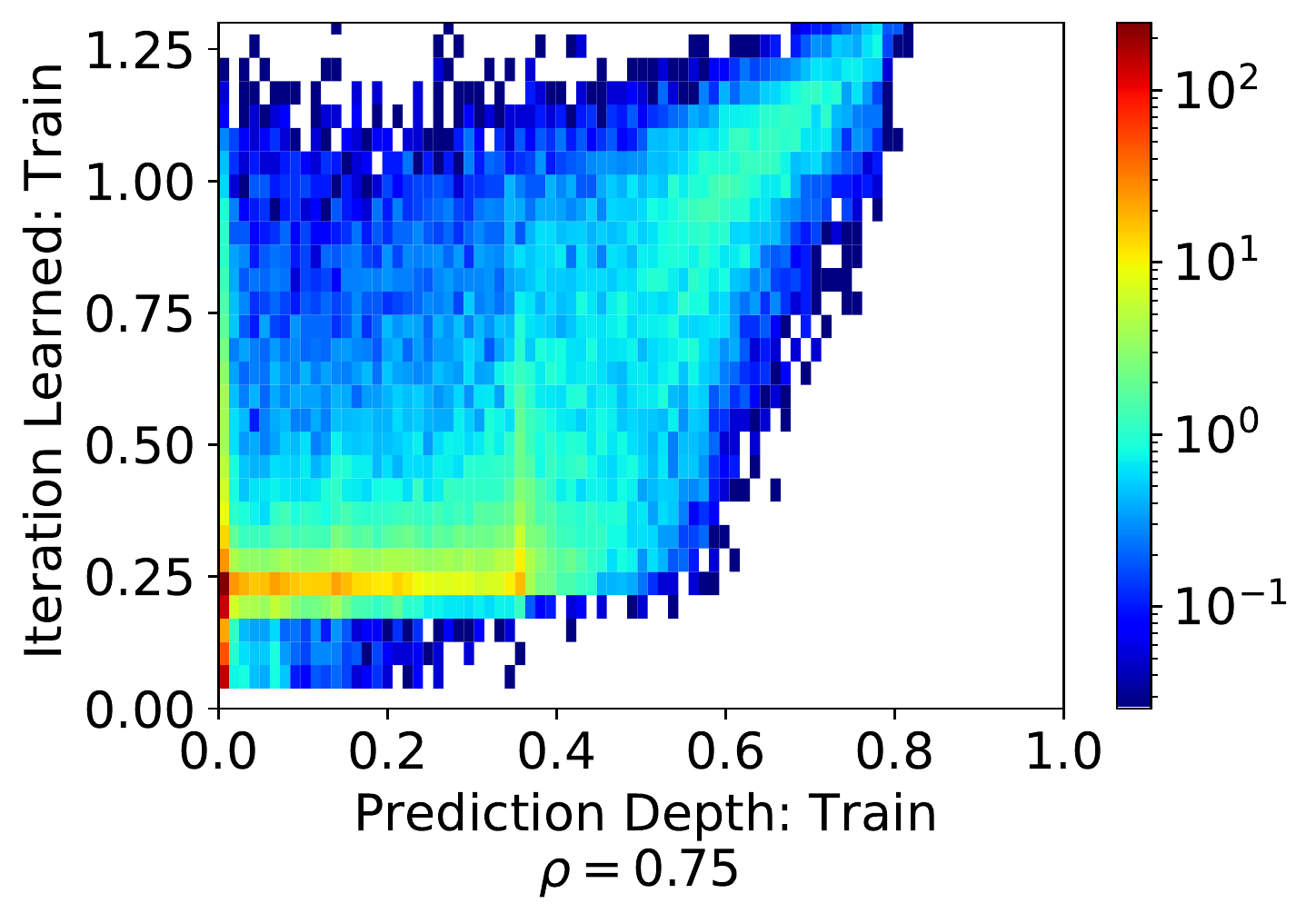}
\end{subfigure}
\begin{subfigure}
         \centering
         \includegraphics[width=0.49\columnwidth]{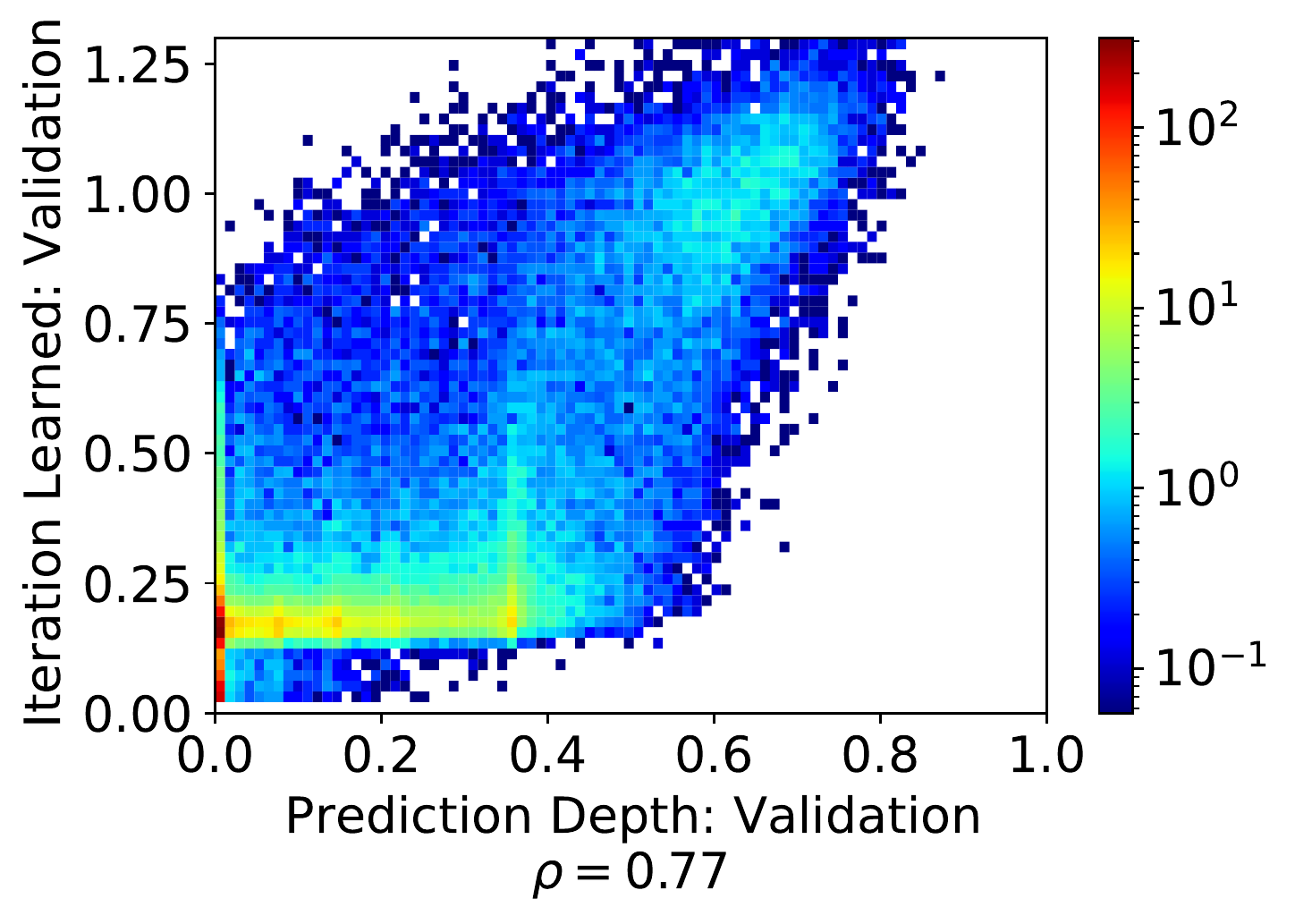}
\end{subfigure}
\begin{subfigure}
         \centering
         \includegraphics[width=0.49\columnwidth]{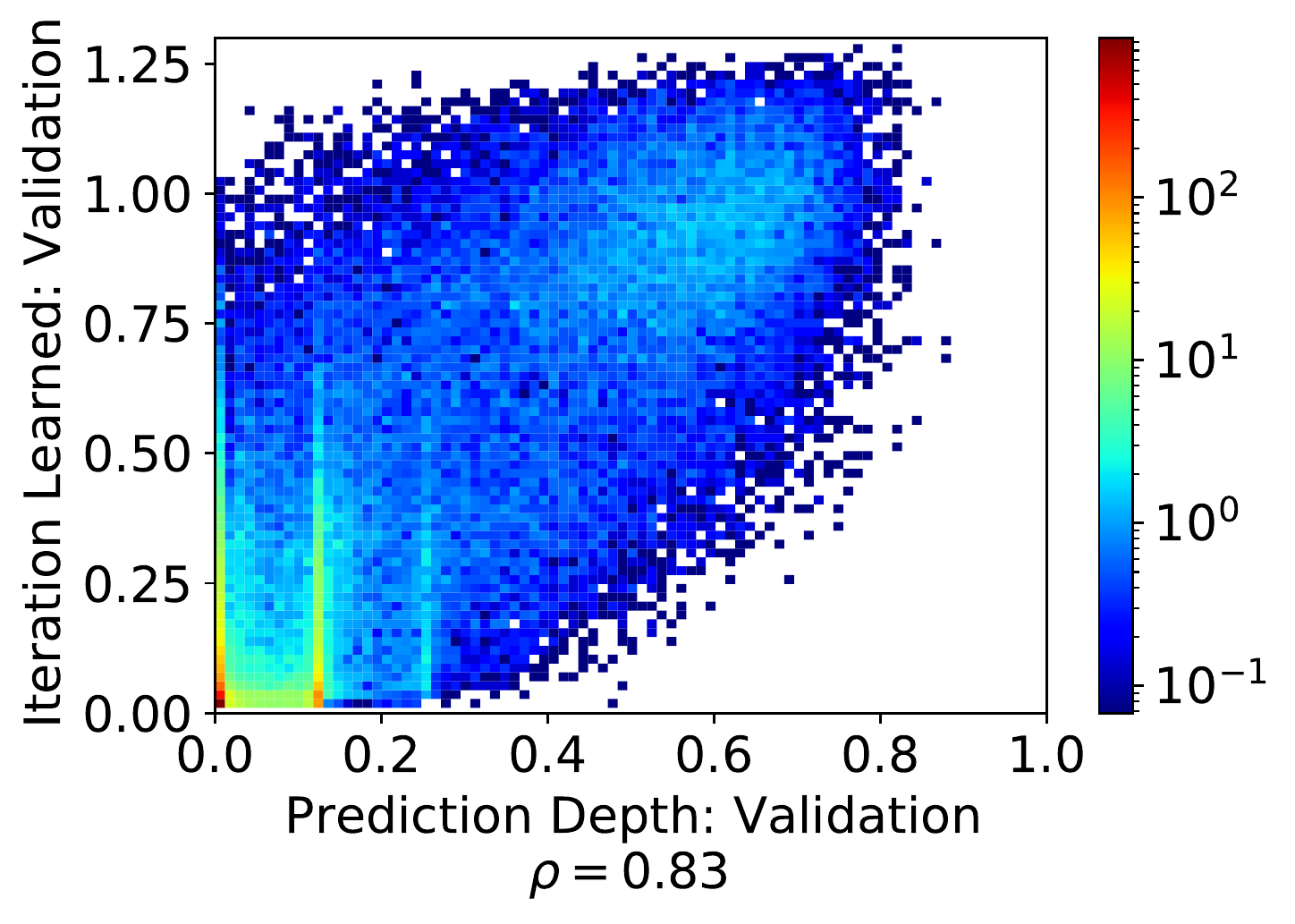}
\end{subfigure}
\begin{subfigure}
         \centering
         \includegraphics[width=0.49\columnwidth]{figures/del_mlp_svhn_ll_vs_itl.pdf}
         \caption{SVHN. Top row: ResNet18. Middle row: VGG16. Bottom row: MLP. Histogram comparing the mean prediction depth to the mean iteration learned when each data point occurs in either the training split (left column) or the validation split (right column). See Appendix~\ref{app:ll_itl} for a description of the experiments performed.
\label{fig:ll_v_itl_4}}
\end{subfigure}
\end{center}
\end{figure}

Figure~\ref{fig:ll_vs_iter_all} reproduces the result shown in Figure~\ref{fig:visual_correspondance} (left) for every architecture and dataset.
To give a more complete picture of the relationship between the prediction depth and the iteration learned, Figures~\ref{fig:ll_v_itl_1} to~\ref{fig:ll_v_itl_4} show histograms of the mean prediction depth and iteration learned for each data point when it occurs in both the training and validation splits.
As described in Appendix~\ref{app:experiments_desc}, for each dataset and architecture we trained 250 models with random 90:10\% validation:train splits. Each time a data point appears in either split we record the prediction depth and the iteration learned. These histograms compare the mean prediction depth to the mean iteration learned for all data points in both the train and validation splits.
The Spearman's Correlation Coefficient is given beneath each plot.

\subsection{Consistency of margin results \label{app:consistency_margins}}

\begin{figure}[ht!]
\begin{center}
\resizebox{1.\textwidth}{!}{%
\begin{subfigure}
         \centering
         \includegraphics[height=3cm]{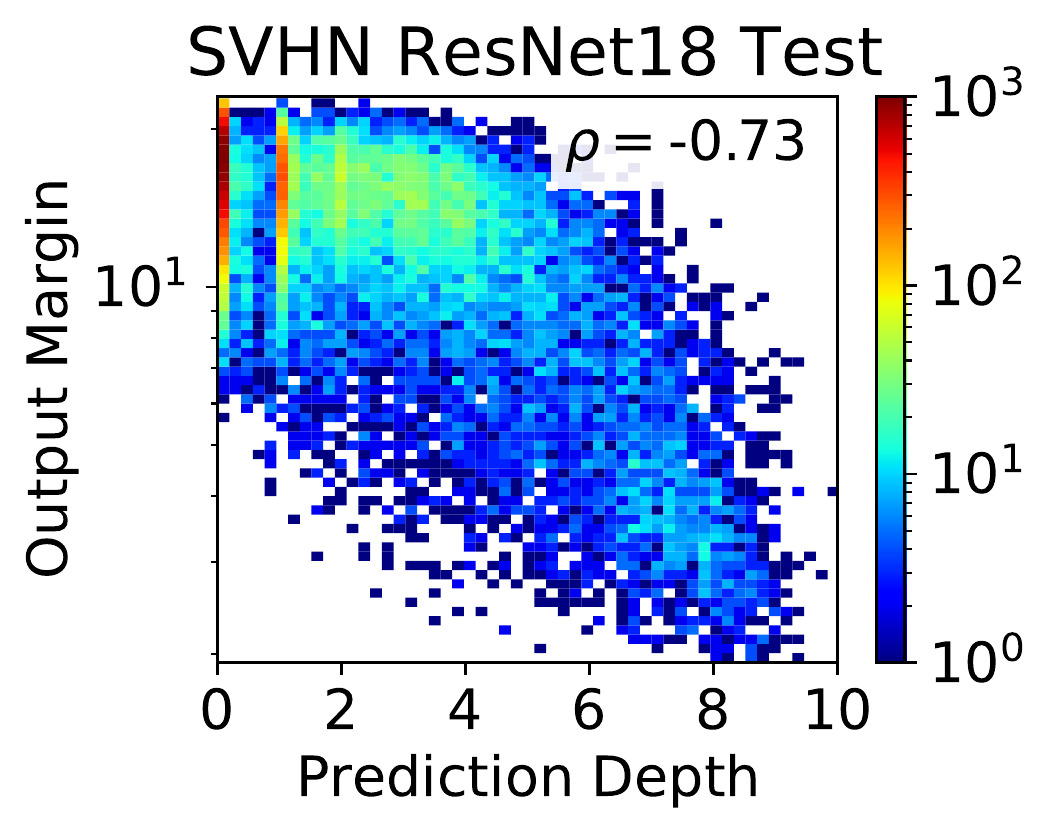}
\end{subfigure}
\begin{subfigure}
         \centering
         \includegraphics[height=3cm]{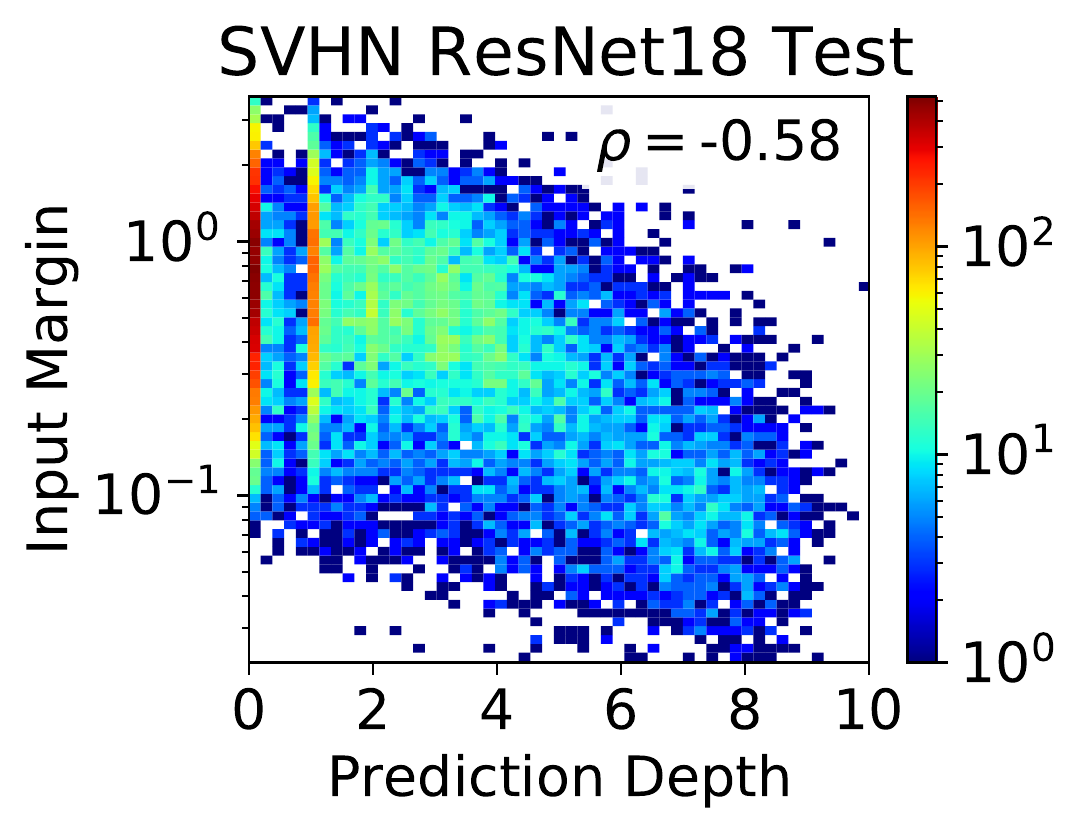}
\end{subfigure}
\begin{subfigure}
         \centering
         \includegraphics[height=3cm]{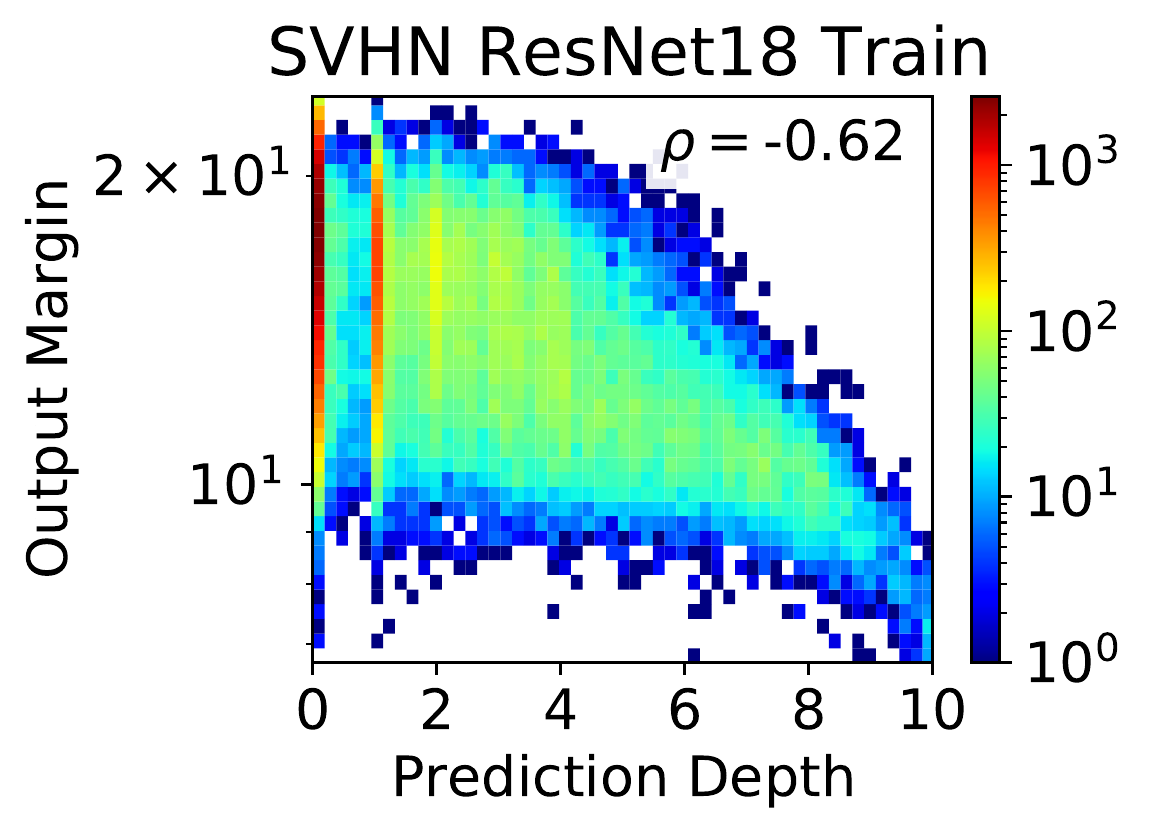}
\end{subfigure}
\begin{subfigure}
         \centering
         \includegraphics[height=3cm]{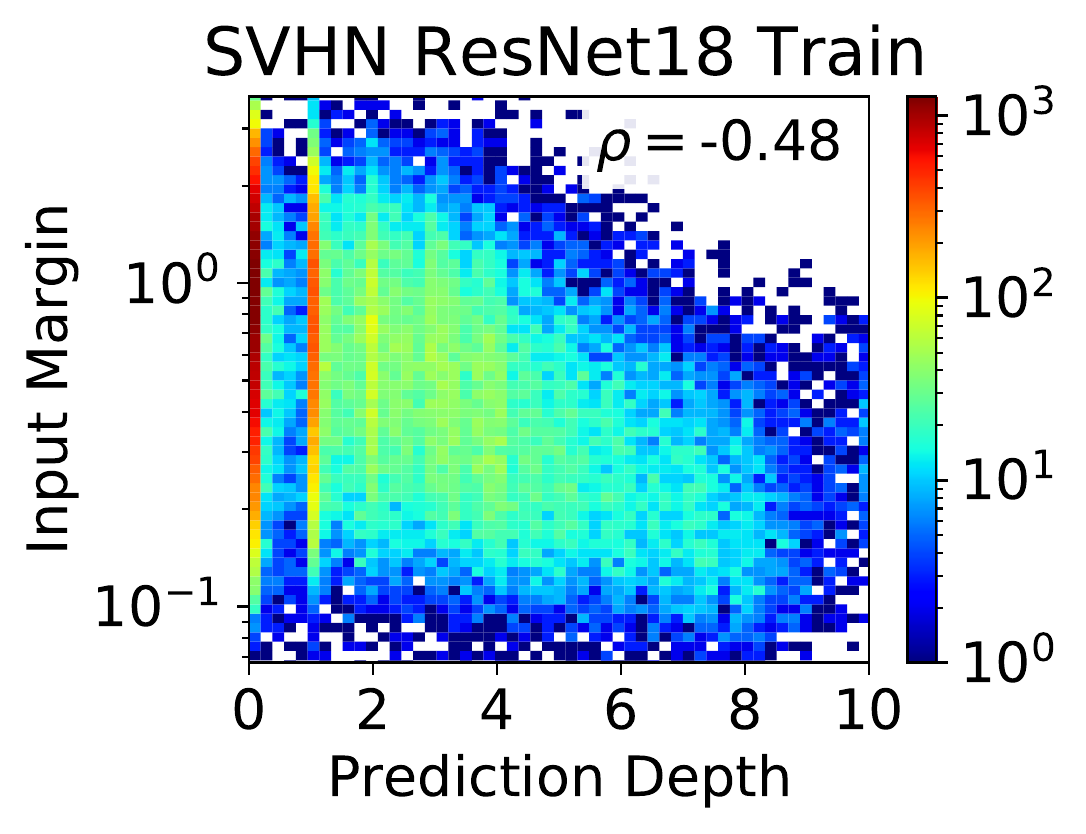}
\end{subfigure}}

\resizebox{1.\textwidth}{!}{%
\begin{subfigure}
         \centering
         \includegraphics[height=3cm]{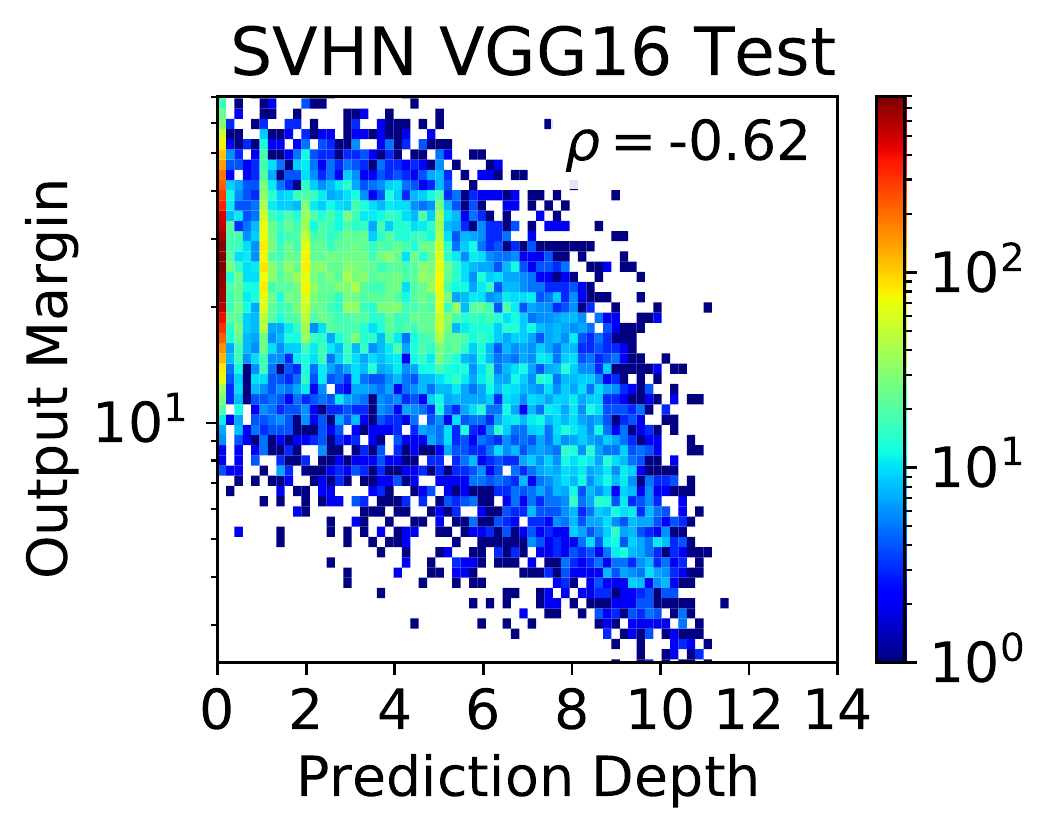}
\end{subfigure}
\begin{subfigure}
         \centering
         \includegraphics[height=3cm]{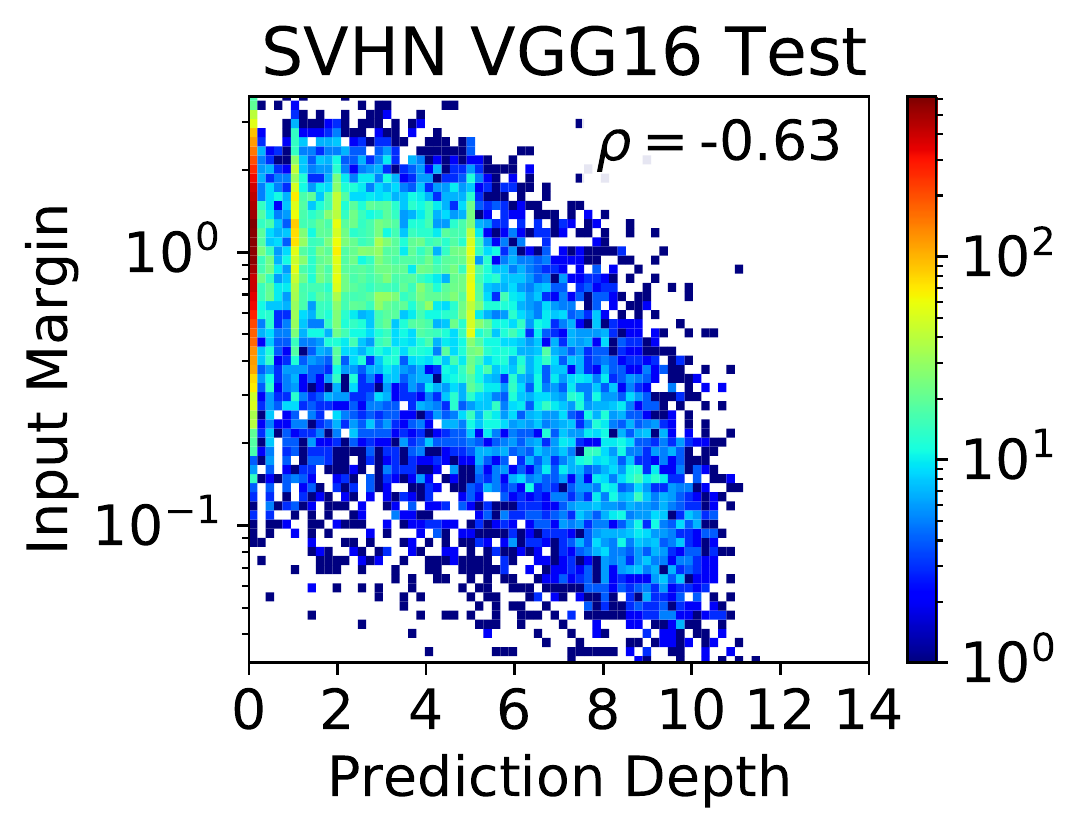}
\end{subfigure}
\begin{subfigure}
         \centering
         \includegraphics[height=3cm]{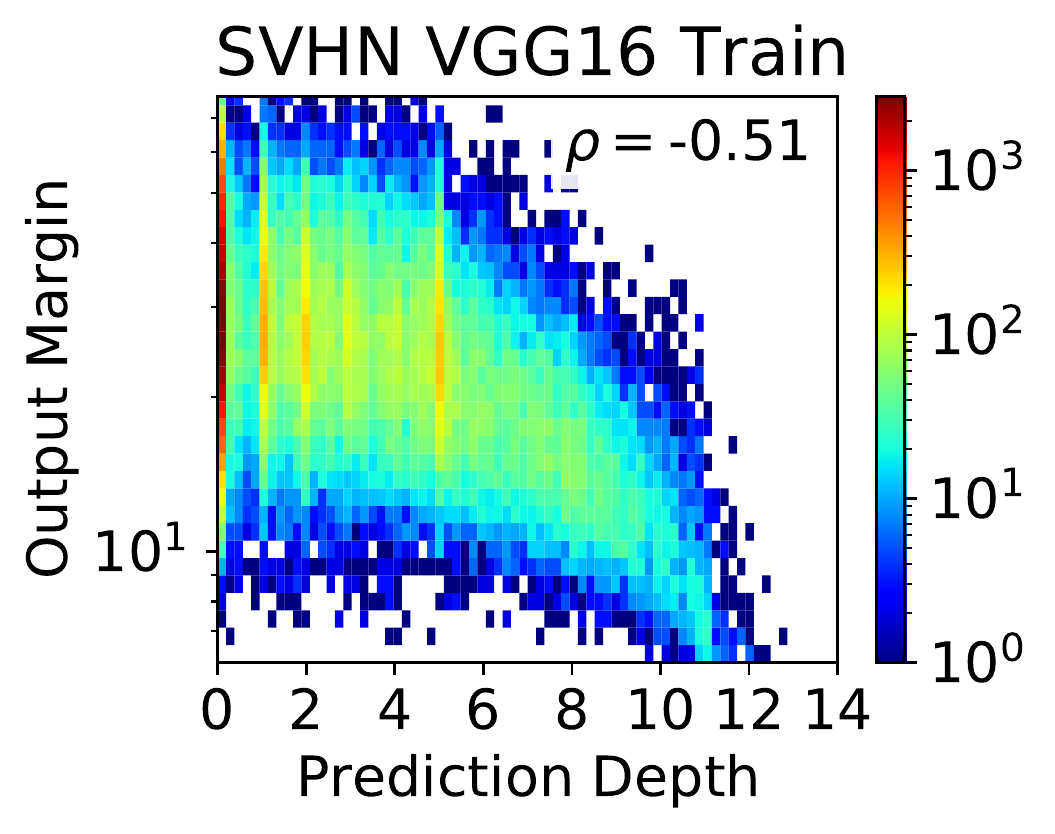}
\end{subfigure}
\begin{subfigure}
         \centering
         \includegraphics[height=3cm]{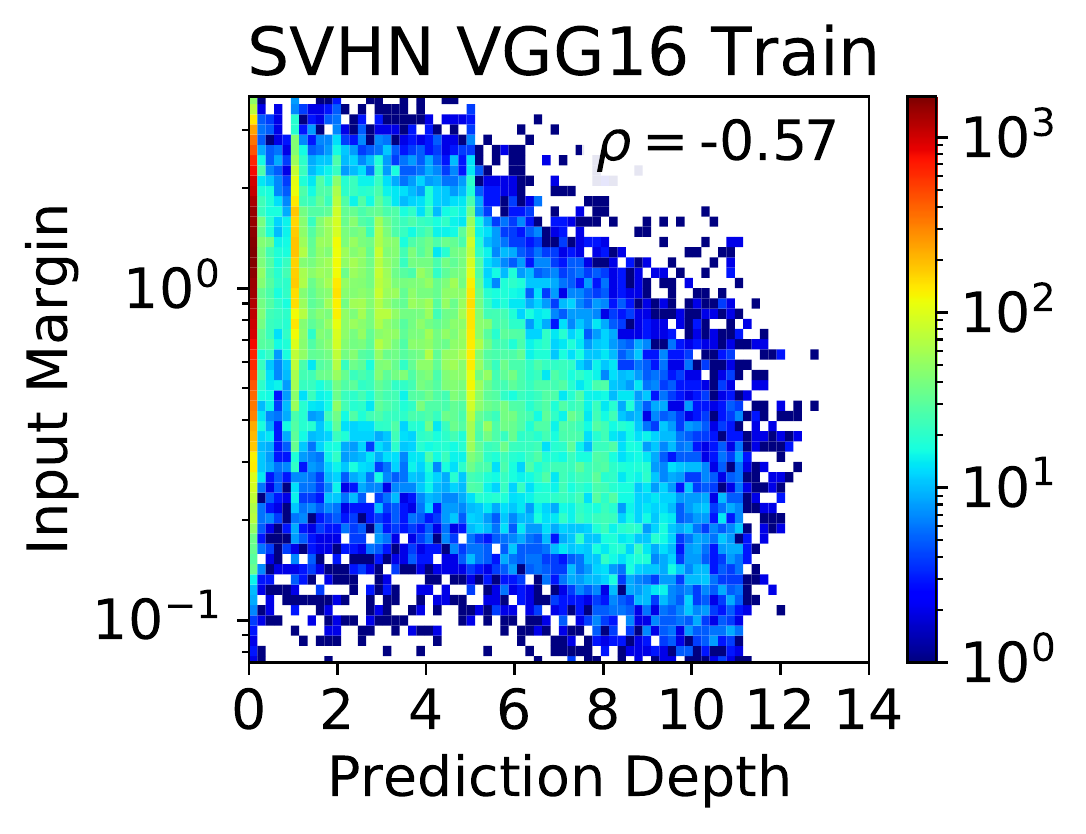}
\end{subfigure}}

\resizebox{1.\textwidth}{!}{%
\begin{subfigure}
         \centering
         \includegraphics[height=3cm]{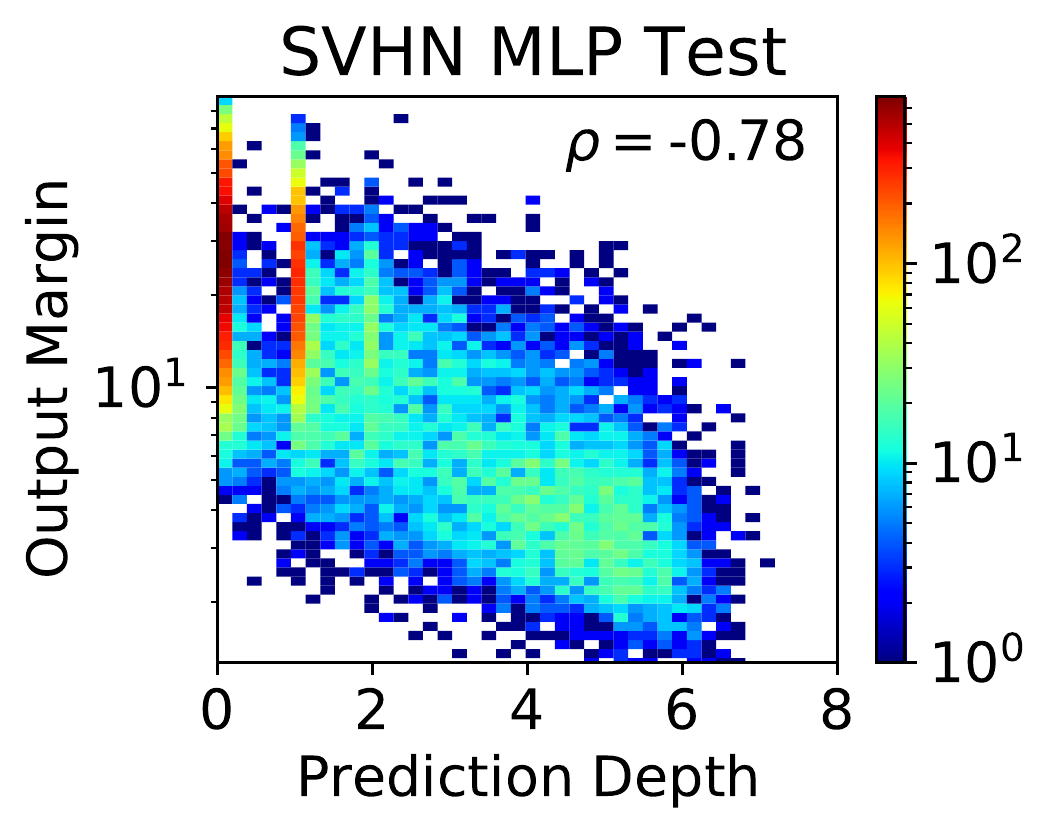}
\end{subfigure}
\begin{subfigure}
         \centering
         \includegraphics[height=3cm]{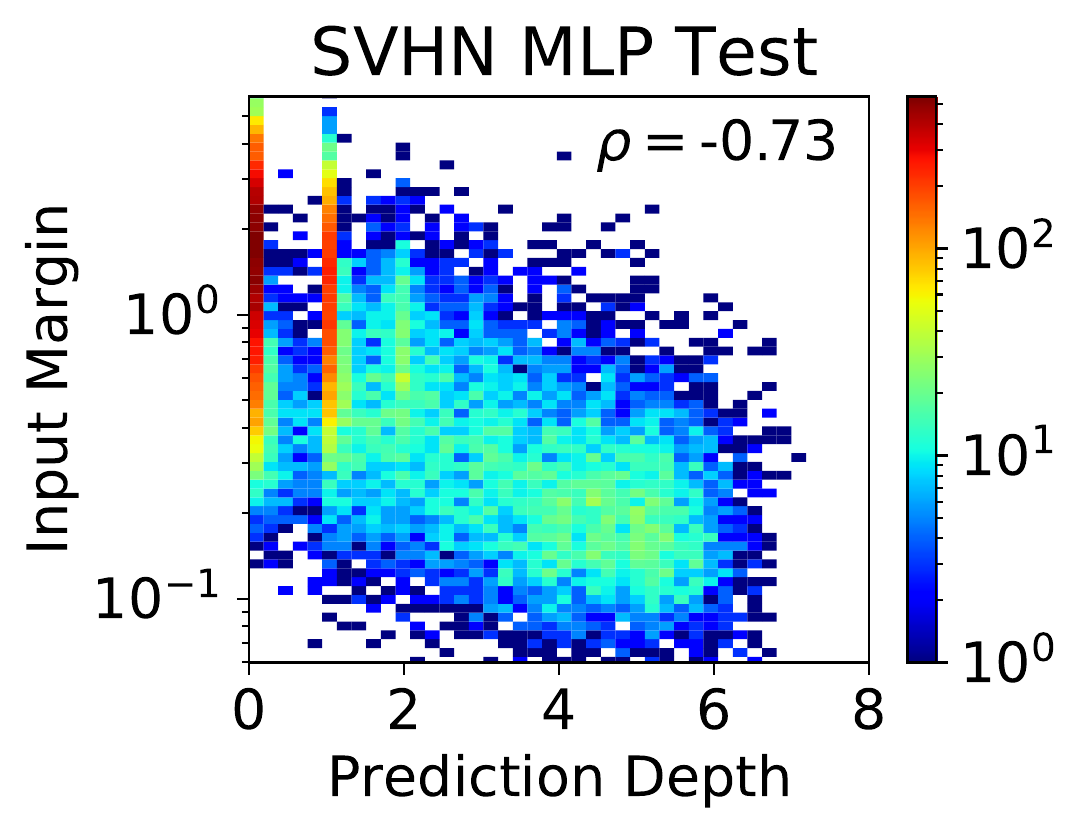}
\end{subfigure}
\begin{subfigure}
         \centering
         \includegraphics[height=3cm]{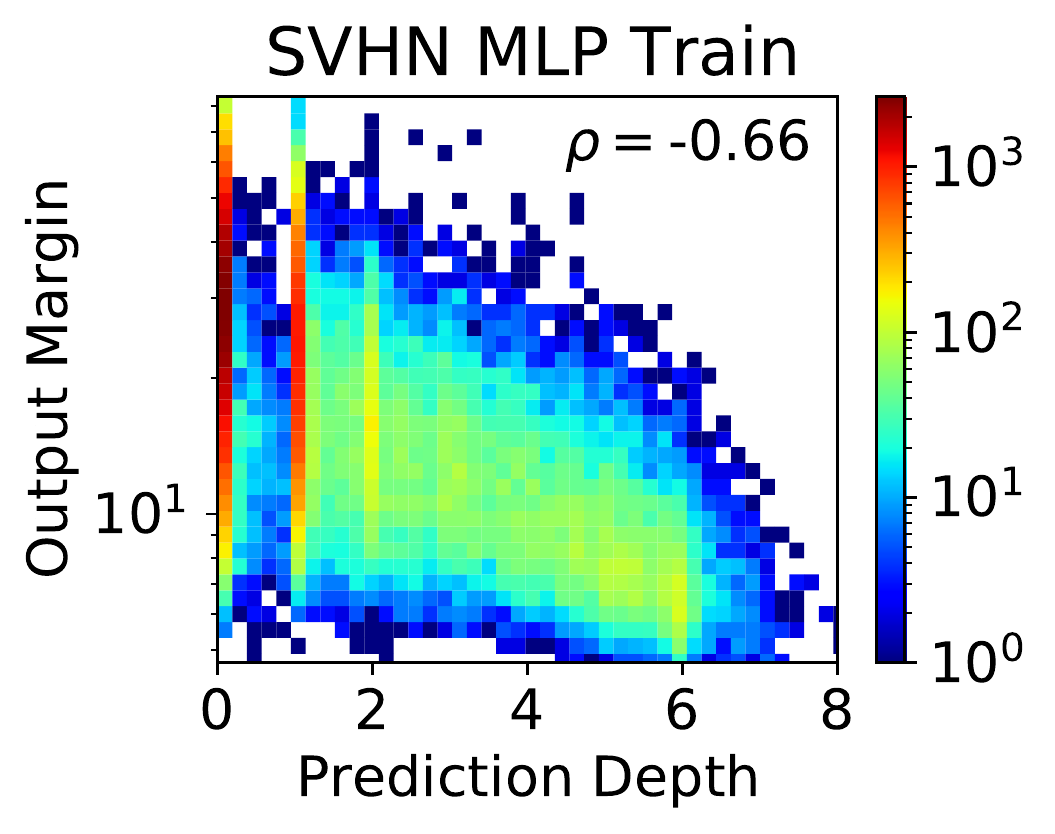}
\end{subfigure}
\begin{subfigure}
         \centering
         \includegraphics[height=3cm]{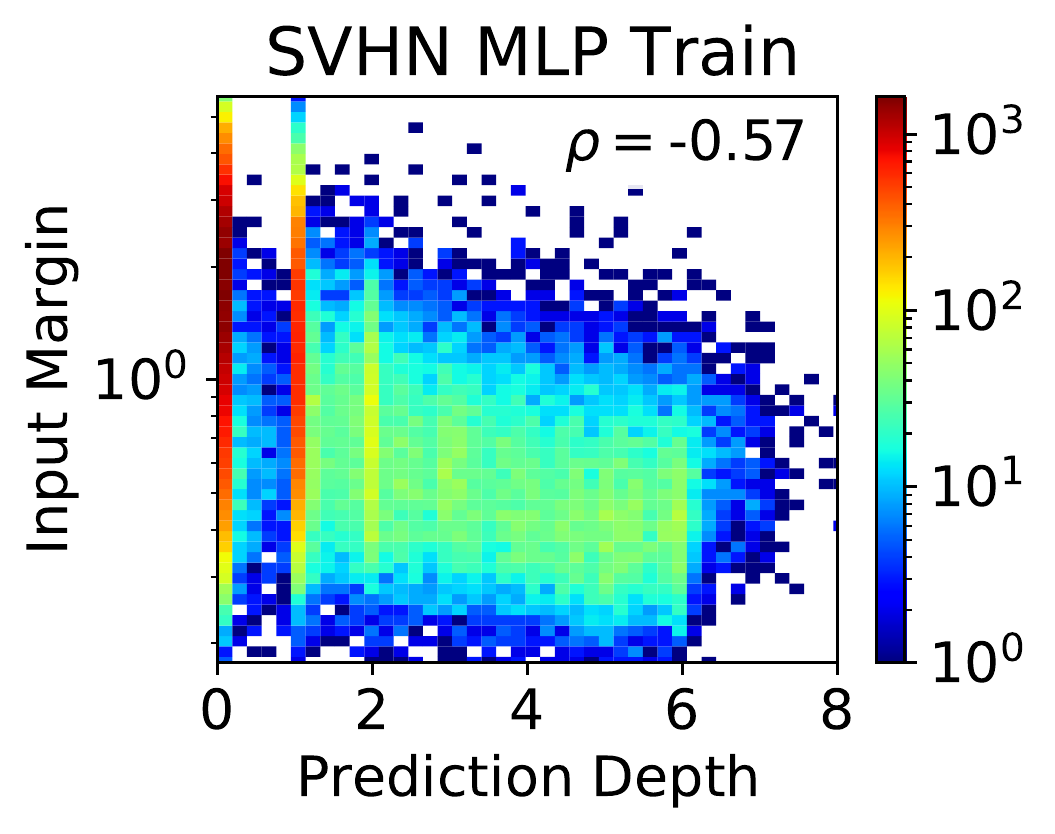}
\end{subfigure}}

\end{center}
\caption{Consistency of Figure~\ref{fig:margins_vs_depth}, showing the correlation between prediction depth, and the input and output margins (log scale) for both the test and training splits of SVHN. The correlation coefficient between the prediction depth and the logarithm of the margin is given in each plot. For each architecture, we train 25 models with different random seeds on the full training split. We record the input and output margins together with the prediction depth for every data point in both the train and test splits. These histograms compare the mean values of each margin to the mean prediction depth for all data points.
\label{fig:svhn_marg_consist}}
\end{figure}

\begin{figure}[ht!]
\begin{center}
\resizebox{1.\textwidth}{!}{%
\begin{subfigure}
         \centering
         \includegraphics[height=3cm]{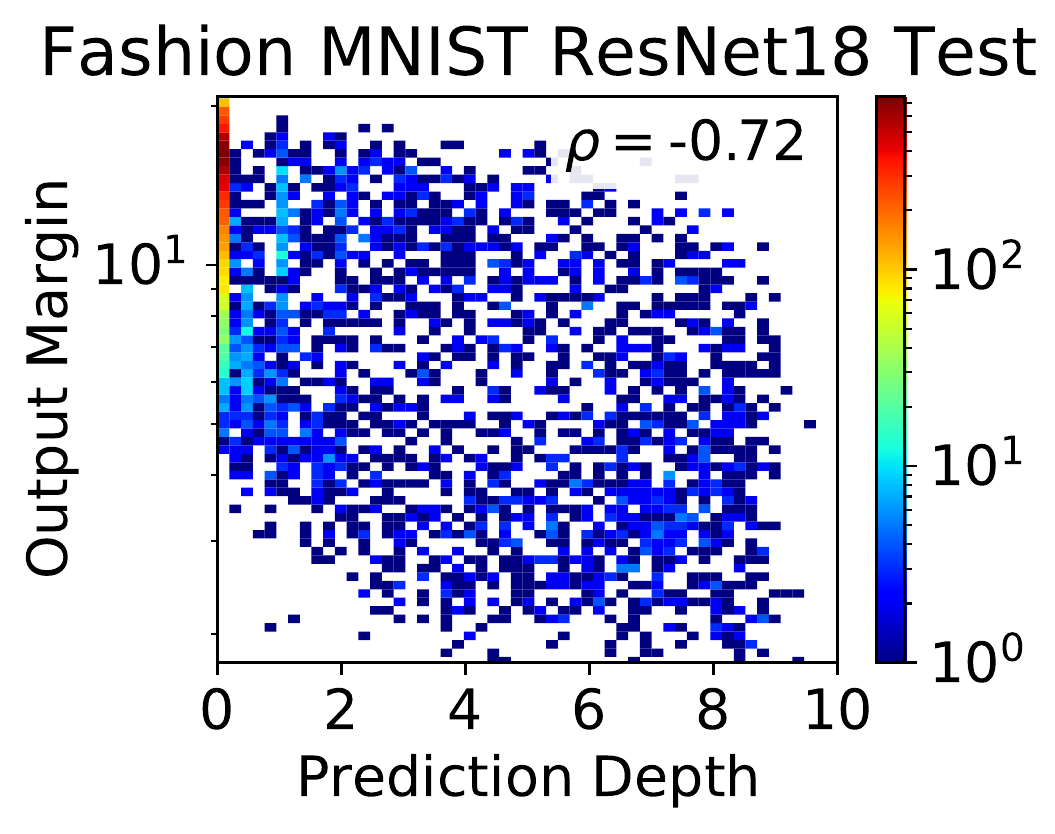}
\end{subfigure}
\begin{subfigure}
         \centering
         \includegraphics[height=3cm]{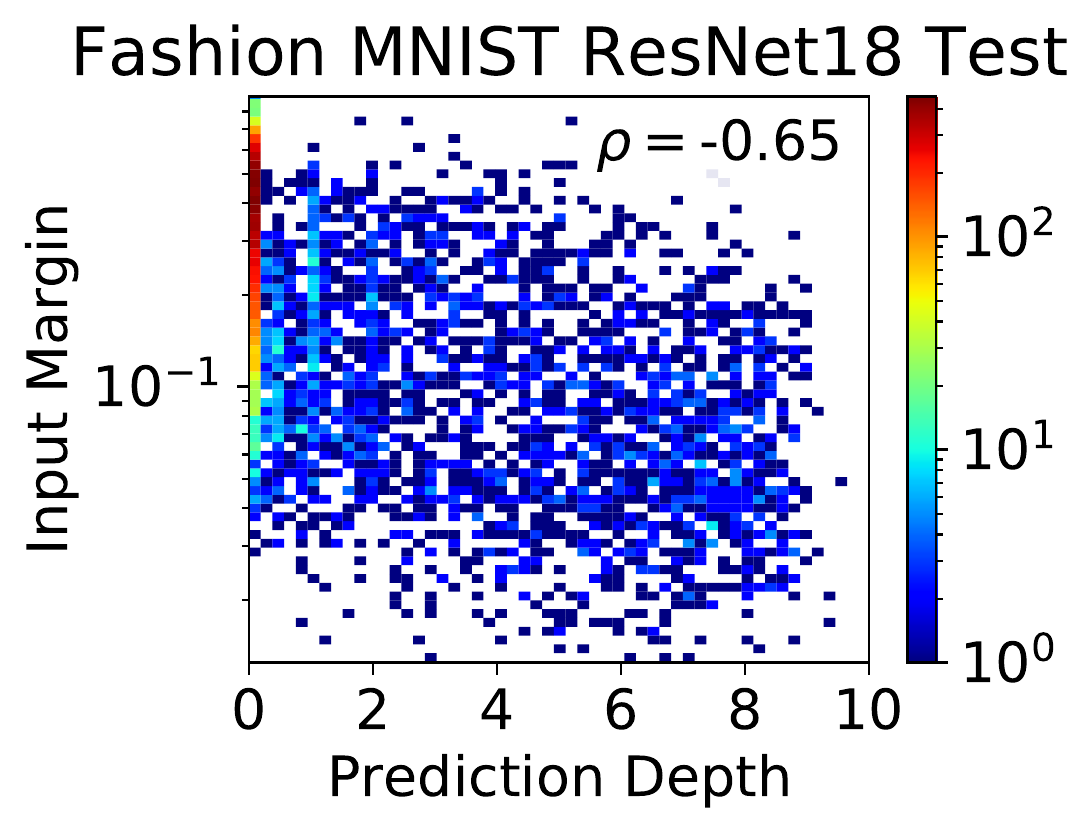}
\end{subfigure}
\begin{subfigure}
         \centering
         \includegraphics[height=3cm]{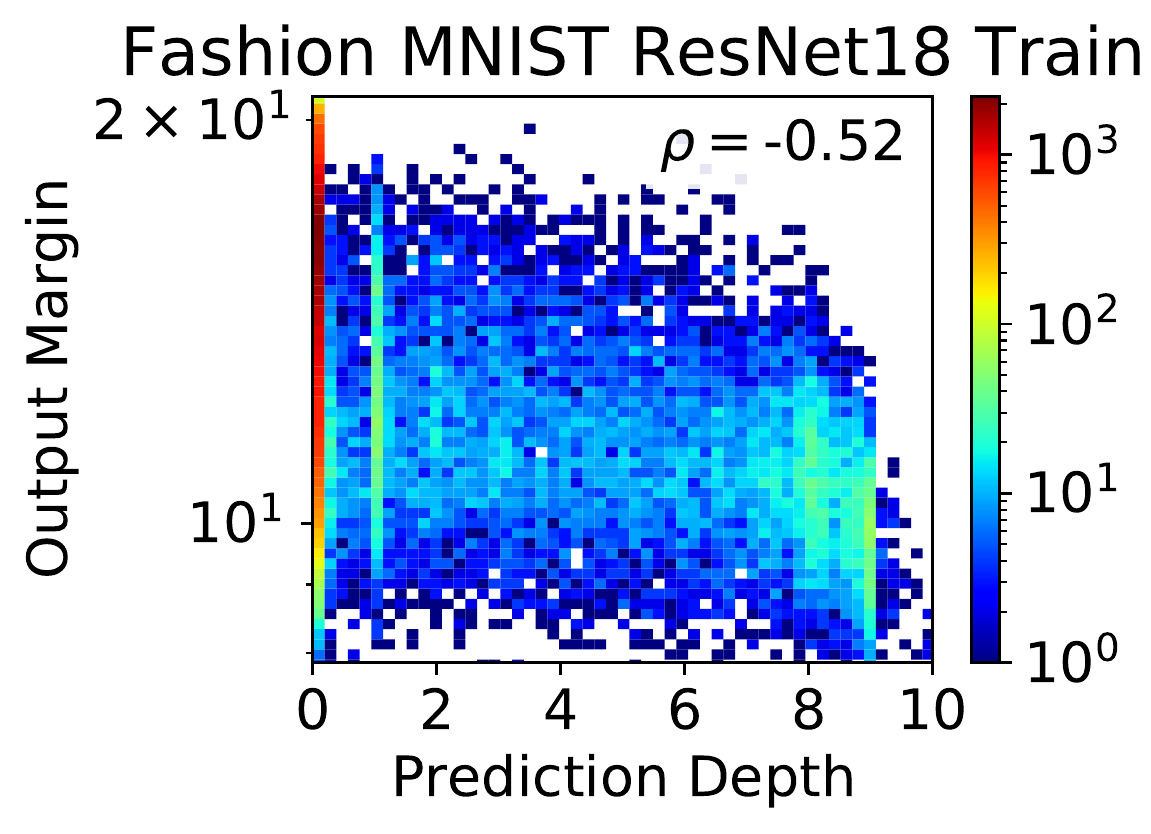}
\end{subfigure}
\begin{subfigure}
         \centering
         \includegraphics[height=3cm]{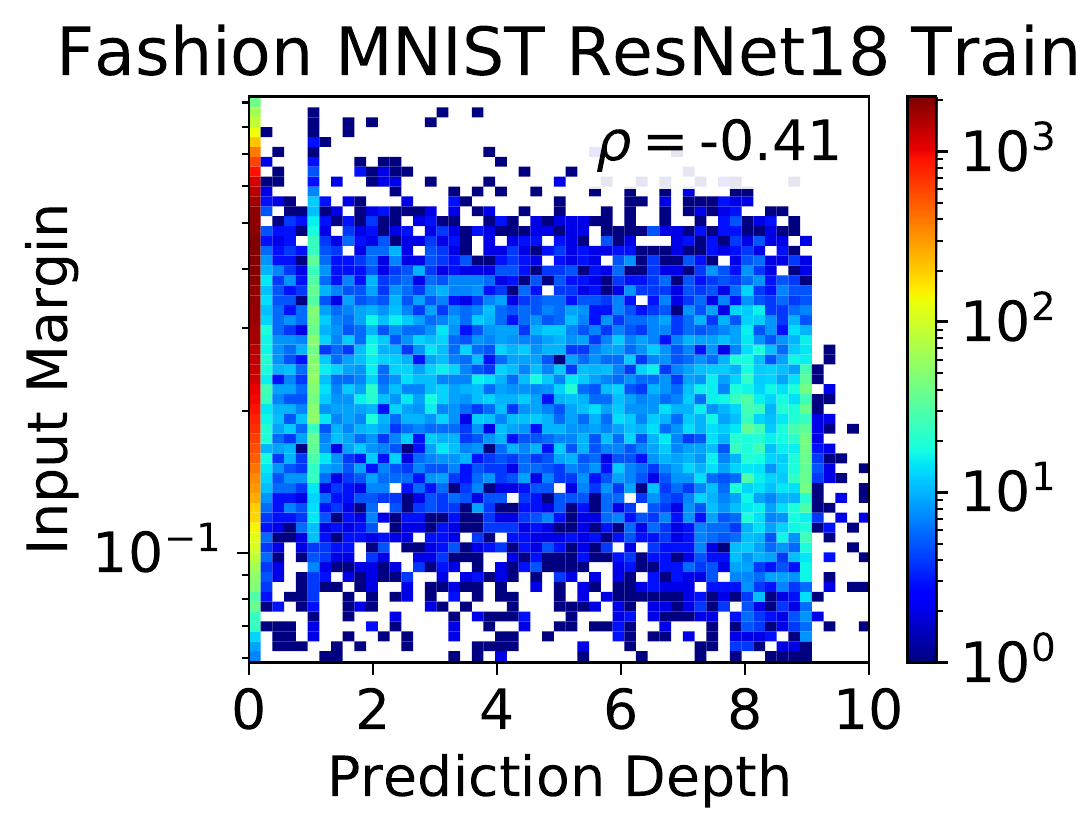}
\end{subfigure}}

\resizebox{1.\textwidth}{!}{%
\begin{subfigure}
         \centering
         \includegraphics[height=3cm]{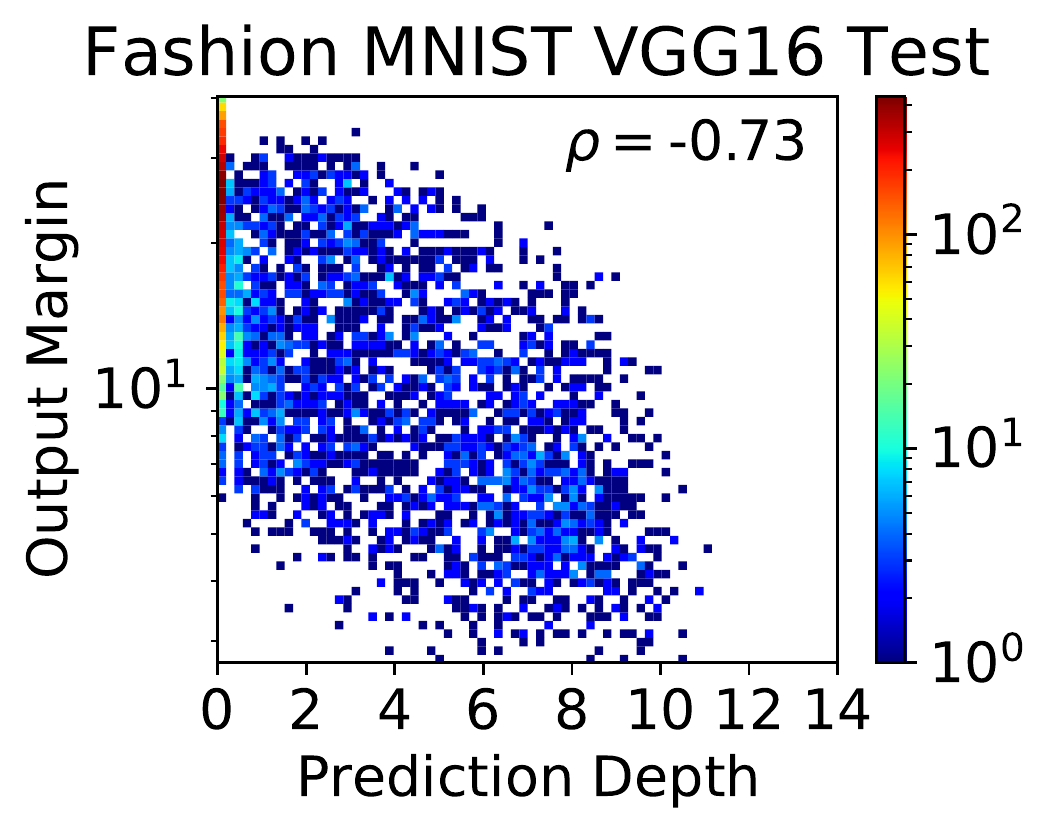}
\end{subfigure}
\begin{subfigure}
         \centering
         \includegraphics[height=3cm]{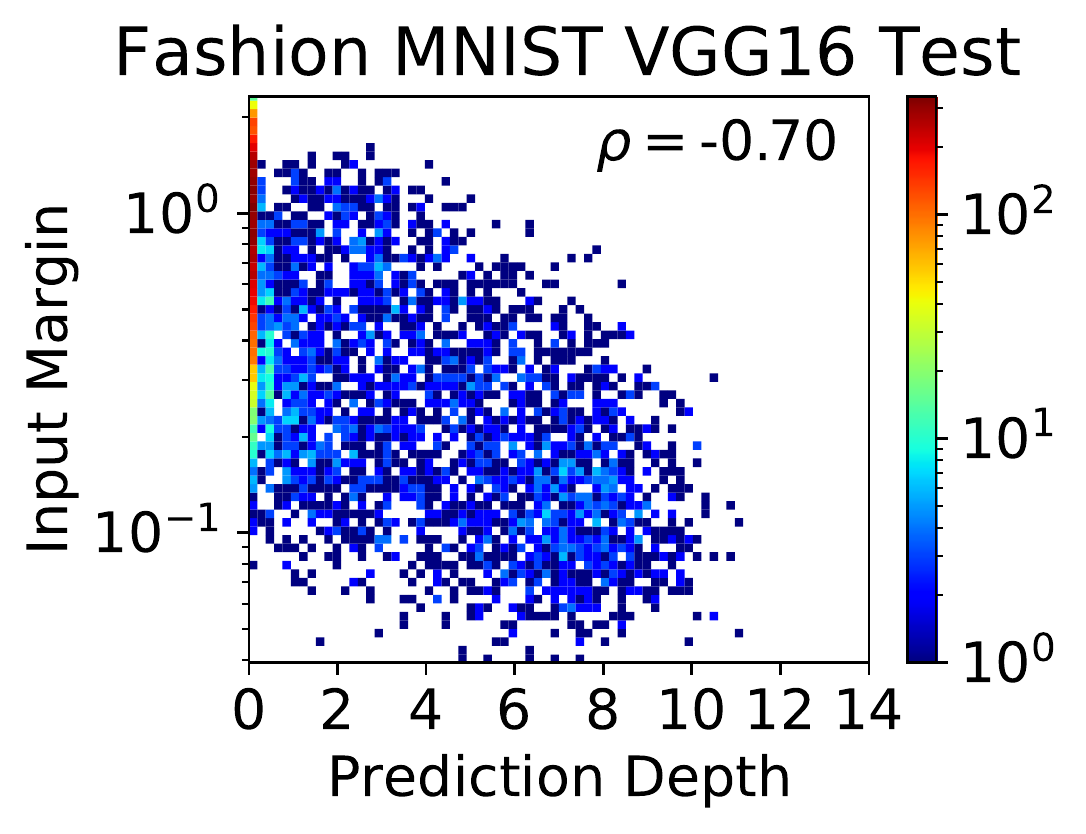}
\end{subfigure}
\begin{subfigure}
         \centering
         \includegraphics[height=3cm]{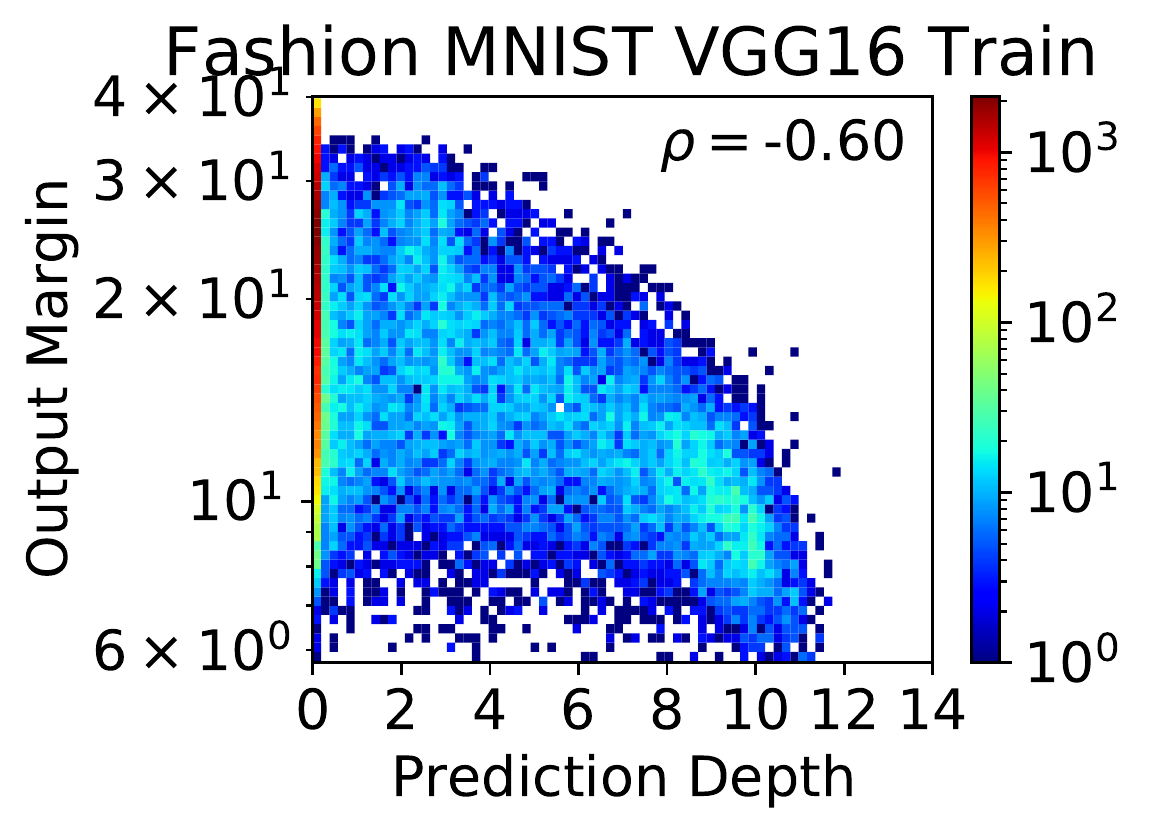}
\end{subfigure}
\begin{subfigure}
         \centering
         \includegraphics[height=3cm]{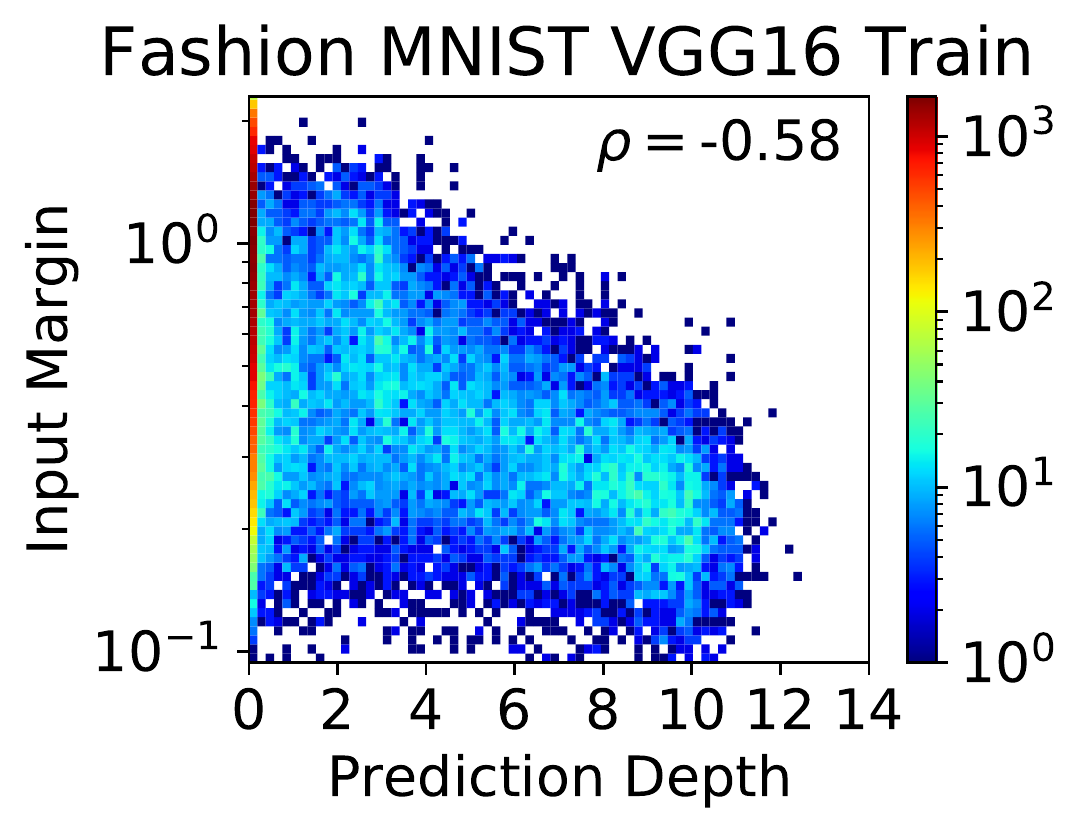}
\end{subfigure}}

\resizebox{1.\textwidth}{!}{%
\begin{subfigure}
         \centering
         \includegraphics[height=3cm]{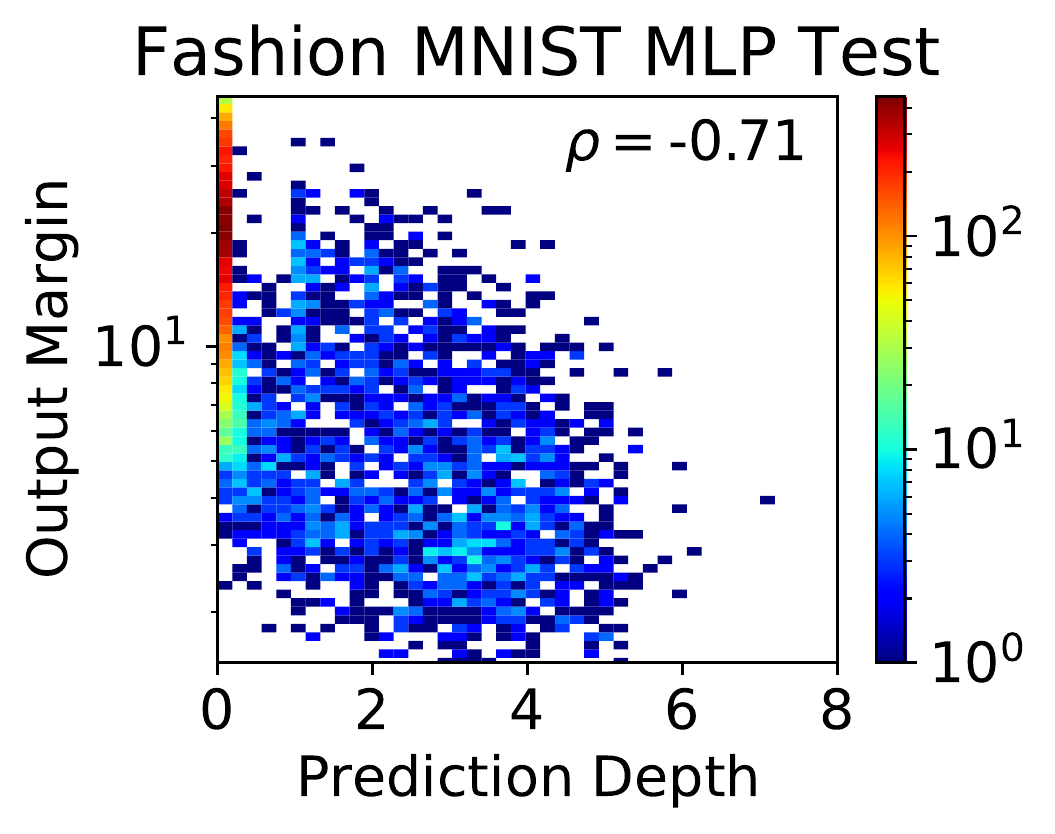}
\end{subfigure}
\begin{subfigure}
         \centering
         \includegraphics[height=3cm]{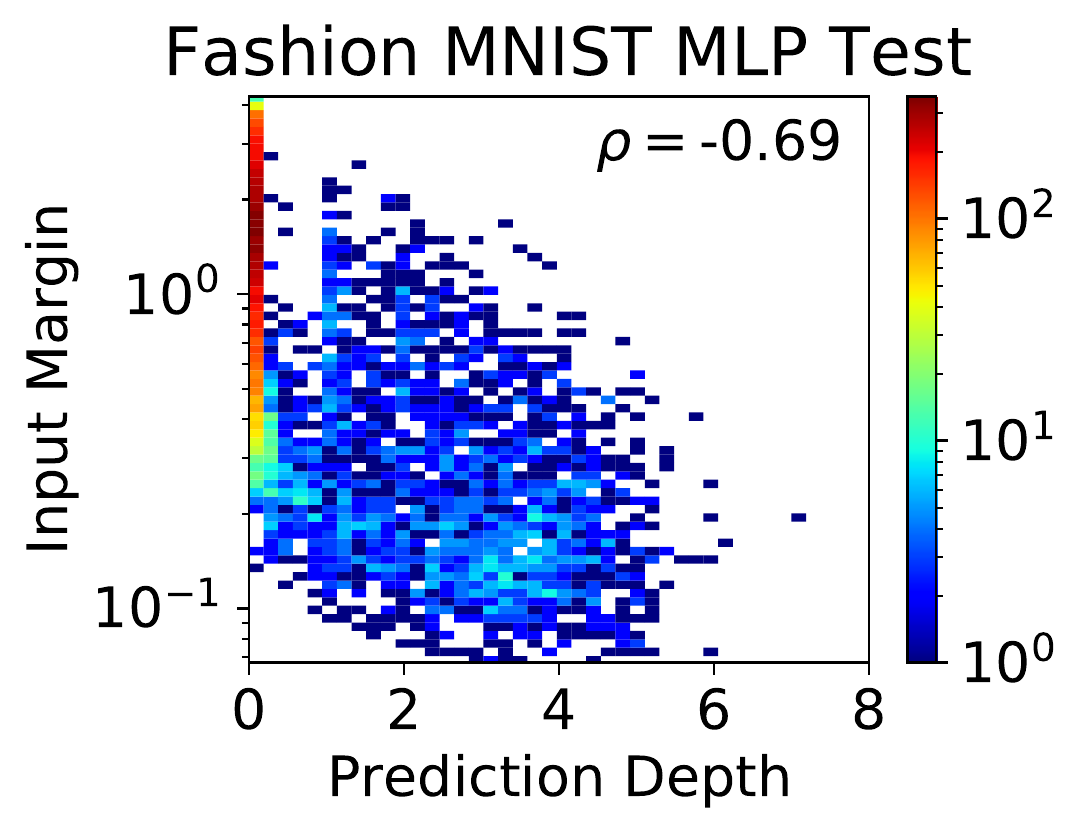}
\end{subfigure}
\begin{subfigure}
         \centering
         \includegraphics[height=3cm]{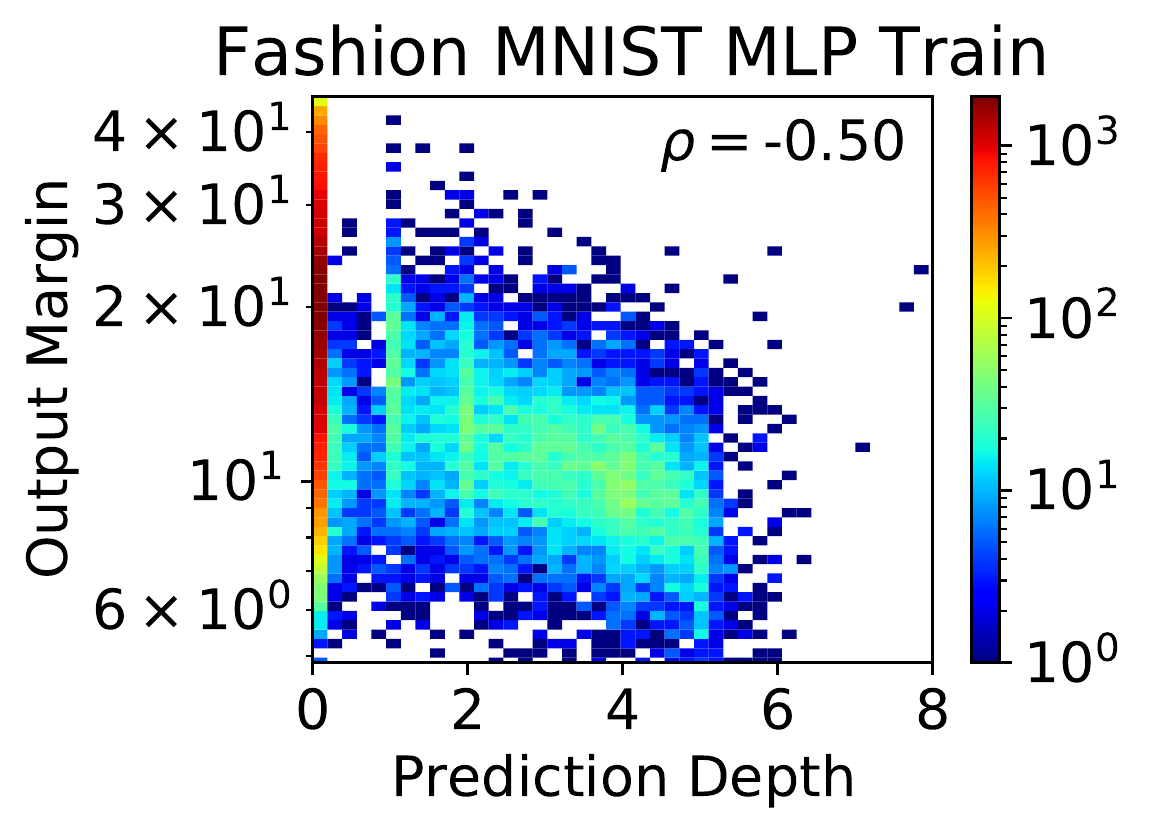}
\end{subfigure}
\begin{subfigure}
         \centering
         \includegraphics[height=3cm]{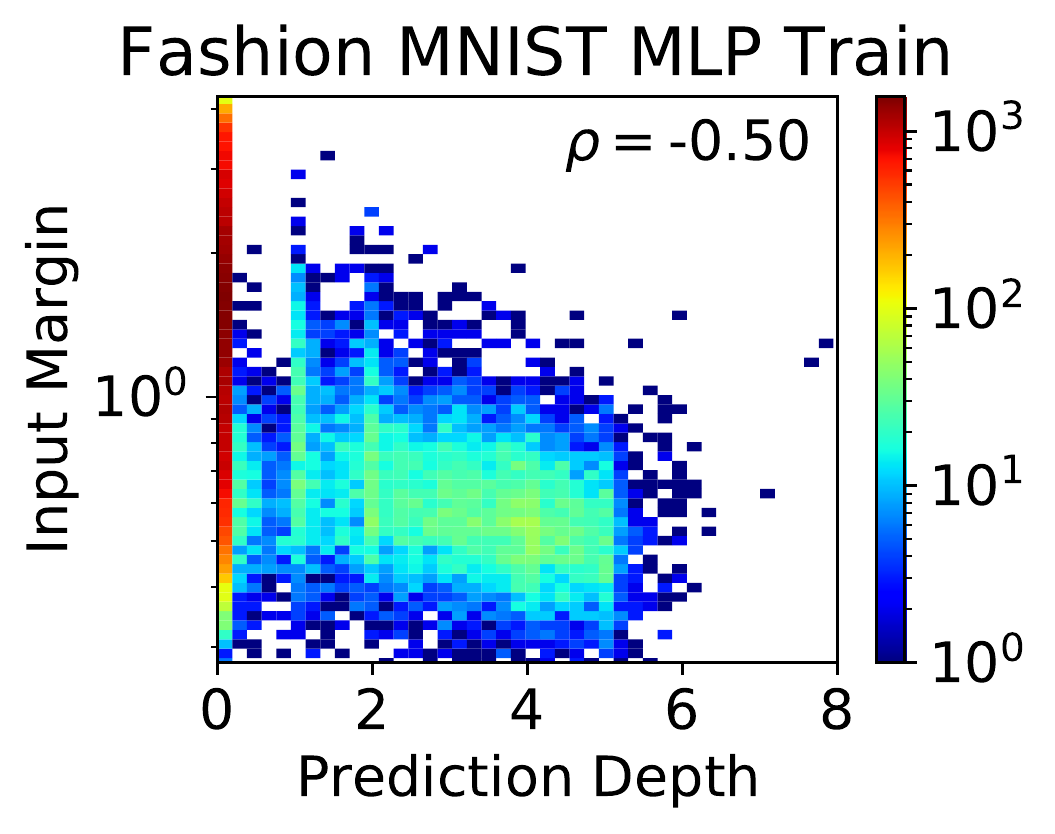}
\end{subfigure}}

\end{center}
\caption{Consistency of Figure~\ref{fig:margins_vs_depth}, showing the correlation between prediction depth, and the input and output margins (log scale) for both the test and training splits of Fashion MNIST. The correlation coefficient between the prediction depth and the logarithm of the margin is given in each plot. For each architecture, we train 25 models with different random seeds on the full training split. We record the input and output margins together with the prediction depth for every data point in both the train and test splits. These histograms compare the mean values of each margin to the mean prediction depth for all data points.
\label{fig:fmnist_marg_consist}}
\end{figure}

\begin{figure}[ht!]
\begin{center}
\resizebox{1.\textwidth}{!}{%
\begin{subfigure}
         \centering
         \includegraphics[height=3cm]{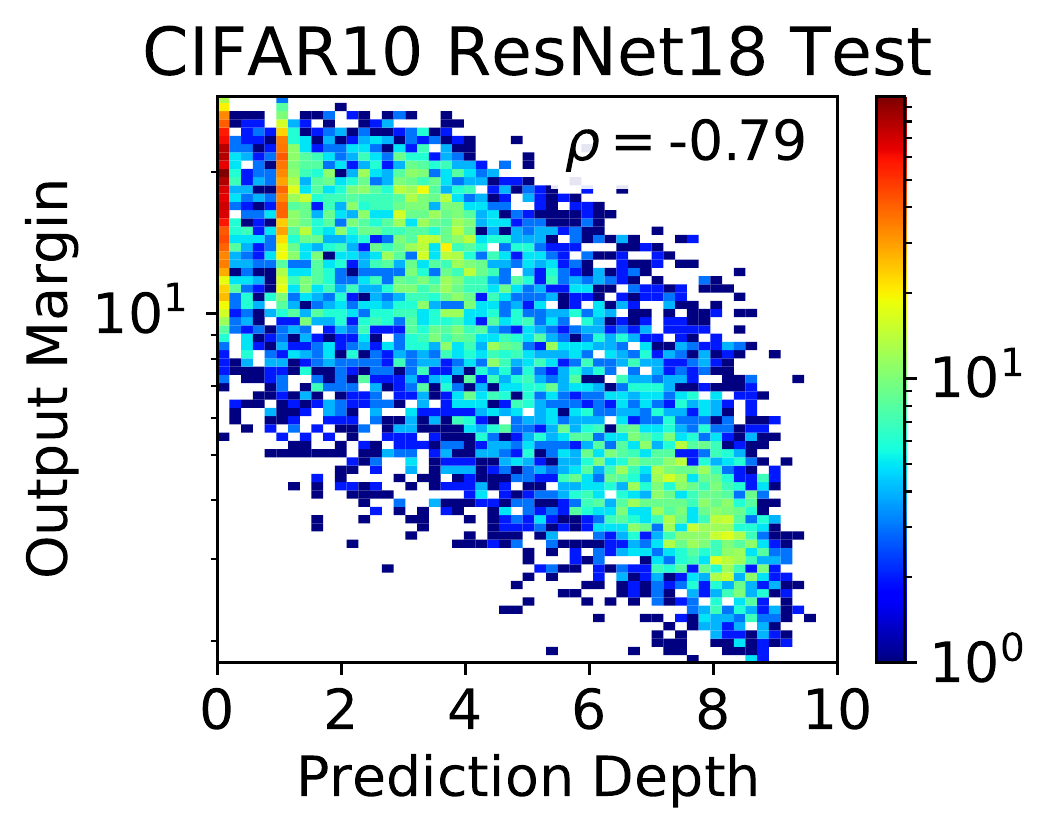}
\end{subfigure}
\begin{subfigure}
         \centering
         \includegraphics[height=3cm]{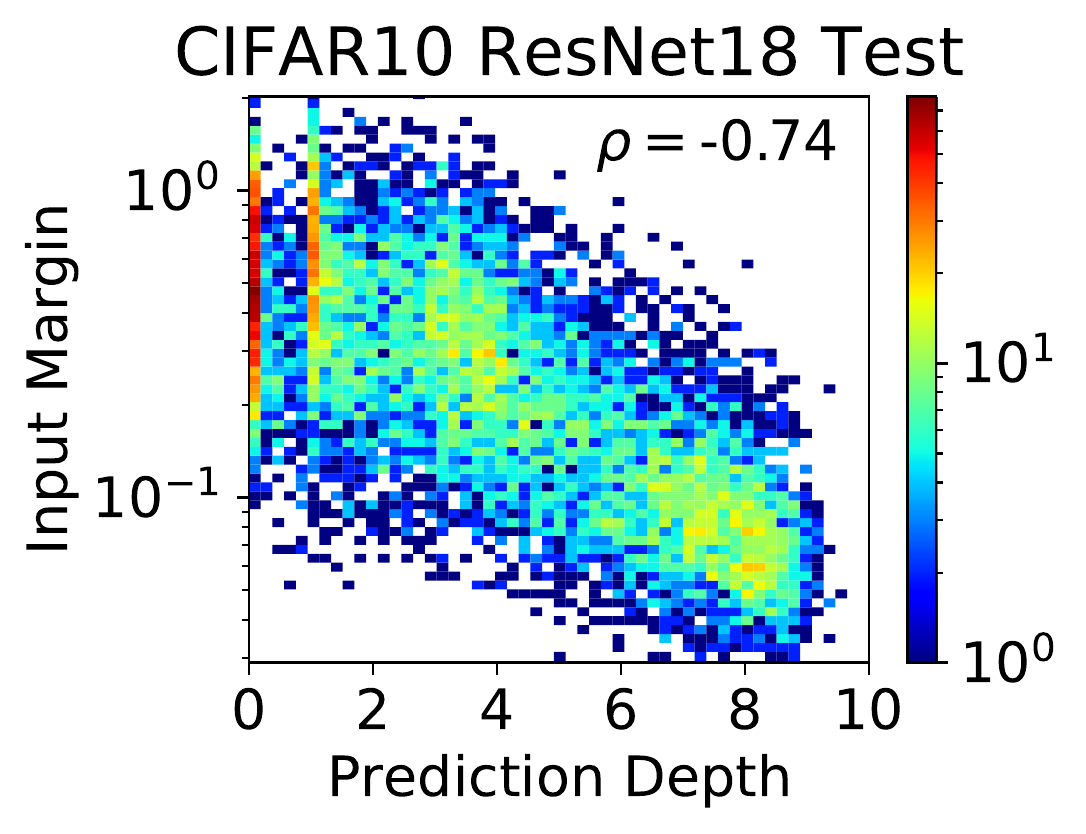}
\end{subfigure}
\begin{subfigure}
         \centering
         \includegraphics[height=3cm]{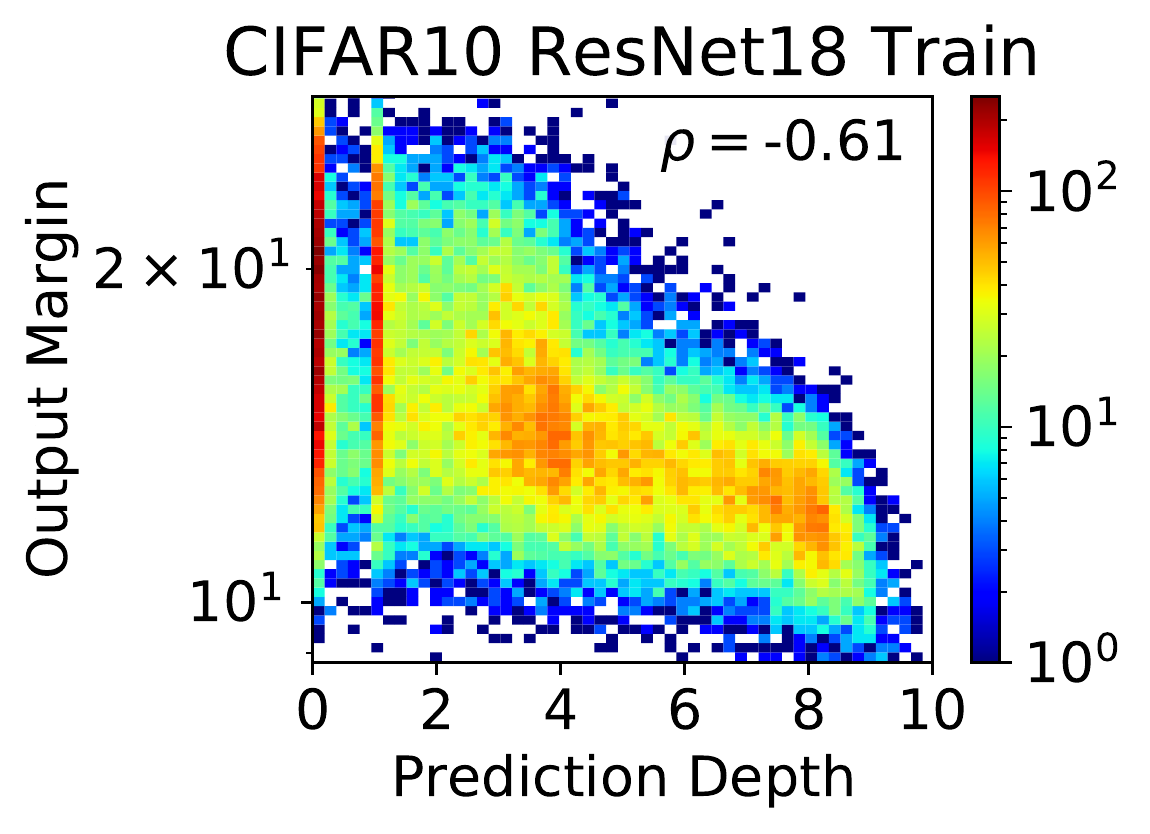}
\end{subfigure}
\begin{subfigure}
         \centering
         \includegraphics[height=3cm]{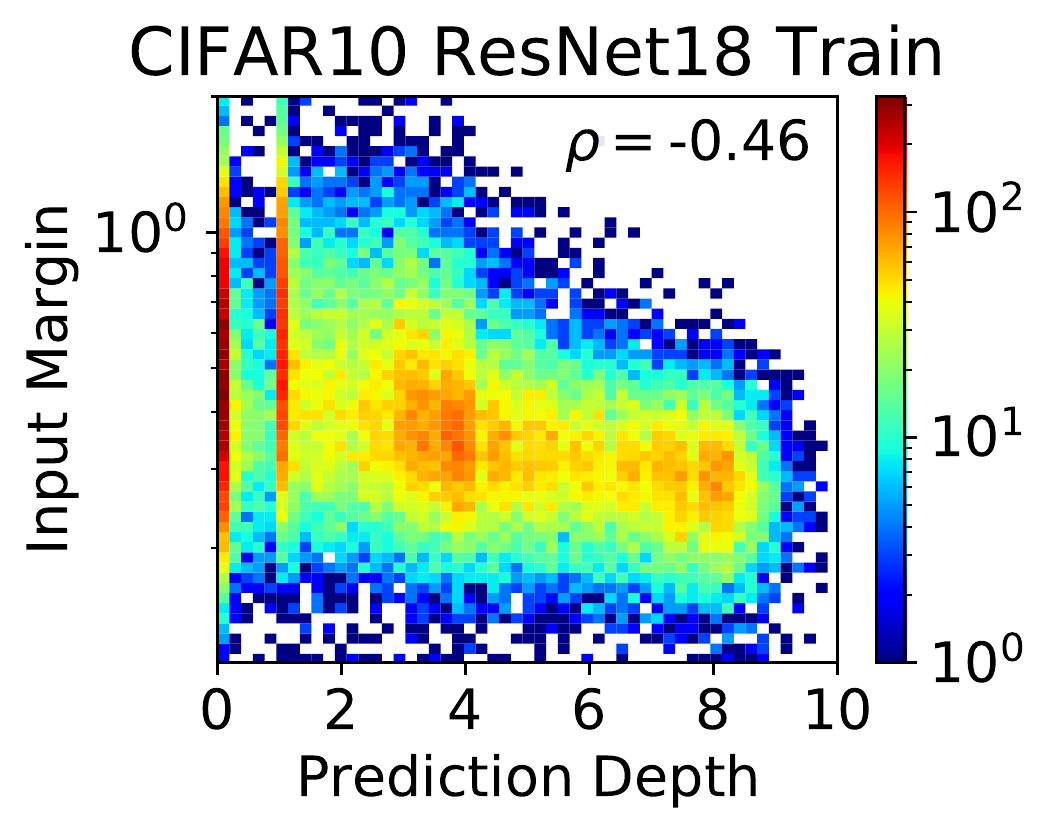}
\end{subfigure}}

\resizebox{1.\textwidth}{!}{%
\begin{subfigure}
         \centering
         \includegraphics[height=3cm]{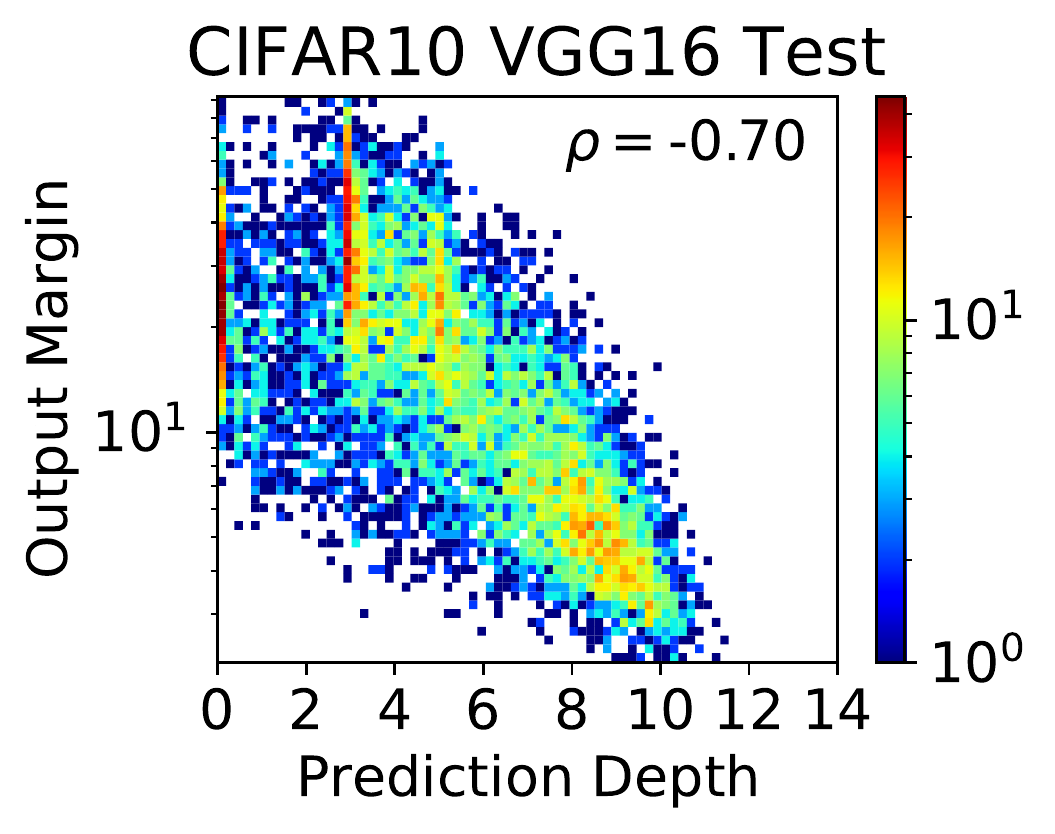}
\end{subfigure}
\begin{subfigure}
         \centering
         \includegraphics[height=3cm]{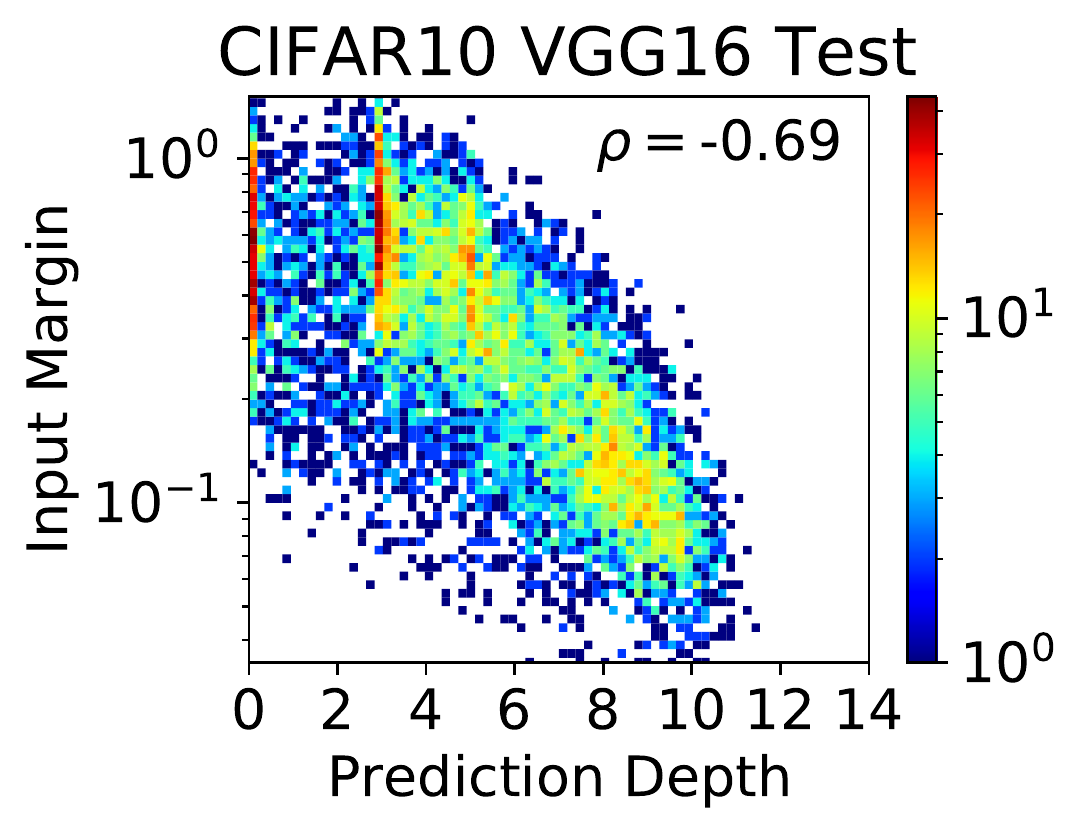}
\end{subfigure}
\begin{subfigure}
         \centering
         \includegraphics[height=3cm]{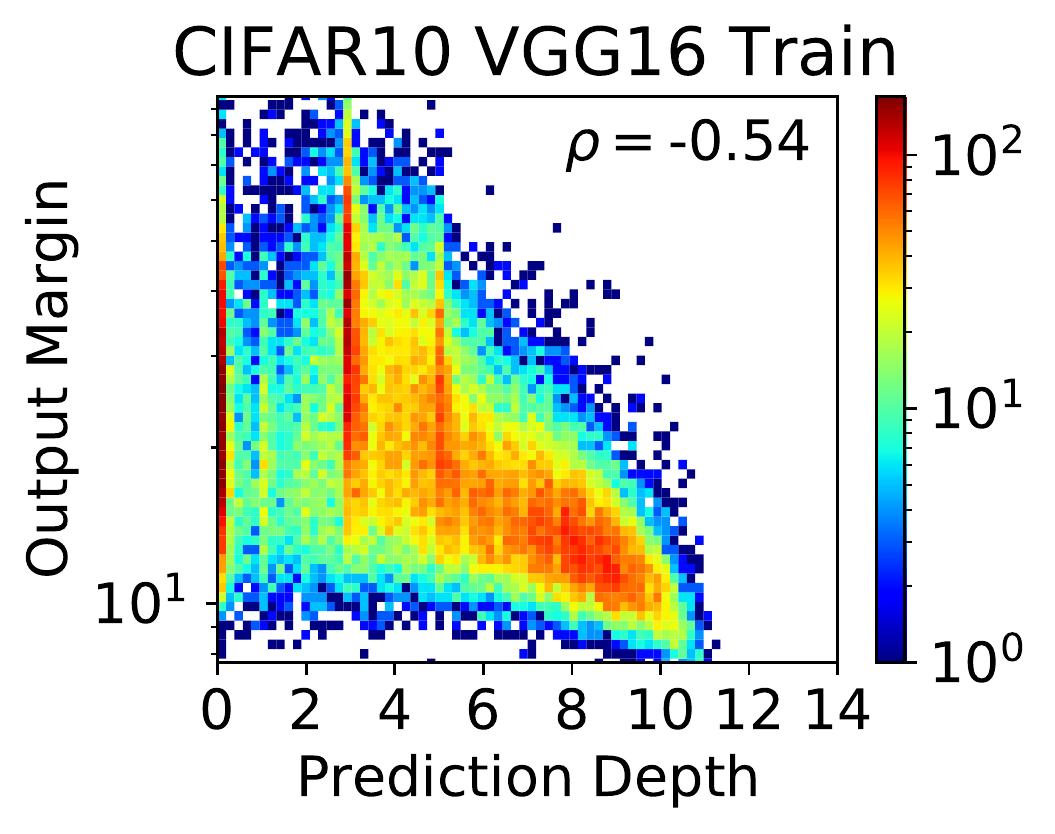}
\end{subfigure}
\begin{subfigure}
         \centering
         \includegraphics[height=3cm]{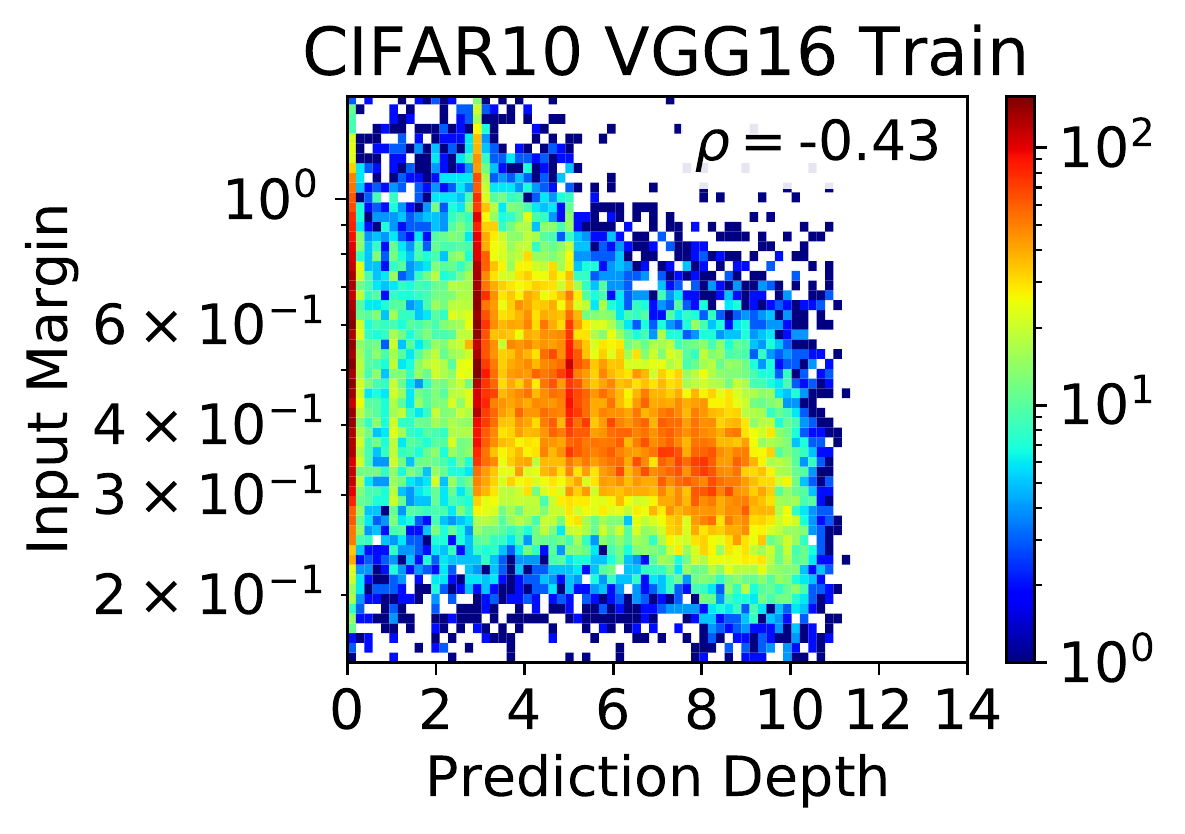}
\end{subfigure}}

\resizebox{1.\textwidth}{!}{%
\begin{subfigure}
         \centering
         \includegraphics[height=3cm]{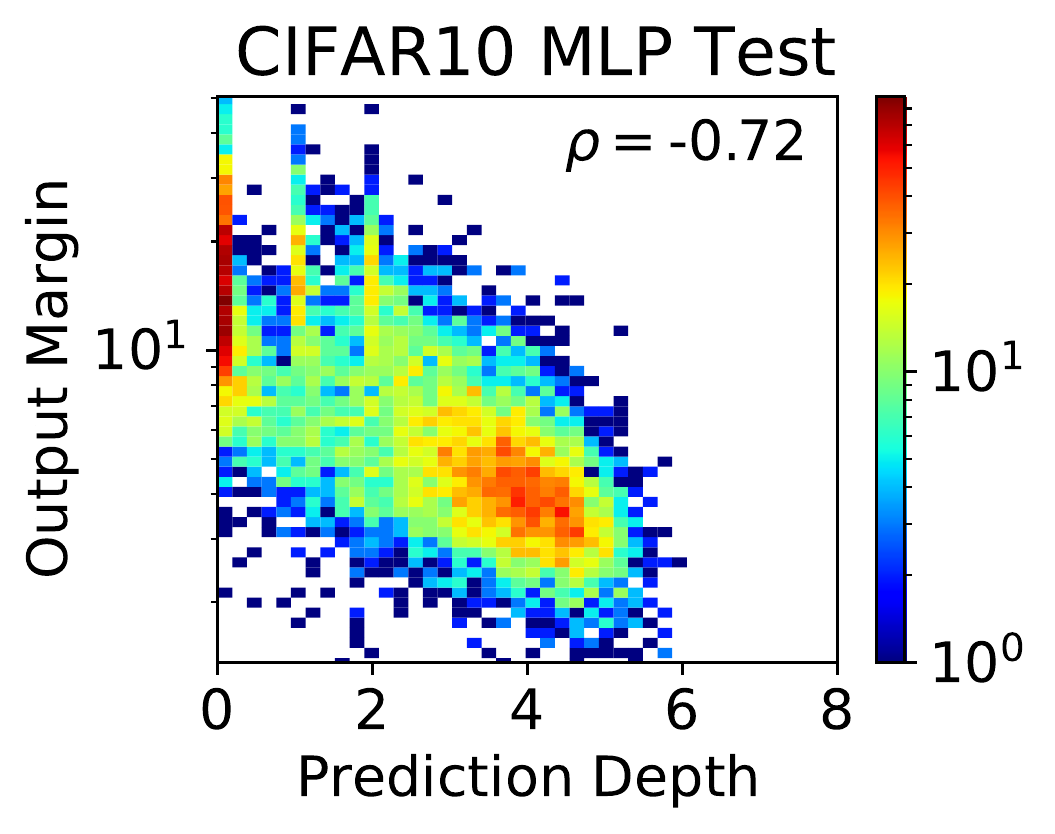}
\end{subfigure}
\begin{subfigure}
         \centering
         \includegraphics[height=3cm]{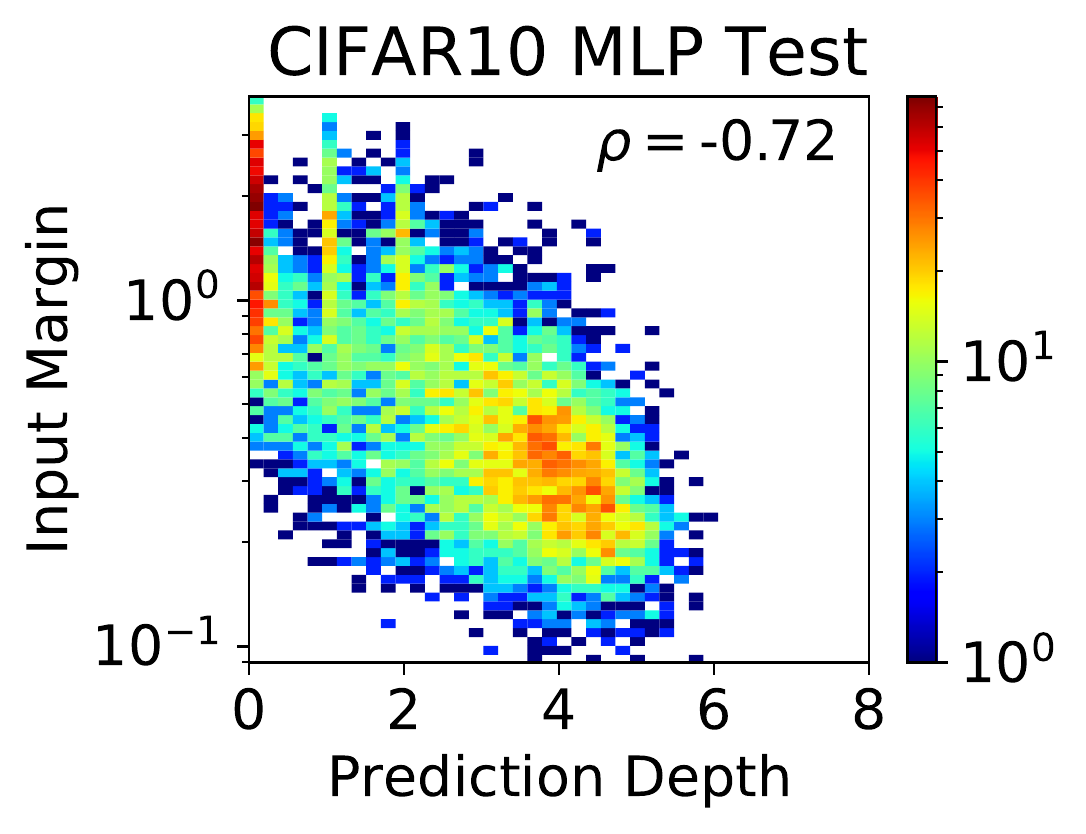}
\end{subfigure}
\begin{subfigure}
         \centering
         \includegraphics[height=3cm]{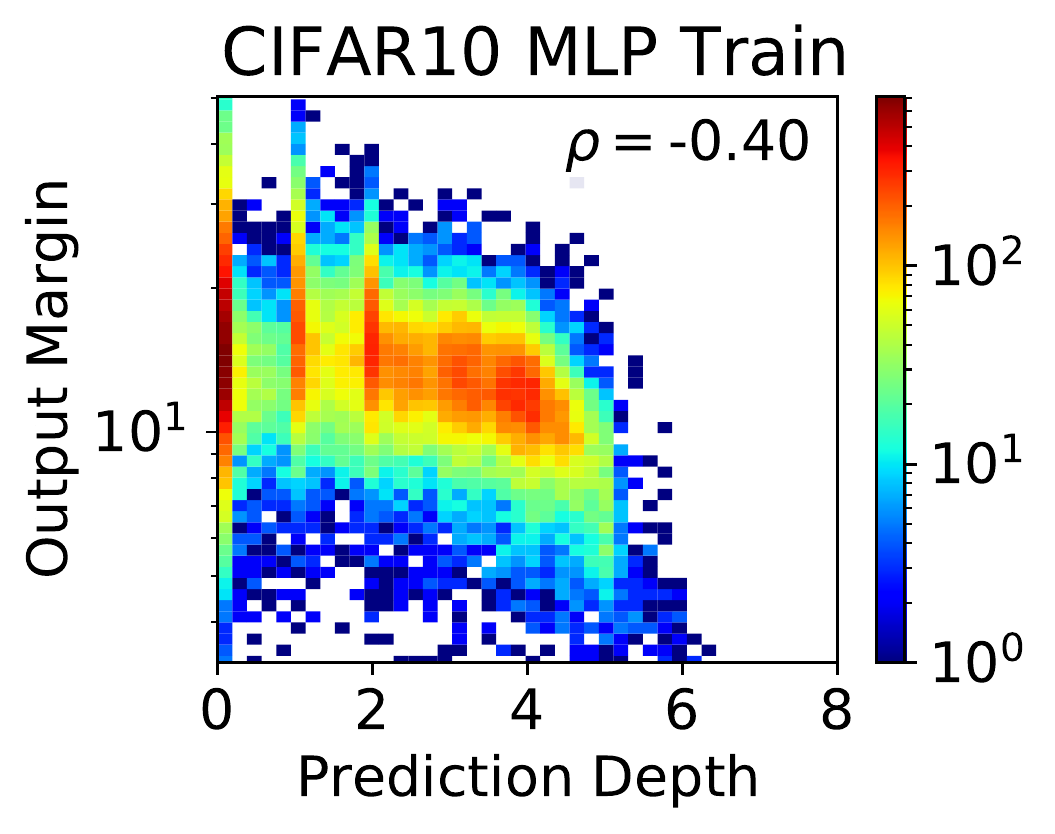}
\end{subfigure}
\begin{subfigure}
         \centering
         \includegraphics[height=3cm]{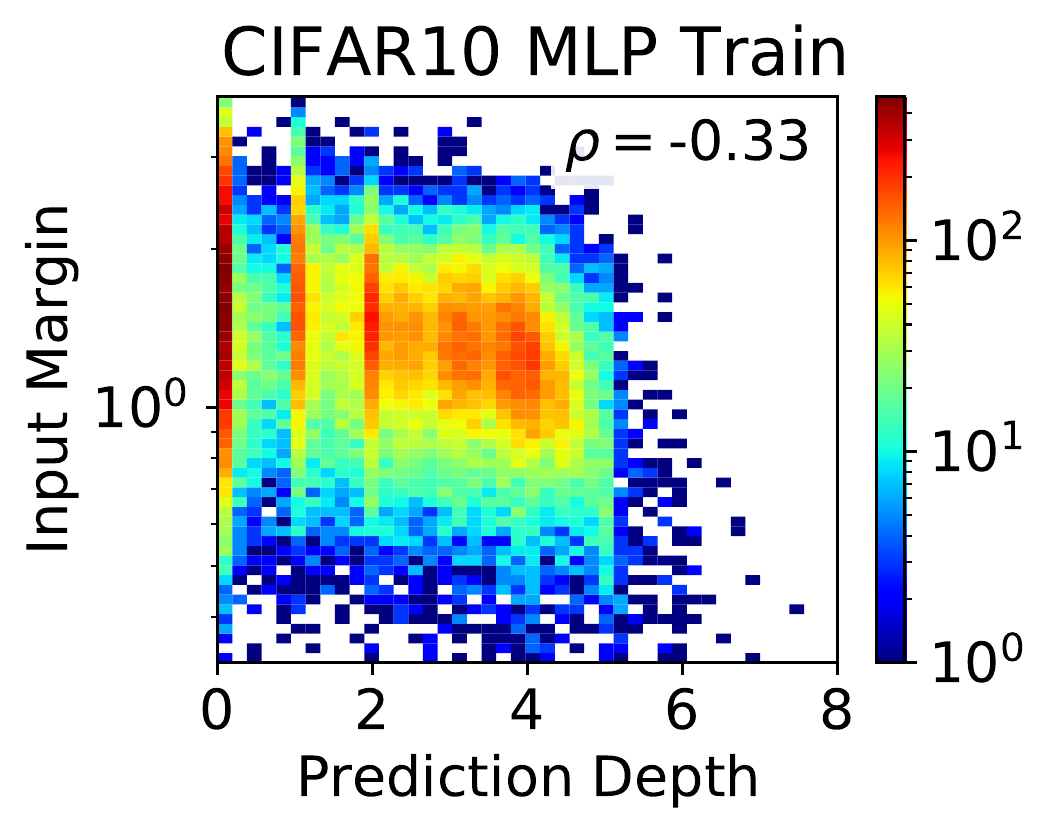}
\end{subfigure}}

\end{center}
\caption{Consistency of Figure~\ref{fig:margins_vs_depth}, showing the correlation between prediction depth, and the input and output margins (log scale) for both the test and training splits of CIFAR10. The correlation coefficient between the prediction depth and the logarithm of the margin is given in each plot. For each architecture, we train 25 models with different random seeds on the full training split. We record the input and output margins together with the prediction depth for every data point in both the train and test splits. These histograms compare the mean values of each margin to the mean prediction depth for all data points.
\label{fig:cifar10_marg_consist}}
\end{figure}

\begin{figure}[ht!]
\begin{center}
\resizebox{1.\textwidth}{!}{%
\begin{subfigure}
         \centering
         \includegraphics[height=3cm]{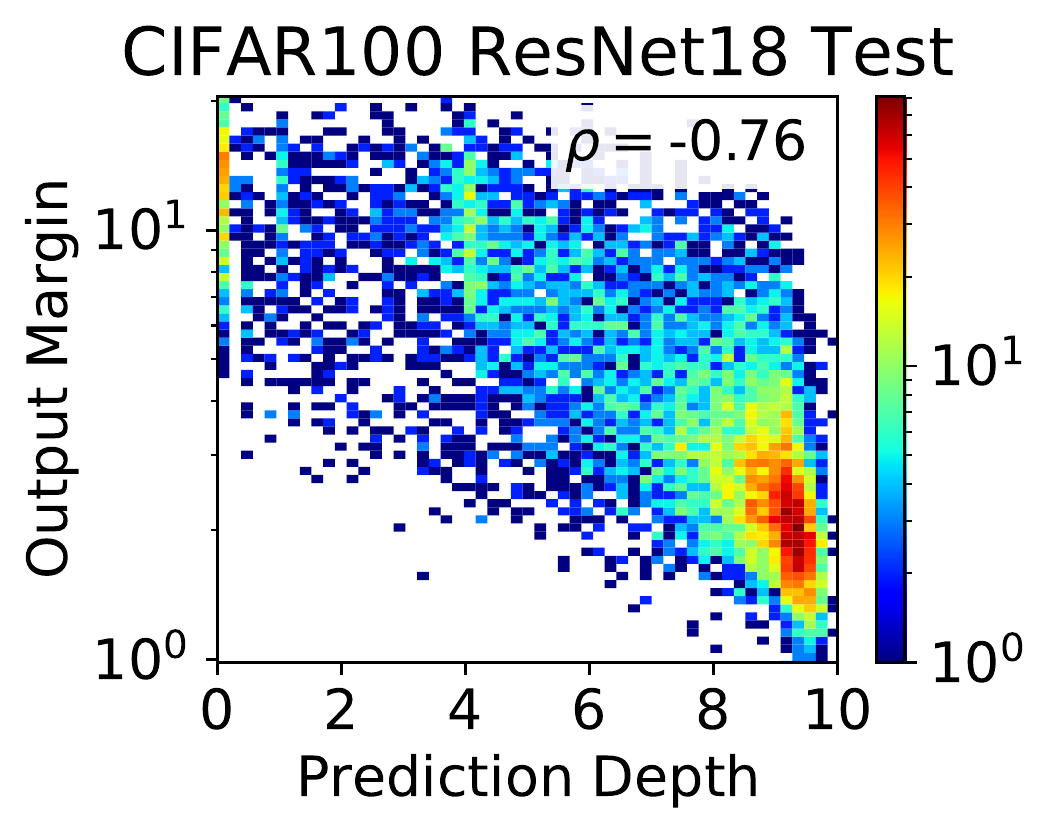}
\end{subfigure}
\begin{subfigure}
         \centering
         \includegraphics[height=3cm]{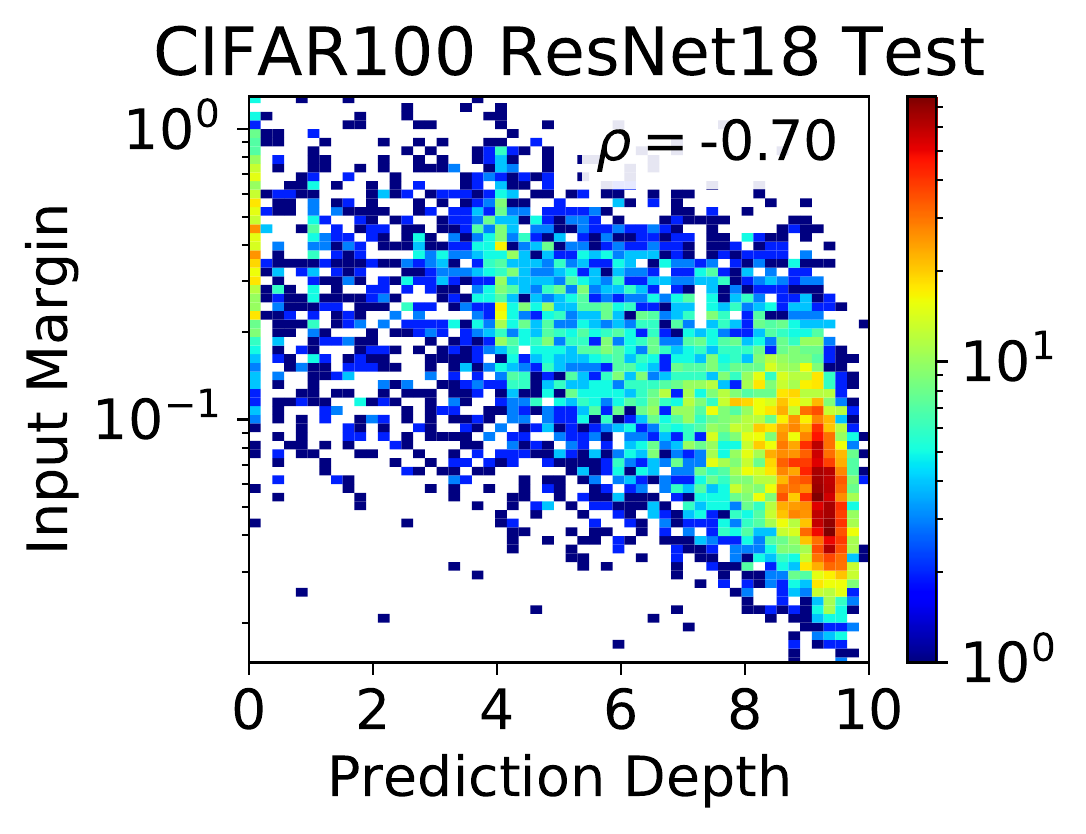}
\end{subfigure}
\begin{subfigure}
         \centering
         \includegraphics[height=3cm]{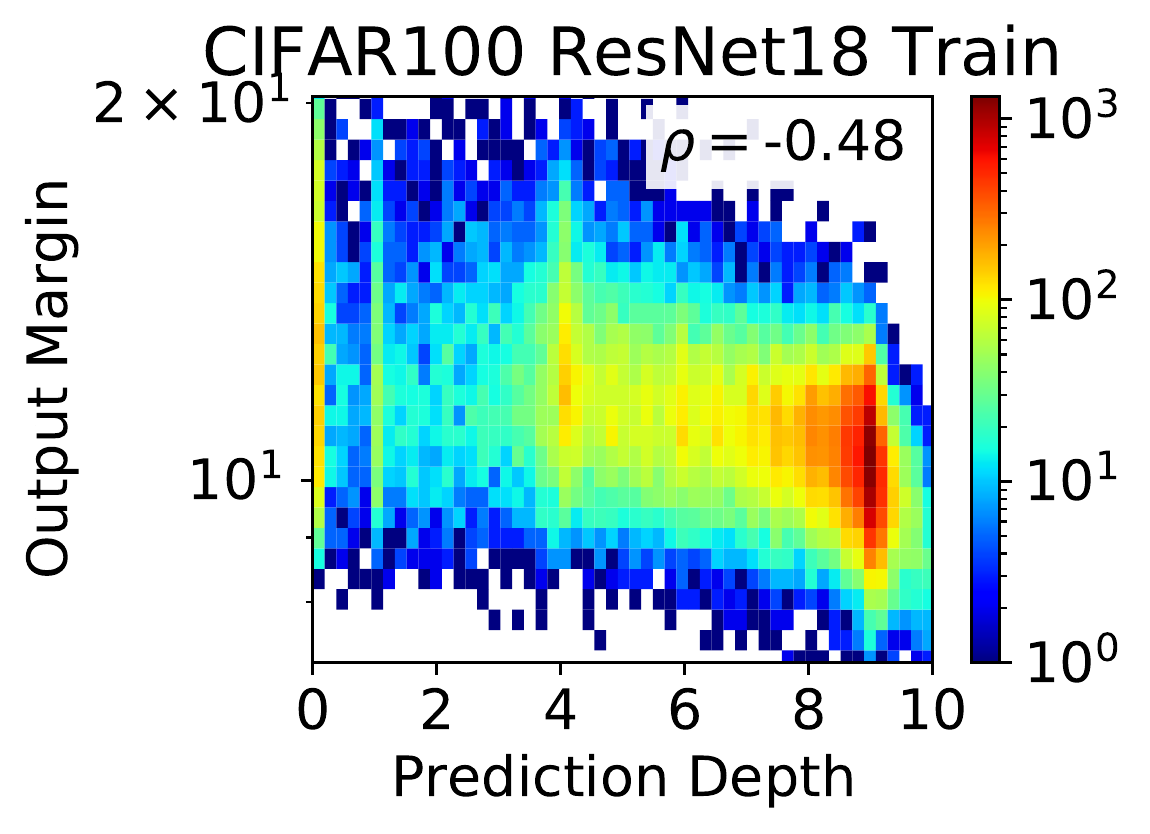}
\end{subfigure}
\begin{subfigure}
         \centering
         \includegraphics[height=3cm]{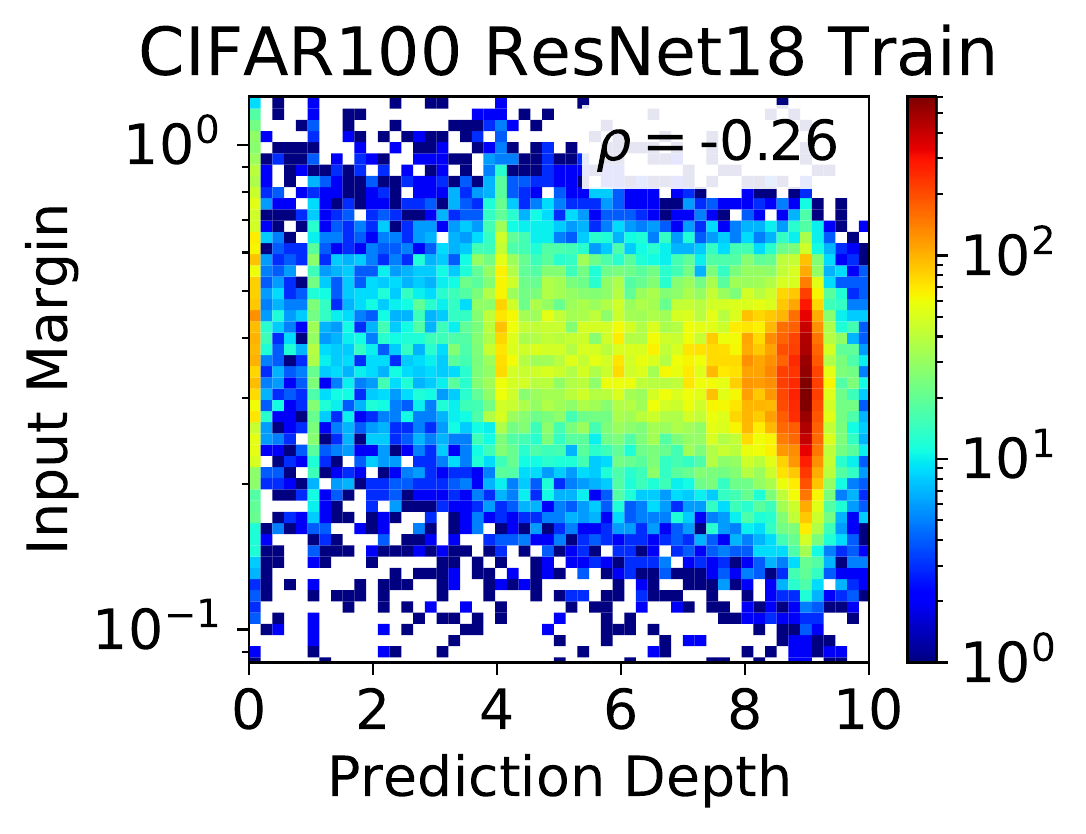}
\end{subfigure}}

\resizebox{1.\textwidth}{!}{%
\begin{subfigure}
         \centering
         \includegraphics[height=3cm]{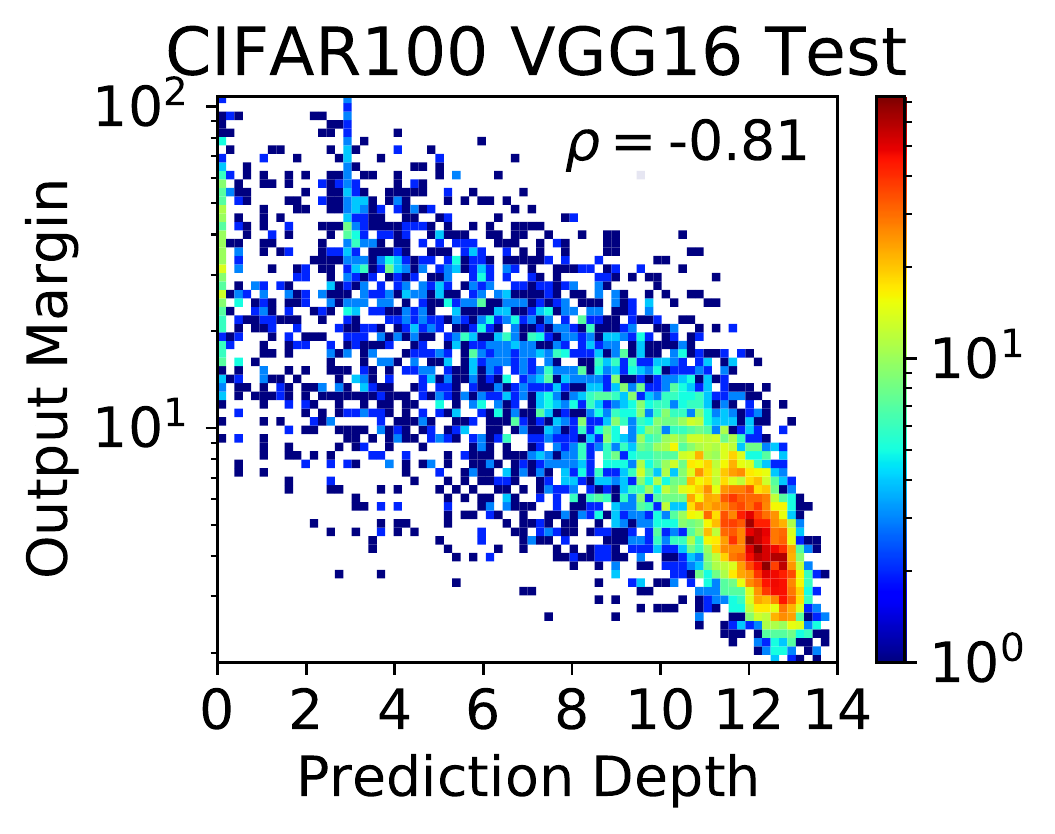}
\end{subfigure}
\begin{subfigure}
         \centering
         \includegraphics[height=3cm]{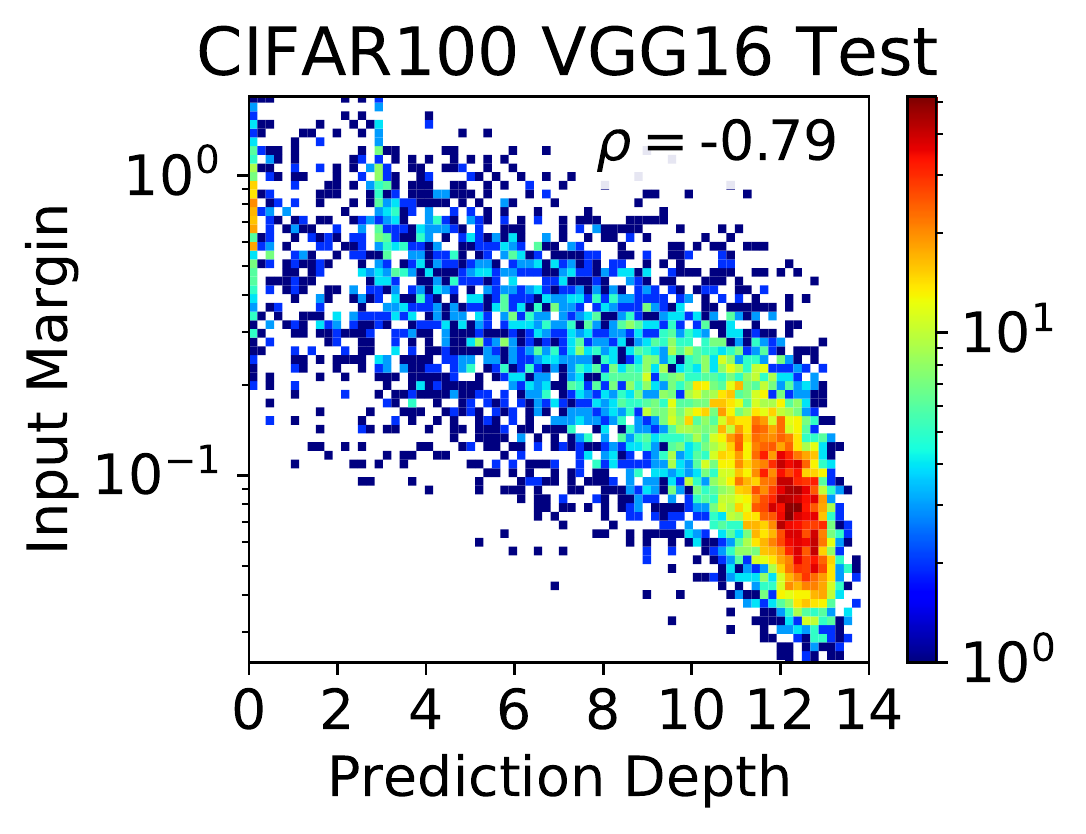}
\end{subfigure}
\begin{subfigure}
         \centering
         \includegraphics[height=3cm]{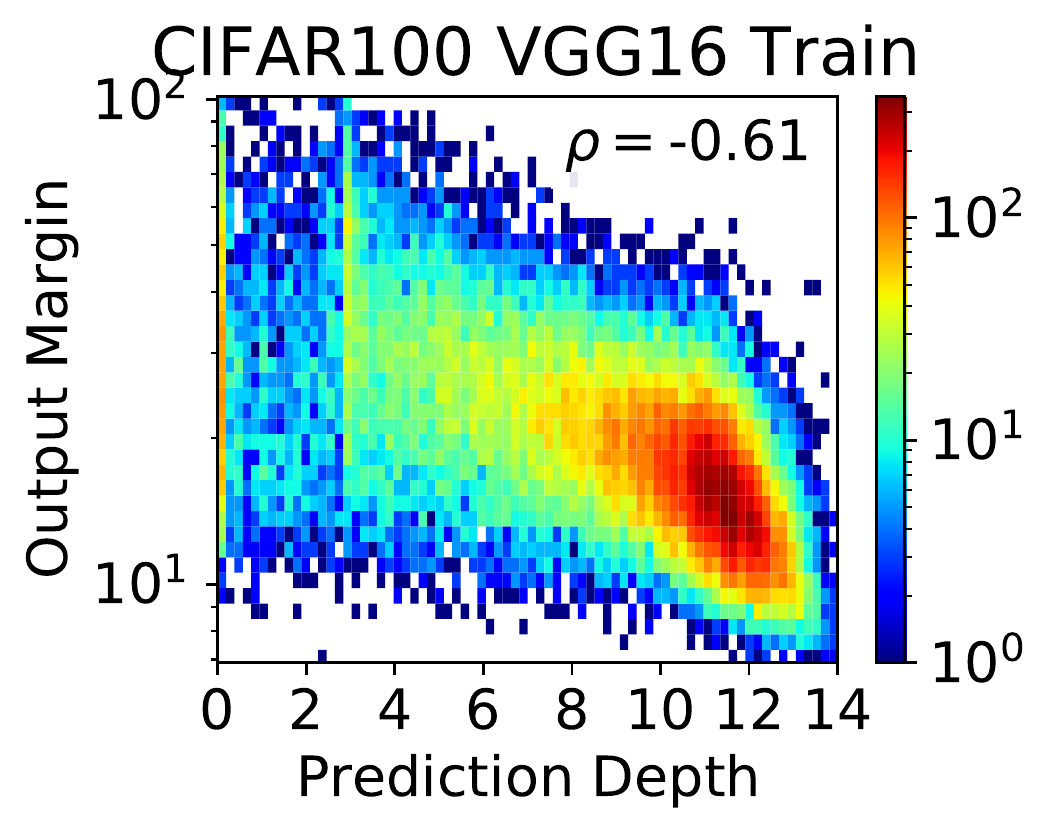}
\end{subfigure}
\begin{subfigure}
         \centering
         \includegraphics[height=3cm]{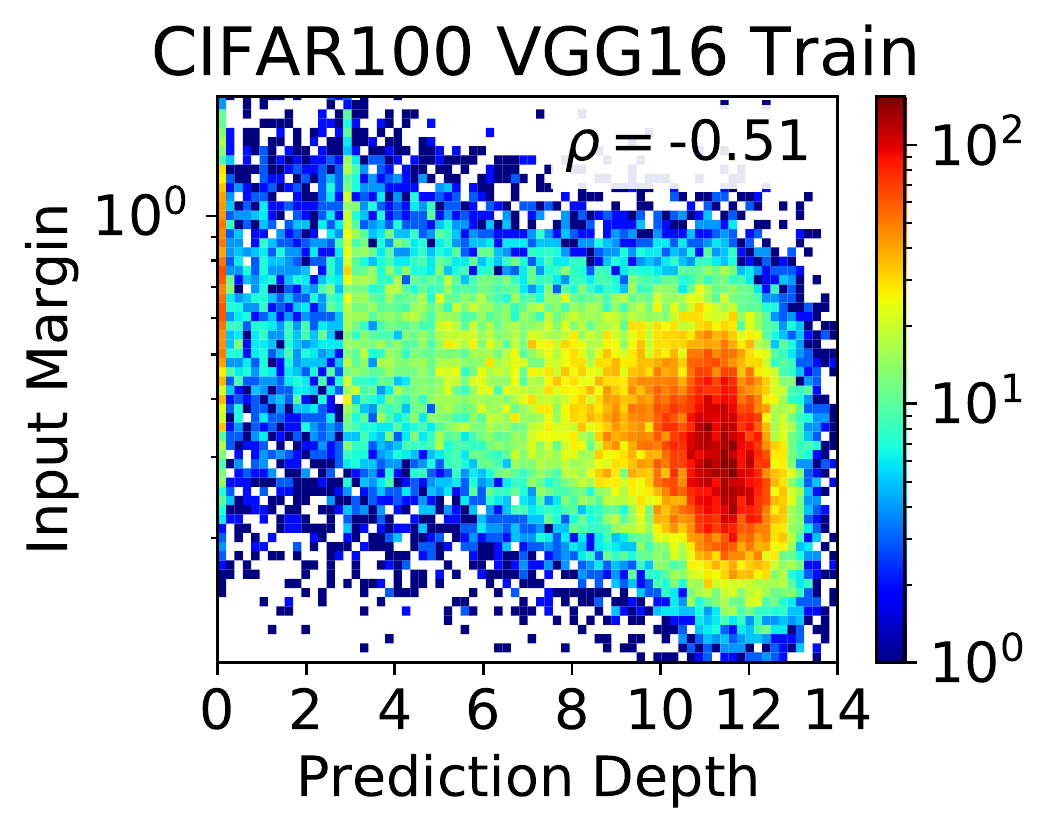}
\end{subfigure}}

\resizebox{1.\textwidth}{!}{%
\begin{subfigure}
         \centering
         \includegraphics[height=3cm]{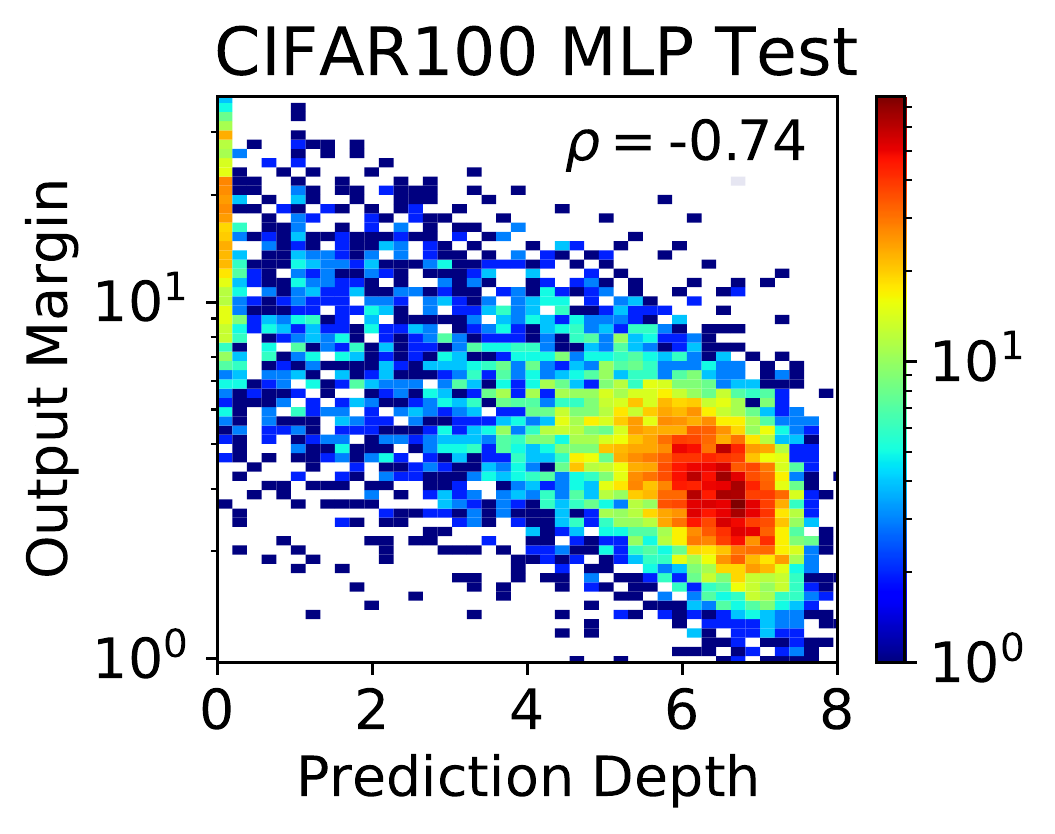}
\end{subfigure}
\begin{subfigure}
         \centering
         \includegraphics[height=3cm]{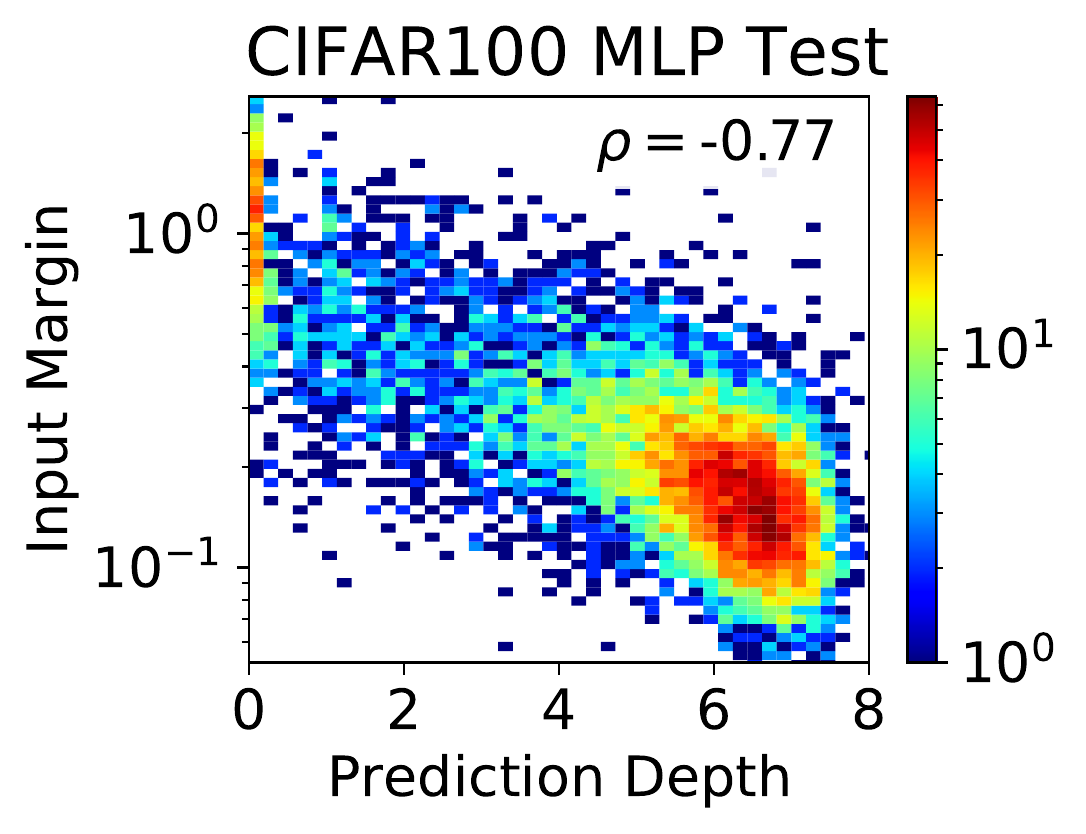}
\end{subfigure}
\begin{subfigure}
         \centering
         \includegraphics[height=3cm]{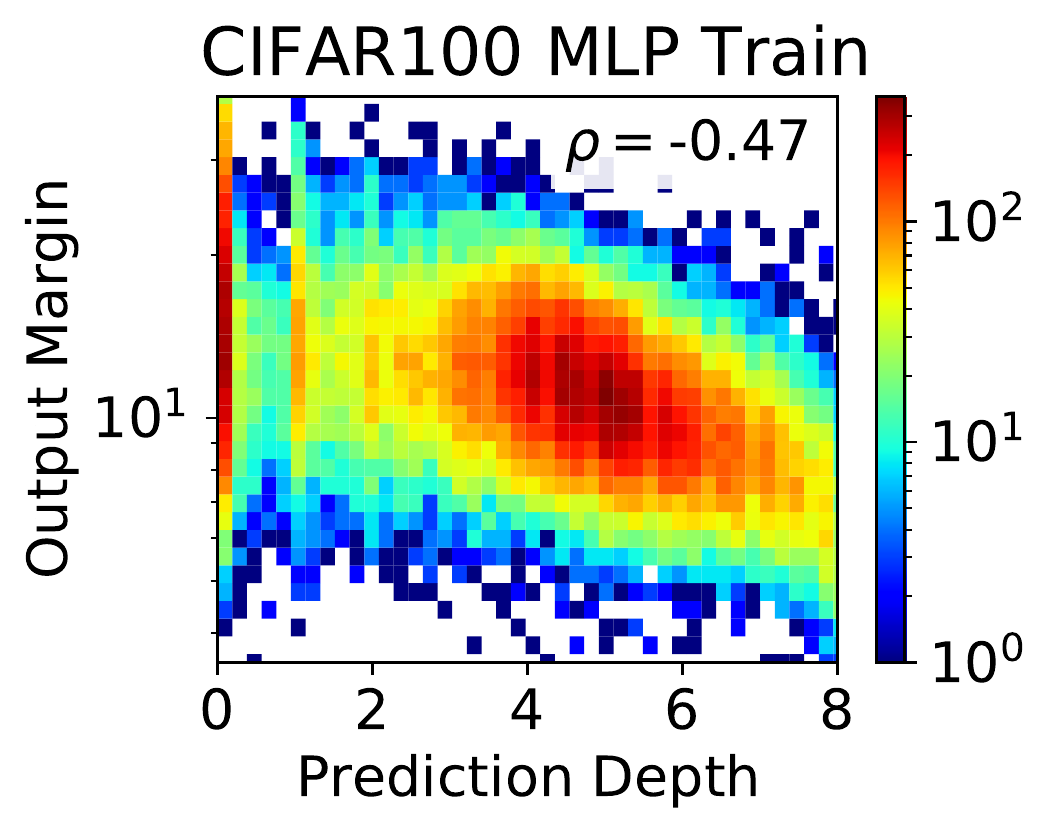}
\end{subfigure}
\begin{subfigure}
         \centering
         \includegraphics[height=3cm]{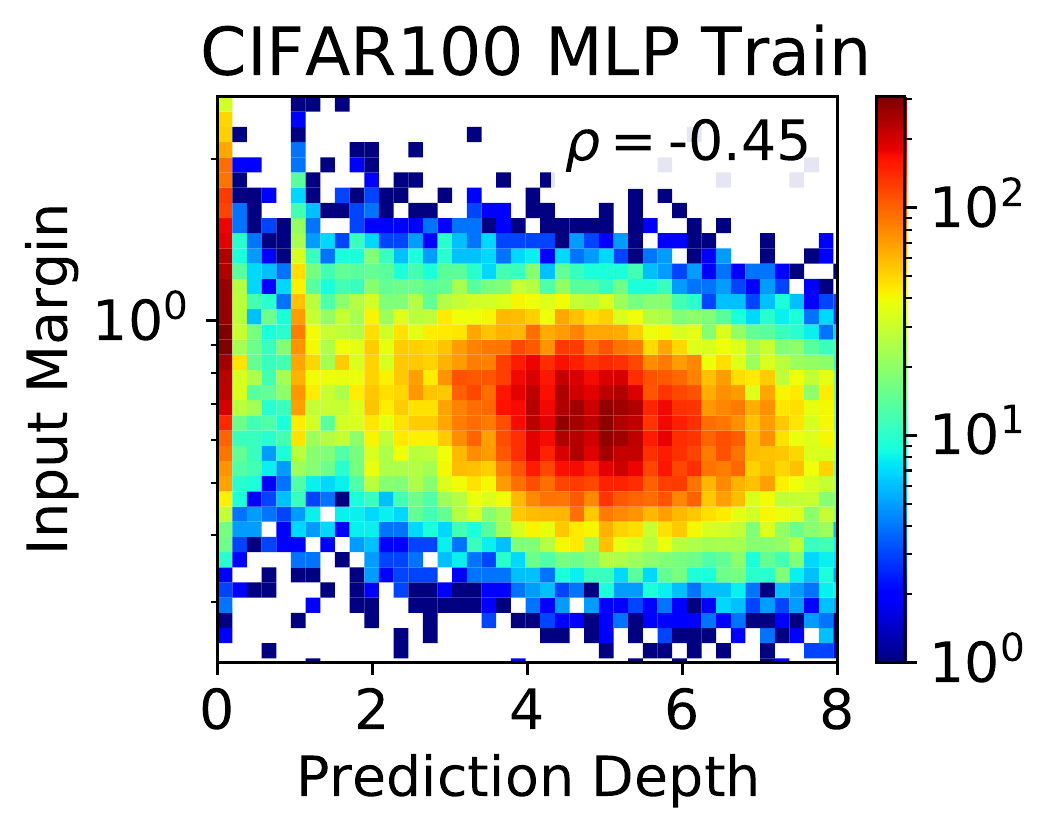}
\end{subfigure}}

\end{center}
\caption{Consistency of Figure~\ref{fig:margins_vs_depth}, showing the correlation between prediction depth, and the input and output margins (log scale) for both the test and training splits of CIFAR100. The correlation coefficient between the prediction depth and the logarithm of the margin is given in each plot. For each architecture, we train 25 models with different random seeds on the full training split. We record the input and output margins together with the prediction depth for every data point in both the train and test splits. These histograms compare the mean values of each margin to the mean prediction depth for all data points.
\label{fig:cifar100_marg_consist}}
\end{figure}

Figures~\ref{fig:svhn_marg_consist} to~\ref{fig:cifar100_marg_consist} reproduce Figure~\ref{fig:margins_vs_depth} (left and middle) for all datasets and architectures in both the training and test splits.

\subsection{Consistent two-dimensional relationship between prediction depths in the training and validation splits \label{app:ll_tvt}}

\begin{figure}[ht!]
\begin{center}
\begin{subfigure}
         \centering
         \includegraphics[width=0.49\columnwidth]{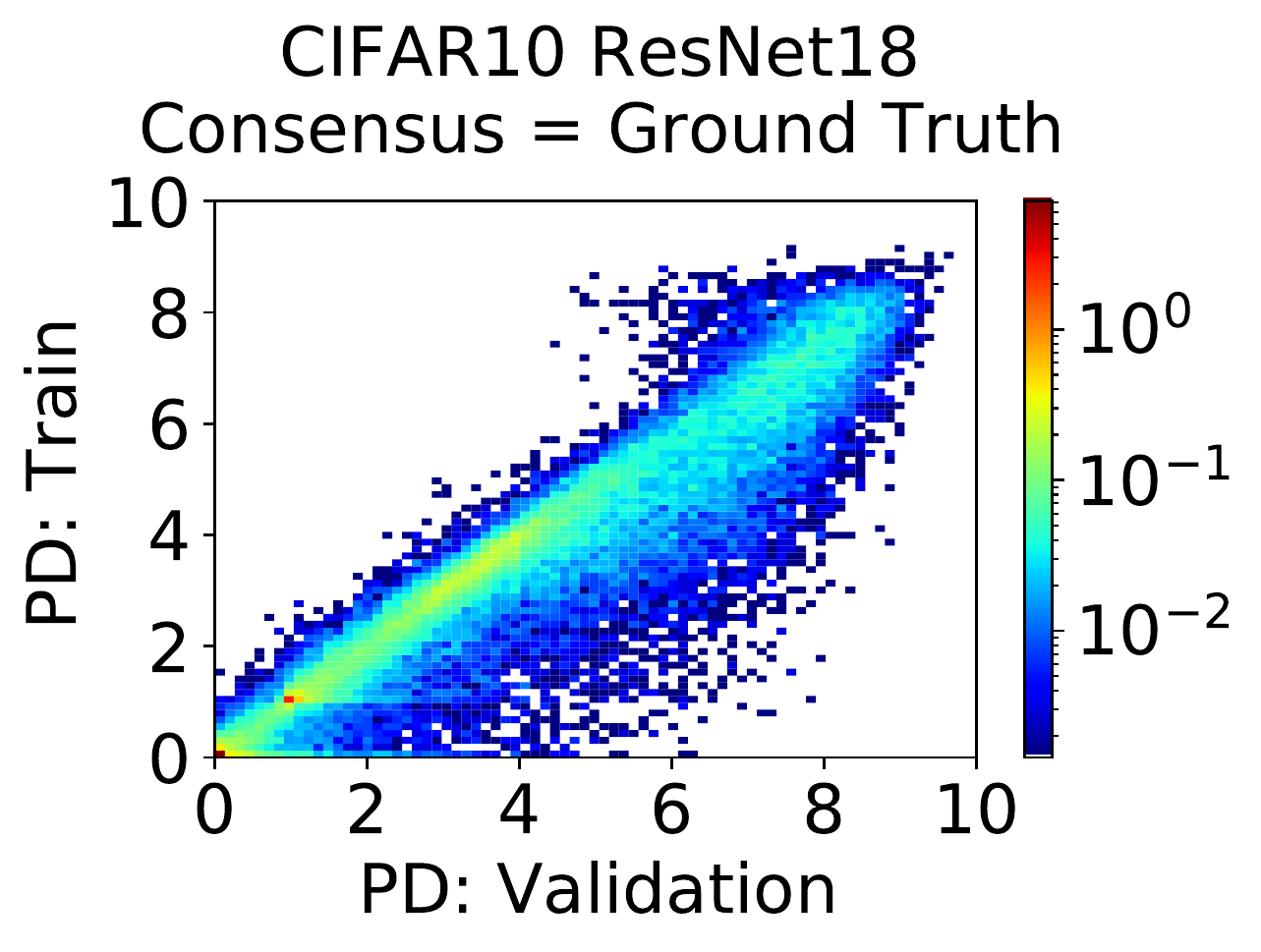}
\end{subfigure}
\begin{subfigure}
         \centering
         \includegraphics[width=0.49\columnwidth]{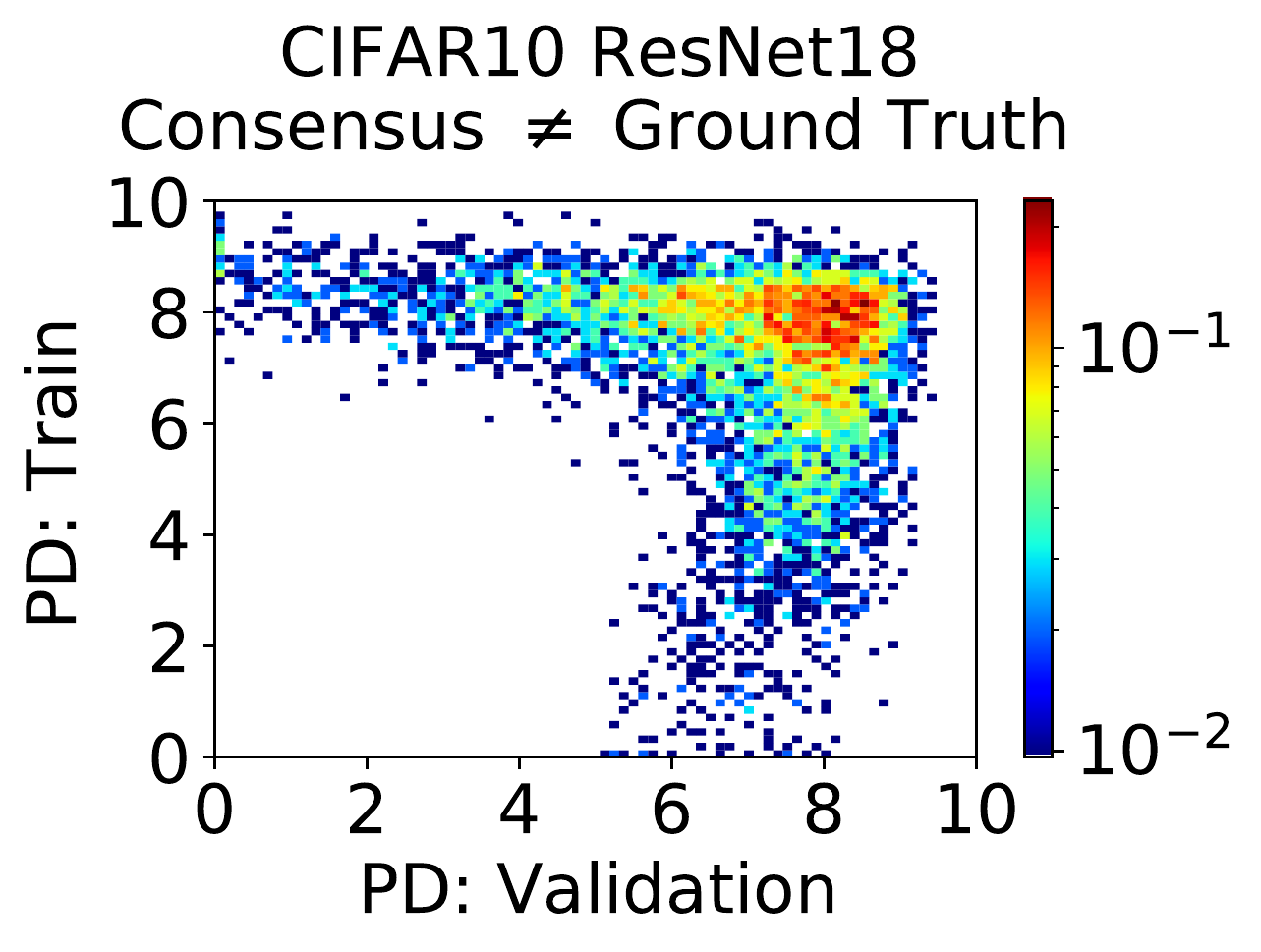}
\end{subfigure}
\begin{subfigure}
         \centering
         \includegraphics[width=0.49\columnwidth]{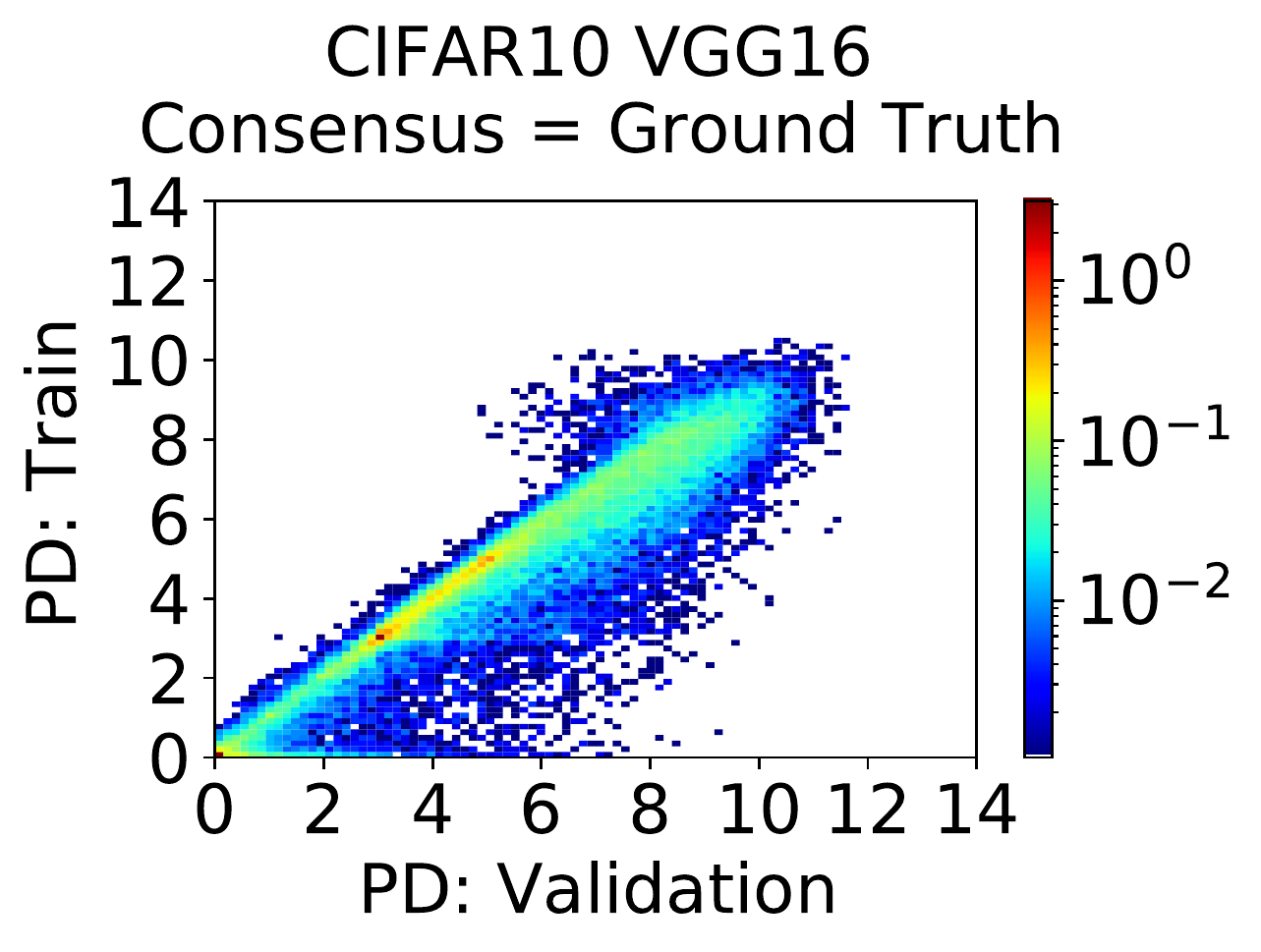}
\end{subfigure}
\begin{subfigure}
         \centering
         \includegraphics[width=0.49\columnwidth]{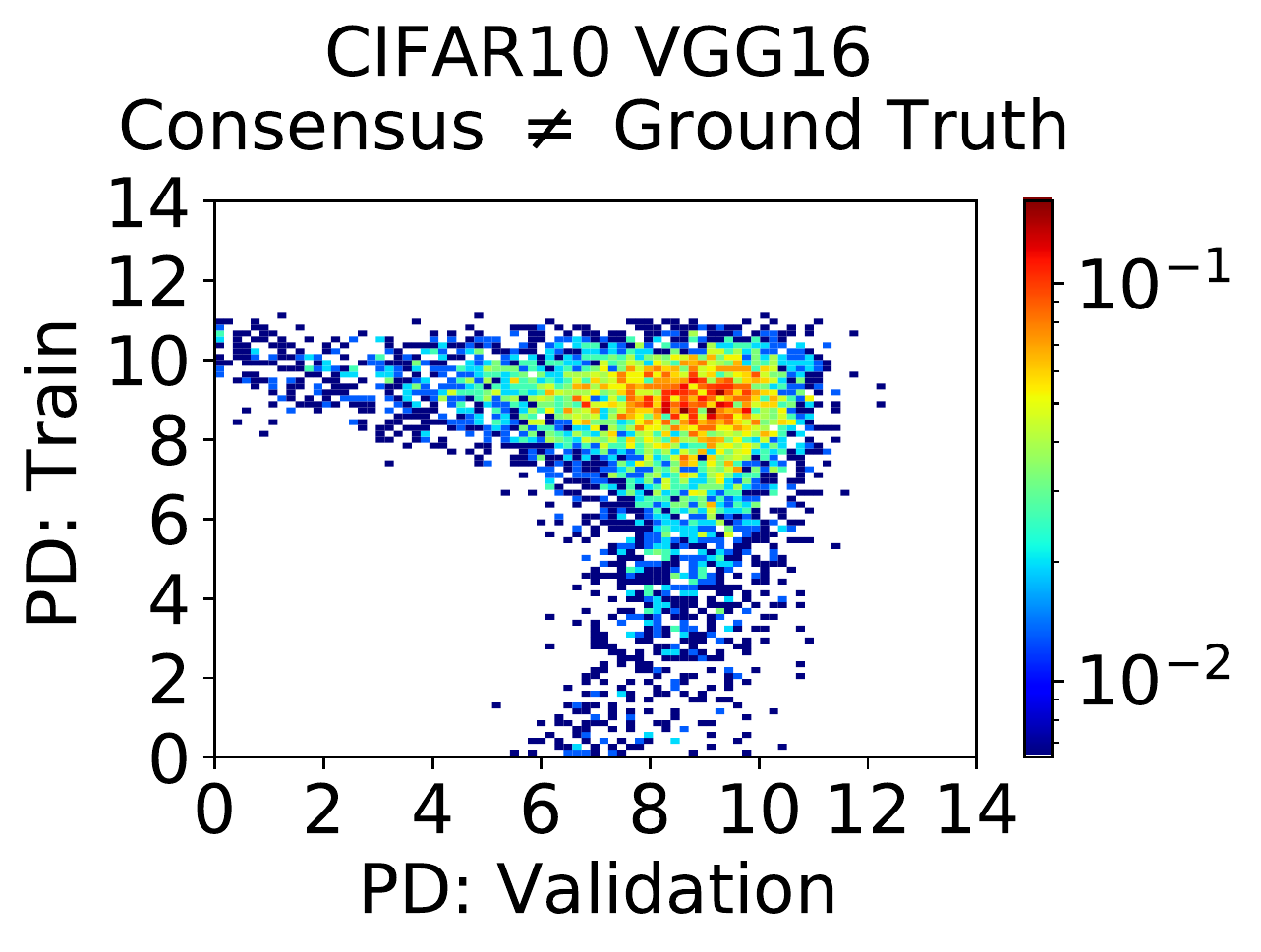}
\end{subfigure}
\begin{subfigure}
         \centering
         \includegraphics[width=0.49\columnwidth]{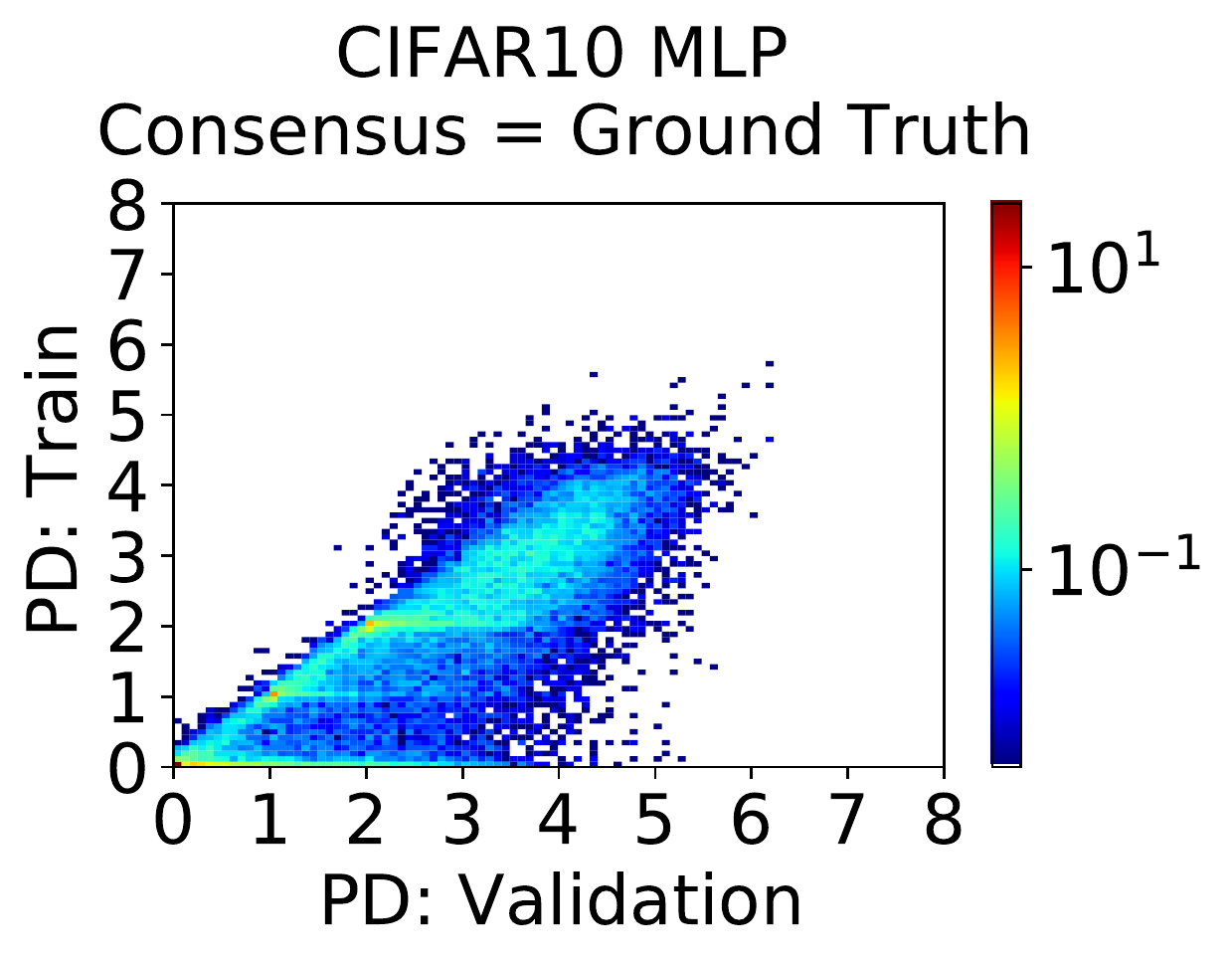}
\end{subfigure}
\begin{subfigure}
         \centering
         \includegraphics[width=0.49\columnwidth]{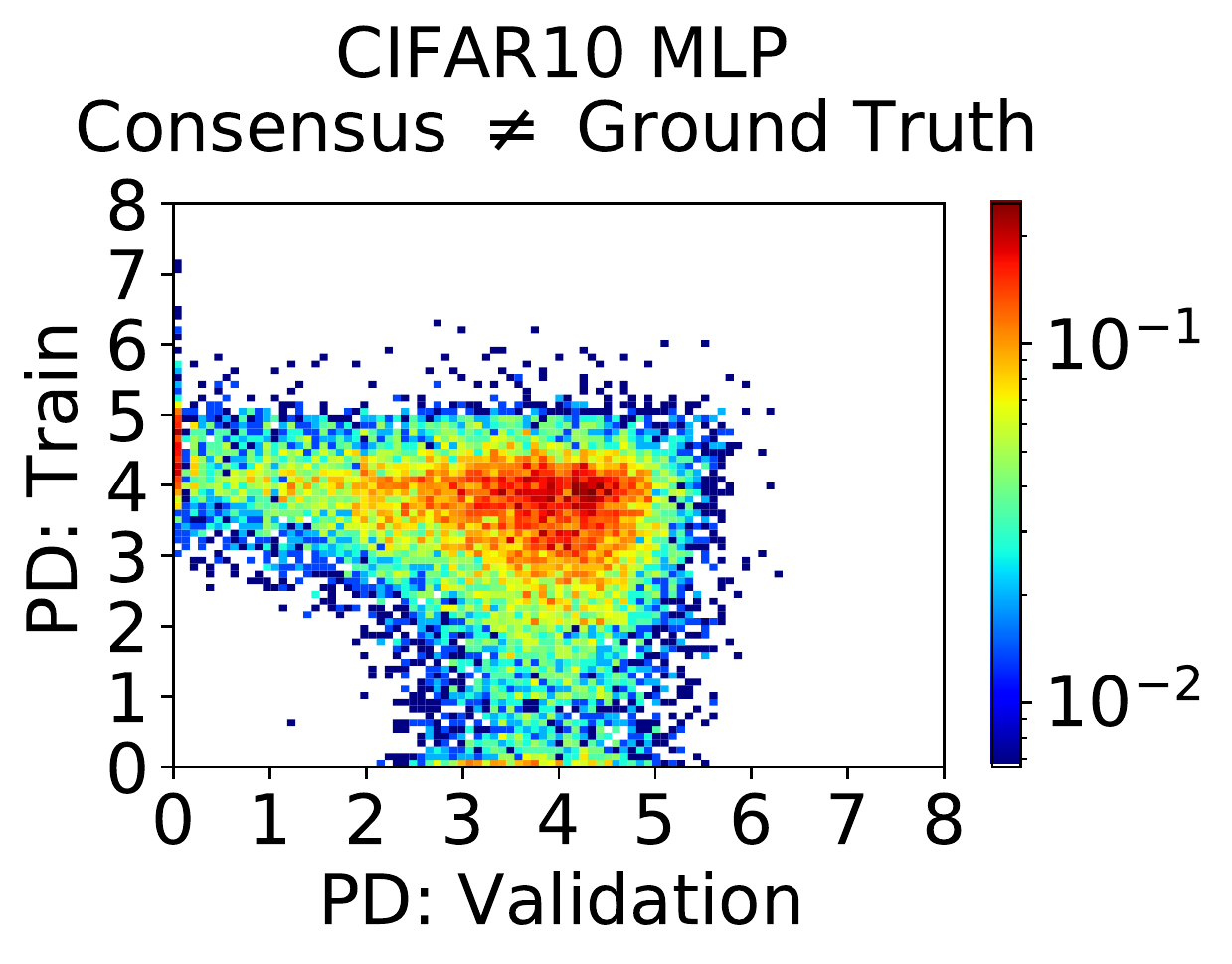}
         \caption{Demonstrating consistency of the histograms shown in Figure~\ref{fig:ll_test_v_train} for all architectures on CIFAR10.
         These histograms compare the mean prediction depth when each data point occurs in either the validation split or the training split. Results are shown separately for data points where the consensus class is the same as or different from the ground truth label. See Appendix~\ref{app:ll_tvt} for a description.
\label{fig:ll_v_ll_0}}
\end{subfigure}
\end{center}
\end{figure}

\begin{figure}[ht!]
\begin{center}

\begin{subfigure}
         \centering
         \includegraphics[width=0.49\columnwidth]{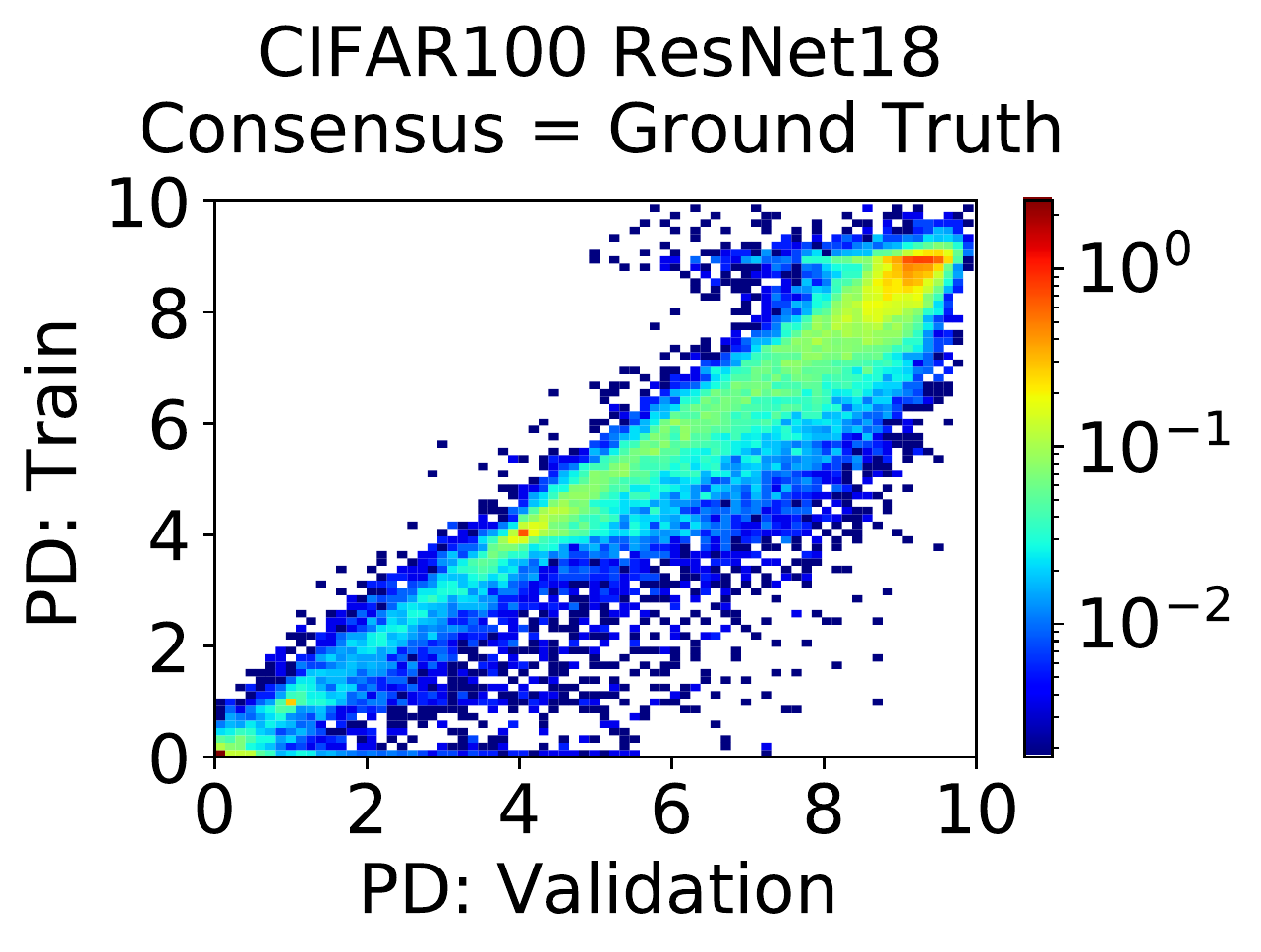}
\end{subfigure}
\begin{subfigure}
         \centering
         \includegraphics[width=0.49\columnwidth]{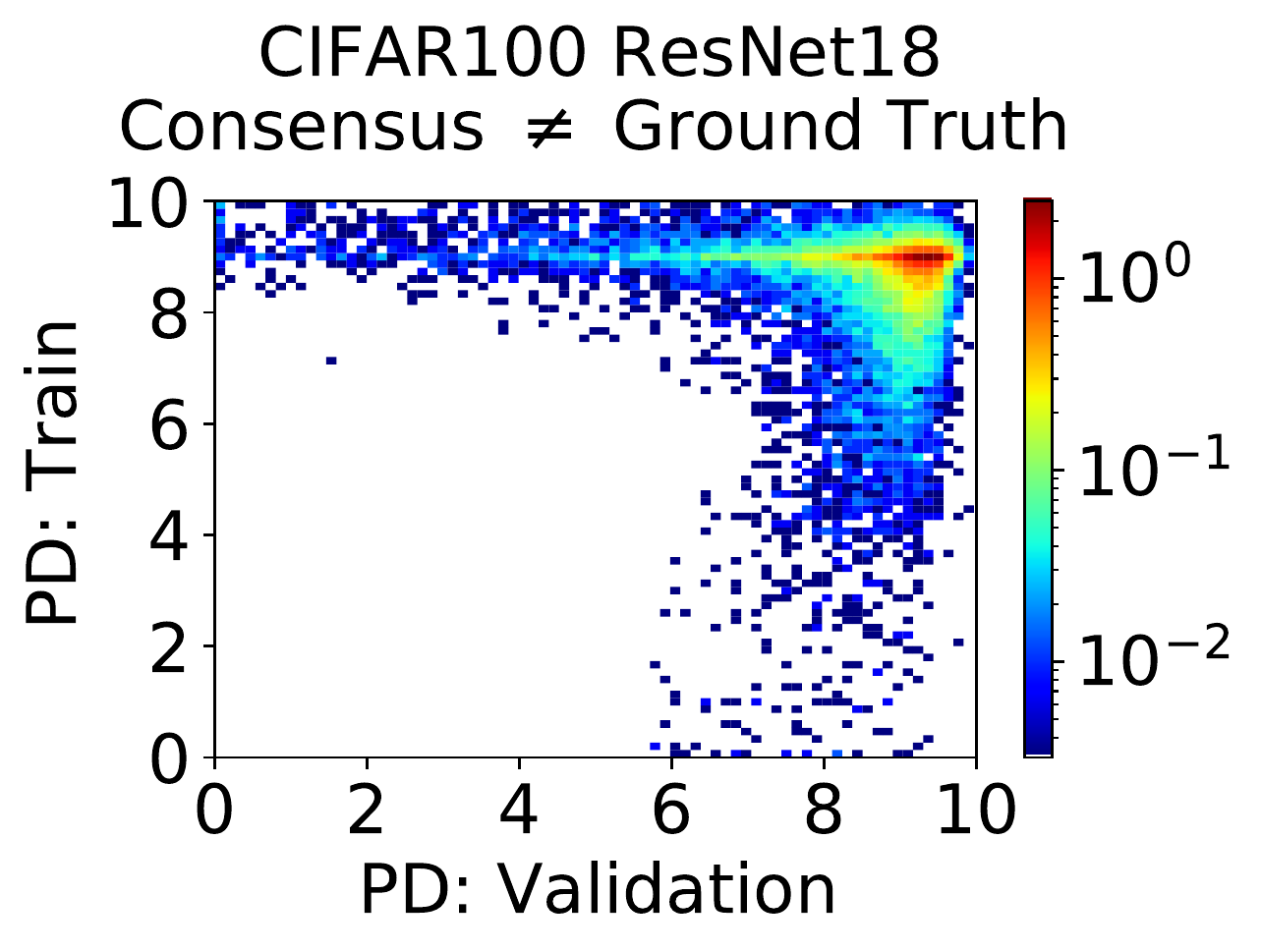}
\end{subfigure}
\begin{subfigure}
         \centering
         \includegraphics[width=0.49\columnwidth]{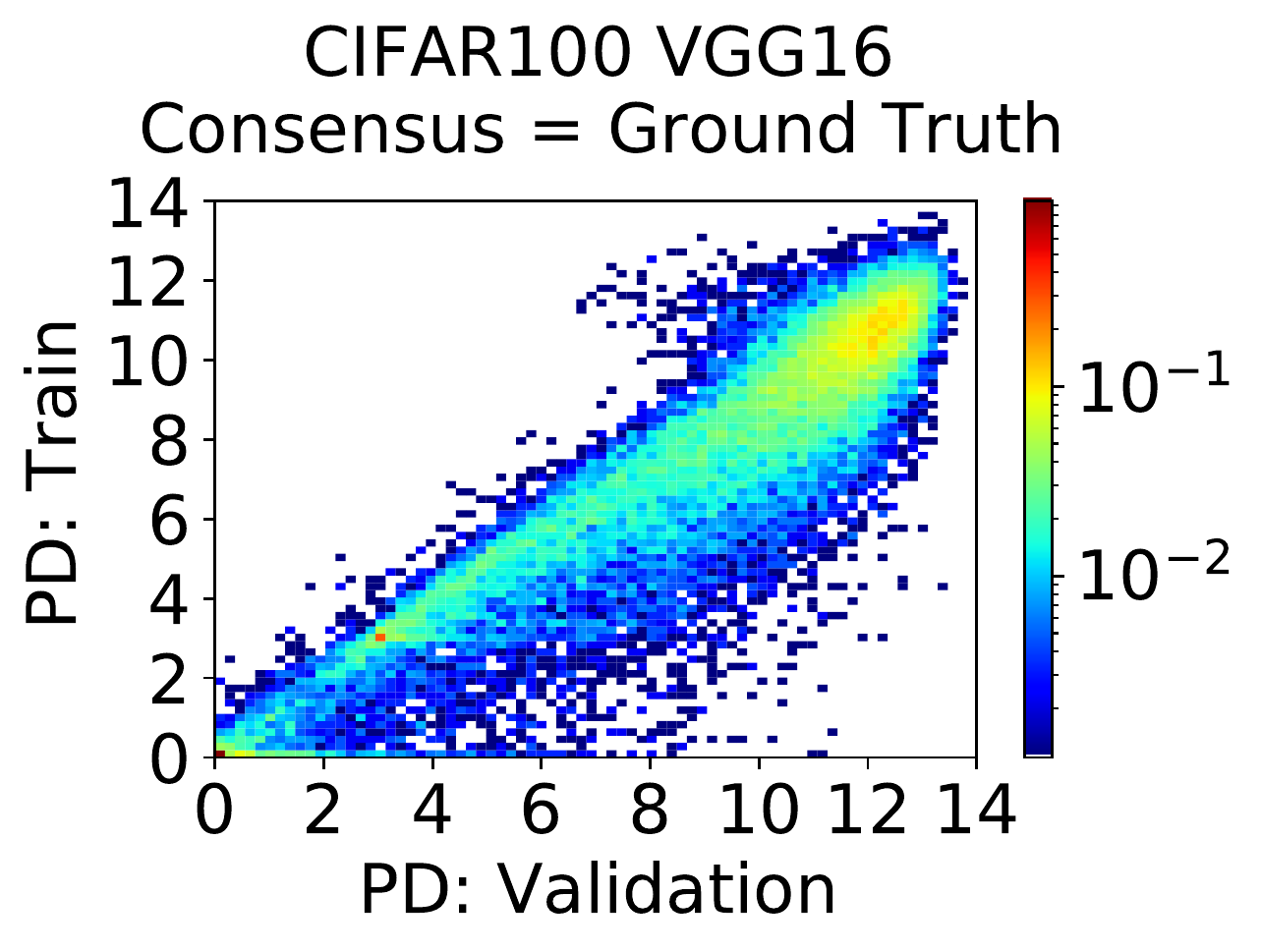}
\end{subfigure}
\begin{subfigure}
         \centering
         \includegraphics[width=0.49\columnwidth]{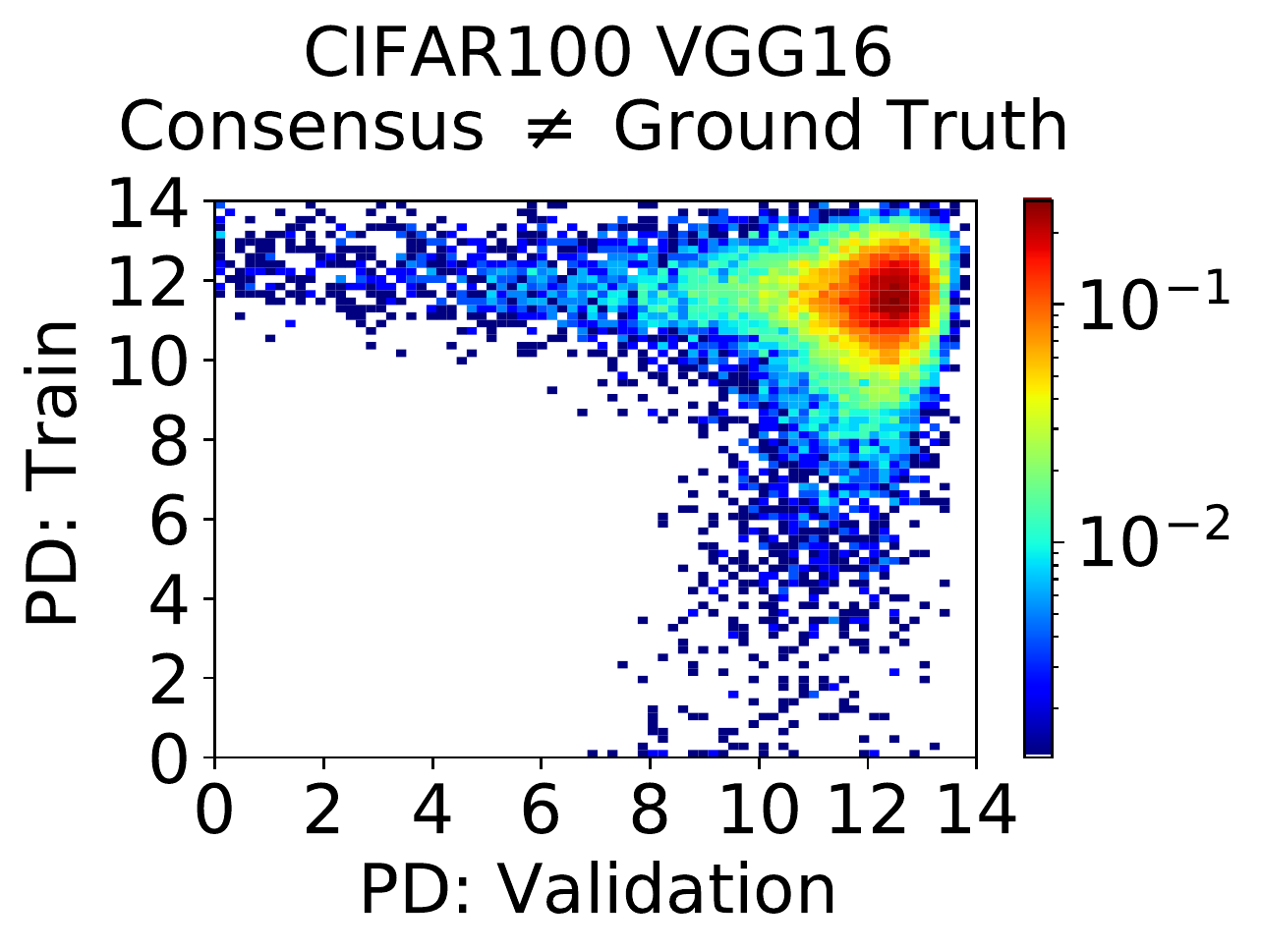}
\end{subfigure}
\begin{subfigure}
         \centering
         \includegraphics[width=0.49\columnwidth]{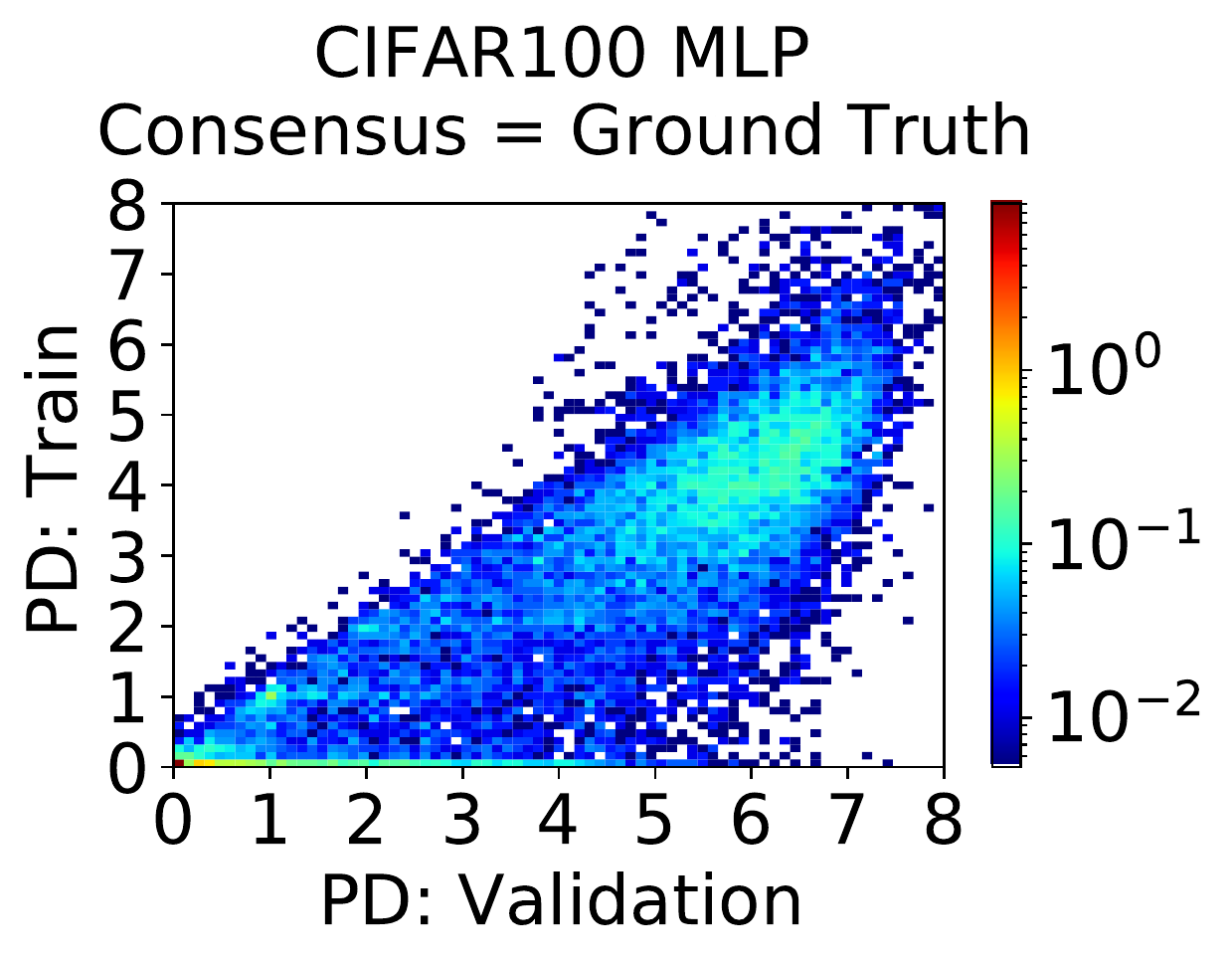}
\end{subfigure}
\begin{subfigure}
         \centering
         \includegraphics[width=0.49\columnwidth]{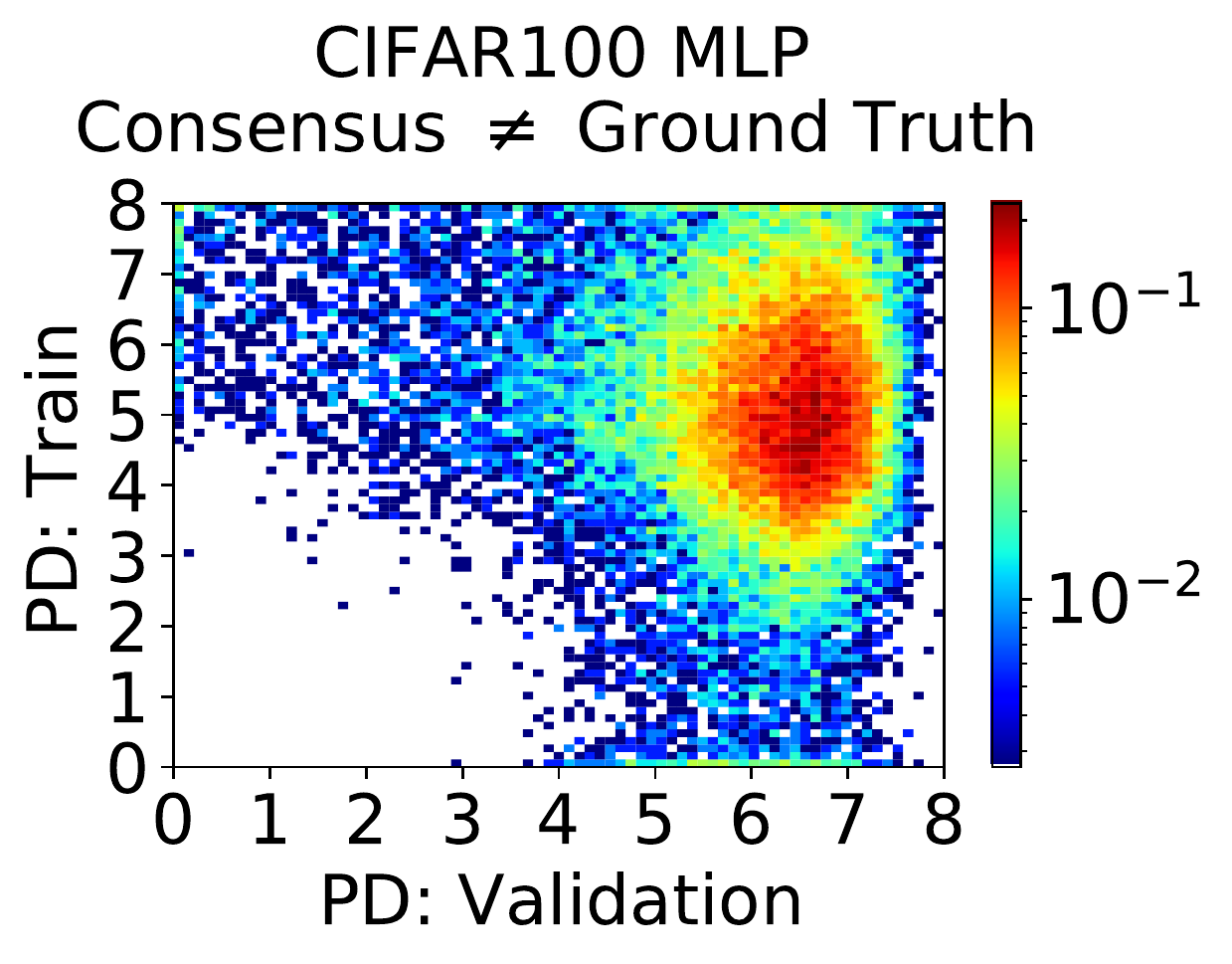}
         \caption{Demonstrating consistency of the histograms shown in Figure~\ref{fig:ll_test_v_train} for all architectures on CIFAR100.
         These histograms compare the mean prediction depth when each data point occurs in either the validation split or the training split. Results are shown separately for data points where the consensus class is the same as or different from the ground truth label. See Appendix~\ref{app:ll_tvt} for a description.}
         \label{fig:ll_v_ll_3}
\end{subfigure}
\end{center}
\end{figure}

\begin{figure}[ht!]
\begin{center}

\begin{subfigure}
         \centering
         \includegraphics[width=0.49\columnwidth]{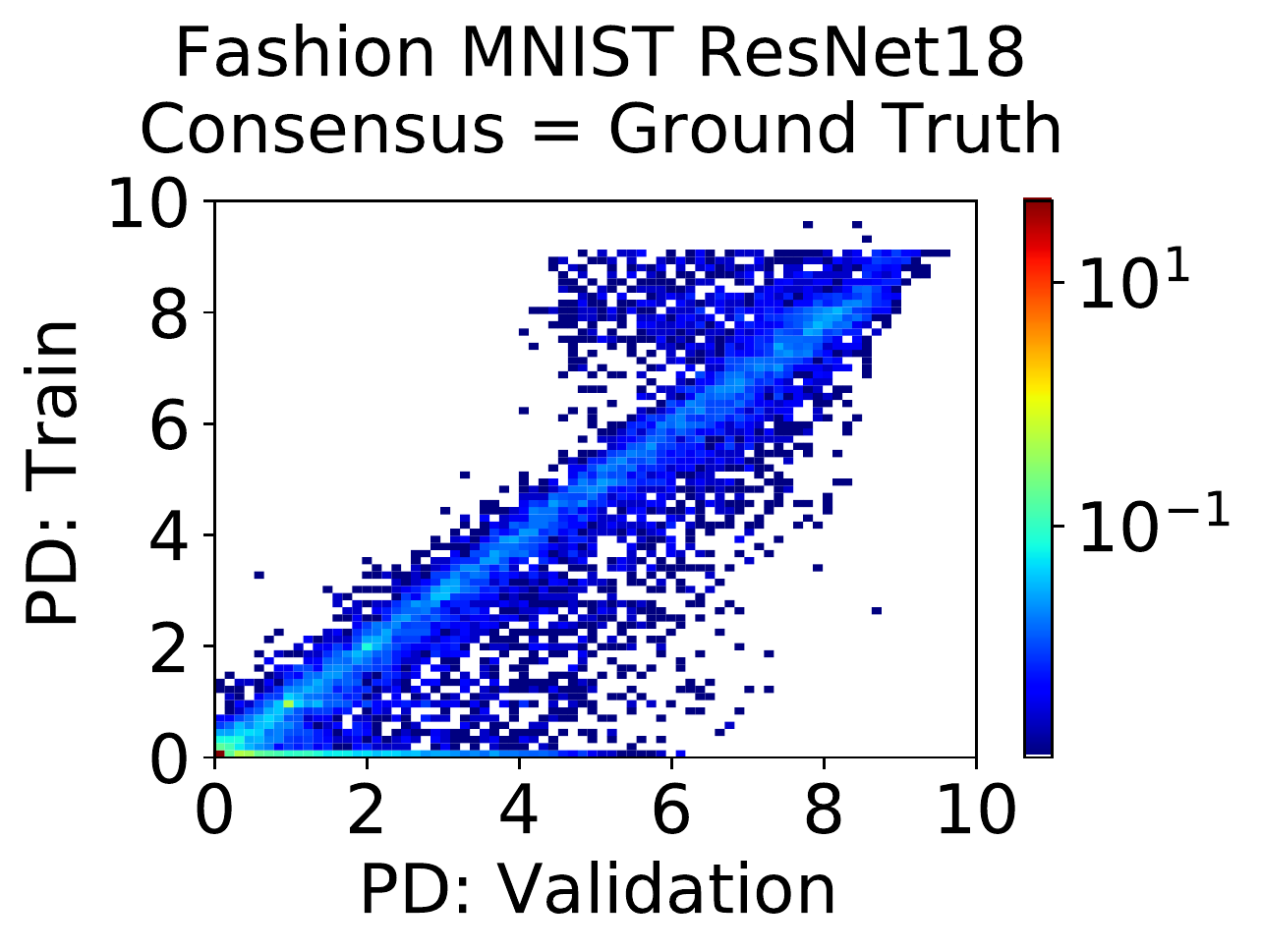}
\end{subfigure}
\begin{subfigure}
         \centering
         \includegraphics[width=0.49\columnwidth]{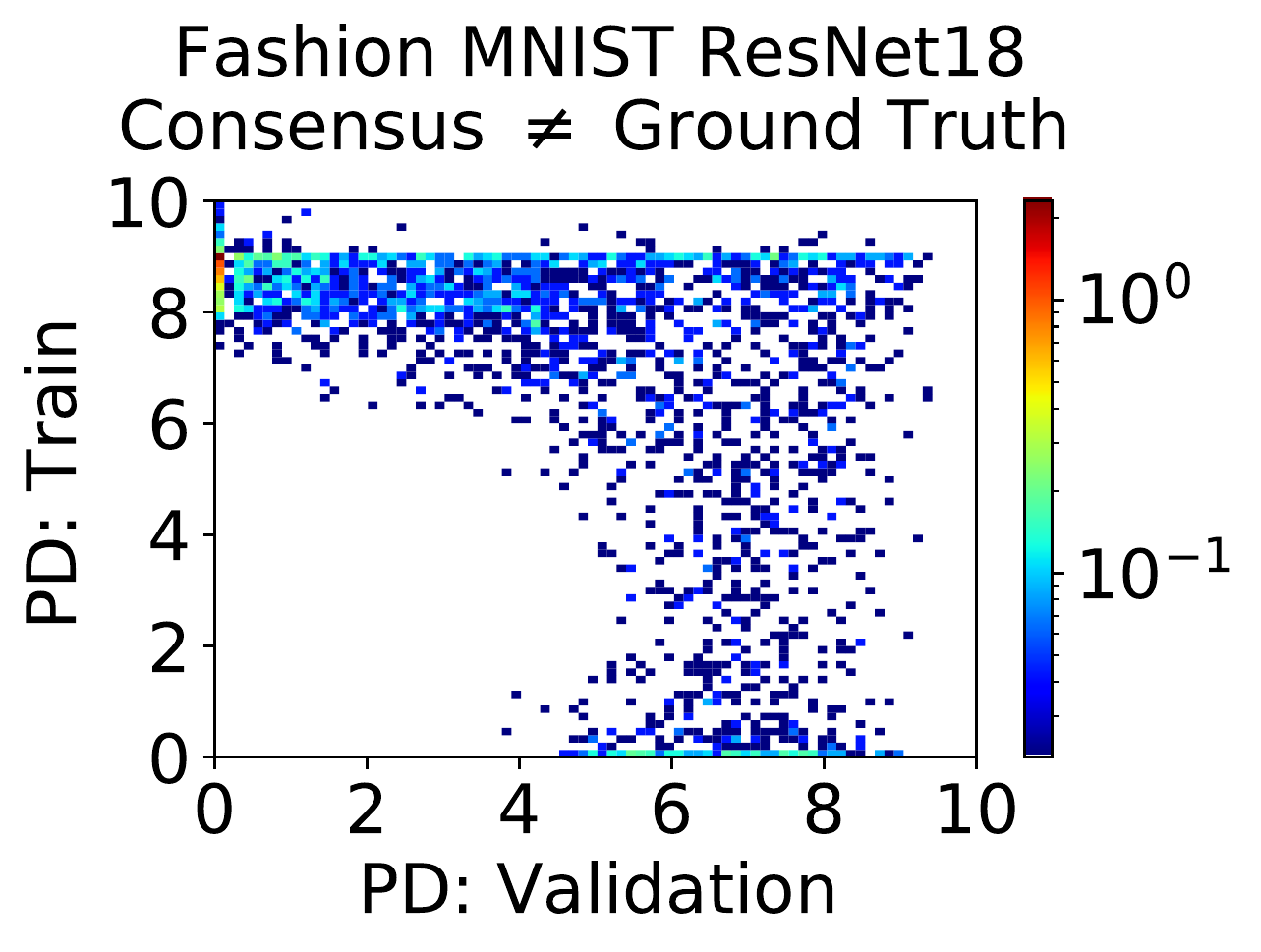}
\end{subfigure}
\begin{subfigure}
         \centering
         \includegraphics[width=0.49\columnwidth]{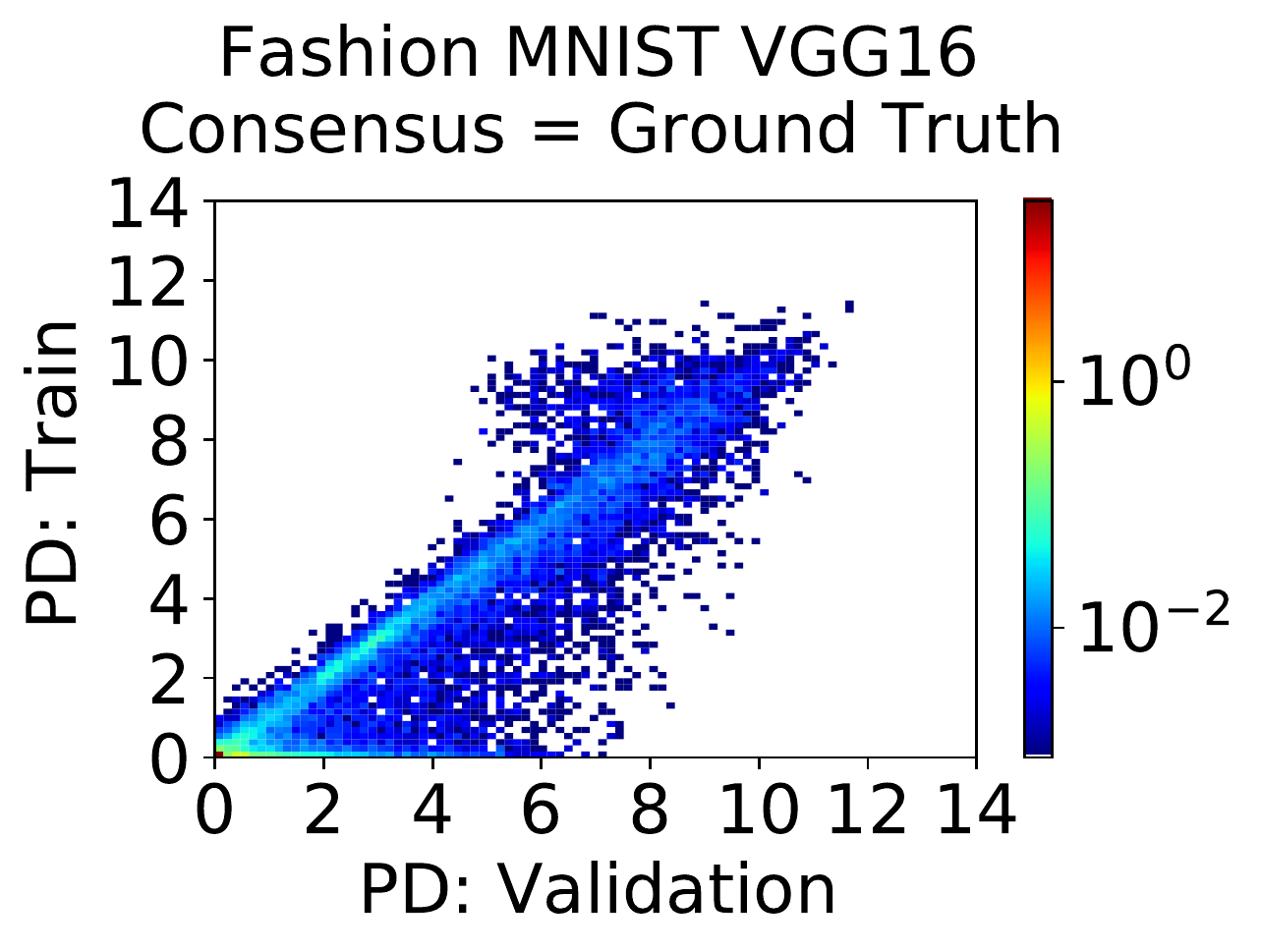}
\end{subfigure}
\begin{subfigure}
         \centering
         \includegraphics[width=0.49\columnwidth]{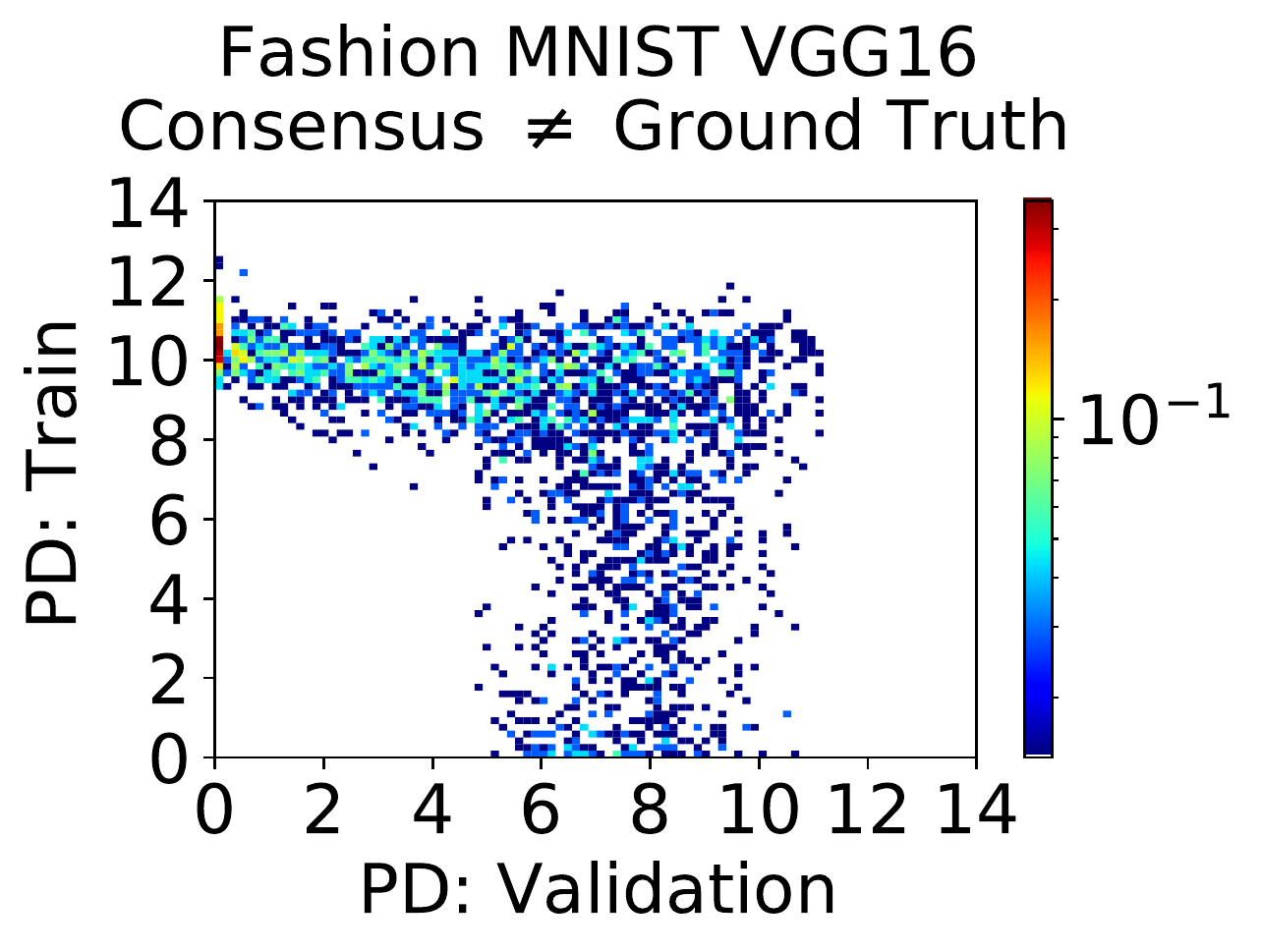}
\end{subfigure}
\begin{subfigure}
         \centering
         \includegraphics[width=0.49\columnwidth]{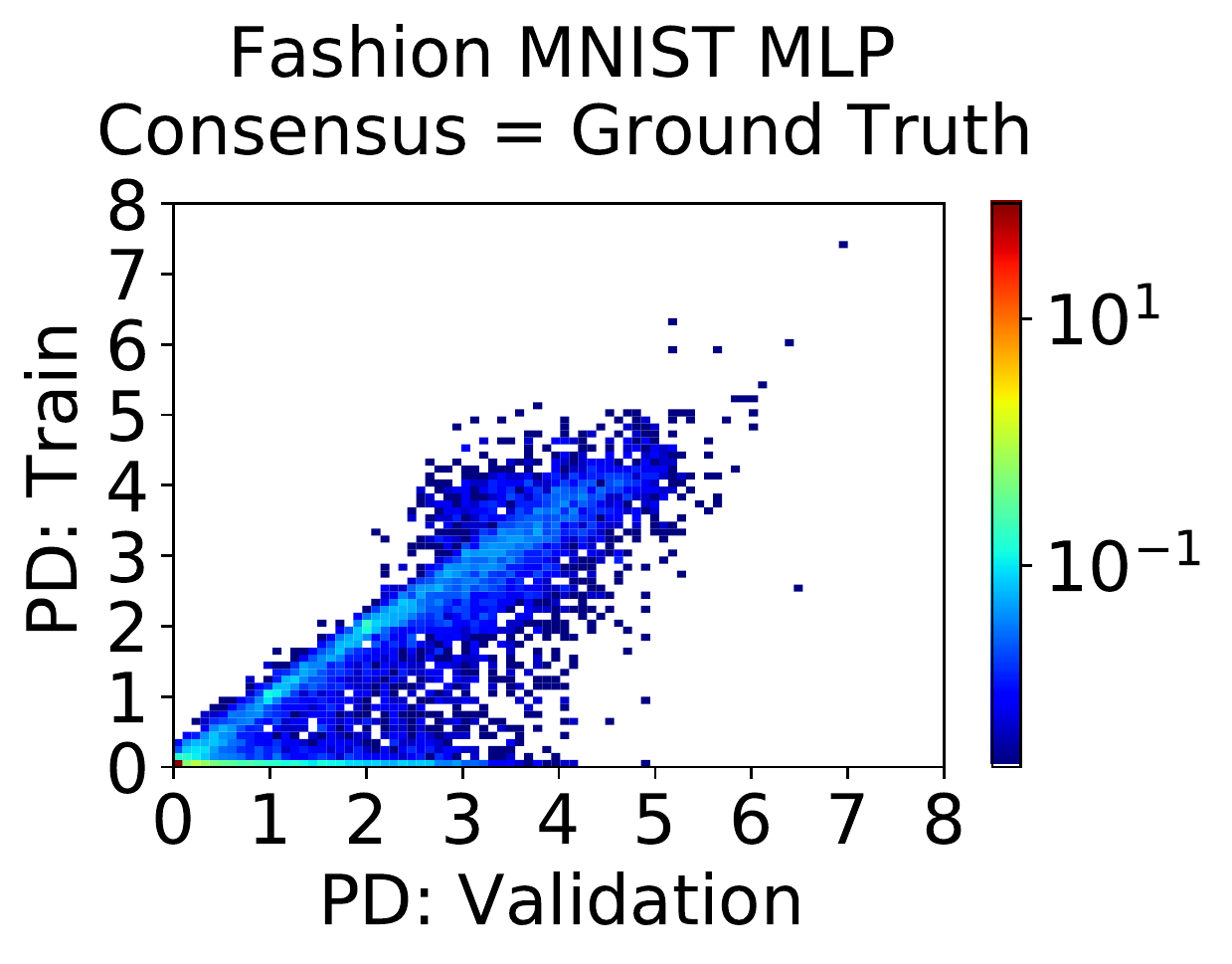}
\end{subfigure}
\begin{subfigure}
         \centering
         \includegraphics[width=0.49\columnwidth]{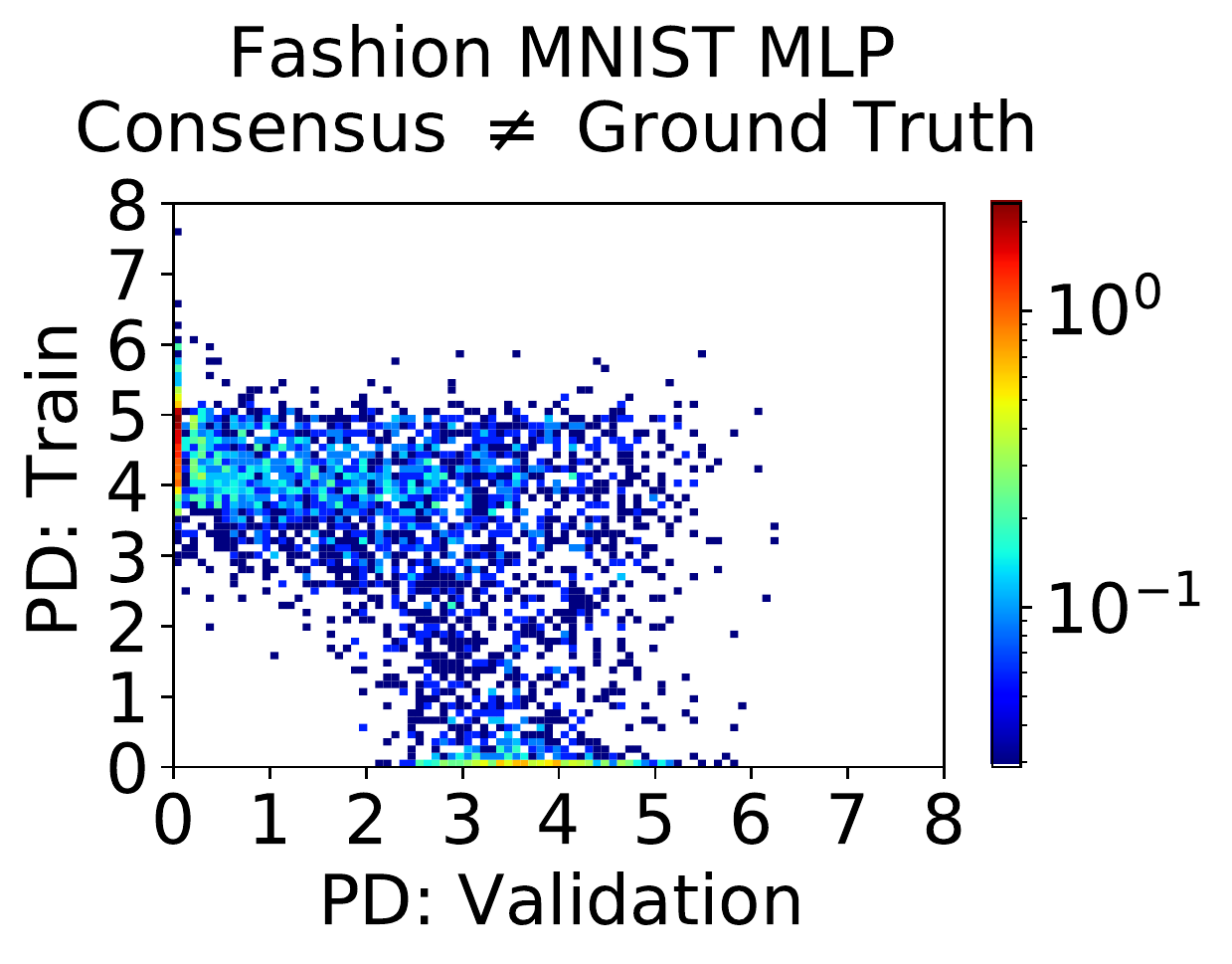}
          \caption{Demonstrating consistency of the histograms shown in Figure~\ref{fig:ll_test_v_train} for all architectures on Fashion MNIST.
         These histograms compare the mean prediction depth when each data point occurs in either the validation split or the training split. Results are shown separately for data points where the consensus class is the same as or different from the ground truth label. See Appendix~\ref{app:ll_tvt} for a description.
         \label{fig:ll_v_ll_6}}
\end{subfigure}
\end{center}
\end{figure}

\begin{figure}[ht!]
\begin{center}
\begin{subfigure}
         \centering
         \includegraphics[width=0.49\columnwidth]{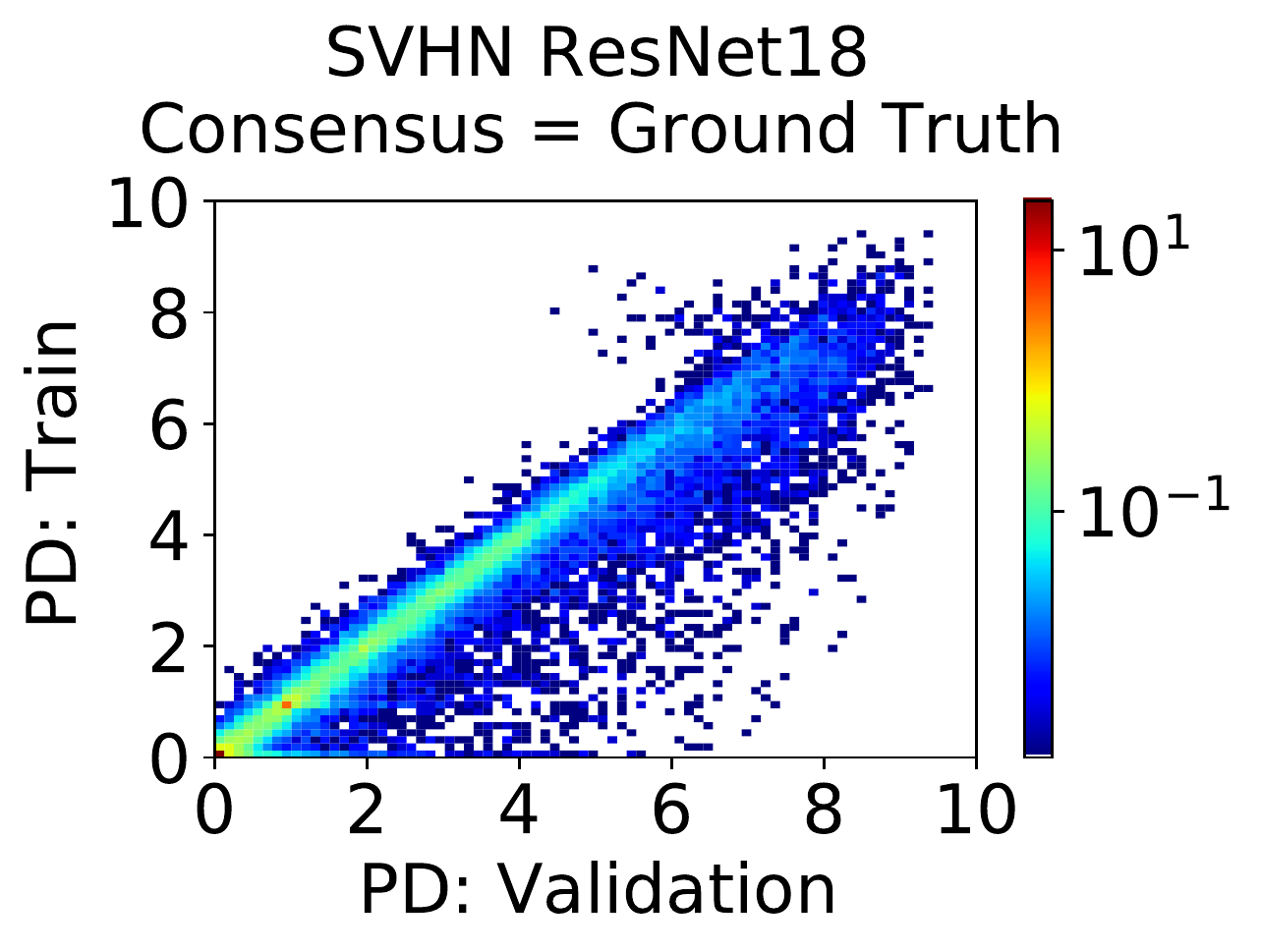}
\end{subfigure}
\begin{subfigure}
         \centering
         \includegraphics[width=0.49\columnwidth]{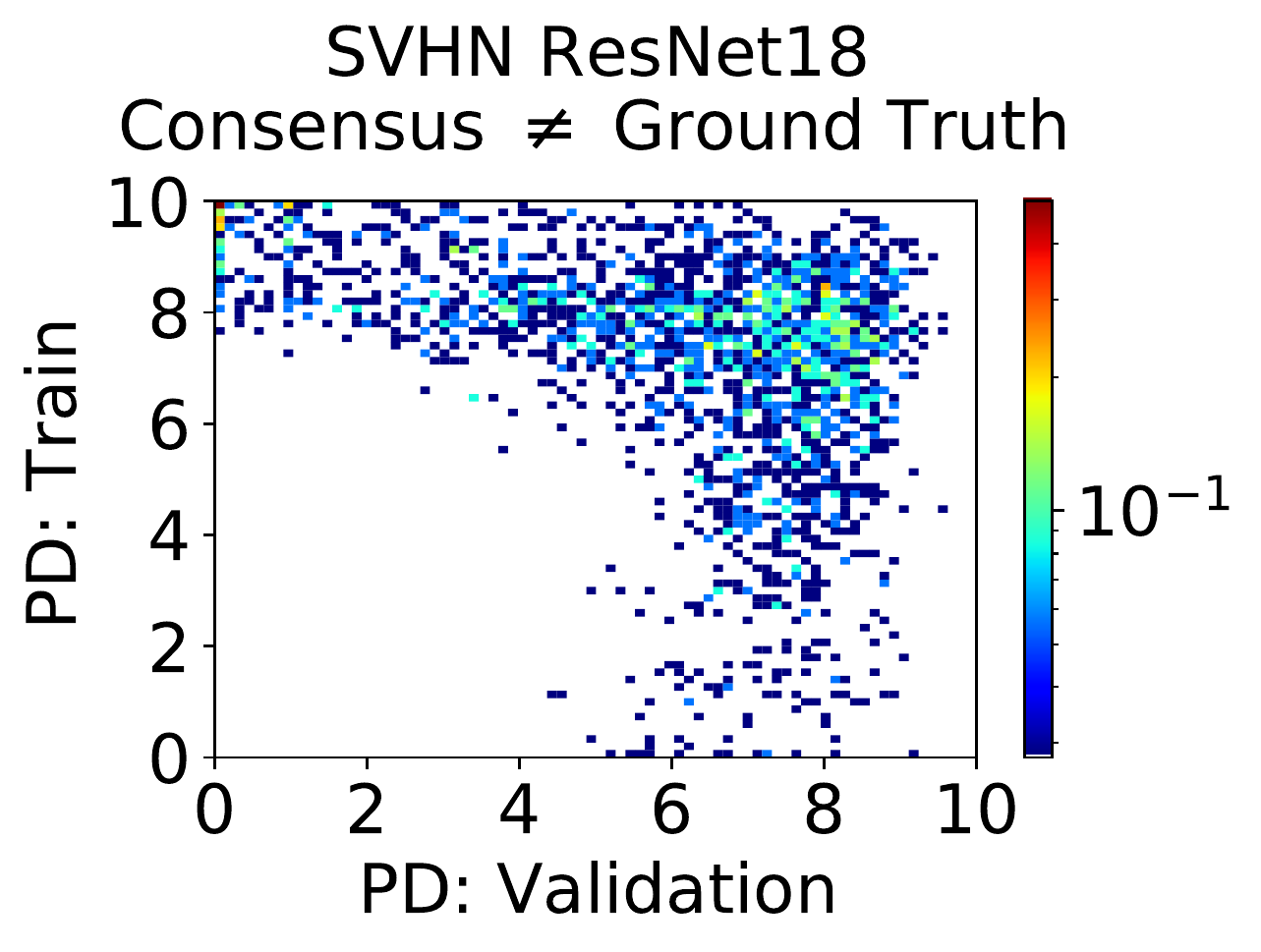}
\end{subfigure}
\begin{subfigure}
         \centering
         \includegraphics[width=0.49\columnwidth]{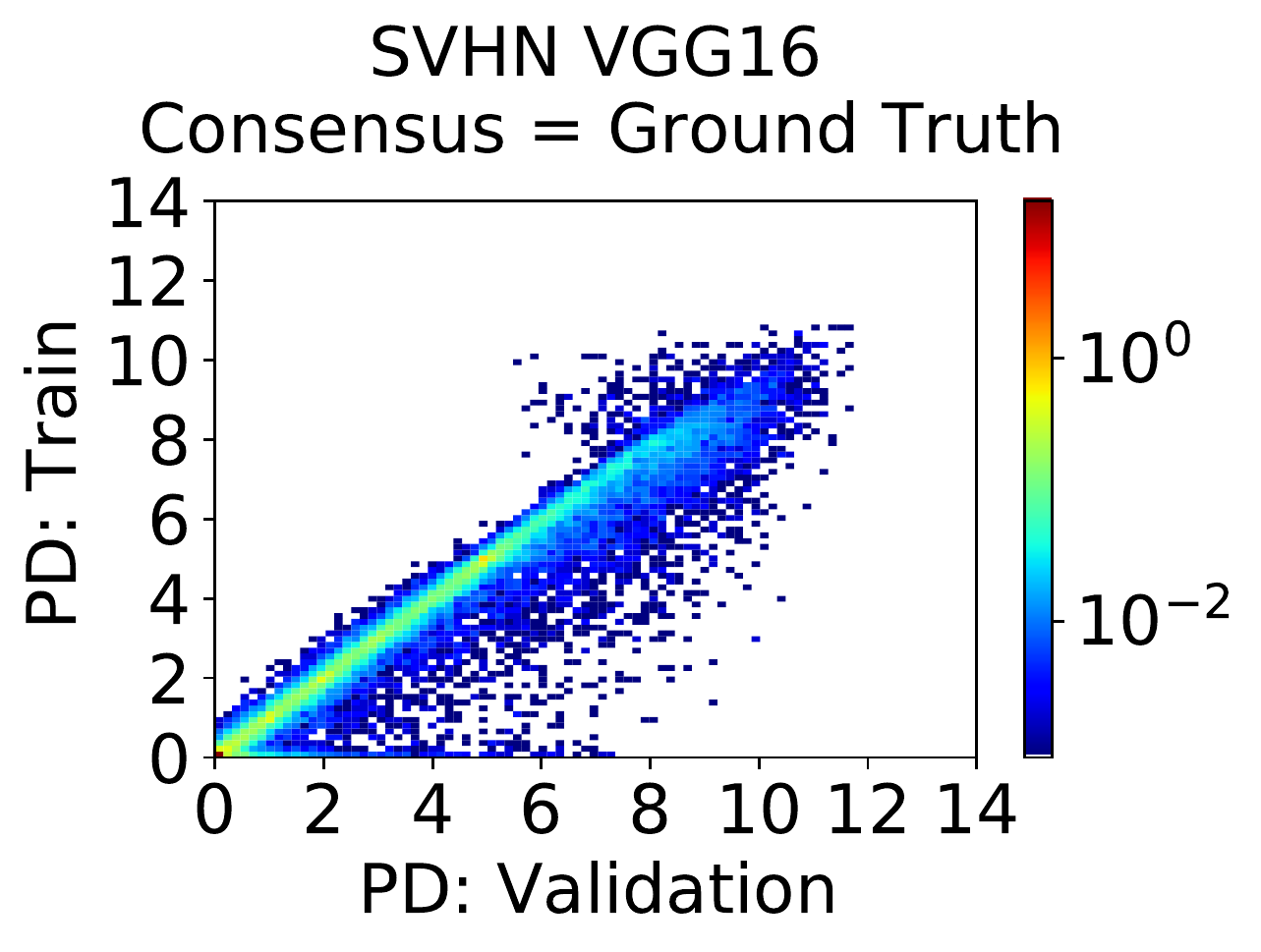}
\end{subfigure}
\begin{subfigure}
         \centering
         \includegraphics[width=0.49\columnwidth]{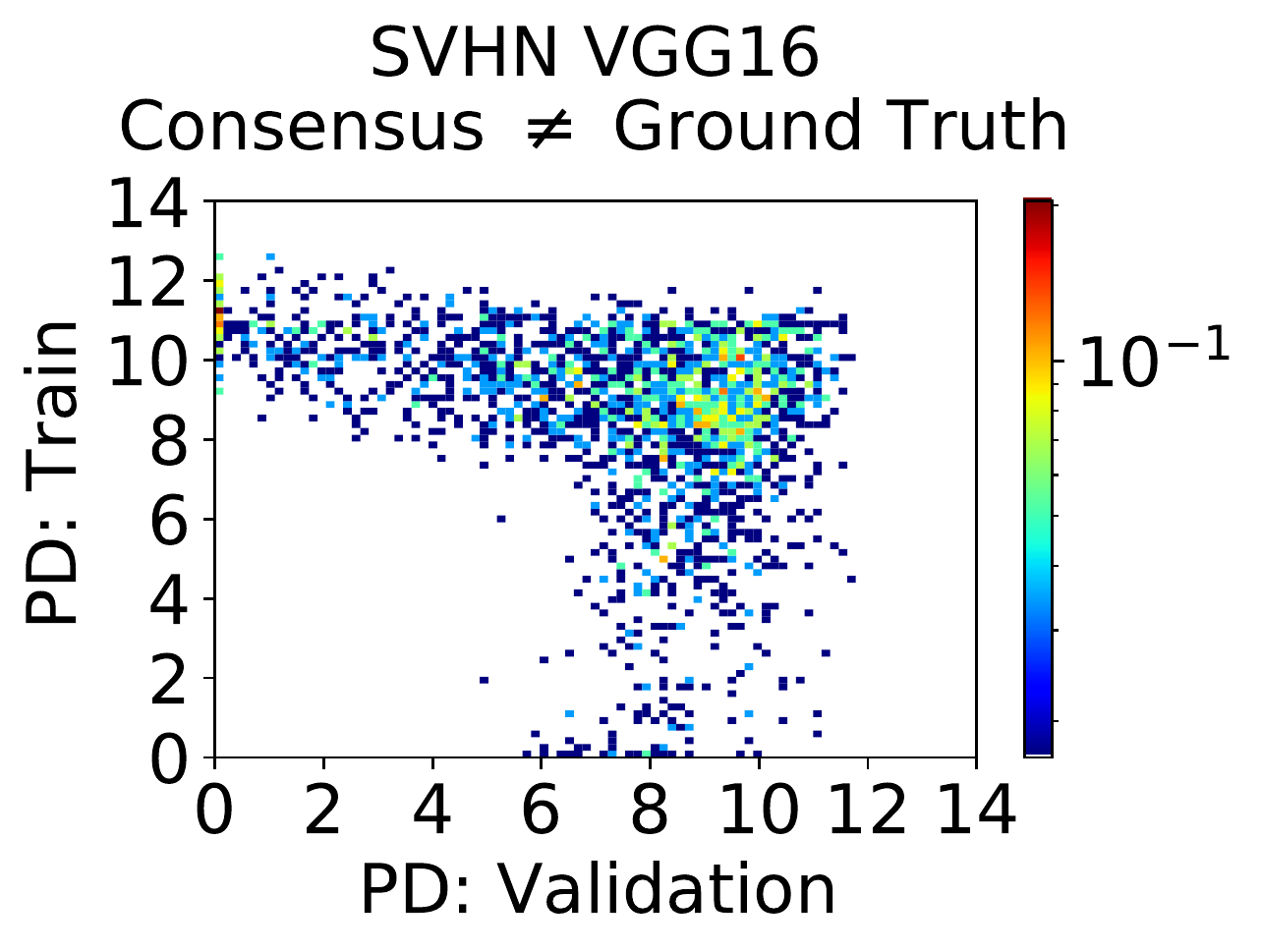}
\end{subfigure}
\begin{subfigure}
         \centering
         \includegraphics[width=0.49\columnwidth]{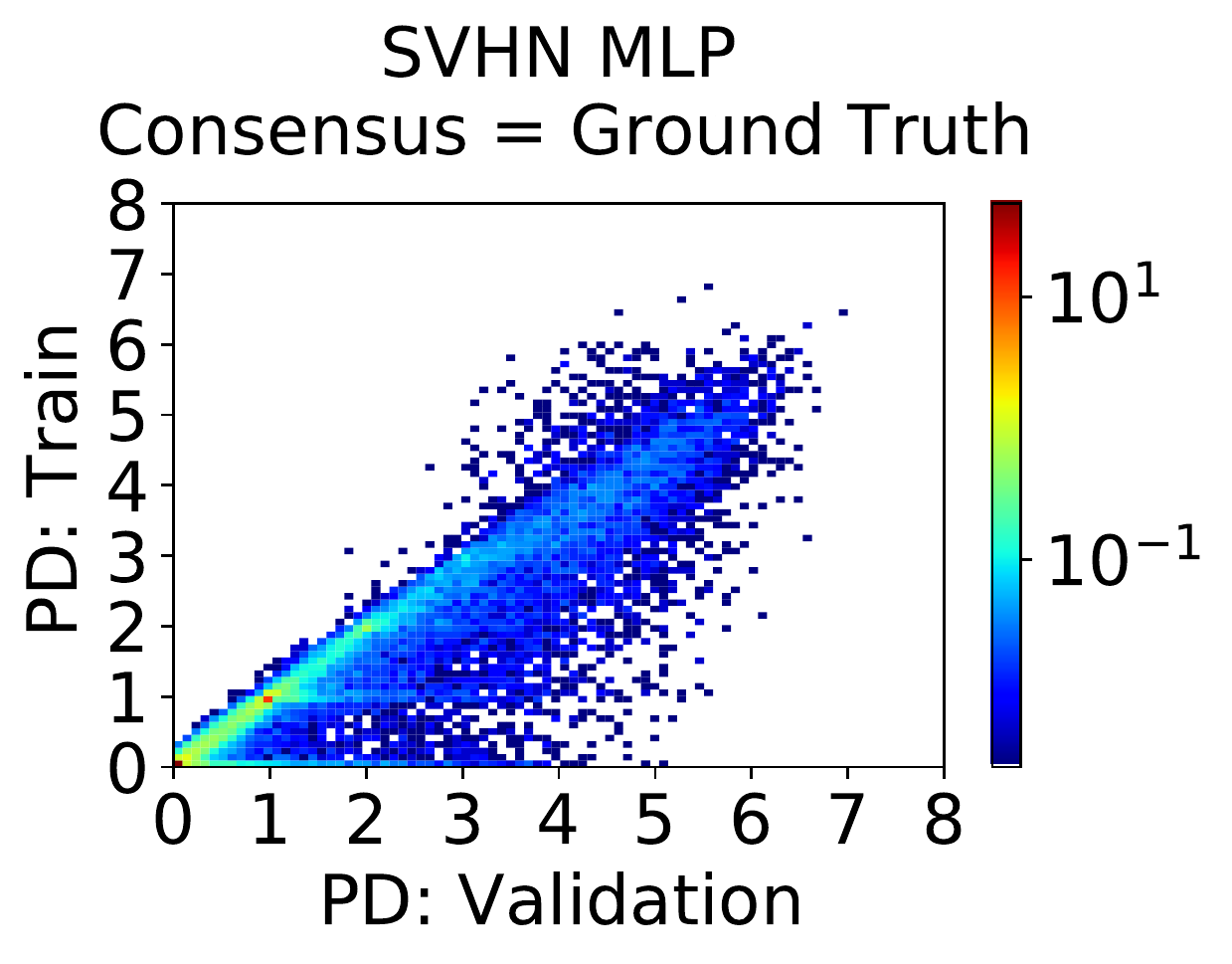}
\end{subfigure}
\begin{subfigure}
         \centering
         \includegraphics[width=0.49\columnwidth]{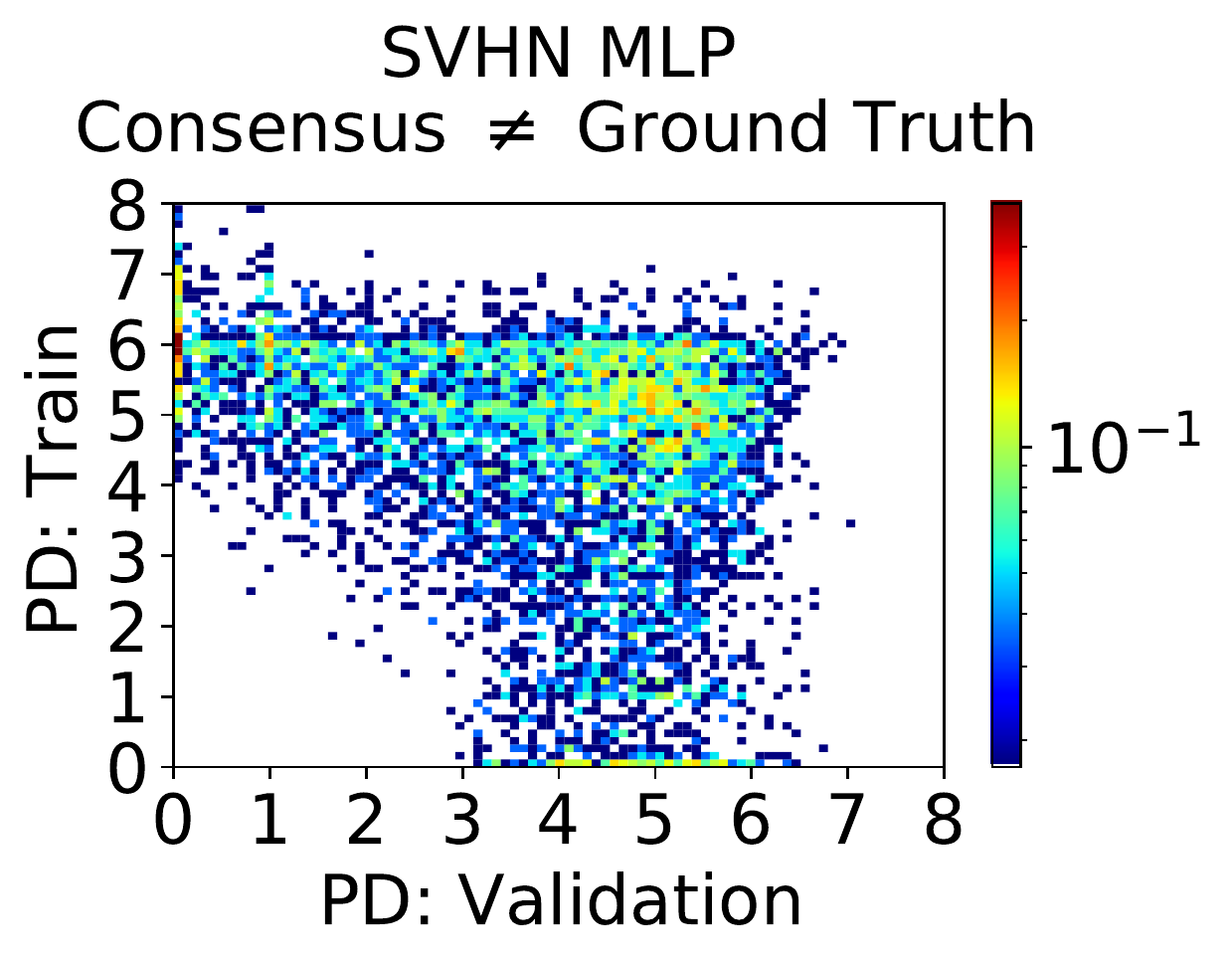}
         \caption{Demonstrating consistency of the histograms shown in Figure~\ref{fig:ll_test_v_train} for all architectures on SVHN.
         These histograms compare the mean prediction depth when each data point occurs in either the validation split or the training split. Results are shown separately for data points where the consensus class is the same as or different from the ground truth label. See Appendix~\ref{app:ll_tvt} for a description.
         \label{fig:ll_v_ll_9}}
\end{subfigure}
\end{center}
\end{figure}

Figures~\ref{fig:ll_v_ll_0} to~\ref{fig:ll_v_ll_9} demonstrate consistency of the histograms shown in Figure~\ref{fig:ll_test_v_train} for all datasets and architectures.
As described in Appendix~\ref{app:experiments_desc}, for each dataset and architecture we trained 250 models with random 90:10\% validation:train splits. Each time a data point appears in either split we record the prediction depth. These histograms compare the mean prediction depths in the two splits for all data points which can be very different from each other, depending on whether the consensus class matches or differs from the ground truth class.

\subsection{Evolution of clustering in the hidden layers for the different forms of example difficulty \label{app:cluster_4forms}}

\begin{figure*}[ht!]
\begin{center} \resizebox{1.\textwidth}{!}{%
\includegraphics[trim=0 0 0 0, clip,height=5cm]{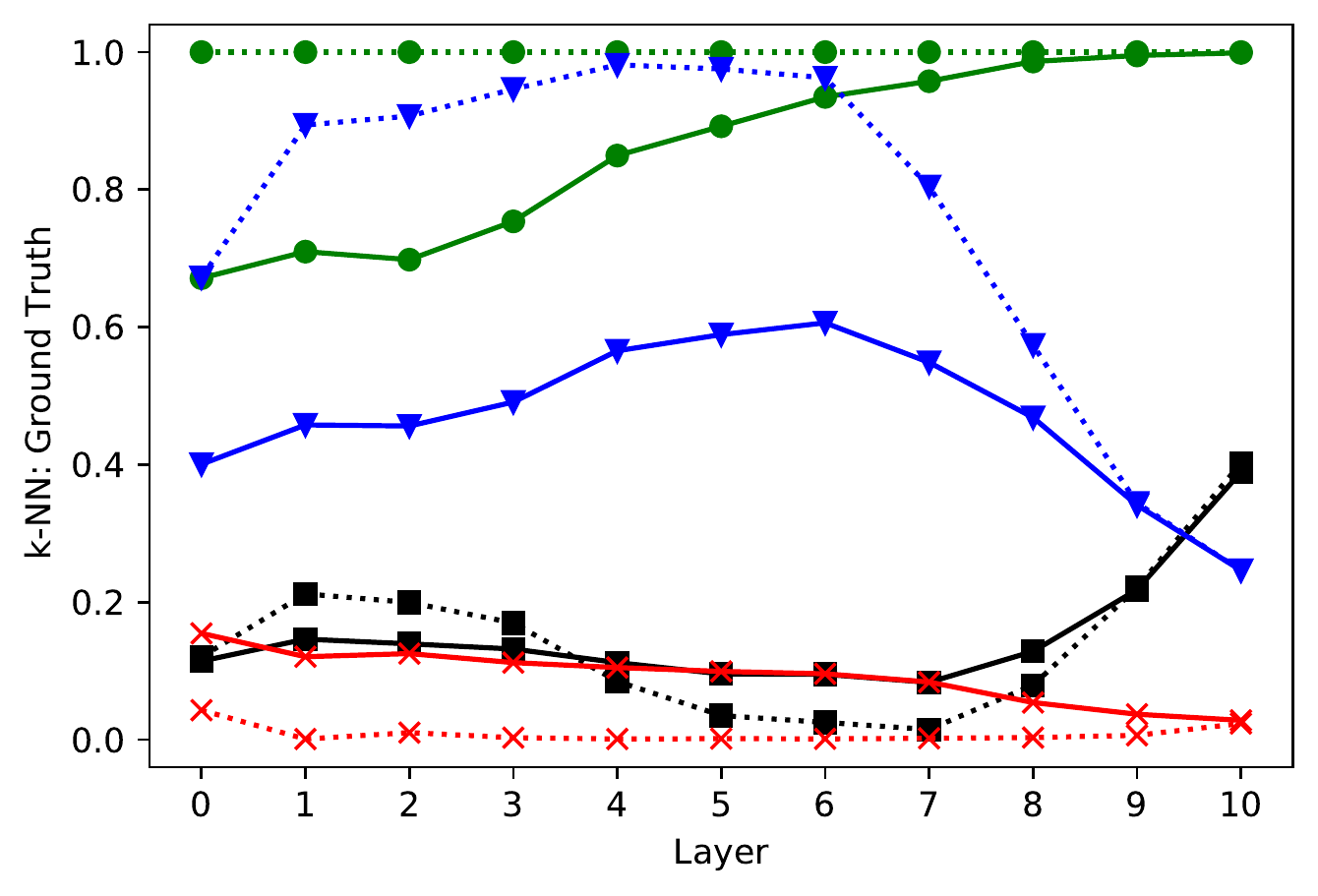}%
\quad
\includegraphics[trim=0 0 0 0, clip,height=5cm]{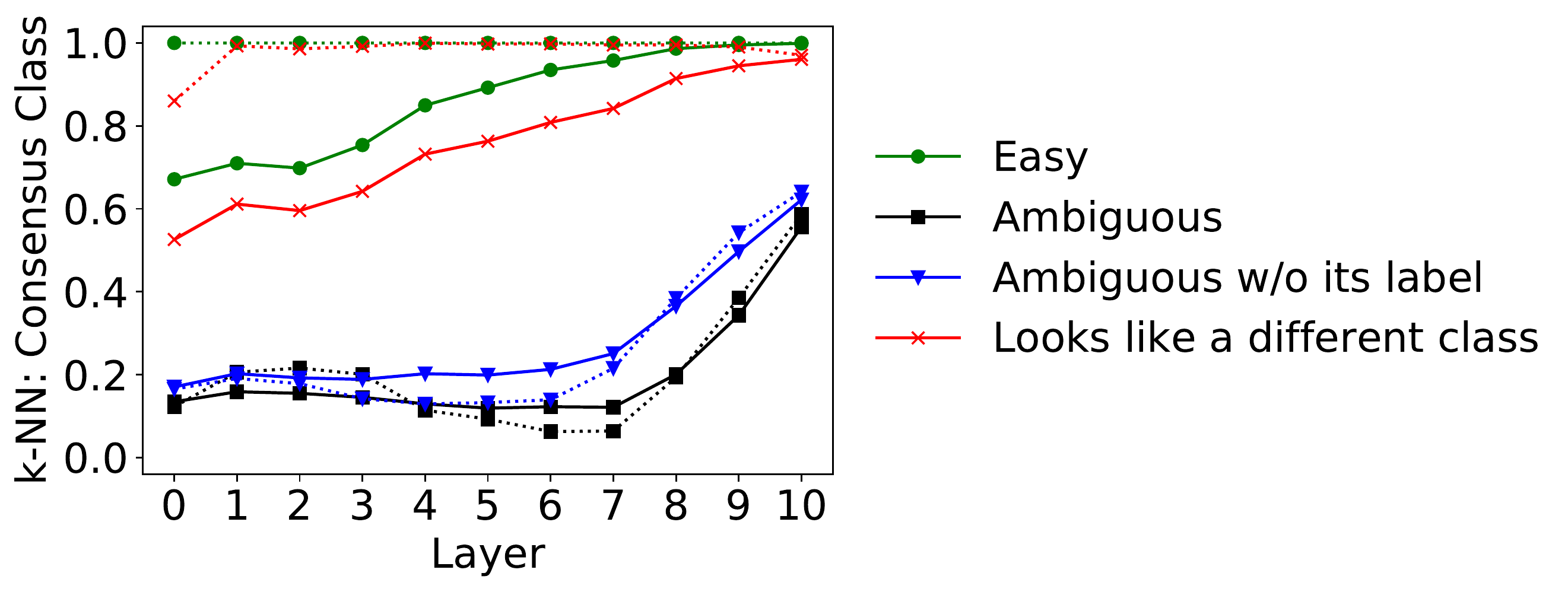}%
}
\end{center}
\vspace*{-7mm}
\caption{Reproducing Figure~\ref{fig:knn_4forms} for ResNet18 on CIFAR10.
\label{fig:cluster_4forms_all_1}
}
\end{figure*}

\begin{figure*}[ht!]
\begin{center} \resizebox{1.\textwidth}{!}{%
\includegraphics[trim=0 0 0 0, clip,height=5cm]{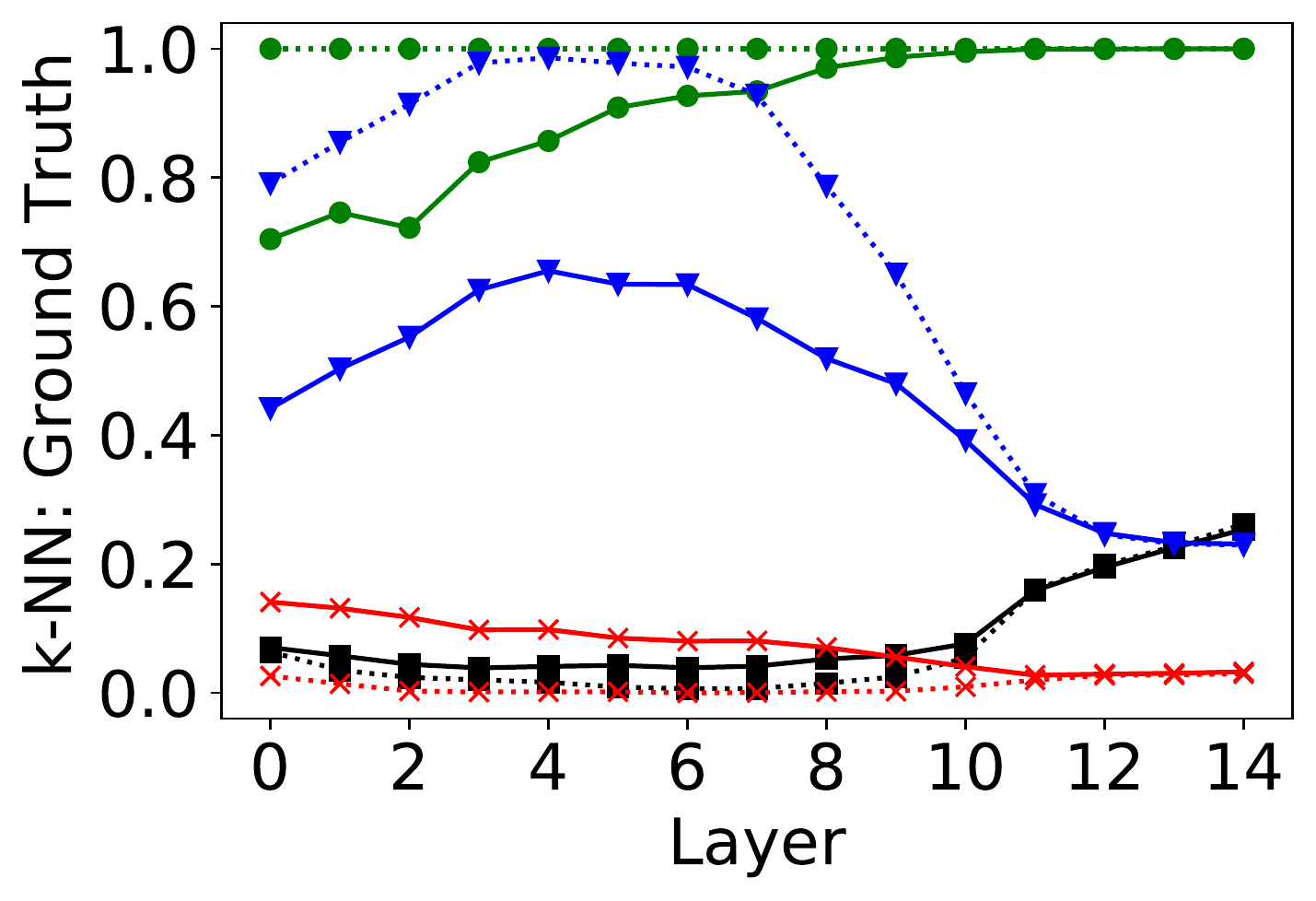}%
\quad
\includegraphics[trim=0 0 0 0, clip,height=5cm]{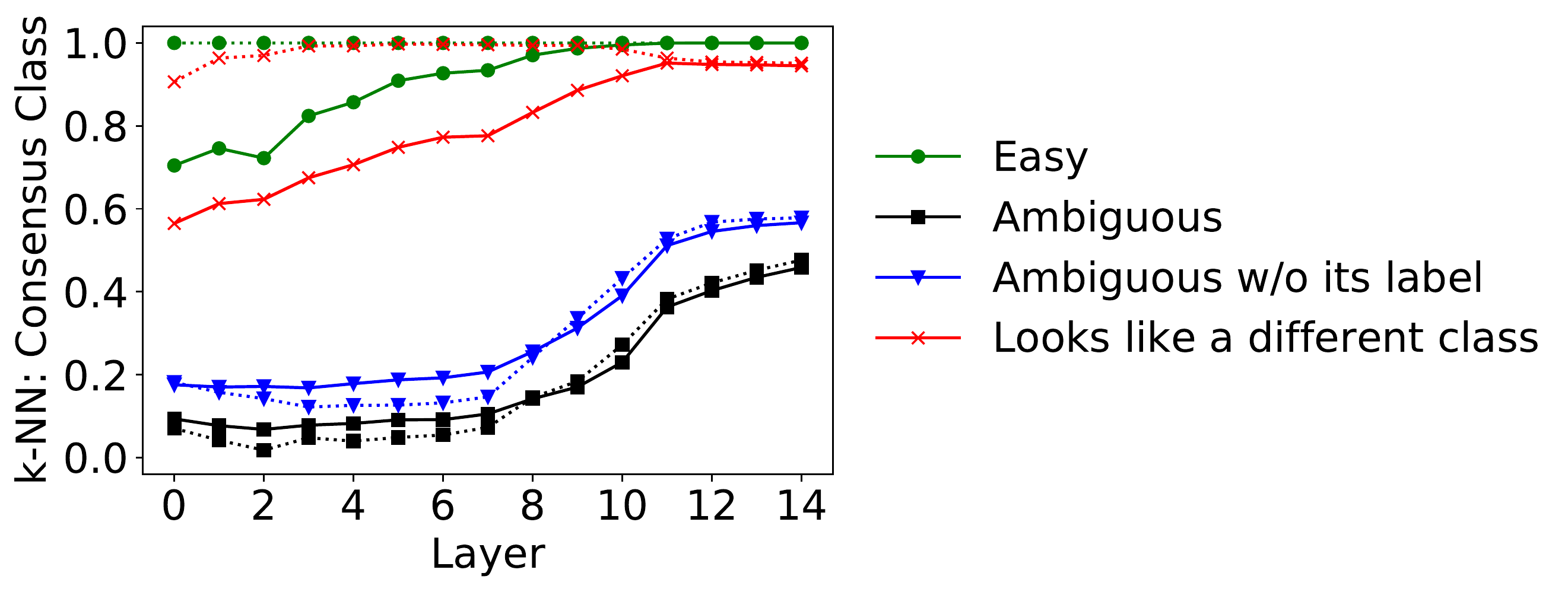}%
}
\end{center}
\vspace*{-7mm}
\caption{Reproducing Figure~\ref{fig:knn_4forms} for VGG16 on CIFAR10.
\label{fig:cluster_4forms_all_2p5}
}
\end{figure*}

\begin{figure*}[ht!]
\begin{center} \resizebox{1.\textwidth}{!}{%
\includegraphics[trim=0 0 0 0, clip,height=5cm]{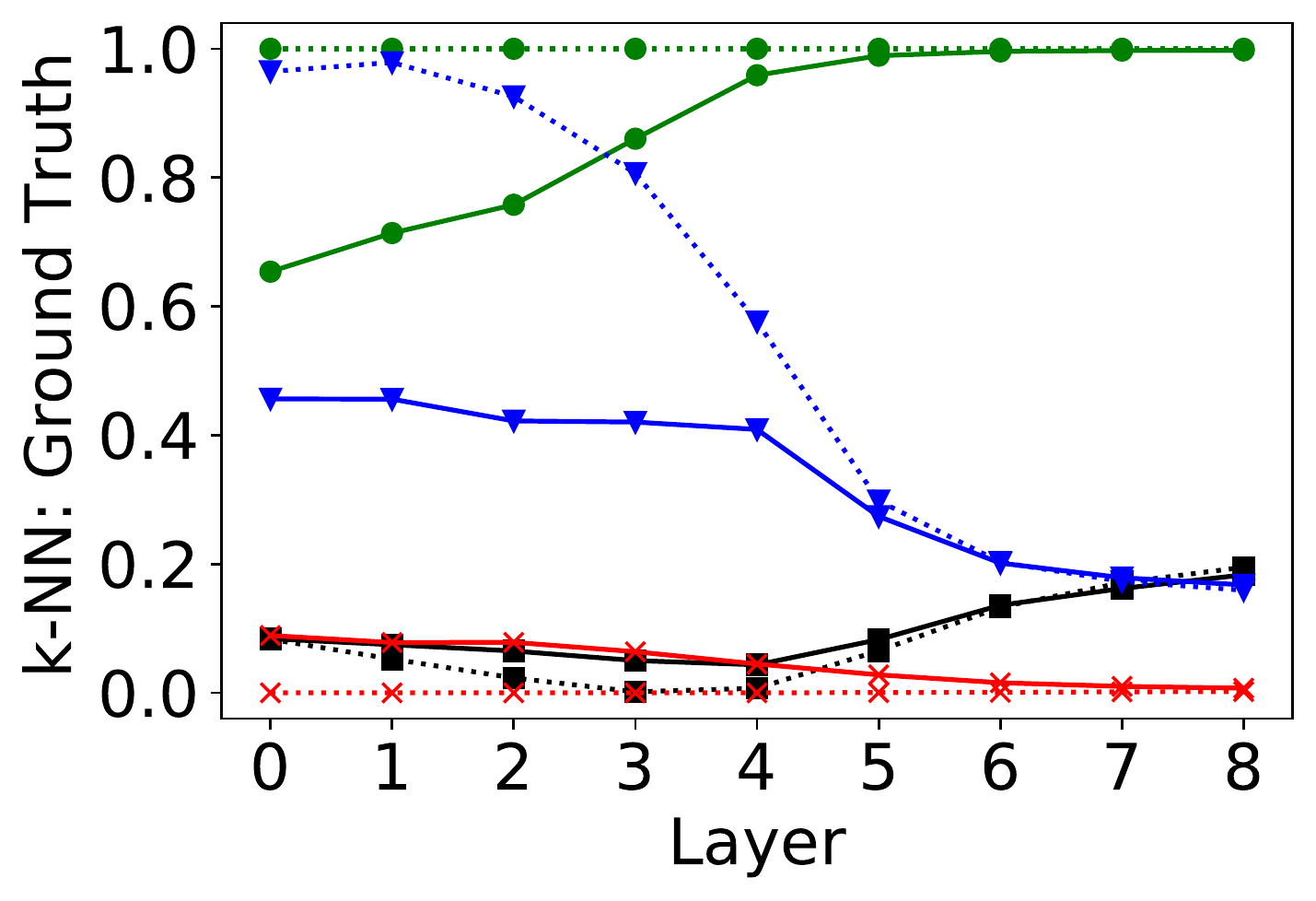}%
\quad
\includegraphics[trim=0 0 0 0, clip,height=5cm]{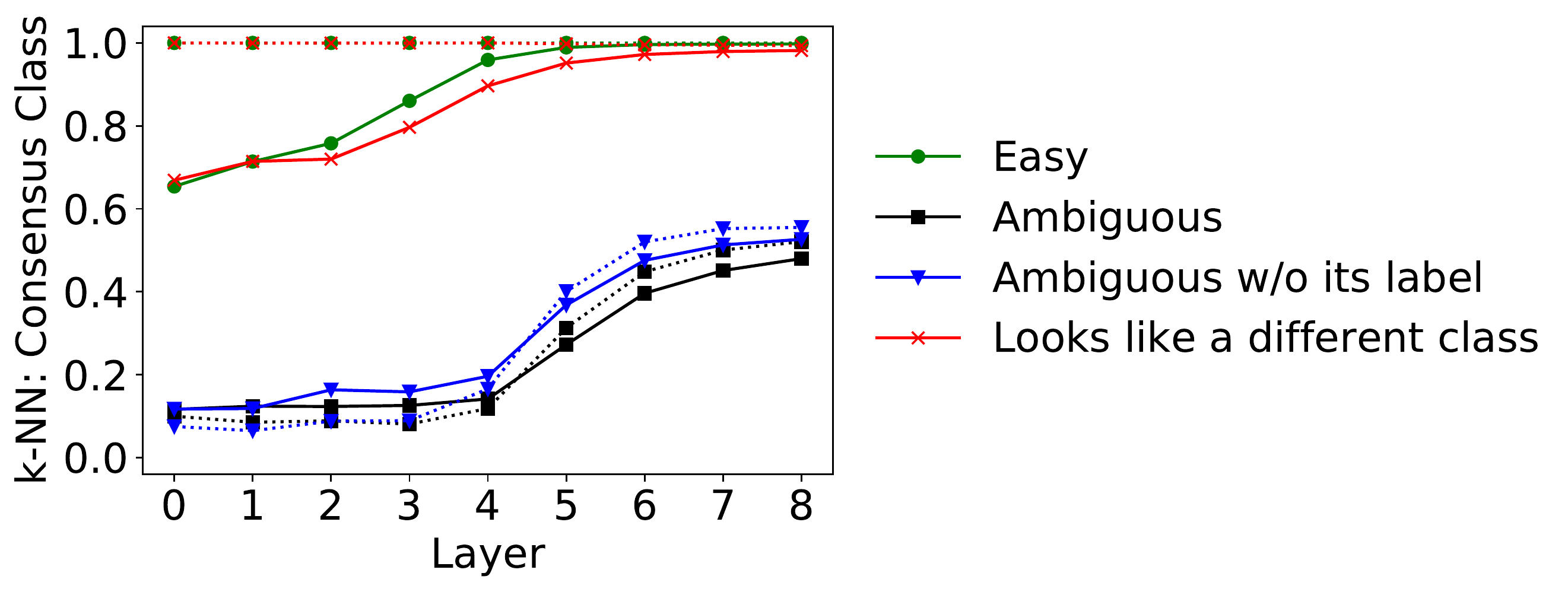}%
}
\end{center}
\vspace*{-7mm}
\caption{Reproducing Figure~\ref{fig:knn_4forms} for MLP on CIFAR10.
\label{fig:cluster_4forms_all_2}
}
\end{figure*}

\begin{figure*}[ht!]
\begin{center} \resizebox{1.\textwidth}{!}{%
\includegraphics[trim=0 0 0 0, clip,height=5cm]{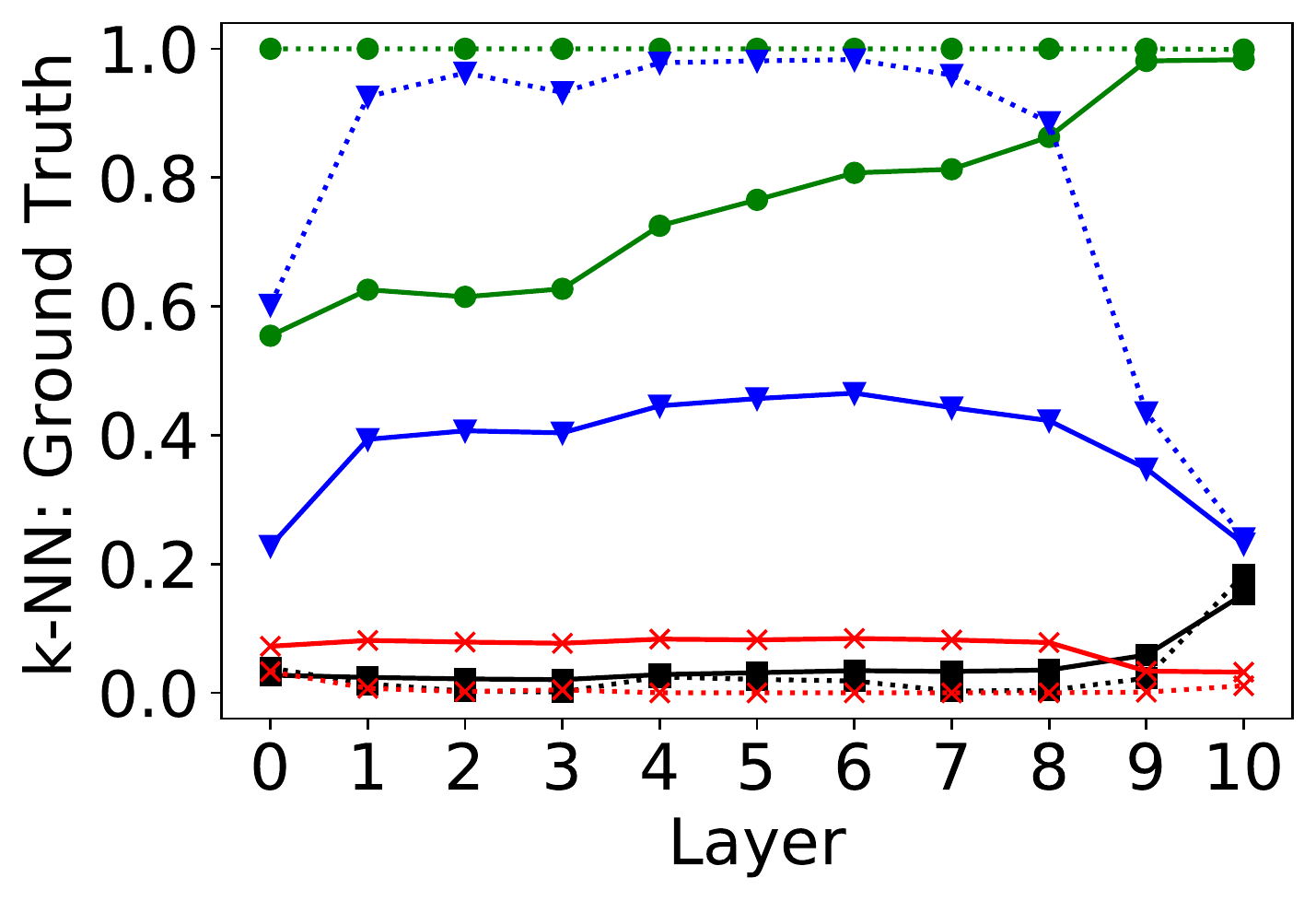}%
\quad
\includegraphics[trim=0 0 0 0, clip,height=5cm]{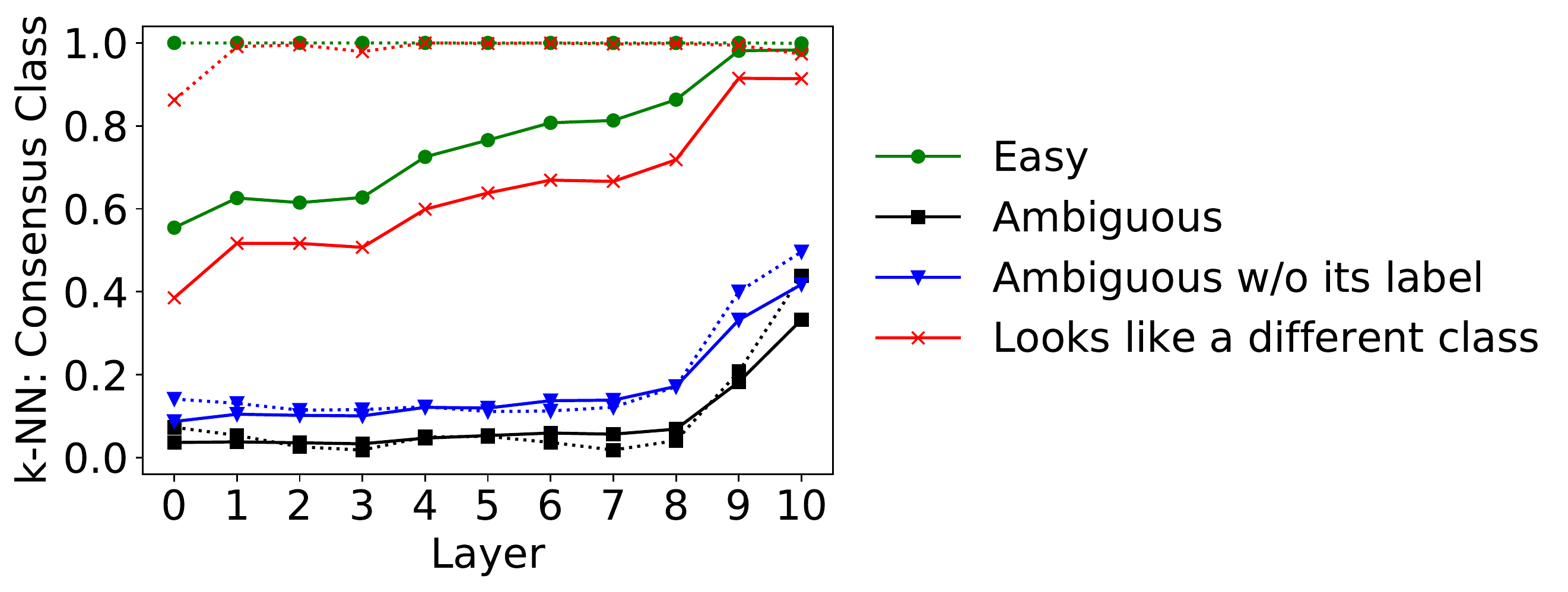}%
}
\end{center}
\vspace*{-7mm}
\caption{Reproducing Figure~\ref{fig:knn_4forms} for ResNet18 on CIFAR100.
\label{fig:cluster_4forms_all_3}
}
\end{figure*}

\begin{figure*}[ht!]
\begin{center} \resizebox{1.\textwidth}{!}{%
\includegraphics[trim=0 0 0 0, clip,height=5cm]{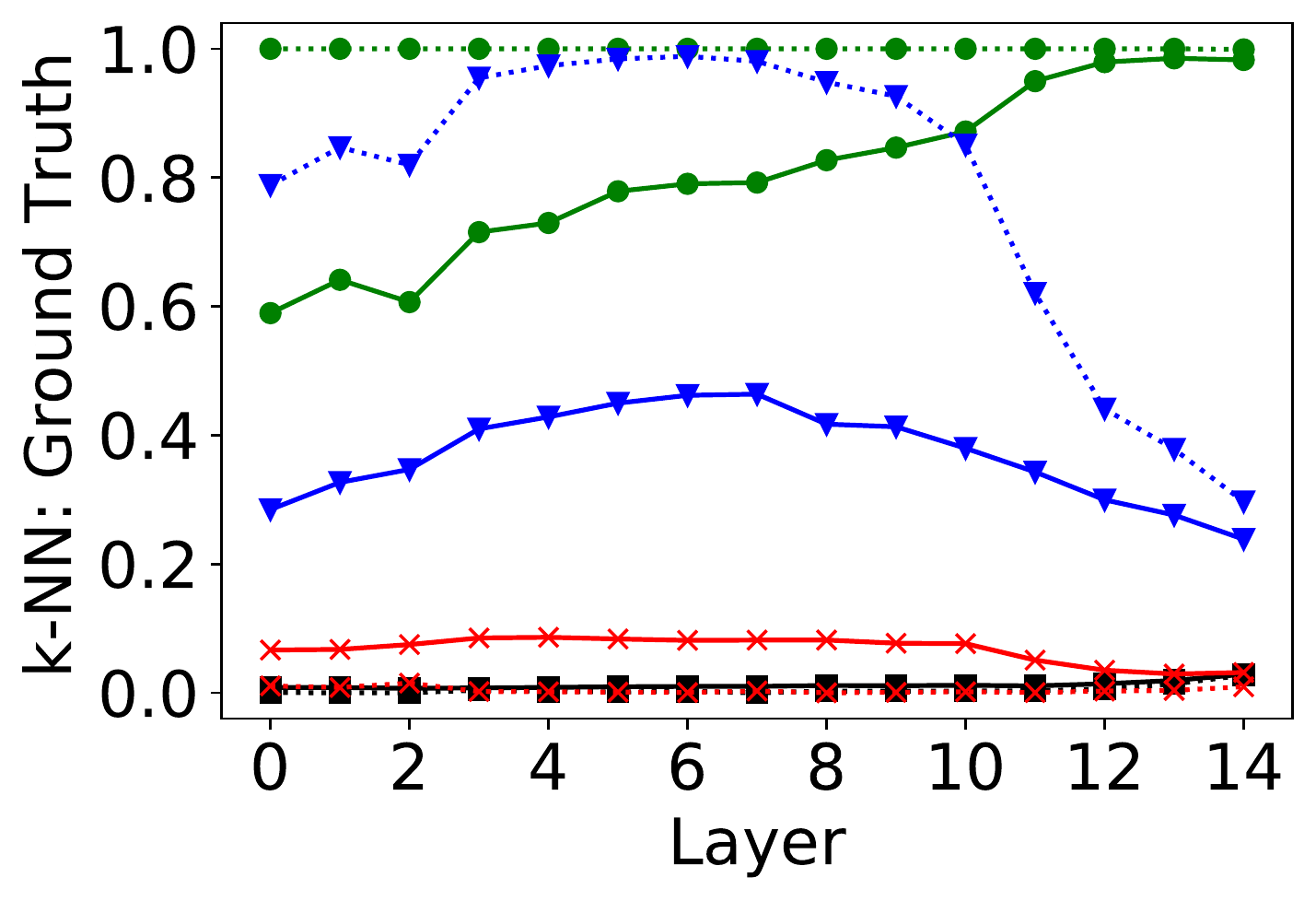}%
\quad
\includegraphics[trim=0 0 0 0, clip,height=5cm]{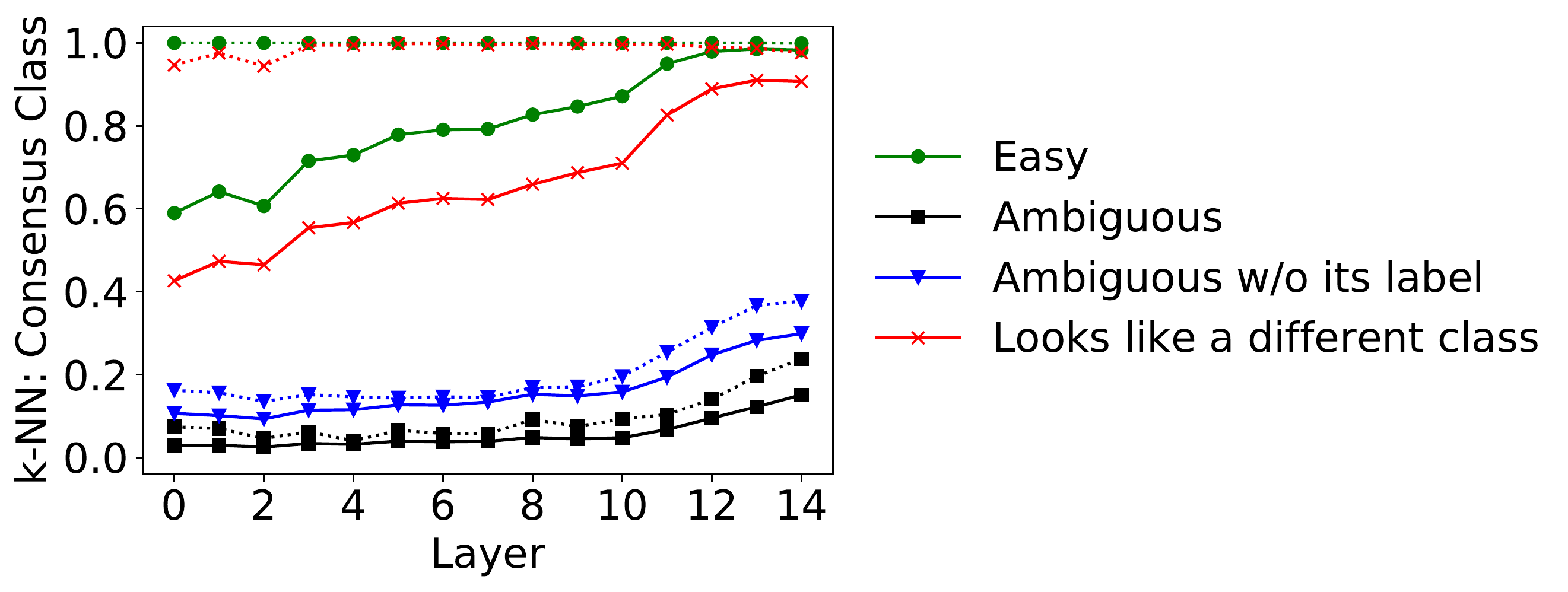}%
}
\end{center}
\vspace*{-7mm}
\caption{Reproducing Figure~\ref{fig:knn_4forms} for VGG16 on CIFAR100.
\label{fig:cluster_4forms_all_4}
}
\end{figure*}

\begin{figure*}[ht!]
\begin{center} \resizebox{1.\textwidth}{!}{%
\includegraphics[trim=0 0 0 0, clip,height=5cm]{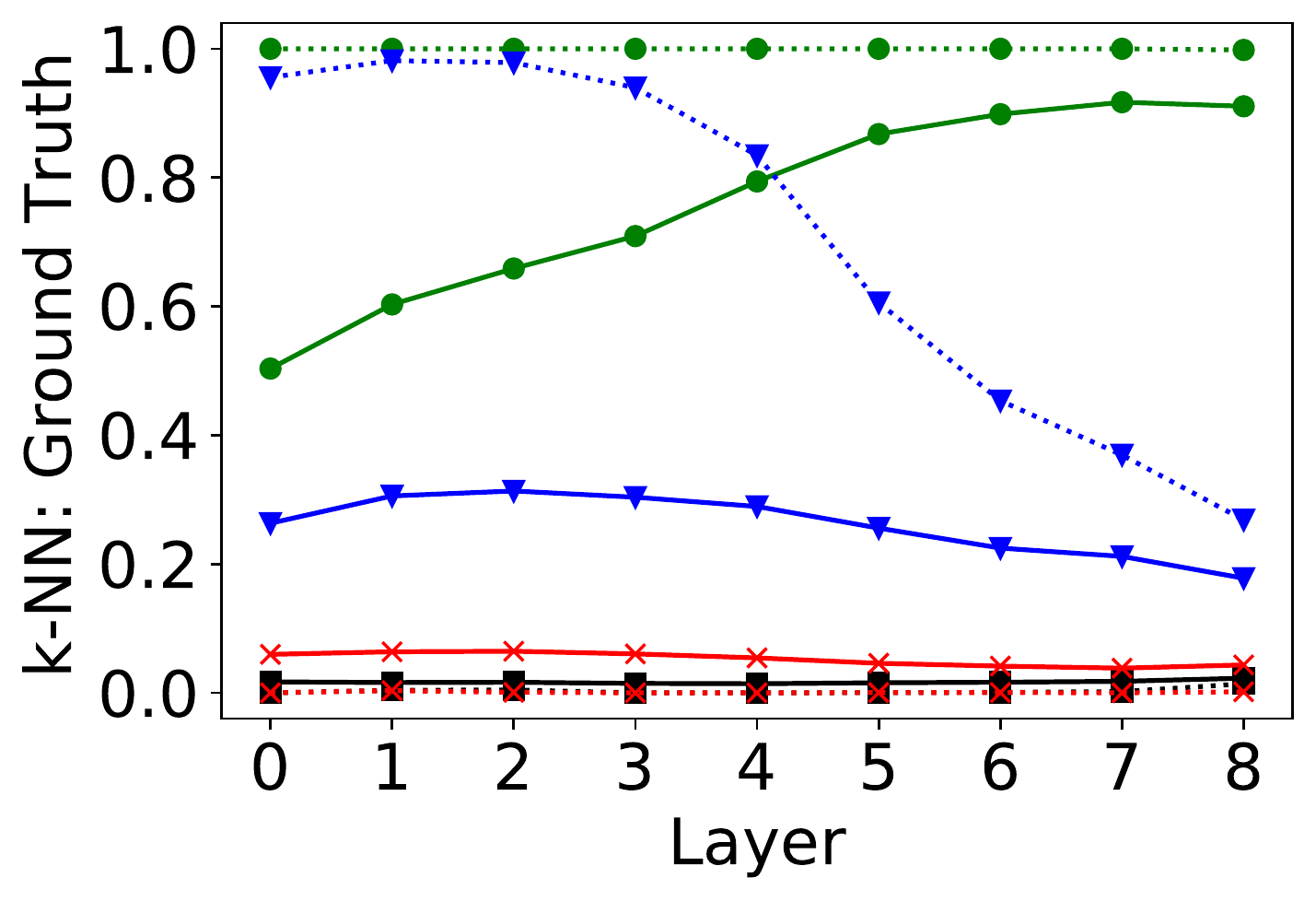}%
\quad
\includegraphics[trim=0 0 0 0, clip,height=5cm]{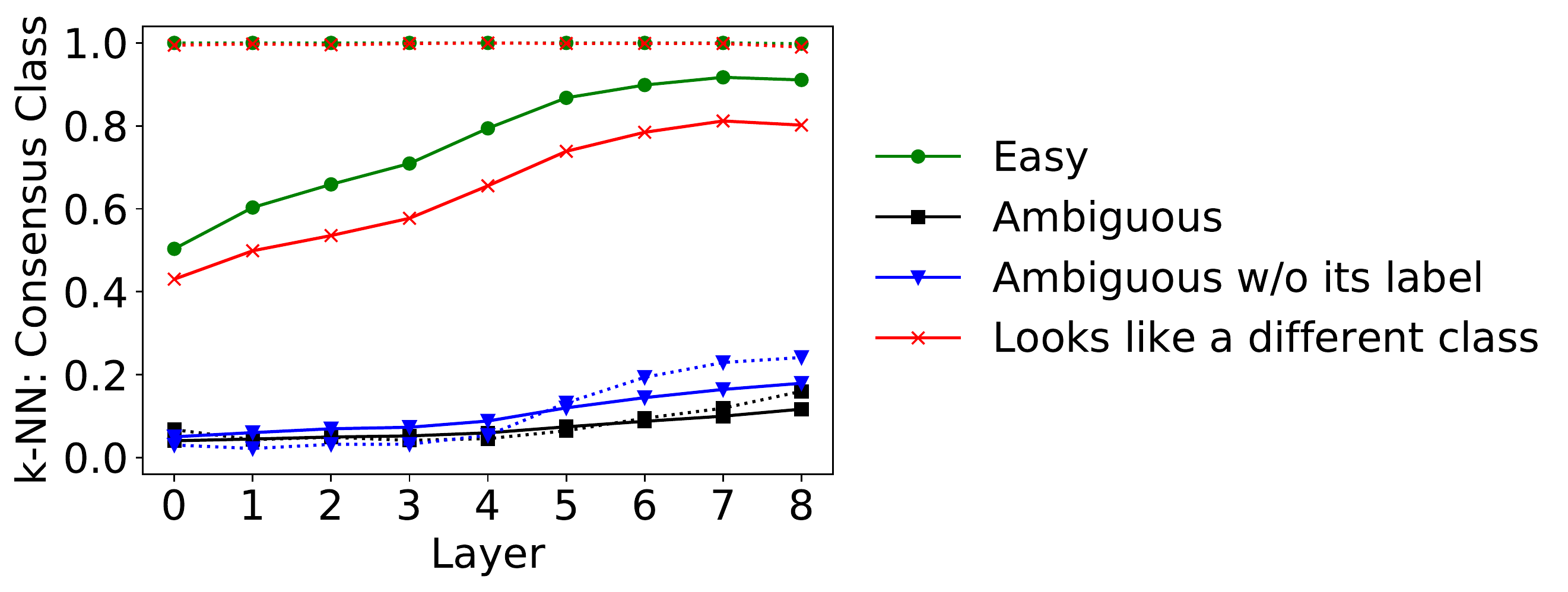}%
}
\end{center}
\vspace*{-7mm}
\caption{Reproducing Figure~\ref{fig:knn_4forms} for MLP on CIFAR100.
\label{fig:cluster_4forms_all_5}
}
\end{figure*}

\begin{figure*}[ht!]
\begin{center} \resizebox{1.\textwidth}{!}{%
\includegraphics[trim=0 0 0 0, clip,height=5cm]{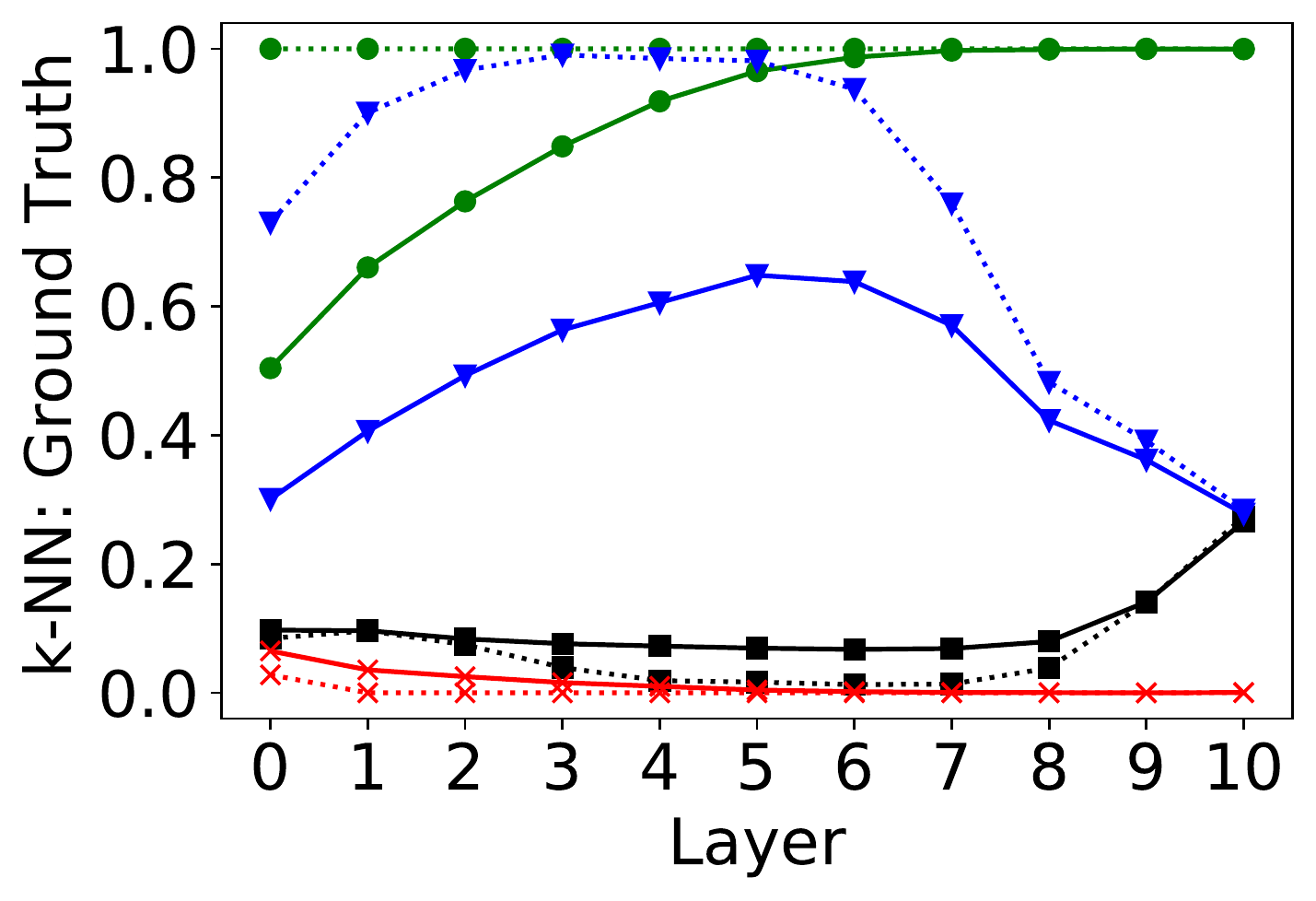}%
\quad
\includegraphics[trim=0 0 0 0, clip,height=5cm]{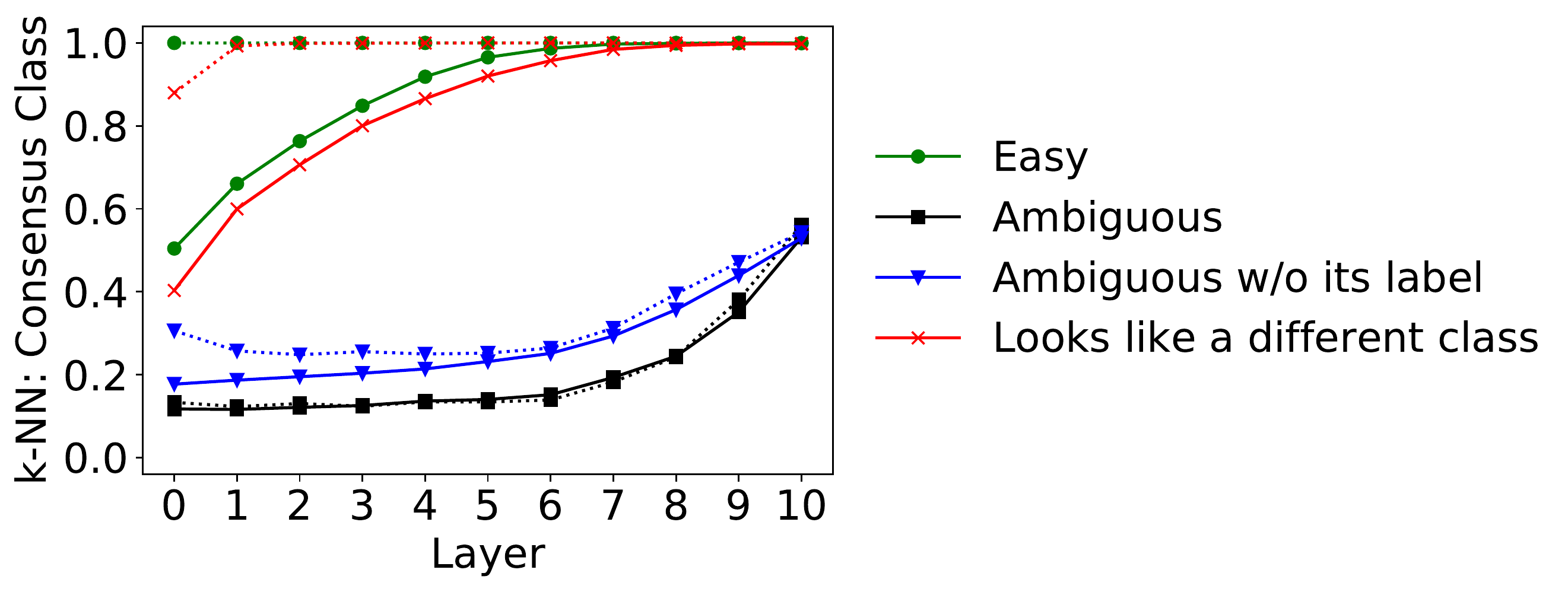}%
}
\end{center}
\vspace*{-7mm}
\caption{Reproducing Figure~\ref{fig:knn_4forms} for ResNet18 on SVHN.
\label{fig:cluster_4forms_all_6}
}
\end{figure*}

\begin{figure*}[ht!]
\begin{center} \resizebox{1.\textwidth}{!}{%
\includegraphics[trim=0 0 0 0, clip,height=5cm]{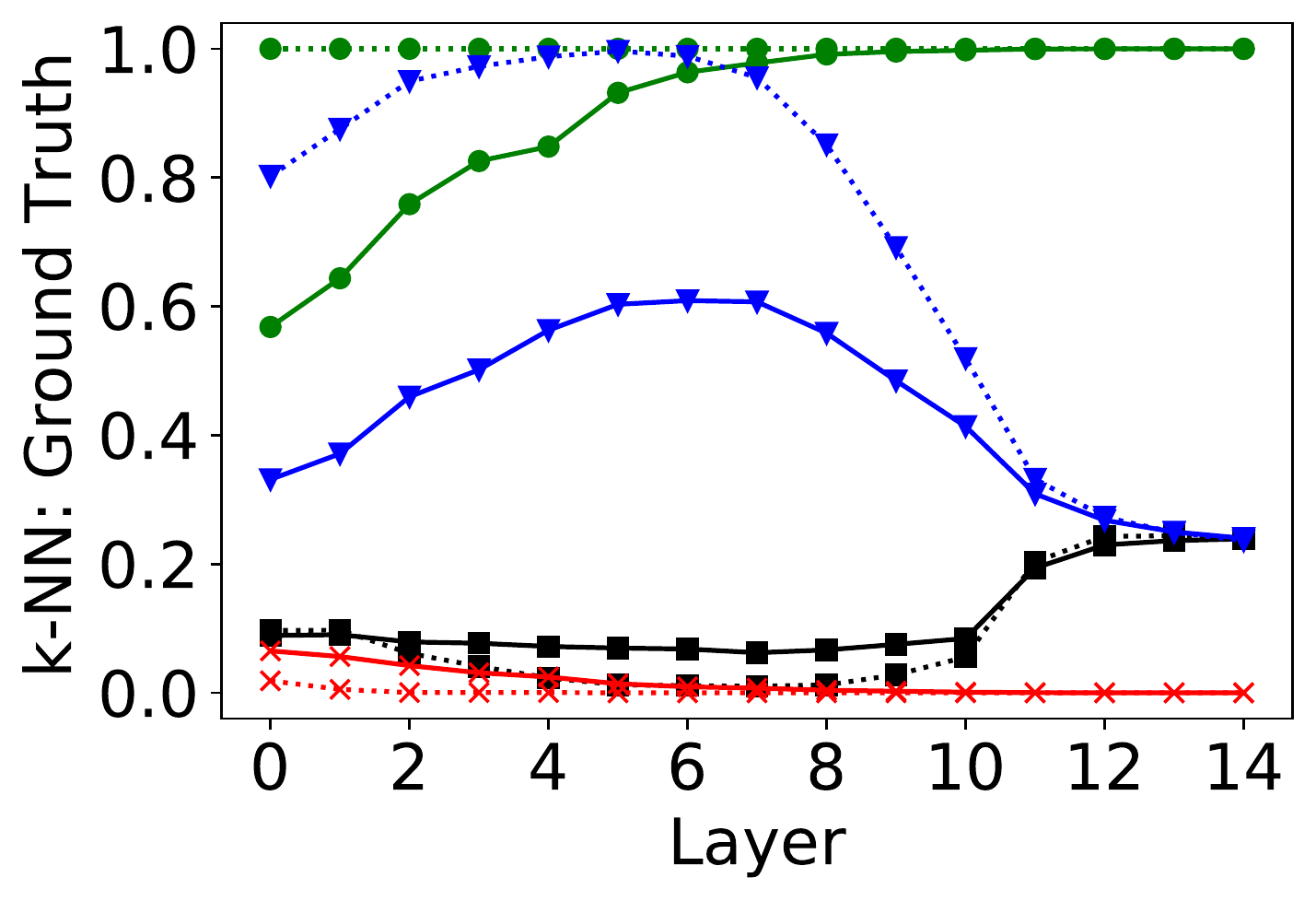}%
\quad
\includegraphics[trim=0 0 0 0, clip,height=5cm]{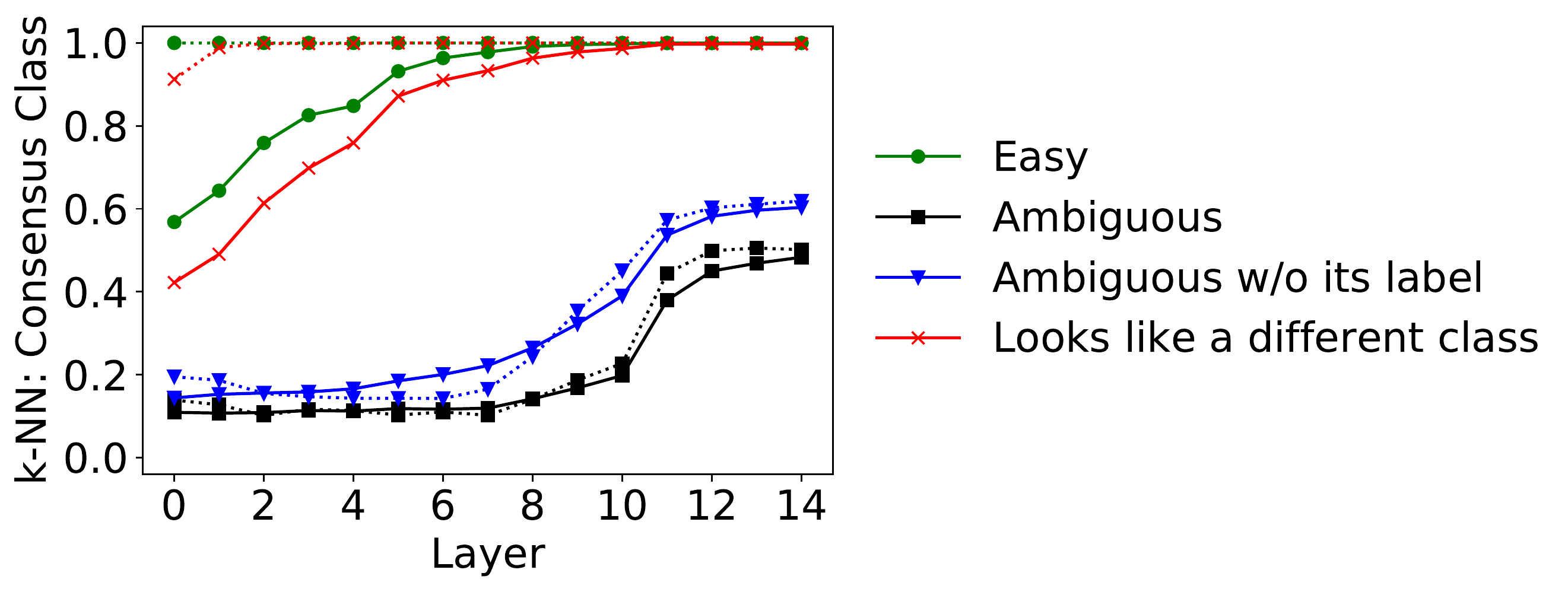}%
}
\end{center}
\vspace*{-7mm}
\caption{Reproducing Figure~\ref{fig:knn_4forms} for VGG16 on SVHN.
\label{fig:cluster_4forms_all_7}
}
\end{figure*}

\begin{figure*}[ht!]
\begin{center} \resizebox{1.\textwidth}{!}{%
\includegraphics[trim=0 0 0 0, clip,height=5cm]{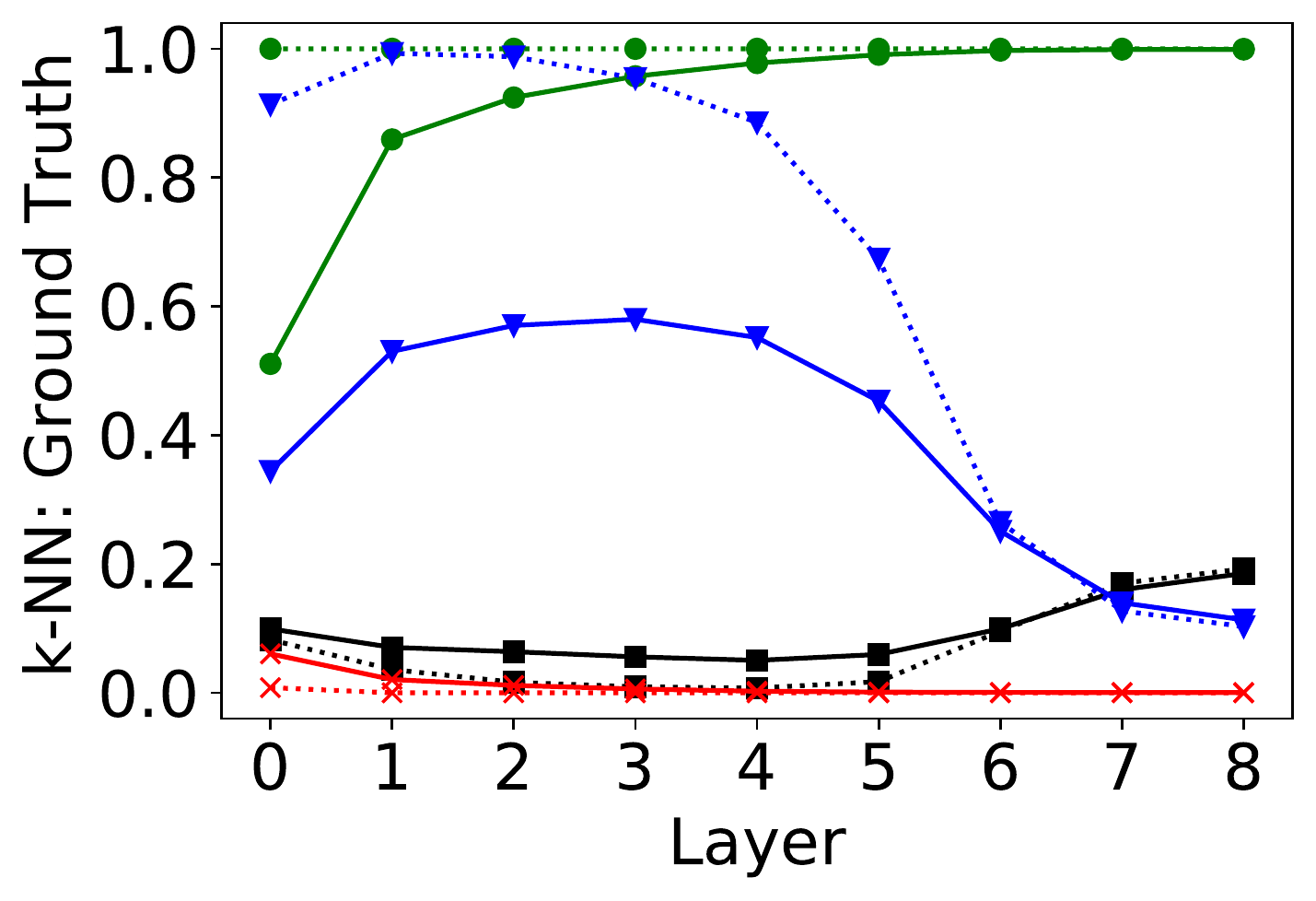}%
\quad
\includegraphics[trim=0 0 0 0, clip,height=5cm]{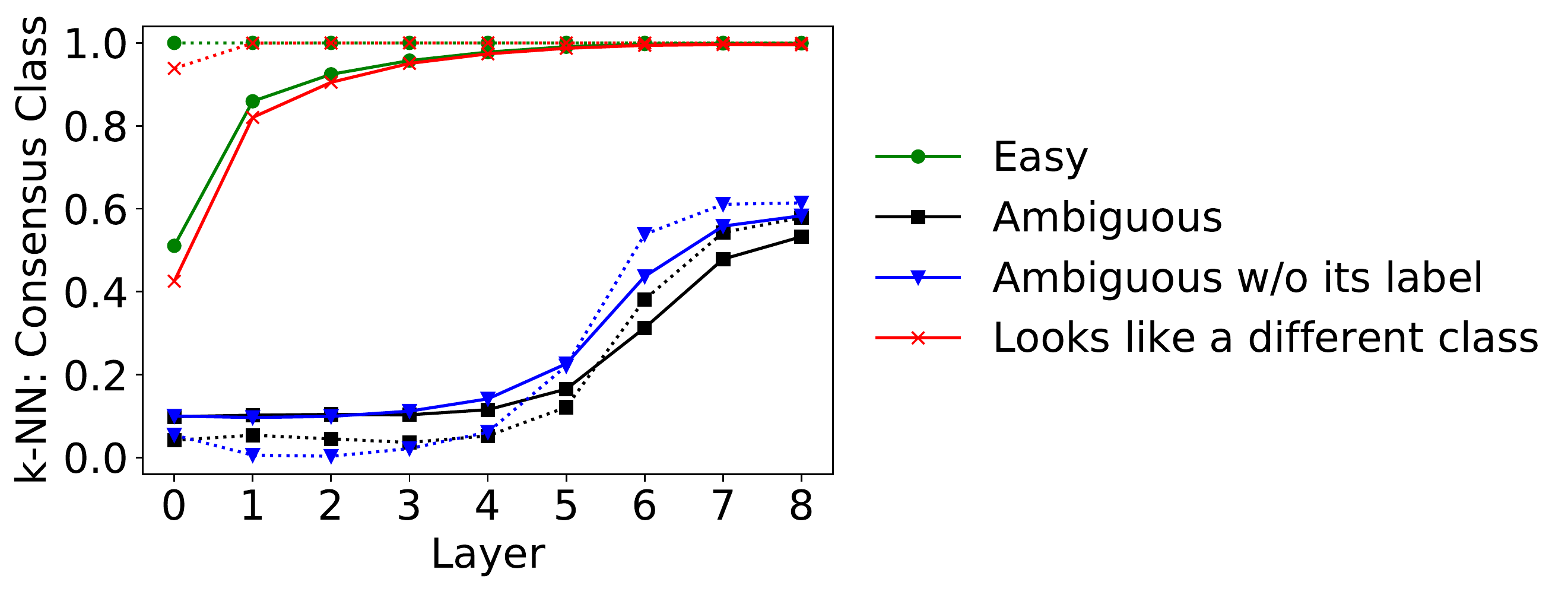}%
}
\end{center}
\vspace*{-7mm}
\caption{Reproducing Figure~\ref{fig:knn_4forms} for MLP on SVHN.
\label{fig:cluster_4forms_all_8}
}
\end{figure*}

\begin{figure*}[ht!]
\begin{center} \resizebox{1.\textwidth}{!}{%
\includegraphics[trim=0 0 0 0, clip,height=5cm]{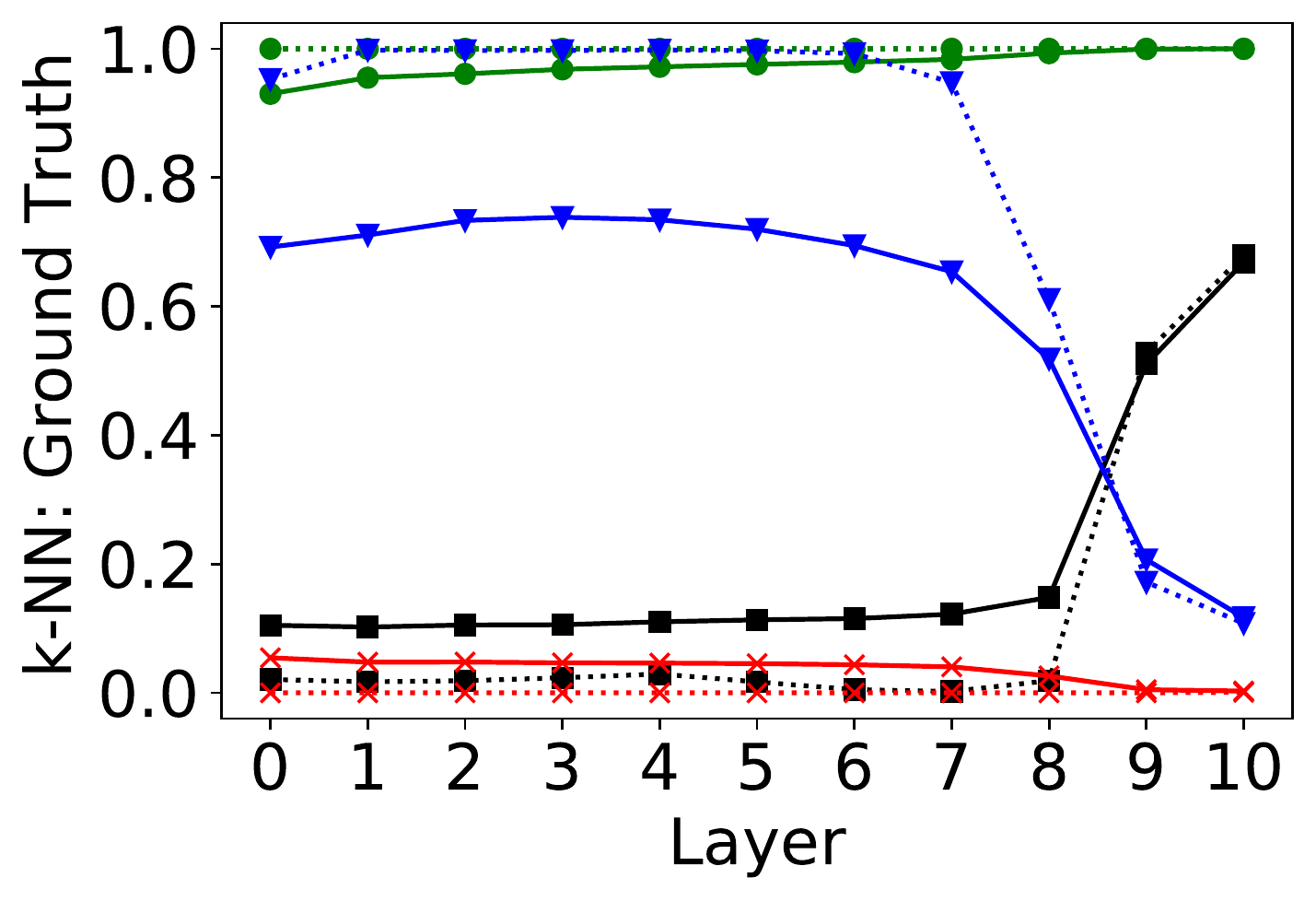}%
\quad
\includegraphics[trim=0 0 0 0, clip,height=5cm]{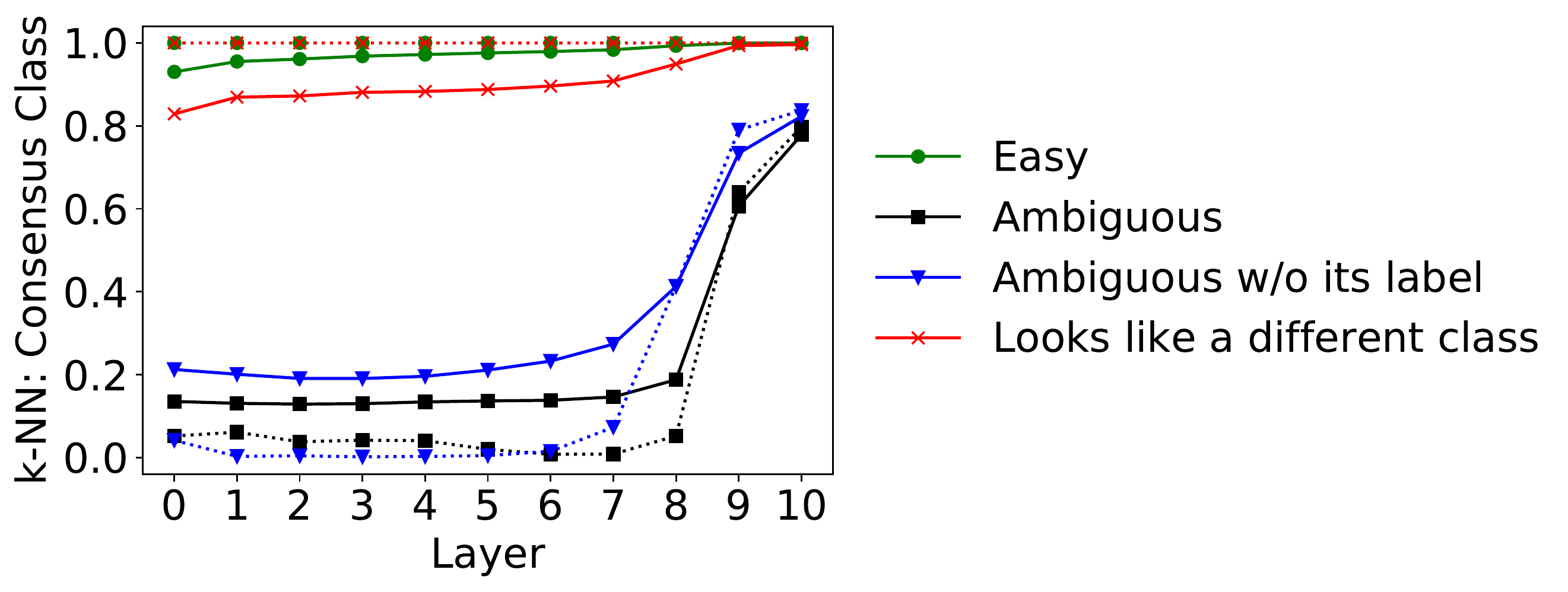}%
}
\end{center}
\vspace*{-7mm}
\caption{Reproducing Figure~\ref{fig:knn_4forms} for ResNet18 on Fashion MNIST.
\label{fig:cluster_4forms_all_9}
}
\end{figure*}

\begin{figure*}[ht!]
\begin{center} \resizebox{1.\textwidth}{!}{%
\includegraphics[trim=0 0 0 0, clip,height=5cm]{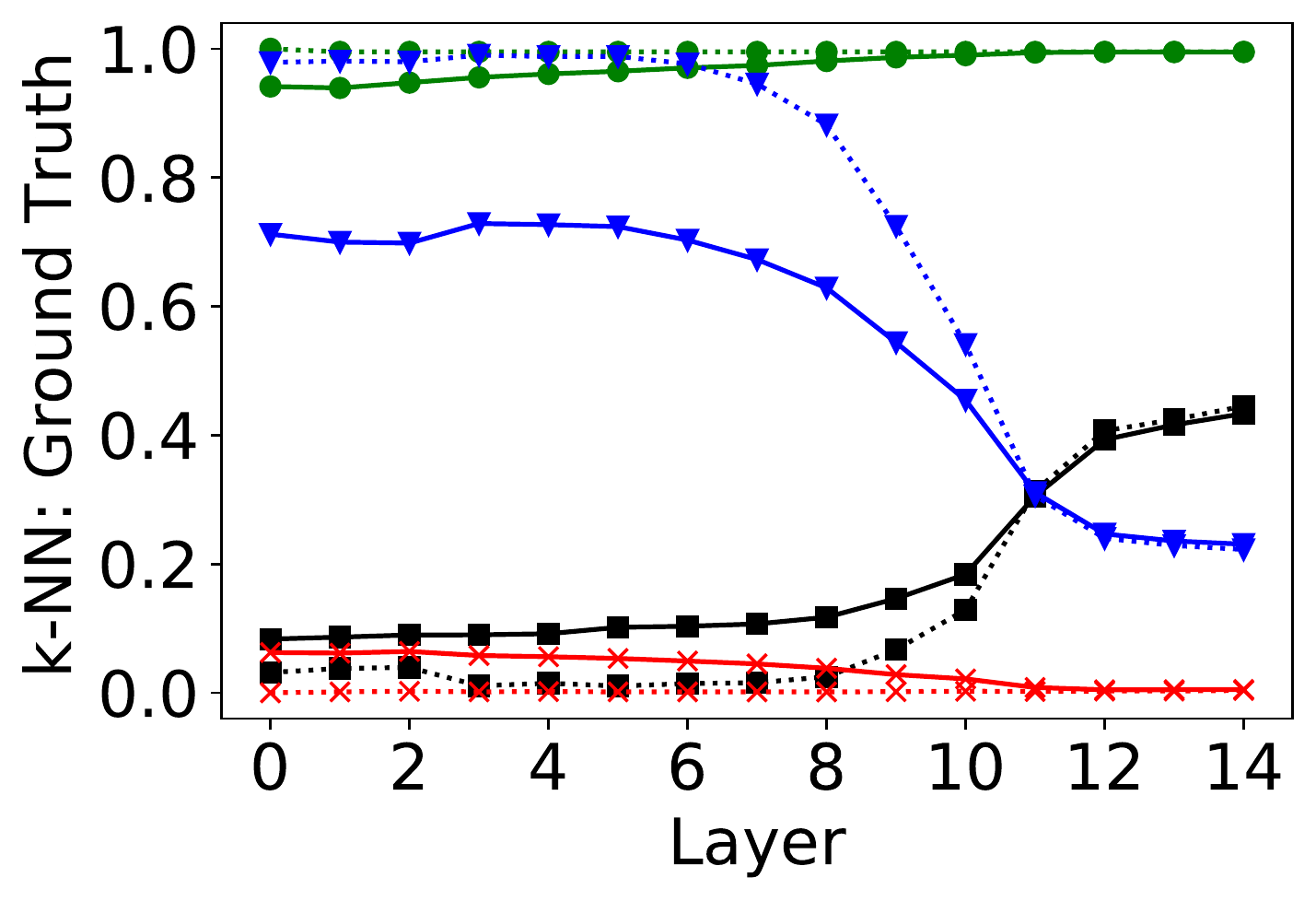}%
\quad
\includegraphics[trim=0 0 0 0, clip,height=5cm]{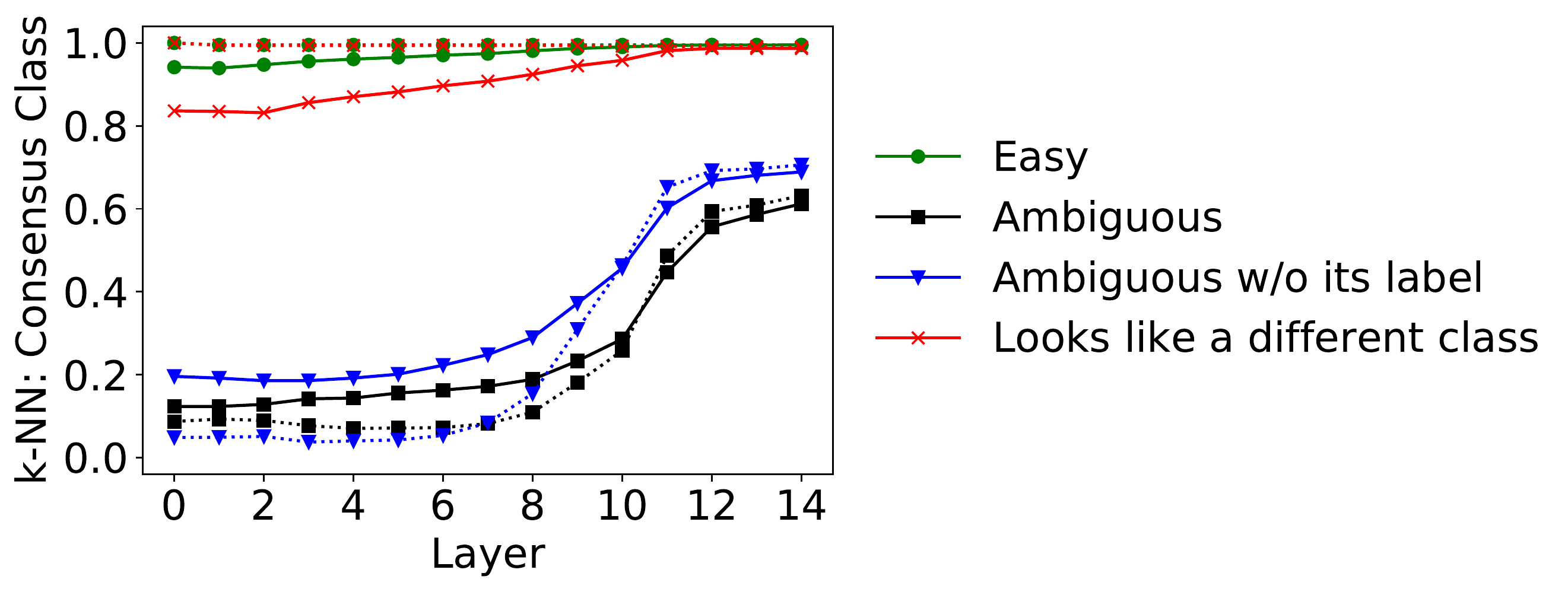}%
}
\end{center}
\vspace*{-7mm}
\caption{Reproducing Figure~\ref{fig:knn_4forms} for VGG16 on Fashion MNIST.
\label{fig:cluster_4forms_all_10}
}
\end{figure*}

\begin{figure*}[ht!]
\begin{center} \resizebox{1.\textwidth}{!}{%
\includegraphics[trim=0 0 0 0, clip,height=5cm]{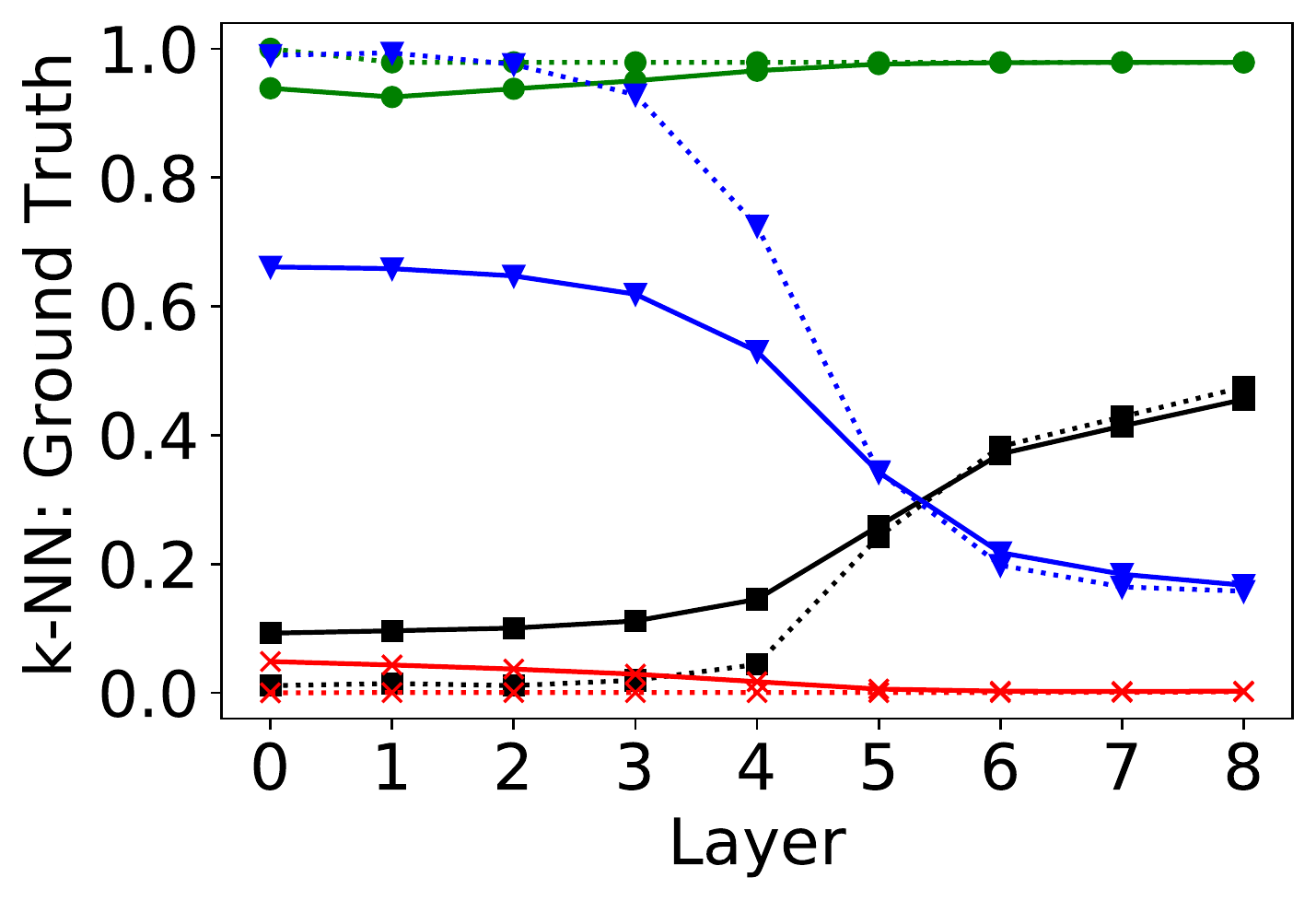}%
\quad
\includegraphics[trim=0 0 0 0, clip,height=5cm]{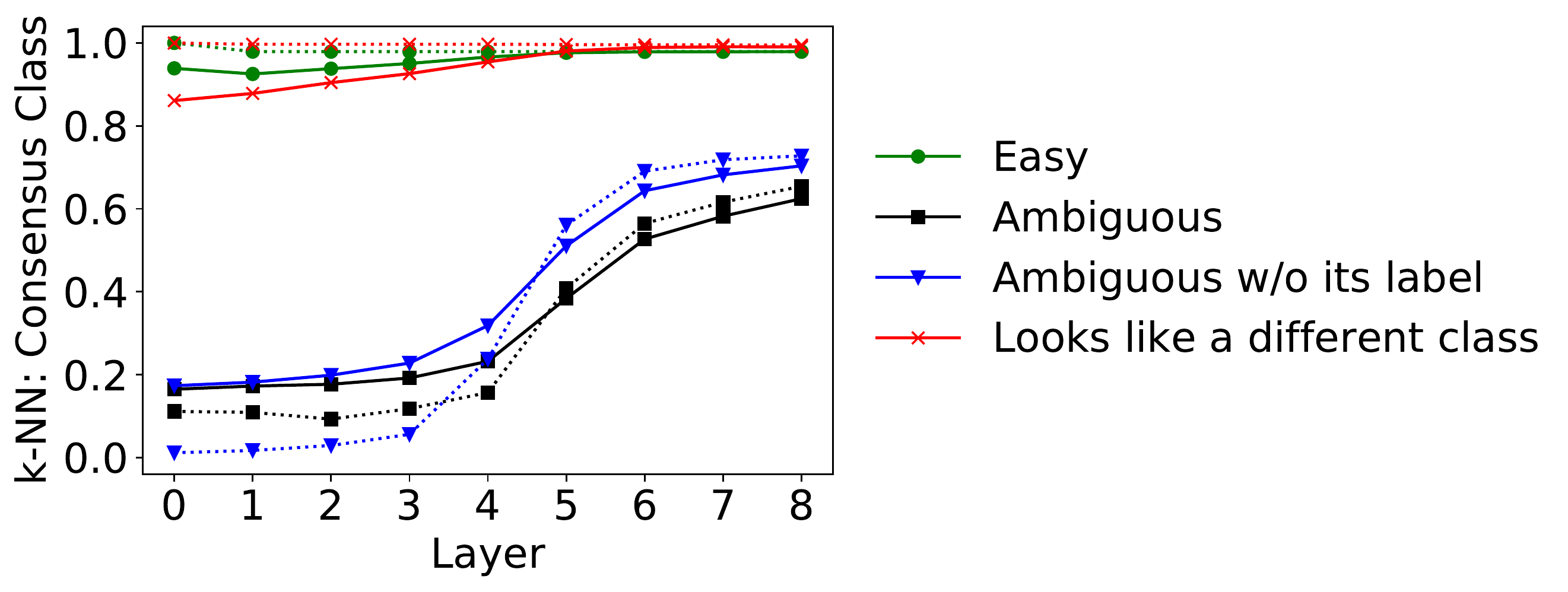}%
}
\end{center}
\vspace*{-7mm}
\caption{Reproducing Figure~\ref{fig:knn_4forms} for MLP on Fashion MNIST.
\label{fig:cluster_4forms_all_11}
}
\end{figure*}

Figures~\ref{fig:cluster_4forms_all_1} to~\ref{fig:cluster_4forms_all_11} reproduce similar behavior to that shown in Figure~\ref{fig:knn_4forms} for all datasets and architectures.
Please see Figure~\ref{fig:knn_4forms} for a detailed description.

\section{Pertinence of example difficulty to topics in machine learning \label{app:exdiff_pertinence}}

We will describe the relevance of our work to distribution shift and robustness; algorithmic fairness, curriculum learning and models that explicitly address heteroscedastic uncertainty.

\begin{description}
\item[Distribution Shift and Robustness:] Recent work has hypothesized that the linear relationship between the performance of a model before and after distribution shift could potentially be explained in a theory based on the difficulty of examples \citep{recht2019imagenet}. Recent work has additionally discussed how examples that belong to a minority group might appear difficult to classify correctly under distribution shift~\citep{nagarajan2020understanding}.
Therefore it seems natural to suppose that the richer picture of example difficulty we introduce could lead to a deeper understanding of distribution shift and aid with the development of more robust algorithms.

\item[Curriculum Learning:] This class of training algorithms exploits additional information about a dataset (obtained in advance) to present easier examples earlier in the training process~\citep{elman1993learning, sanger1994neural, bengio2009curriculum}. 
Different notions of difficulty have been the subject of several related studies~\citep{bengio2009curriculum,forgetting19,hacohen2019curriculum} and it has been shown that (neglecting the cost of obtaining the curriculum) following a curriculum can improve training time significantly, particularly for large training data \citep{wu2020curricula}.
We envisage that richer, more effective curricula could be designed by distinguishing different forms of example difficulty. This could, for example, be achieved setting the curriculum according to a each data point's location in Figure~\ref{fig:ll_test_v_train}.

\item[Algorithmic Fairness:]
We have seen that mislabeled data is processed similarly to data that simply looks mislabeled to the algorithm (both ``look like a different class''). This presents a fairness challenge when filtering ``noisy labels''.
Similarly, we have seen that examples of rare subgroups (which are essential to include in the training set for robustness~\citep{feldman2020neural} and fairness~\citep{hooker2020characterising} are processed similarly to truly ``ambiguous'' inputs.
Finding ways to deal with ``label noise'' without biasing against these subgroups remains an open challenge. 
In further work, we anticipate that examining datasets in an enlarged space of different example difficulty measures~\citep{chiyuan_cscores,forgetting19, carlini2019distribution, hooker2019compressed, lalor2018understanding, hooker_exdiff_gradvar} may allow algorithms that distinguish between these different sources of label noise to reach higher accuracy and to be fairer.

\item[Heteroscedastic Uncertainty:]
There are a class of models with two heads, one to model the mean and the other the uncertainty of the prediction (E.g.~\citet{kendall2017uncertainties, kendall2018multi}).
These models learn to become uncertain on difficult inputs and treat example difficulty as a one-dimensional quantity.
It seems highly likely that this uncertainty will lead to the model down-weighting examples of rare subgroups in the data.
We suggest that methods for modeling uncertainty could additionally be tasked with estimating the location of a training point in Figure~\ref{fig:ll_test_v_train}.
It seems plausible to suppose that new models able to distinguish the form of an example's difficulty could later be refined to be fairer, more accurate and better  calibrated.
\end{description}

\section{Alternative Definitions for Prediction Depth \label{app:ll_alt_defs}}

Instead of using the network's final prediction on a data point to assign the prediction depth, one could instead use the ground truth label.
This would require a different rule for assigning a prediction depth to validation data points that are incorrectly classified as compared to data points that are correctly classified.
We consider our definition to be simpler than combining two separate rules. 

One could alternatively have defined the prediction depth for each example by first leaving it out of the training set, and then training networks of different depths to identify the number of layers required to classify it correctly. 
In fact, architectures of different depths have different inductive biases, so the relative difficulty of inputs can become inverted with changing depth~\citep{shallow_examples_first}.
Such an approach would be expensive but could lead to a rich picture of how example difficulty changes with architecture.

Another potential approach would have been to use a linear classifier such as Logistic Regression in the embedding spaces. 
Indeed linear probes, logistic regression and SVM probes have been previously applied to the hidden spaces of DNNs (E.g.~\citet{cohen2018dnn,alain2016understanding}). 

\begin{figure}[ht]
\begin{center}
         \includegraphics[width=0.49\columnwidth]{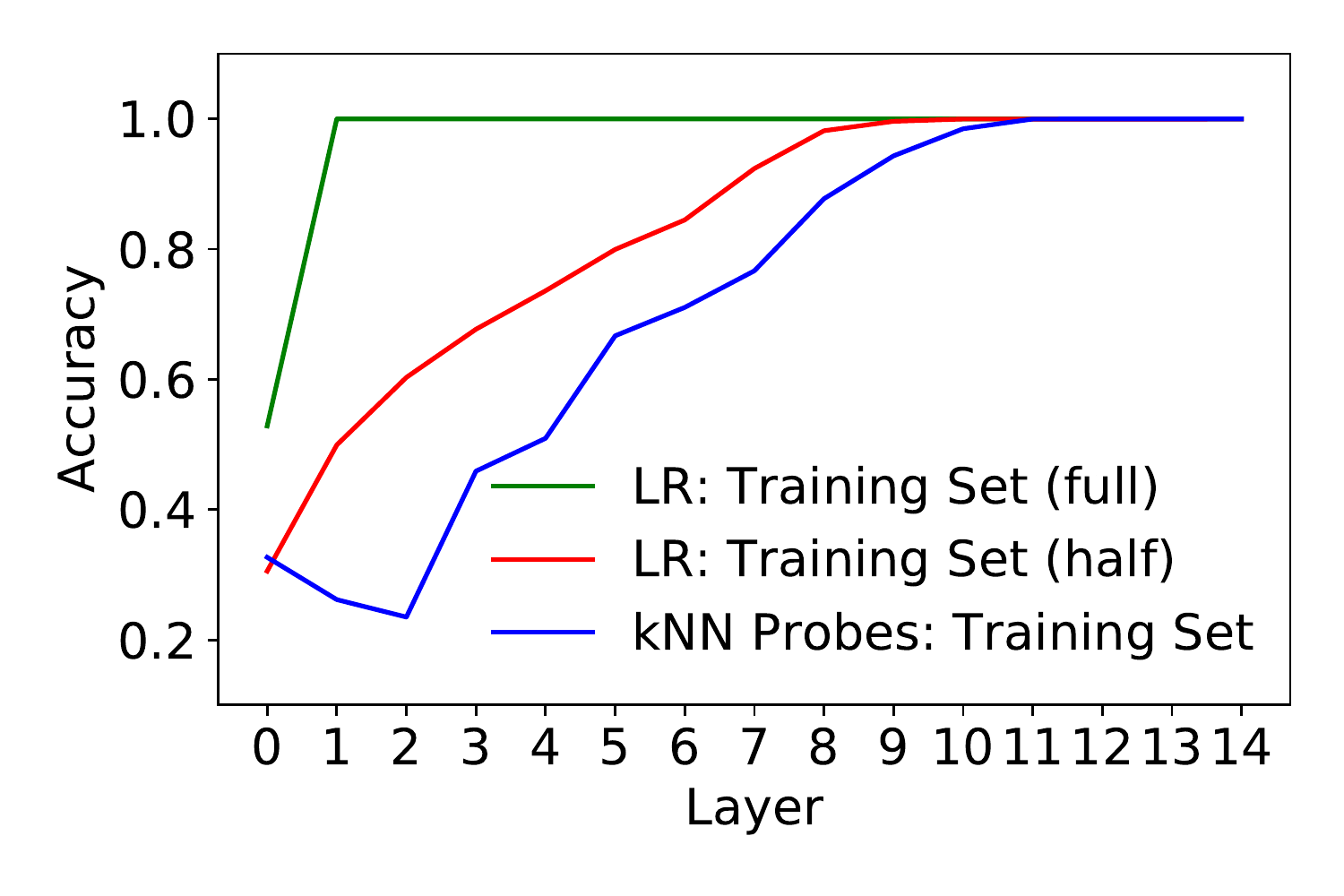}
         \caption{
         Comparison of k-NN probe and Logistic Regression (LR) probe accuracies for VGG16 trained on CIFAR10.
         LR is already able to divide the training set into linearly separated classes after the first convolutional operation.
         In red we show the accuracy of LR probes trained on a random subset (half) of the data and predicting on the other half. These results are converged (closely repeatable between different trained VGG16 models).
\label{fig:knn_lr}}
\end{center}
\end{figure}

Figure~\ref{fig:knn_lr} compares the behavior of k-NN probes and Logistic Regression (LR) probes after the convolution operations of VGG16 with CIFAR10.
LR is able to completely separate the training set after the first convolution operation. 
We also show the behavior when training LR on a random 50\% of the dataset and predicting on the other half.
k-NN shows lower accuracy until the classes become entirely clustered.
We chose k-NN probes for this investigation.

\end{document}